\documentclass[5p]{elsarticle}

\pdfoutput=1

\usepackage{mathrsfs} 
\usepackage{amsmath}
\usepackage{graphicx}
\usepackage{verbatim}
\usepackage{pgfplots}
\usepackage[justification=centering]{caption}
\usepackage{enumitem}
\usepackage{subcaption}
\usepackage{lineno}
\usepackage{hyperref}
\usepackage[graphicx]{realboxes}
\usepackage{xcolor}
\usepackage{bm}
\usepackage{rotating}
\usepackage{mathtools}
\usepackage{float}
\usepackage{leftidx}
\usepackage{multicol}	
\usepackage{multirow}	
\usepackage{longtable}	
\usepackage{threeparttablex}    
\usepackage{array}
\usepackage{booktabs}
\usepackage{color}
\usepackage{tikz}
\usepackage{pgfplots}
\usepackage{pgf-umlsd}
\usepackage{ifthen}
\usepackage{amssymb}
\usepackage{pdflscape}
\usepackage{makecell}
\usepackage{pifont}

\setlist[itemize]{noitemsep, nolistsep}
\modulolinenumbers[5]
\usepgfplotslibrary{groupplots}

\newcommand{\cmark}{\ding{51}}%
\newcommand{\xmark}{\ding{55}}%
\newcommand{\etal}{\textit{et al. }}

\journal{Robotics and Autonomous Systems}

\begin{document}

\begin{frontmatter}

\title{On Deep Learning Techniques to Boost Monocular Depth Estimation for Autonomous Navigation}

\author[address1]{Raul de Queiroz Mendes\corref{corauthor}}
\ead{raulmendes@usp.br}
\cortext[corauthor]{Corresponding author}
\author[address1]{Eduardo Godinho Ribeiro}
\ead{eduardogr@usp.br}
\author[address1]{Nicolas dos Santos Rosa}
\ead{nicolas.rosa@usp.br}
\author[address1]{Valdir Grassi Jr}
\ead{vgrassi@usp.br}

\address[address1]{Department of Computer and Electrical Engineering, S\~{a}o Carlos School of Engineering, University of S\~{a}o Paulo, Brazil}


\begin{abstract}
Inferring the depth of images is a fundamental inverse problem within the field of Computer Vision since depth information is obtained through 2D images, which can be generated from infinite possibilities of observed real scenes. Benefiting from the progress of Convolutional Neural Networks (CNNs) to explore structural features and spatial image information, Single Image Depth Estimation (SIDE) is often highlighted in scopes of scientific and technological innovation, as this concept provides advantages related to its low implementation cost and robustness to environmental conditions. In the context of autonomous vehicles, state-of-the-art CNNs optimize the SIDE task by producing high-quality depth maps, which are essential during the autonomous navigation process in different locations. However, such networks are usually supervised by sparse and noisy depth data, from Light Detection and Ranging (LiDAR) laser scans, and are carried out at high computational cost, requiring high-performance Graphic Processing Units (GPUs). Therefore, we propose a new lightweight and fast supervised CNN architecture combined with novel feature extraction models which are designed for real-world autonomous navigation. We also introduce an efficient surface normals module, jointly with a simple geometric 2.5D loss function, to solve SIDE problems. We also innovate by incorporating multiple Deep Learning techniques, such as the use of densification algorithms and additional semantic, surface normals and depth information to train our framework. The method introduced in this work focuses on robotic applications in indoor and outdoor environments and its results are evaluated on the competitive and publicly available NYU Depth V2 and KITTI Depth datasets.
\end{abstract}


\end{frontmatter}


\section{Introduction}

Autonomous navigation technologies are increasingly present in the academy, industries, agriculture and, to a lesser extent, on the streets and homes to aid in everyday tasks. However, many challenges still need to be solved to make autonomous vehicles a reality for the population. One of these challenges involves the improvement of perception mechanisms that integrate the robotic platforms of the autonomous system. These mechanisms are responsible for mapping the environment in which the vehicle is inserted, helping it to comprehend the surrounding 3D space \cite{mancini2017toward}.

In order to understand the 3D space, the perception system must be able to estimate the distance in which the objects are positioned in the scene. From this, the autonomous vehicle may avoid obstacles and perform safer traffic on navigable surfaces of different scenarios such as cities and roads. Therefore, it is essential that robotic platforms have the ability to explore depth cues of the environment, which can also benefit other Computer Vision tasks such as plane estimation \cite{wang2016surge}, occluding contours estimation \cite{ramamonjisoa2019sharpnet}, object detection \cite{chang2019deep}, Visual Odometry (VO) \cite{loo2019cnn} and Simultaneous Localization and Mapping (SLAM) \cite{tateno2017cnn}.

RGBD sensors, Light Detection and Ranging (LiDAR), stereo cameras, RADAR and SONAR are widely commercialized technologies for depth sensing. However, there are limitations regarding the use of these technologies for depth perception applications. LiDAR sensors are cost-prohibitive and perform sparse and noisy depth measurements over long distances, Kinect sensors are sensitive to light and have a lower distance measurement limit when compared to LiDAR, stereo cameras fail in reflective and low texture regions, whereas RADAR and SONAR are generally used as auxiliaries. 

Thus, the use of monocular cameras, in the context of robotic perception, becomes advantageous, once they demand a smaller installation space, are low cost, capture RGB images, are robust to environmental conditions, are efficient in terms of energy consumption and are a widespread technology \cite{ma2018sparse}. 

Recently, the advance of Convolutional Neural Networks (CNNs), in addition to the availability of databases built with structured-light and LiDAR sensors, motivated the emergence of new works in the Single Image Depth Estimation (SIDE) area. However, state-of-the-art SIDE approaches use sparse and noisy ground truth to train networks and do not fully benefit from how depth, semantics and surface normals can be added to improve the performance of the method regarding speed and accuracy. Moreover, deep models applied to the SIDE task still present significant estimation errors which indicate that the area is open to new solutions.

In this work, we aim to introduce a Fully Convolutional Network (FCN), with an encoder-decoder structure, to tackle single-view depth estimation problems in robotic applications. For the training of this network, we centered on developing a simple 2.5D loss function that focuses on geometric cues of the image. Such a loss is used in conjunction with a novel module that estimates surface normals based on the depth prediction step.

At the same time, this work also shows how some Deep Learning methods may be applied to improve the results of our framework with respect to the quality and accuracy of the predictions. Among such methods, we employ: depth completion \cite{van2019sparse,ma2019self}, semantic segmentation \cite{guizilini2020semantically}, surface normals estimation \cite{yin2019enforcing}, Atrous Spatial Pyramid Pooling (ASPP) \cite{chen2017deeplab}, skip-connections \cite{lee2019big}, multi-scale models \cite{xu2017multi} and multi-layered deconvolutions \cite{alhashim2018high}.

The developed pipeline is tested and evaluated indoors using the NYU Depth V2 dataset \cite{silberman2011indoor} and outdoors using the KITTI Depth dataset \cite{uhrig2017sparsity}. Moreover, our work provides comprehensive surveys over the monocular depth estimation and depth completion areas and we compare the proposed method with other approaches in the SIDE literature with the aid of metrics that are widely disseminated.

The main contributions of this paper are:
\begin{itemize}
    \item Extensive and detailed surveys on monocular depth estimation and depth completion;
    \item A simple CNN architecture, along with four feature extraction models, that can be applied for both SIDE and correlated tasks at a high frame rate;
    \item A lightweight module that is capable of estimating surface normals with state-of-the-art accuracy;
    \item A practical 2.5D loss function that is employed in the depth and surface normals experiments;
    \item Ablation studies considering variations in loss functions, network structures, fine-tuning techniques and additional training information; 
    \item Application of different indoor and outdoor datasets and evaluation on the publicly available and competitive KITTI Depth \cite{uhrig2017sparsity} and NYU Depth V2 \cite{silberman2012indoor}.
\end{itemize}


\section{Depth Estimation: A Brief Survey}

There is a considerable amount of classic methods in the literature that tackle monocular depth inference problems \cite{hoiem2008closing, hoiem2007recovering, hedau2009recovering}. However, with the development of more complex and efficient deep CNNs \cite{huang2017densely, howard2017mobilenets, xie2017aggregated, hu2018squeeze}, several recent works started to exploit such networks, besides graphic models, to extract dense feature maps from RGB images in SIDE applications.

For the presented task, typical CNN models have a fully convolutional architecture with an encoder-decoder structure. The encoder extracts feature maps from the input and the decoder retrieves full resolution images. The training process is usually performed by minimizing a regression loss function such as the Mean Squared Error (MSE) \cite{fu2018deep} and the Mean Absolute Error (MAE) \cite{godard2017unsupervised}. Regarding the type of learning, SIDE models can be trained in three different manners: supervised, semi-supervised and self-supervised.

\subsection{Supervised Learning}

This type of SIDE approach uses sparse and noisy reference depth maps, constructed through point clouds from laser scans, to train supervised deep CNNs. 

Eigen \etal \cite{eigen2014depth} presented a fundamental work in which the developed architecture is composed of two stacks. The first stack estimates the scene depth from a global point of view and the second stack performs local refinements. The authors also introduced a scale-invariant loss function (SILog) that highlights the depth relationships of the image pixels, instead of focusing on the general scale.

With another fundamental work, Laina \etal \cite{laina2016deeper} introduced an FCN composed of up-projection and residual learning mechanisms \cite{he2016deep} to predict accurate depth maps in a faster way. The authors also presented the benefits of using the Huber reverse loss function (BerHu) \cite{zwald2012berhu,girshick2015fast} to train SIDE networks.

Exploring the benefits of graphic models, Liu \etal \cite{liu2015deep} combined a CNN with a continuous Conditional Random Field (CRF) and proposed a new superpixel pooling method that addresses the problem of information loss after successive downsampling operations. Cao \etal \cite{cao2017estimating} used classification methods, as opposed to common regression techniques, through the discretization of the ground truth (depth bins). Their results show that SIDE by classification can surpass the conventional regression. However, the accuracy depends on the number of depth bins since the loss function is not able to differentiate depth values within a bin. 

Inspired by this idea, Liao \etal \cite{liao2017parse} employed a regression loss combined with a classification function to improve accuracy. The authors also introduced a CNN that outputs depth maps from RGB images concatenated with reference depth data from 2D laser scans. 

Using different approaches, Xu \etal \cite{xu2017multi} proposed a network structure that integrates a CNN with a fusion module composed of continuous CRFs. Fu \etal \cite{fu2018deep} developed an ordinal regression method based on a spacing-increasing discretization technique. Exploiting transfer learning strategies, Alhashim \etal \cite{alhashim2018high} employed the DenseNet-169 \cite{huang2017densely} model pretrained on the ILSVRC \cite{russakovsky2015imagenet} in a new FCN architecture. 

Recently, Lee \etal \cite{lee2019big} introduced a deep network architecture that applies the DenseNet-161 \cite{huang2017densely} and the ResNext-101 \cite{xie2017aggregated} as encoder. Coupled with the encoder structure, the authors used a dense multi-scale feature learning module (Dense ASPP), followed by a decoder with convolution and upconvolution blocks, downsampling and concatenation operations and Local Planar Guidance layers (LPG). 

Exploring temporal features, Mancini \etal \cite{mancini2017toward} introduced Fully Connected Long Short-Term Memory (FC-LSTM) \cite{hochreiter1997long} layers in a CNN to address the monocular depth prediction task. However, to use image sequences as input to an FC-LSTM, they need to be transformed into 1D vectors, which impairs the network's learning of spatial relationships. Conv-LSTM \cite{xingjian2015convolutional}, on the other hand, is a type of neural layer that allows the extraction of spatial and temporal features. This layer is employed in the works of Kumar \etal \cite{cs2018depthnet} and Wang \etal \cite{wang2019recurrent}.

Applying attention-based techniques, which are adaptive and allow only features that have meaningful information to be focused on during training, Xu \etal \cite{xu2018structured} proposed a multi-scale CNN composed of an attention-guided CRF module to perform single-view depth prediction. Following a resembling idea, Chen \etal \cite{chen2019attention} introduced an attention-based context aggregation CNN that aggregates contextual information at pixel and global levels from the feature maps. 

\subsubsection{SIDE and Semantic Segmentation}

Semantic segmentation and monocular depth estimation are fundamental and correlated tasks in Computer Vision, since, from them, it is possible to infer scene geometry, artifacts scale and location, the scene’s structure and the distances at which objects are located \cite{jiao2018look}. Moreover, the aforementioned tasks benefit mutually, as understanding the depth of the scene helps semantic segmentation during categorization, especially for the case when there are different artifacts with similar structures and at different depths \cite{wang2015towards}. On the other hand, semantic labels provide geometric and perspective information that assists the depth estimation of each object in the scene, improving mainly the predictions at high gradient regions (borders). 

Classic approaches already employ methods addressing semantic segmentation to estimate depth \cite{hoiem2011recovering, liu2010single} and RGBD images to obtain precise semantic classes \cite{silberman2012indoor, gupta2014learning}. Using CNNs, Eigen \etal \cite{eigen2015predicting} introduced a deep multi-scale network, which, with minor modifications, can predict depth, surface normals and semantic maps. Benefiting from graphic models, Mousavian \etal \cite{mousavian2016joint} developed a jointly CNN and CRF to estimate depth and semantics. Wang \etal \cite{wang2015towards} also tackled SIDE and semantics simultaneously using two CNNs and a Hierarchical CRF. Going further, Jiao \etal \cite{jiao2018look} achieved impressive accuracy approaching both tasks with a new CNN, whose training is based on minimizing an attention-guided loss function. 

\subsubsection{SIDE and Surface Normals Estimation}

The first studies on 3D understanding sought to obtain, among other information, illumination changes and color intensities \cite{zhang1999shape} and volumetric shapes \cite{binford1971visual}. Due to their limitations, later works focused on other types of geometric cues such as segments and vanishing points \cite{hoiem2007recovering} and super-pixels \cite{saxena2007learning}. However, these methods depend on volumetric relationships and structural constraints. With the aid of datasets generated by RGBD sensors, later methods sought to estimate the 2.5D layout of scenes through depth and surface normals estimations, which are highly correlated. Some of these works were developed by Silberman \etal \cite{silberman2012indoor} and Fouhay \etal \cite{fouhey2013data,fouhey2014unfolding}.

In the deep learning scenario, Wang \etal \cite{wang2016surge} introduced a framework that estimates depth and surface normals through a four-stream CNN, coupled with a dense CRF, which retrieves planar surfaces and edges, as well as partial depth and surface normals. Also in a multitask context, Qi \etal \cite{qi2018geonet} created a network with two branches, called normal-to-depth and depth-to-normal, for depth and surface normals estimations. Based on the affinity rate between pixel pairs produced in correlated tasks, Zhang \etal\cite{zhang2019pattern} proposed a CNN that explores similarity patterns between corresponding pixels, through an affinity matrix, to generate depth, semantics and surface normals maps simultaneously.

Rather than using multi-scale decoding structures, Yin \etal \cite{yin2019enforcing} developed a CNN that performs SIDE and benefits from 3D geometric features of the scene through virtual normals. From another perspective, Ramamonjisoa \etal\cite{ramamonjisoa2019sharpnet} introduced the SharpNet, which predicts depth, surface normals and occluding contours, as well as a loss function that forces consistency between depth and occluding contours and between depth and surface normals. 

\subsection{Self-supervised Learning}

With the premise that supervised methods demand a large amount of ground truth depth data, which require significant time and effort to be produced, self-supervised learning strategies began to be developed. The emergence of self-supervised methods showed that it is possible to train SIDE models using only synchronized image pairs from stereo rigs \cite{garg2016unsupervised, godard2017unsupervised, pillai2019superdepth} or sequences of frames \cite{zhou2017unsupervised}.

\subsubsection{Training with Stereo Images}

With a breakthrough work, Garg \etal \cite{garg2016unsupervised} proposed a CNN that is fed with pairs of stereo images and retrieves depth maps. The model learns the nonlinear transformations necessary to recover the depth map using a loss function equivalent to the photometric difference between images. Godard \etal \cite{godard2017unsupervised} also formulated a deep CNN, called Monodepth, which receives only the left image of the stereo pair during testing. During training, their network learns to infer the disparity maps of both camera images. Going further, Godard \etal \cite{godard2019digging} improved Monodepth with a loss function able to reduce the number of artifacts, to deal with occluded pixels and to disregard pixels that do not change in a sequence of images.

With a semi-supervised scope, the pipelines proposed by Kuznietsov \etal \cite{kuznietsov2017semi} and Amiri \etal \cite{amiri2019semi} employ supervised learning from sparse data produced by LiDAR sensors, as well as self-supervised training reasoned by stereo pairs. On the other hand, the pipelines proposed in the works of Yang \etal \cite{yang2018deep} and Andraghetti \etal \cite{andraghetti2019enhancing} use a self-supervised training strategy, based on stereo images, and also a supervised training procedure assisted by VO. Furthermore, Ramirez \etal \cite{ramirez2018geometry} was the first to consider stereo self-supervised learning along with the supervision of semantic classes.

Following the stereo matching setup, Luo \etal \cite{luo2018single} and Tosi \etal \cite{tosi2019learning} presented SIDE frameworks that synthesize the right view of the stereo pair from the original left image. In a multi-task approach, Li \etal \cite{li2018undeepvo} introduced a deep network that estimates depth and 6D-camera pose with the assistance of a loss function that addresses the spatial and temporal aspects of the incoming image pairs. Also, Babu \etal \cite{babu2018undemon} developed a framework that predicts depth maps and camera pose with the help of a loss based on the Charbonnier penalty \cite{charbonnier1994two}. This penalty expression is used in both spatial and temporal reconstruction errors that constitute the objective function.

Other stereo-based methods resort to trinocular training \cite{poggi2018learning}, synthetic training data \cite{guo2018learning}, generative adversarial training \cite{cs2018monocular,aleotti2018generative,gupta2019unsupervised}, uncertainty handling \cite{chen2020self,poggi2020uncertainty}, occlusion handling \cite{gonzalezbello2020forget} and knowledge distillation \cite{pilzer2018unsupervised}. Network architectures that are lighter and faster for real-time applications in embedded systems \cite{poggi2018towards,peluso2019enabling,elkerdawy2019lightweight} are also applied in the stereo matching self-supervised approach of SIDE.

\subsubsection{Training with Monocular Video}

With the first efforts to perform self-supervised depth estimation from monocular video, Zhou \etal \cite{zhou2017unsupervised} proposed a self-supervised model that performs SIDE and camera pose estimation. On the other hand, Yang \etal \cite{yang2018unsupervised} developed a CNN-based method that estimates depth and surface normals through edge recognition. In a later work, Yang \etal \cite{yang2018lego} introduced an architecture that retrieves depth maps, surface normals and edge representations using a regularization method named 3D as-smooth-as-possible.

Also with a multitask approach, Yin \etal \cite{yin2018geonet} proposed a self-supervised framework that predicts depth, optical flow and camera pose through an adaptive geometric consistency loss function, which is robust to occlusions and outliers. In parallel, Mahjourian \etal \cite{mahjourian2018unsupervised} introduced a method that estimates depth and ego-motion from temporally consistent adjacent images using a 3D geometric loss function. Wang \etal \cite{wang2018learning} leveraged depth normalization operations and a differentiable version of a direct VO algorithm. Showing promising results on SIDE and VO, the method from Casser \etal \cite{casser2019depth} can model dynamic objects present in the scene with the support of instance segmentation masks. 

Other SIDE strategies, self-supervised by monocular videos, employ semantics, flow and camera motion information \cite{teed2018deepv2d,casser2019unsupervised,ranjan2019competitive,chen2019self,guizilini2020semantically}, as well as other types of geometric constraints \cite{chen2019self, guizilini2019packnet,sharma2019unsupervised} and lightweight CNNs \cite{liu2020mininet}.

The monocular video approach is challenging since the deep network needs to estimate the camera transformations through the sequence of input images, in addition to the depth of the scene. CNNs, trained with stereo images, do not need to estimate the camera pose, as they explore the correspondence of both images of the stereo pair. However, as a disadvantage, faults are prone to occur in areas close to occlusions \cite{godard2019digging}.

\subsection{Depth Completion}

Depth completion is distinct from the SIDE task, as the former is focused on the densification of sparse and noisy point cloud data, obtained from light-structured sensors, LIDAR sensors or SLAM/Structure from Motion (SfM) algorithms \cite{cheng2019cspn++}. In addition to filling data into depth channel spaces without information (depth inpainting) \cite{herrera2013depth, zhang2018probability}, the depth completion task also aims at depth denoising \cite{zhang2016fast, shen2013layer} and depth super-resolution \cite{yu2013shading, lu2015sparse}.

The first depth completion strategies typically resorted to handcrafted features to infer dense depth or disparity images \cite{qiu2019deeplidar}. In this context, Ku \etal \cite{ku2018defense} presented an unguided depth completion algorithm, based on inversion, expansion and morphological closing. On the other hand, the method of Schneider \etal \cite{schneider2016semantically} exploits the guidance of semantic labels and edge maps to estimate dense images. However, such classic approaches cannot generalize well for different types of scenes, demanding fine adjustments of specific parameters for each circumstance. With that in mind, these classic algorithms were outperformed by novel learning-based models, which can be divided regarding the presence or absence of a guidance RGB image.

\subsubsection{Depth Completion from Sparse Samples}

In this category, depth completion techniques retrieve dense depth maps from sparse depth samples. Uhrig \etal \cite{uhrig2017sparsity} introduced a sparse convolutional network composed of sparse convolutional layers that handle sparsity through the use of observation masks. Such masks hold the valid pixels information from the input feature maps, which are propagated by pooling operations.

Enhancing the normalized convolution approach, Eldesokey \etal \cite{eldesokey2019confidence} proposed an unguided CNN that receives sparse depth and confidence maps as input. This network employs new normalized convolutions that infer confidence maps and transmit them to the adjacent layer. Chodosh \etal \cite{chodosh2018deep} used established concepts of sparse representation learning (dictionary learning) in a neural network pipeline, based on the alternating direction neural network, to address the task of depth completion. 

Huang \etal \cite{huang2019hms} developed a CNN that benefits from observation masks and sparsity-invariant operations such as upsampling, average and convolution. Recently, sparsity-invariant convolutions \cite{uhrig2017sparsity} are being revisited and complemented by novel mask aware operations to handle the RGB guided depth completion task \cite{yan2020revisiting}. Other few works are focused on closing the gap between unguided and guided depth completion methods \cite{lu2020depth}.

\subsubsection{Guided Depth Completion}

The approaches that deal with guided depth completion leverage RGB image cues to densify sparse and noisy depth maps \cite{tang2019learning}. Ma \etal \cite{ma2018sparse} proposed a CNN that is fed with color images and sparse depth data sampled with the Bernoulli probability strategy. This deep model can predict dense depth maps from different numbers of target samples. In a later work, Ma \etal \cite{ma2019self} introduced a self-supervised pipeline that handles sparse data and also receives sequences of RGB images as input. 

Addressing both SIDE and depth completion, Cheng \etal \cite{cheng2018depth} designed a framework that employs a Convolutional Spatial Propagation Network (CSPN). This CNN propagates through the application of recurrent convolutions in a fixed neighborhood and, due to this, it can learn the affinity matrix related to the pixels of the output depth map from the depth estimation stage. Posteriorly, Cheng \etal \cite{cheng2019cspn++} enhanced the CSPN's depth completion performance through an adaptive technique that infers some network's hyper-parameters.

Inspired by the CSPN, the network architecture proposed by Park \etal \cite{park2020non} learns the affinities of the estimated depth map pixels, and its confidences, in a non-local neighborhood. In another variation, Xu \etal \cite{xu2020deformable} designed a deformable spatial propagation network that propagates with adaptive receptive fields and affinity matrices. 

On the other hand, Chen \etal \cite{chen2019learning} developed a depth completion CNN that learns 2D-3D features in a multi-level way, while Li \etal \cite{li2020multi} used blocks of hourglass networks that are fed with sparse samples and guided by the encoder features. Recently, Tang \etal \cite{tang2019learning} designed a guided method that employs spatially variable convolution kernels. Other works also benefit from morphological operators \cite{dimitrievski2018learning}, additional surface normals information \cite{qiu2019deeplidar,xu2019depth}, phase masks \cite{wu2019phasecam3d} and cross-guidance techniques \cite{lee2020deep}.

\subsection{Final Considerations}

From Table \ref{tab:litside}, we can see that state-of-the-art supervised CNNs reach a prediction speed that does not exceed 20 fps, using powerful Graphic Processing Units (GPUs) and feature extraction networks with no less than 25M trainable parameters. However, comparing the results obtained in \cite{casser2019depth,godard2019digging,liu2020mininet,poggi2020uncertainty} with \cite{fu2018deep,lee2019big,yin2019enforcing}, we have that supervised methods still obtain more accurate depth predictions. Furthermore, self-supervised methods, in some cases, require VO techniques \cite{wang2018learning,casser2019unsupervised,andraghetti2019enhancing}, semi-supervised pipelines \cite{kuznietsov2017semi,klodt2018supervising,tosi2019learning}, both images from the stereo pair during training \cite{garg2016unsupervised,pilzer2018unsupervised,gonzalezbello2020forget}, as well as optical flow, semantic and surface normals cues \cite{yang2018lego,zou2018df,ranjan2019competitive,guizilini2020semantically} to obtain better results.

Therefore, this work aims to explore the benefits of supervised networks, along with other deep learning techniques, to solve problems that involve SIDE. The next section shows how the current proposal is correlated to state-of-the-art FCN architectures, which employ supervised learning to perform monocular depth estimation. Other deep learning methods, which include the tasks of semantic segmentation, surface normals estimation and depth completion, are also covered.

\begin{table*}[htb!]
\centering
\caption{Specifications of SIDE state-of-the-art works. The symbol $\dagger$ indicates the total number of trainable parameters of the encoder, while $\ddagger$ denotes the total size of the framework}
\label{tab:litside}
\resizebox{1.0\textwidth}{!}{
\begin{tabular}{c|c|c|c|c|c}
\hline
\textbf{Method} & \textbf{Dataset} & \textbf{GPU} & \textbf{Network encoder} & \textbf{Parameters} & \textbf{Prediction speed} \\\hline \hline
BTS \cite{lee2019big} & KITTI Depth / NYU Depth V2 & NVIDIA GTX 1080 Ti & DenseNet-161 & 47\ M$\ddagger$ & $16\ fps$ \\
DORN \cite{fu2018deep} & KITTI Depth / Make3D / NYU Depth V2 & NVIDIA Titan X Pascal & ResNet-101 & 110\ M$\ddagger$ & $2\ fps$\\
VNL \cite{yin2019enforcing} & KITTI Depth / NYU Depth V2 & Huawei GPU-accelerated Cloud Server & ResNext-101 $(32\times4d)$ & 44\ M$\dagger$ & $2\ fps$\\
DenseDepth \cite{alhashim2018high} & KITTI Depth / NYU Depth V2 & NVIDIA TITAN Xp & DenseNet-169 & 42\ M$\ddagger$ & -\\
ACAN \cite{chen2019attention} & KITTI Depth / NYU Depth V2 & NVIDIA GTX 1080 Ti & ResNet-101 & 44\ M$\dagger$ & -\\
PAP-Depth \cite{zhang2019pattern} & KITTI Depth / SUNRGBD / NYU Depth V2 & NVIDIA P40 & ResNet-50 & 25\ M$\dagger$ & $5\ fps$\\
DABC \cite{li2018deep} & ScanNet / KITTI Depth & NVIDIA GTX 1080 & ResNext-101 $(64\times4d)$ & - & $1\ fps$\\
APMoE \cite{kong2019pixel} & NYU Depth V2 / BSDS500 / Stanford-2D-3D / Cityscapes & NVIDIA Titan X & ResNest-50 & 25\ M$\dagger$ & $5\ fps$\\
CSWS \cite{li2018monocular} &  NYU Depth V2 / KITTI Depth & NVIDIA Tesla Titan X & ResNet-152 & 60\ M$\dagger$ & $5\ fps$\\
Cao \etal\cite{cao2017estimating} &  NYU Depth V2 / KITTI Depth / SUNRGBD & - & ResNet-152 & 82\ M$\ddagger$ & -\\
Kuznietsov \etal \cite{kuznietsov2017semi} & KITTI Depth & NVIDIA GTX 980 Ti & ResNet-50 & 80\ M$\ddagger$ & -\\
LSIM \cite{goldman2019learn} & KITTI Depth & NVIDIA Titan X & ResNet-50 & 25\ M$\dagger$ & $12\ fps$\\
DHGRL \cite{zhang2018deeph} & KITTI Depth / Make3D / NYU Depth V2 & NVIDIA K80 & ResNet-50 & 25\ M$\dagger$ & $5\ fps$\\
SDNet \cite{ochs2019sdnet} & KITTI Depth / Cityscapes & NVIDIA Titan Xp & ResNet-50 & 25\ M$\dagger$ & $5\ fps$\\
monoResMatch \cite{tosi2019learning} & KITTI Depth / Cityscapes & NVIDIA 2080 Ti & monoResMatch & 42\ M$\ddagger$ & $29\ fps$\\
monoResMatch \cite{tosi2019learning} & KITTI Depth / Cityscapes & Jetson TX2 & monoResMatch & 42\ M$\ddagger$ & $1\ fps$\\
3Net \cite{poggi2018learning} & KITTI Depth / Cityscapes & NVIDIA 2080 Ti & ResNet-50 & 25\ M$\dagger$ & $49\ fps$\\
3Net \cite{poggi2018learning} & KITTI Depth / Cityscapes & Jetson TX2 & ResNet-50 & 25\ M$\dagger$ & $2\ fps$\\
VOMonodepth \cite{andraghetti2019enhancing} & KITTI Depth & NVIDIA 2080 Ti & ResNet-50 & 25\ M$\dagger$ & $41\ fps$\\
VOMonodepth \cite{andraghetti2019enhancing} & KITTI Depth & Jetson TX2 & ResNet-50 & 25\ M$\dagger$ & $1\ fps$\\
Monodepth \cite{godard2017unsupervised} & KITTI Depth / Make3D & NVIDIA Titan X & ResNet-50 & 31\ M$\ddagger$ &$28\ fps$\\
Monodepth2 \cite{godard2019digging} & KITTI Depth & NVIDIA Titan X & ResNet-18 & 14\ M$\ddagger$ & -\\
SA-Attention \cite{zhou2019unsupervised} & KITTI Depth / Make3D & - & ResNet-50 & 34\ M$\ddagger$ & -\\
Struct2depth \cite{casser2019depth} & KITTI Depth / Cityscapes / Fetch Indoor Navigation & GTX 1080Ti & Struct2depth & 14\ M$\ddagger$ & $30\ fps$\\
\hline
\end{tabular}
} 
\end{table*}


\section{Method}
\label{metodologia}

In this section, we exploit the concepts behind our novel supervised monocular depth estimation method. We aim to introduce the details from the proposed baseline CNN architecture, which is divided into the encoder and decoder well-defined phases. Along with such architecture, $4$ different encoder models are presented. The developed models are used in the ablation studies step so that we may verify the one that best fits our framework. We also cover a significant variety of loss functions that are employed in the monocular depth estimation studies. 

Later, the pipelines that account for semantic segmentation, surface normals estimation and depth completion are described. Through small modifications in the proposed CNN, the former approach uses some pre-trained weights for semantic segmentation to enhance the depth inference accuracy. In the second strategy, a new surface normals estimation module is developed, which leverages the retrieved depth maps, the feature maps produced by the entire baseline and edge constraints. Furthermore, a simple geometric loss function is introduced to handle both depth and surface normals predictions efficiently. We add a gaussian blur technique to the proposed module to verify its robustness to noisy inputs.

In the last pipeline, we employ the sampling strategy based on the categorical distribution to densify sparse and noisy depth maps for indoor and outdoor environments. All the variations of the baseline network can be executed at a frame rate above $20$. 

Posteriorly, we introduce the experimental phases that cover the implementation details, the datasets used and the evaluation criteria applied to our results. Finally, we present the ablation studies, with the quantitative and qualitative analysis, and the conclusion.

\subsection{Network Architecture}

\begin{figure*}[]
\centering
\includegraphics[width=\textwidth]{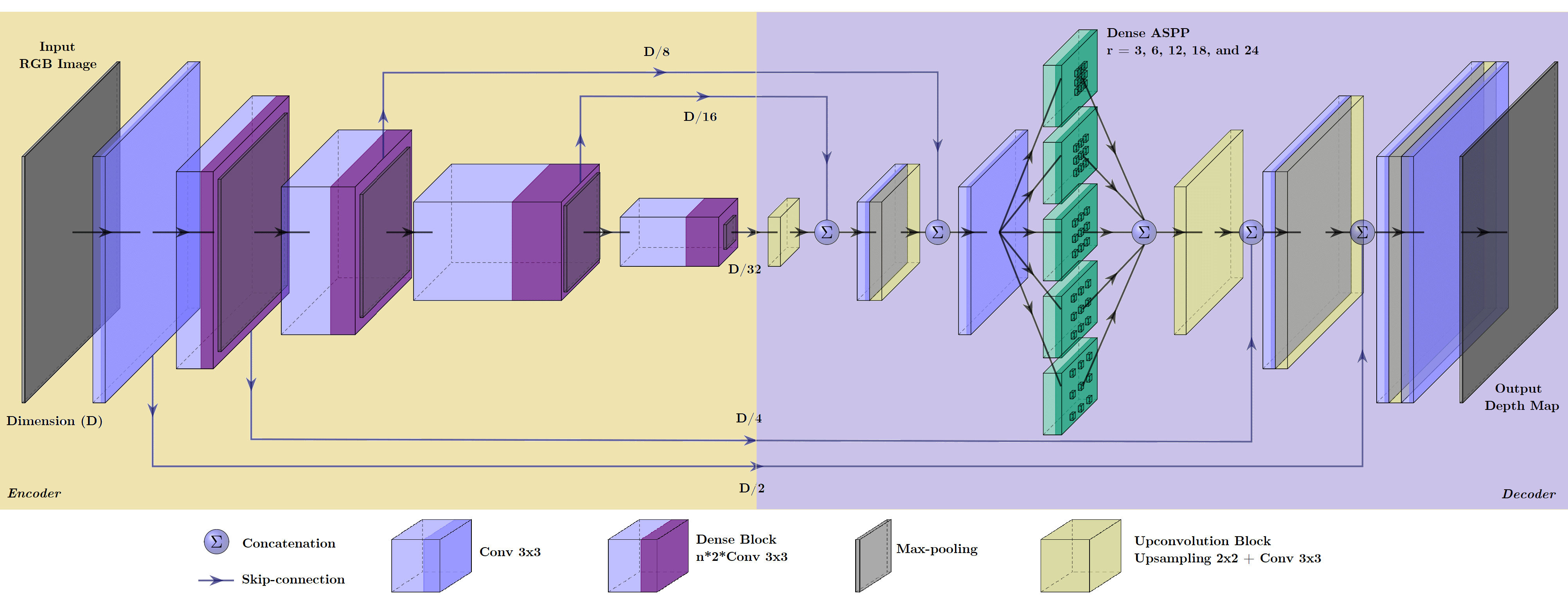}  
\caption{Proposed baseline of the DenseSIDENet}
\label{fig28}
\end{figure*}

The proposed supervised FCN, named  DenseSIDENet (DSN), is inspired by the framework of Lee \etal\cite{lee2019big} since their model is state-of-the-art for the SIDE task. Compared to the most accurate method of Lee \etal\cite{lee2019big}, ours can reach $56 \times$ fewer trainable parameters. Fig. \ref{fig28} illustrates the baseline of the DenseSIDENet highlighting the encoder-decoder structure.

We implemented as encoder the following lightweight feature extraction CNNs: MobileNet \cite{sandler2018mobilenetv2}, DenseNet-121 \cite{huang2017densely} and new variations of the robotic grasp detection network proposed by Ribeiro \etal \cite{ribeiro2019fast}. The aforementioned networks have $0.6M$ (up to $4M$), $6M$ and $0.3M$ (up to $9M$) trainable parameters, respectively, which makes it possible to evaluate the framework at a high frame rate without losing accuracy.

\begin{figure}[]
\centering
\includegraphics[width=\linewidth]{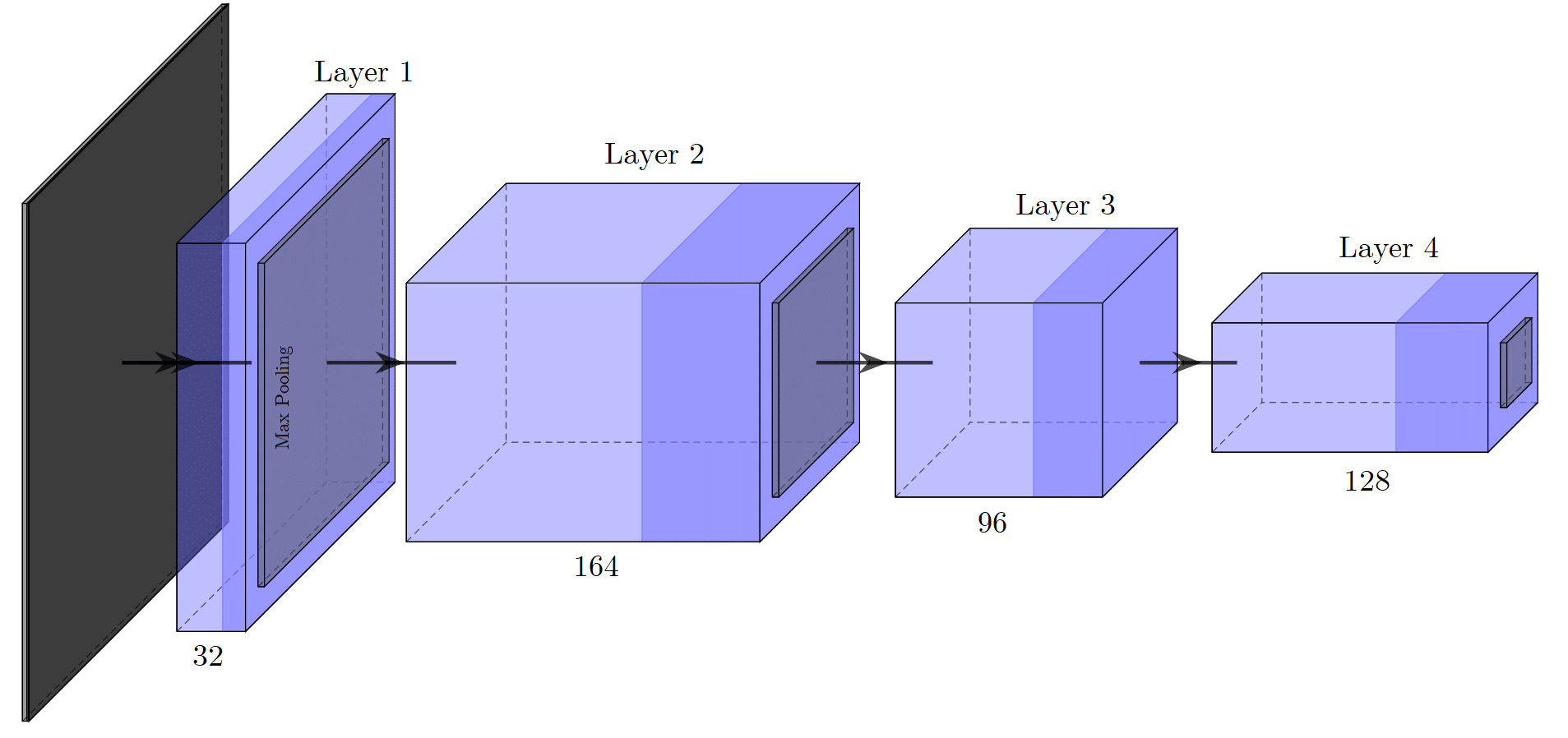}\\  
\caption{Modified structure of the CNN proposed by Ribeiro \etal \cite{ribeiro2019fast} for the task of robotic grasping detection} 
\label{figgrasp}
\end{figure}

The CNN depicted in Fig. \ref{figgrasp} refers to one of the proposed variations of Ribeiro's model \cite{ribeiro2019fast}. Due to its efficiency and size, it is designed for the DSN encoder stage. Particularly, the number of filters ($ \{32, 164, 96, 128 \}$) and the kernel size ($3\times 3$) of each convolution are maintained from the original network. However, the flatten operation, the FC layers and the output layer are removed. 

We also developed a dense configuration of the network of Fig. \ref{figgrasp}. The standard structure of the DenseNet presents a sequence of dense convolution blocks interspersed with transition layers which are composed of convolution and average pooling operations. Therefore, layers $1$, $2$, $3$ and $4$ of the architecture in Fig. \ref {figgrasp} are redesigned to resemble dense blocks alternated with transition layers. However, the filter ratio ($\{32, 164, 96, 128\}$), the kernel size ($3\times 3$) and the max-pooling operations are preserved, according to what is proposed by Ribeiro \etal\cite{ribeiro2019fast}. 

Table \ref{tab:netarch} details $3$ lightweight versions of the proposed dense architecture. For each model, $k$ refers to the growth rate of the filters, which is added at the end of the dense blocks, $\theta$ corresponds to the filter's compression and $t$ refers to the total number of trainable parameters. Beyond the differences presented in Table \ref{tab:netarch}, the stages of the proposed models are initialized with the following filter distribution: $32$ filters for the initial convolution, $164$ filters for the dense blocks $1$, $2$ and $3$, $96$ filters for the transition layers $1$, $2$ and $3$, $128$ filters for the dense block $4$. The first convolutions of the dense blocks are set up with $2\times$ the current number of filters.

\begin{table*}[]
\centering
\caption{Composition of the proposed models to be applied in the feature extraction stage of the DenseSIDENet baseline. For all models, $k$ indicates the filters growth rate, $\theta$ is the filters compression and $t$ is the total number of parameters}
\label{tab:netarch}
\resizebox{1.0\textwidth}{!}{
\begin{tabular}{@{}c|c|c|c@{}}
\hline
\textbf{Layers} & \textbf{Model 1} $(k=6,\ \theta=1,\ t=1M)$ & \textbf{Model 2} $(k=16,\ \theta=1,\ t=3M)$ & \textbf{Model 3} $(k=24,\ \theta=1,\ t=9M)$ \\ 
\hline \hline
Initial Convolution &$3\times3$ conv, stride $2$ &$3\times3$ conv, stride $2$ &$3\times3$ conv, stride $2$ \\\hline
Initial Pooling &$3\times3$ max pooling, stride $2$ &$3\times3$ max pooling, stride $2$ &$3\times3$ max pooling, stride $2$ \\\hline
Dense Block $1$ &$\begin{bmatrix}
3\times3 \textrm{ conv, stride } 1\\
3\times3 \textrm{ conv, stride } 1
\end{bmatrix}$ $\times6$
&$\begin{bmatrix}
3\times3 \textrm{ conv, stride } 1\\
3\times3 \textrm{ conv, stride } 1
\end{bmatrix}$ $\times6$
&$\begin{bmatrix}
3\times3 \textrm{ conv, stride } 1\\
3\times3 \textrm{ conv, stride } 1
\end{bmatrix}$ $\times6$
\\\hline
Transition Layer $1$ &\begin{tabular}{c}
  $3\times3$ conv, stride 1 \\
  $2\times2$ max pooling, stride 2
\end{tabular}
&\begin{tabular}{c}
  $3\times3$ conv, stride 1 \\
  $2\times2$ max pooling, stride 2
\end{tabular}
&\begin{tabular}{c}
  $3\times3$ conv, stride 1 \\
  $2\times2$ max pooling, stride 2
\end{tabular}
\\\hline
Dense Block $2$ &$\begin{bmatrix}
  3\times3 \textrm{ conv, stride } 1\\
  3\times3 \textrm{ conv, stride } 1
\end{bmatrix}$ $\times12$
&$\begin{bmatrix}
3\times3 \textrm{ conv, stride } 1\\
3\times3 \textrm{ conv, stride } 1
\end{bmatrix}$ $\times12$
&$\begin{bmatrix}
3\times3 \textrm{ conv, stride } 1\\
3\times3 \textrm{ conv, stride } 1
\end{bmatrix}$ $\times12$
\\\hline
Transition Layer $2$ &\begin{tabular}{c}
  $3\times3$ conv, stride $1$ \\
  $2\times2$ max pooling, stride $2$
\end{tabular}
&\begin{tabular}{c}
  $3\times3$ conv, stride $1$ \\
  $2\times2$ max pooling, stride $2$
\end{tabular}
&\begin{tabular}{c}
  $3\times3$ conv, stride $1$ \\
  $2\times2$ max pooling, stride $2$
\end{tabular}
\\\hline
Dense Block $3$ &$\begin{bmatrix}
3\times3 \textrm{ conv, stride } 1\\
3\times3 \textrm{ conv, stride } 1
\end{bmatrix}$ $\times18$
&$\begin{bmatrix}
3\times3 \textrm{ conv, stride } 1\\
3\times3 \textrm{ conv, stride } 1
\end{bmatrix}$ $\times18$
&$\begin{bmatrix}
3\times3 \textrm{ conv, stride } 1\\
3\times3 \textrm{ conv, stride } 1
\end{bmatrix}$ $\times24$
\\\hline
Transition Layer $3$ &\begin{tabular}{c}
  $3\times3$ conv, stride $1$ \\
  $2\times2$ max pooling, stride $2$
\end{tabular}
&\begin{tabular}{c}
  $3\times3$ conv, stride $1$ \\
  $2\times2$ max pooling, stride $2$
\end{tabular}
&\begin{tabular}{c}
  $3\times3$ conv, stride $1$ \\
  $2\times2$ max pooling, stride $2$
\end{tabular}
\\\hline
Dense Block $4$ &$\begin{bmatrix}
3\times3 \textrm{ conv, stride } 1\\
3\times3 \textrm{ conv, stride } 1
\end{bmatrix}$ $\times16$
&$\begin{bmatrix}
3\times3 \textrm{ conv, stride } 1\\
3\times3 \textrm{ conv, stride } 1
\end{bmatrix}$ $\times14$
&$\begin{bmatrix}
3\times3 \textrm{ conv, stride } 1\\
3\times3 \textrm{ conv, stride } 1
\end{bmatrix}$ $\times18$
\\
\hline
\end{tabular}
}
\end{table*}

Attached to the DSN encoder, a pair of upconvolution, concatenation and convolution blocks replace the last pooling layers of the feature extraction network to prevent the feature maps resolution from decreasing. Afterwards, a Dense ASPP \cite{yang2018denseaspp} is implemented according to Lee \etal \cite{lee2019big}, who also utilize the dilated rates ${3, 6, 12, 18, 24}$.

Since this module consists of densely connected dilated convolution layers, it does not suffer from the same degradation problems as the original ASPP. Thus, the densely connected ASPP is applied to extract multi-scale features and, at the same time, to avoid the excessive reduction of the receptive field.

Due to this, the DSN decoder stage processes larger feature maps, with multiple-sized receptive fields, and performs less upsampling operations to recover the original image resolution $(D)$ at the output. Fig. \ref{fig28} shows that the decoder stage does not add highly complex neural layers that could increase the number of trainable parameters and decrease the DSN's prediction speed.

As also can be noticed in Fig. \ref{fig28}, the upconvolution block is formed by a convolution layer with $3\times3$ kernel, which is preceded by a nearest neighbors upsampling operation that retrieves feature maps with $2\times$ increased resolution. In the framework, such blocks are followed by batch normalization operations and, subsequently, by skip-connections/concatenations. The first upconvolution block is initialized with only $256$ filters which are decreased by a fraction of $2$ until the last layer. 

Throughout the network structure, ReLU \cite{agarap2018deep} and eLU \cite{clevert2015fast} activation functions are employed, with the exception of the output layer that uses the sigmoid function. The output depth map of the DSN is multiplied by the maximum depth value comprising the dataset used. The network is fed with a single RGB image with $352\times704$ and $416\times544$ pixels and predicts a depth map with $352\times1216$ and $480\times640$ pixels for the outdoor \cite{uhrig2017sparsity,cordts2016cityscapes} and indoor \cite{silberman2012indoor,song2015sun,janoch2013category,xiao2013sun3d} datasets respectively. 

\subsection{SIDE Loss Functions}
\label{fdc}

The losses Scale-invariant Error ($\mathcal{L}_{SILog}$) \cite{eigen2014depth}, modified Scale-invariant Error ($\mathcal{L}_{SILog}^{+}$) \cite{lee2019big}, Huber ($\mathcal{L}_{Huber}$) \cite{carvalho2018regression}, BerHu ($\mathcal{L}_{BerHu}$) \cite{zwald2012berhu}, Mean Absolute Error ($\mathcal{L}_{1}$) \cite{godard2017unsupervised}, Mean Squared Error ($\mathcal{L}_{2}$) \cite{fu2018deep}, Charbonnier ($\mathcal{L}_{charbo}$) \cite{sun2014quantitative} and Log-cosh ($\mathcal{L}_{cosh}$) \cite{chen2018log} are used in the training phase of the network depicted in Fig. \ref{fig28}. The pixels without depth information are disregarded by the loss functions, as they do not influence the adjustment of the network weights.

The $\mathcal{L}_{SILog}$ expression (Eq. \ref{eq:silog_original}) was introduced in \cite{eigen2014depth} and has often been used in monocular depth estimation CNNs \cite{lee2019big, cs2018depthnet, carvalho2018regression} since it is an efficient function to determine the relationships between the depths of a scene without the influence of its scale \cite{eigen2014depth}. Such a loss function, modified in \cite{lee2019big}, can be expressed by Eq. \ref{silog}, in which $y^*_i$ represents the ground truth depth values, $y_i$ refers to the prediction of the network for the $i-th$ pixel and $N$ is equal to the total number of samples.

\begin{equation}
        \mathcal{L}_{SILog}(h) = \frac{1}{N}\sum_{i}^{}h_{i}^{2} - \frac{\lambda}{N^{2}}\left(\sum_{i}^{}h_{i}\right)^{2}, 
        \label{eq:silog_original}
\end{equation}
\begin{equation*}
        h_{i} = \log( y_{i}) - \log( y_{i}^{*}),\quad\lambda = 0.85
\end{equation*}

\begin{equation}
        \mathcal{L}_{SILog}^{+} = \theta\sqrt{\mathcal{S}(h)},
        \label{silog}
\end{equation}
\begin{equation*}
       \mathcal{S}(h)=\frac{1}{N}\sum_{i=1}^{N}h_{i}^{2}-\left(\frac{1}{N}\sum_{i=1}^{N}h_{i}\right)^{2}+(1-\lambda)\left(\frac{1}{N}\sum_{i=1}^{N}h_{i}\right)^{2},
\end{equation*}
\begin{equation*}
        h_i = \log( y_{i}) - \log( y_{i}^{*}),\quad\lambda = 0.85, \quad \theta=10.0
\end{equation*}

BerHu, the inverse function of $\mathcal{L}_{Huber}$ (Eq. \ref{huber}), is also applied in this work due to its advantages related to the combination of the losses $\mathcal{L}_1$ and $\mathcal {L}_2$. Among these benefits, residuals ($|y_i - y^*_i|$) with high values are penalized by the function $\mathcal{L}_2$ which is sensitive to outliers. On the other hand, residuals with low values are penalized by the $\mathcal{L}_1$ function, which returns greater depth values than those returned by the $\mathcal{L}_2$. The BerHu penalty can be expressed by Eq. \ref{berhu} with the default $\kappa=5$.

\begin{equation}
        \mathcal{L}_{Huber}(h)=
        \left\{
        \begin{array}{l}
            |h| \;\;\;\;\;\;\;\;\,|h|\geq g,\\
            \frac{h^{2}+g^{2}}{2g}
            \quad|h|<g,\\
        \end{array}
        \right.
        \label{huber}
\end{equation}
\begin{equation*}
        h=\log( y)-\log( y^{*}),\quad g=\frac{1}{\kappa}\max_{i}\left(|\log( y_{i})-\log( y_{i}^{*})|\right)
\end{equation*}

\begin{equation}
        \mathcal{L}_{BerHu}(h)=
        \left\{
        \begin{array}{l}
            |h| \;\;\;\;\;\;\;\;\,|h|\leq g,\\
            \frac{h^{2}+g^{2}}{2g}
            \quad|h|>g\\
        \end{array}
        \right.
        \label{berhu}
\end{equation}
\begin{equation*}
        h=\log( y)-\log( y^{*}),\quad g=\frac{1}{\kappa}\max_{i}\left(|\log( y_{i})-\log( y_{i}^{*})|\right)
\end{equation*}

In addition to the $\mathcal{L}_{SILog}$, $\mathcal{L}_{SILog}^{+}$, $\mathcal{L}_{BerHu}$ and $\mathcal{L}_{Huber}$ expressions, the $\mathcal{L}_{1}$ and $\mathcal{L}_{2}$ loss functions, represented by Eqs. \ref{l1} and \ref{l2}, respectively, are employed during training, as they are widely addressed by single-view depth estimation networks \cite{godard2017unsupervised,ma2018sparse,fu2018deep}. The Charbonnier penalty, described by Eq. \ref{charbo}, has been applied to problems involving regression, mainly in flow estimation approaches \cite{sun2014quantitative} and self-supervised SIDE frameworks \cite{babu2018undemon}. The parameters $a$ and $\alpha$ of $\mathcal{L}_{charbo}(h)$ are adjusted according to what is established by the work of Babu \etal \cite{babu2018undemon}.

\begin{equation}
        \mathcal{L}_{1}(h) = \frac{1}{N}\sum_{i=1}^{N}\left|h_{i}\right|, 
        \quad
        h_{i} = \log( y_{i}) - \log( y_{i}^{*})
        \label{l1}
\end{equation}

\begin{equation}
        \mathcal{L}_{2}(h) = \frac{1}{N}\sum_{i=1}^{N}\left(h_{i}\right)^{2}, 
        \quad
        h_{i} = \log( y_{i}) - \log( y_{i}^{*})
        \label{l2}
\end{equation}

\begin{equation}
        \mathcal{L}_{charbo}(h) = \frac{1}{N}\sum_{i=1}^{N}\left(h_{i}^{2}+\alpha^{2}\right)^{a},
        \label{charbo}
\end{equation}
\begin{equation*}
        h_{i} = \log( y_{i}) - \log( y_{i}^{*}), \quad a=0.45, \quad \alpha=10^{-3}
\end{equation*}

The Log-cosh loss (Eq. \ref {logcosh}) presents an inverse behavior to that of BerHu, operating as $\mathcal{L}_1$ for high residuals and as $\mathcal{L}_2$ for low residuals. This makes its behavior similar to the Huber function, yet the $\mathcal{L}_{cosh}(h)$ has the advantage of being totally differentiable.

\begin{equation}
        \mathcal{L}_{cosh}(h) = \sum_{i=1}^{N}\log\left(\cosh{\left(h_{i}\right)}\right),
        \quad
        h_{i} = y_{i} - y_{i}^{*}
        \label{logcosh}
\end{equation}

Based on the presented losses, we aim to use the one that produces the best results in the tests of the proposed network, replacing the term $\mathscr{L}(y, y^{*})$ of the attention-based function at distant depths, Eq. \ref{eqatten}, which is introduced in the work of Jiao \etal \cite{jiao2018look}. According to the authors, this function allows the network to focus on regions where the availability of pixels with depth information is reduced, once they are unevenly distributed.

The constant $\beta$, the attention term $\gamma$ in logarithmic scale and the function $\mathcal{L}_{a}$ in summation setup are the modifications proposed over the original expression \cite{jiao2018look}. In our approach, $\beta$ is set to $10.0$ and $1.0$ for the indoor and outdoor datasets, respectively, due to scale variations.

\begin{equation}
        \mathcal{L}_{a}=\sum_{i=1}^{N}\left(\gamma + \epsilon\right)\mathscr{L}(y,y^{*}),
        \label{eqatten}
\end{equation}
\begin{equation*}
        \gamma=\frac{\log(y_{i}^{*})}{\max_{i}(\log(y_{i}^{*}))},\quad \epsilon=1.0-\frac{\min(\log(\beta y_{i}),\log(\beta y_{i}^{*}))}{\max(\log(\beta y_{i}),\log(\beta y_{i}^{*}))}
\end{equation*}

Finally, the loss functions $\mathcal{L}_{1}^{+}$ and $\mathcal{L}_{2}^{+}$ are similar to the $\mathcal{L}_{1}$ and $\mathcal{L}_{2}$ expressions respectively. However, the former ones are computed only by the total sum of the equation results, instead of the average calculation. Based on experimental analysis, these variations of $\mathcal{L}_{1}$ and $\mathcal{L}_{2}$ are applied only when training the network with the indoor datasets.

\subsection{SIDE with Semantic Cues Pipeline}

The influence of additional semantic information on the proposed FCN is analyzed through fine-tunning. According to such a strategy, the DSN is firstly pre-trained for semantic segmentation and part of the computed weights are transferred to the monocular depth estimation framework. 

The loss function that is used for the semantic segmentation task can be represented by Eq. \ref{cceloss}, where $K$ is equal to the total number of categories, $y_i^*$ is the ground truth in the one-hot encoding format, $y_i$ is the network prediction, after the application of the softmax activation function, and $t$ represents the elements of the input feature map.

\begin{equation}
        \mathcal{L}_{cce}(y,y^{*}) = -\sum_{i=1}^{K}y_{i}^{*}\log(y_{i}),
        \quad
        y_i = \frac{e^{t_i}}{\sum_j^Ke^{t_j}}
        \label{cceloss}
\end{equation}

Only part of the weights obtained with the semantic segmentation framework is fine-tuned in the depth estimation phase due to the structural differences between the two CNNs, especially in the last layers.

\subsection{SIDE with Surface Normals Cues Pipeline}

Differently from previous works \cite{wang2016surge,qi2018geonet,zhang2019pattern} and closer to the approach of Yin \etal\cite{yin2019enforcing}, we leverage the correlation between depth and surface normals estimation in only one framework. Thus, we do not employ any extra branches or other supplementary models to exploit 3D geometric constraints from surface normals. Contrariwise, they are inferred from the output depth map and the same neural layers used for SIDE.     

To train our CNN for surface normals estimation, the ground truth is obtained as in the works of Fouhey \etal\cite{fouhey2013data}, Qi \etal\cite{qi2018geonet} and Zhang \etal\cite{zhang2019pattern}. Therefore, firstly, the point cloud in 3D space is recovered from the depth maps, assuming the pinhole camera model according to Eq. \ref{pinhole}. In such an equation, $(\upsilon_i, \nu_i)$ refers to the pixel positions in the image, $(x_i, y_i, z_i)$ represents the 3D coordinates of the point cloud, where the values of $z_i$ are equal to the depths of the scene $\rho_i$, $(o_x, o_y)$ is the location of the optical axis and $f_x$ and $f_y$ are the focal lengths of the axes $x$ and $y$ respectively.

\begin{equation}
        x_i = \frac{z_i\left(\upsilon_i-o_x\right)}{f_x}, \quad
        y_i = \frac{z_i\left(\nu_i-o_y\right)}{f_y}, \quad
        z_i = \rho_i
        \label{pinhole}
\end{equation}

As proposed by Qi \etal\cite{qi2018geonet}, after reconstructing the point cloud, we determine the tangent planes to the cloud depth data by defining a neighborhood of points of size $m$, which belongs to the same plan. Once computed the local planes, the surface normal vectors $s=\left(s_x, s_y, s_z\right)$ can be obtained through the relationship $Ds=c$, where $D_{m\times 3}$ is the coordinate matrix of the point cloud and $c_{m\times 1}$ is a vector with constant values. In order to minimize $||Ds-c||_{2}^{2}$, the least-squares solution $\hat{s}_{3\times m}$ is applied to the normal equations system of Eq. \ref{eq:MMQ}.

\begin{equation}
        D^{T} D \hat{s} = D^{T} c
        \label{eq:MMQ}
\end{equation}

From Eq. \ref{eq:MMQ}, we can get to Eq. \ref {eq:MMQ2} which is used to directly calculate the normalized surface normal vectors. Following what was established by Qi \etal\cite{qi2018geonet}, the vector $c_{m\times 1}$ is defined with unit values.

\begin{equation}
        \hat{s} = \frac{(D^{T} D)^{-1}D^{T} c}{||(D^{T} D)^{-1}D^{T} c||_2}
        \label{eq:MMQ2}
\end{equation}

The developed module shares the same feature maps produced in the SIDE step, yet they are concatenated with the predicted depth map and with the output from the Sobel edge detection algorithm \cite{nixon2019feature}. This algorithm is computationally efficient, it is robust to the noise present in the input grayscale image and it retrieves the high gradient features of the input which helps the proposed CNN to better reason about depth in the boundaries of the objects. Eqs. \ref{eqconv} and \ref{eqmagang} describe the $H_x$ and $H_y$ masks used by the Sobel filter to calculate the gradients approximation of the image $I$ (Eq. \ref{eqgrad}). With the achieved gradient vector, its magnitude $H$ and direction $\Psi$ are computed to determine the pixels in border regions, which follow the relationship $H(i, j)>\tau$ wherein $\tau$ is equal to the threshold value. 

\begin{equation}
I_x = 
\frac{\partial I}{\partial x}, \quad
I_y =
\frac{\partial I}{\partial y}
\label{eqgrad}
\end{equation}

\begin{equation}
I_x(i,j)=(I*H_x)(i,j), 
\quad
I_y(i,j)=(I*H_y)(i,j) 
\label{eqconv}
\end{equation}

\begin{equation}
\begin{split}
H_x = \begin{bmatrix}
-1 & 0 & 1\\
-2 & 0 & 2\\
-1 & 0 & 1
\end{bmatrix}, \quad
H_y = -H_x^T,
\\
H = \sqrt{I_x^2 + I_y^2}, \quad
\Psi = arctan\left(\frac{I_y}{I_x}\right)
\end{split}
\label{eqmagang}
\end{equation}

To validate the noise robustness of the surface normals module, a convolutional layer with Gaussian kernel \cite{afshari2017gaussian} is implemented before the Sobel edge detector. This convolutional layer is responsible for filtering the noise present in the input grayscale image $I$, making the output of the edge detection step less affected by them. The proposed linear Gaussian filter is free from second peaks in the frequency domain and can be expressed by Eq. \ref{gauss} wherein $I_g$ is the filtered image and $G$ refers to the $m\times n$ kernel that obeys a Gaussian distribution with $\sigma = 1.76$.

\begin{equation}
\fontsize{9pt}{10pt}\selectfont
I_g(i,j)=(I*G)(i,j)=\sum_{m}\sum_{n}G(m,n)I(i-m,j-n),
\label{gauss}
\end{equation}
\begin{equation*}
G(m,n) = \frac{1}{2\pi\sigma^2}e^{-\frac{m^2+n^2}{2\sigma^2}}
\end{equation*}

For the joint approach of depth and surface normals estimation, we designed the geometric loss function $\mathcal{L}_{2.5D}$, whose expression comprises the functions $\mathscr{L}(y,y^{*})$ and $\Phi(n,n^{*})$. The former loss is replaced by the one that generates the most accurate depth predictions. The latter loss computes the cosine similarity between the valid pixels from the surface normals predictions $n$ and the ground truth $n^*$. Also, $\Phi(n,n^{*})$ is weighted by a proportionality constant $\psi$.

\begin{equation}
        \mathcal{L}_{2.5D} = \mathscr{L}(y,y^{*}) + \psi(\Phi(n,n^{*})),
        \label{l25d}
\end{equation}
\begin{equation*}
        \Phi(n,n^{*}) = 1 - \left(\frac{\left<n,n^* \right>}{||n||_2||n^*||_2}\right)
\end{equation*}

\subsection{Depth Completion Pipeline}

This work also aims to analyze the behavior of the results produced by the network when it is trained and tested with additional depth data at the input, which can be obtained through low-cost 2D LiDARs and the outputs of visual odometry algorithms \cite{ma2018sparse}. 

We simulate visual odometry algorithms or depth sensors by a sampling technique which incorporates the multinomial distribution with the number of independent experiments $n=1$. Thus, only a random sampling attempt of a sparse depth data $\chi$ is performed for the ground truth pixels paired with the network input images. This special case of the multinomial distribution is called categorical distribution, in which the number of categories is fixed at $k = 2$. The Eq. \ref{prob1} regards the mass probability function associated with the random variable $\chi$.

\begin{equation}
        F(\chi~|~\varphi) = \prod_{i=0}^{k-1} \varphi_{i}^{\chi_{i}},
        \quad
        \varphi_{i} > 0, \quad \sum_{i=0}^{k-1} \varphi_{i} = 1
        \label{prob1}
\end{equation}

The probabilities $\varphi_{1}$ and $\varphi_{0}$ refer, respectively, to the success and failure of sampling a sparse depth data. Therefore, as proposed by Ma \etal \cite{ma2018sparse}, $ \varphi_{1} = \frac{d}{d ^ *} $ and $ \varphi_ {0} = 1 - \frac {d} {d ^ *} $ wherein $ d $ and $ d ^ * $ represent the desired amount of samples and the amount of valid data available per depth map respectively.


\section{Experiments}
\label{results}

\subsection{Implementation Details}

To implement and train the proposed network, a server with $4.20$GHz CPU, $2$TB HD, $200$GB SSD, Ubuntu $16.04$ operating system, Tensorflow $1.13.1$ development framework and two NVIDIA Titan X GPUs (both are used in the training phase and only one in the test phase). The Adam optimization algorithm \cite{kingma2014adam} is standard for all experiments, with $\alpha=10^{-3}$, $\beta_1=0.9$, $\beta_2=0.999$ and $\epsilon=10^{-8}$, as proposed by Kingma \etal \cite{kingma2014adam}. 

Furthermore, the learning rate is set to $10^{-4}$ with a polynomial decay at a rate of $0.9$. In each experiment, the CNN is trained for $50$ epochs with batch size $16$. However, the batch size is set to $32$ in the training steps that involve the MobileNetV2-50, the network of Fig. \ref{figgrasp} and the proposed Model $1$.

Although the feature maps generated by the first CNN convolution layers present primitive visual information, the parameters of the batch normalizations and the weights of the first two convolutions of the feature extraction network are not fixed, as occurs in the methods proposed by Fu \etal\cite{fu2018deep} and Lee \etal\cite{lee2019big}.

To avoid overfitting and artificially increase the number of training samples, three types of online data augmentation are employed. Among them, the random horizontal flip operations, with a probability of $50\%$, and rotation in the intervals of $\left[-2.5^{\circ}, 2.5^{\circ}\right]$ and $\left[-1.0^{\circ}, 1.0^{\circ}\right]$ for the indoor and outdoor datasets respectively. Brightness $[0.75, 1.25]$, color $[0.9, 1.1]$ and contrast $[0.9, 1.1]$ distortions are also considered with a probability of $50\%$.

\subsection{Datasets}

\subsubsection{Outdoor}
\label{kittid}

In previous works of the SIDE literature, the KITTI Raw \cite{geiger2013vision}, whose images are obtained directly from the sparse point clouds generated by a LiDAR sensor, is used as ground truth. However, recently, the KITTI Vision Benchmark Suite has released the official KITTI Depth dataset, which is composed of denser depth maps due to the combination of $11$ laser scans \cite{uhrig2017sparsity}. The KITTI Depth consists of stereo images in gray and color scales, which are captured by two pairs of stereo cameras that compose the autonomous navigation platform capable of traversing external environments to collect data.

Moreover, the dataset provides depth measurements generated by the scans of the Velodyne HDL-64E LiDAR in the form of point clouds. Another relevant aspect of the KITTI Depth is the denser ground truth, compared to the point clouds, which presents a grayscale image format, whose pixels provide the depth information of the left and right scenes from the stereo cameras.

Both the stereo images and the densified ground truth have a typical resolution of $375\times1242$ pixels and are divided into $61$ scenes for training and testing, which comprise the categories ``city", ``residential", ``road" and ``campus". At first, the KITTI Raw \cite{geiger2013vision} was subdivided into $32$ different training scenes and $29$ test scenes by Eigen \etal\cite{eigen2014depth}. The training subdivision contains $23488$ images and the testing set comprises $697$ images. Bringing the Eigen Split to the KITTI Depth \cite{uhrig2017sparsity}, we aim to use $23158$ images randomly selected from the total $23488$ training samples.

For the semantic segmentation task, the KITTI Vision Benchmark provides $200$ manually annotated maps for training and $200$ test samples with a typical resolution of $375\times1242$ pixels. Due to this, additional training data from the Cityscapes dataset  \cite{cordts2016cityscapes} are included. The Cityscapes consists of $5K$ semantic maps (with a resolution of $1024\times2048$ pixels), which are formed by fine annotations that sum up $30$ different classes. From the $5K$ images, $2975$, $1525$ and $500$ are reserved for training, testing and validation respectively. Therefore, in this work, the KITTI's $200$ semantic training maps are complemented by $3475$ Cityscapes samples.  

Corresponding to the semantic maps, the Cityscapes dataset also provides the same number of disparity maps, which can be used to reconstruct their respective depth maps. Due to the availability of disparity maps for training, testing and validation, $5K$ depth maps reconstructed from the Cityscapes are used to pre-train the proposed framework.

We also apply the KITTI Odometry dataset \cite{Geiger2012CVPR} in the VO studies with the proposed CNN. Such a dataset is composed of $22$ sequences with stereo images of outdoor scenarios. However, only the first $11$ sequences have their respective reference trajectories, while the remaining ones are used for benchmark evaluation. As proposed by Loo \etal\cite{loo2019cnn}, we focus on the stereo sequences with available ground truth.  

\subsubsection{Indoor}
\label{nyu}

The NYU Depth V2 \cite{silberman2012indoor} is composed of pairs of RGB images and depth maps, with a standard resolution of $480\times640$ pixels, captured by the camera and depth sensor of a Microsoft Kinect. Altogether, the official dataset is formed by $240K$ pairs of RGBD training samples that integrate $464$ indoor scenes. As other SIDE state-of-the-art works \cite{chen2019attention, yin2019enforcing, alhashim2018high}, we apply the official subdivision of the NYU Depth V2 data which comprises $249$ training scenes and $215$ test scenes. 

From the $249$ training scenes, totaling $120K$ samples, $24931$ images are selected to train the proposed CNN. On the other hand, from the $215$ test scenes, $654$ images are pre-defined to test our framework. Additionally, $8587$ samples, from the SUNRGBD \cite{song2015sun}, Berkeley B3DO \cite{janoch2013category} and SUN3D \cite{xiao2013sun3d} datasets are employed in pre-training phases. 

To address the semantic segmentation task, the NYU Depth V2 provides $1449$ images, with $35064$ different objects, that present semantic annotations with $894$ different categories. We also add $10336$ samples, with their corresponding semantic labels, from the SUNRGBD dataset \cite{song2015sun} to the $1449$ available training images.  

The surface normals ground truth, for both indoor and outdoor datasets, is produced according to what is proposed in the work of Fouhey \etal\cite{fouhey2013data}.

\subsection{Evaluation Metrics}

The proposed FCN is tested outdoors according to the Eigen Split. However, as this subdivision is originally proposed for the KITTI Raw dataset \cite{geiger2013vision}, only $652$ of the $697$ test images present ground truth in the KITTI Depth \cite{uhrig2017sparsity}. Furthermore, the network predictions are evaluated considering the central crop established by Garg \etal\cite{garg2016unsupervised} and distance caps of $0-50m$ and $0-80m$. For the evaluation of the framework in indoor environments, the central crop proposed in \cite{eigen2014depth} is used in the predictions that correspond to the 654 test images from the NYU Depth V2.

Therefore, in order to present quantitative results, the Scale-invariant Error (Eq. \ref{eei}), Threshold (Eq. \ref{limite}), Absolute Relative Difference (Eq. \ref{dra}), Squared Relative Difference (Eq. \ref{drq}), Log10 (Eq. \ref{log10}), Mean Absolute Error (Eq. \ref{ema}), Linear RMSE (Eq. \ref{rmselin}) and Log RMSE (Eq. \Ref{rmselog}), are applied in the evaluation phase wherein $T$ is equivalent to the total amount of pixels with depth information:

\begin{equation}
        \textit{$\frac{1}{|T|}\sum\nolimits_{y \in T}\left(y-y^{*}\right)^2-\frac{1}{2T^2}\left(\sum\nolimits_{y \in T}\left(y-y^{*}\right)\right)^2$}
        \label{eei}
\end{equation}
\begin{equation}
        \textit{$\%$ of $y_{i}$ s.t. $\max\left(\frac{y_{i}}{y_{i}^{*}}, \frac{y_{i}^{*}}{y_{i}}\right)$ = $\delta$ $<$ $threshold$}
        \label{limite}
\end{equation}
\begin{equation}
        \textit{$\frac{1}{|T|}\sum\nolimits_{y \in T}\frac{|y-y^{*}|}{y^{*}}$}
        \label{dra}
\end{equation}
\begin{equation}
        \textit{$\frac{1}{|T|}\sum\nolimits_{y \in T}\frac{||y-y^{*}||^{2}}{y^{*}}$}
        \label{drq}
\end{equation}
\begin{equation}
        \textit{$\frac{1}{|T|}\sum\nolimits_{y \in T}|\log_{10} y-\log_{10} y^{*}|$}
        \label{log10}
\end{equation}
\begin{equation}
        \textit{$\frac{1}{|T|}\sum\nolimits_{y \in T}|y-y^{*}|$}
        \label{ema}
\end{equation}
\begin{equation}
        \textit{$\sqrt{\frac{1}{|T|}\sum\nolimits_{y \in T}||y-y^{*}||^{2}}$}
        \label{rmselin}
\end{equation}
\begin{equation}
        \textit{$\sqrt{\frac{1}{|T|}\sum\nolimits_{y \in T}||\log y-\log y^{*}||^{2}}$}
        \label{rmselog}
\end{equation}

The surface normals maps retrieved by our pipeline are evaluated according to the metrics introduced by Fouhey \etal\cite{fouhey2013data}, which are commonly used by recent works \cite{bansal2016marr,wang2016surge,qi2018geonet,zhang2019pattern}. Such metrics consider the angular difference for each valid pixel between the estimated surface normals and the ground truth by calculating the mean, median, root mean squared error (RMSE) and the precision of the estimated pixels. This accuracy is measured by the percentage of vectors whose angular difference does not exceed a threshold $\widetilde{n}=\{11.25^{\circ}, 22.5^{\circ}, 30^{\circ}\}$.

\subsection{Outdoor Ablation Studies}

\subsubsection{Preliminary Analysis}

Firstly, regarding depth completion approaches to enhance the depth estimation results, we used the closing morphological operation to partially densify the reference maps of the KITTI Depth \cite{uhrig2017sparsity}, generating the KITTI Morphological dataset. Along with the closing technique, we applied the RGB\textunderscore guide\&certainty method, introduced by Gansbeke \etal\cite{van2019sparse}, and the depth completion strategy of Hilbert Maps \cite{dos2019sparse} to produce the KITTI Completed and the KITTI Continuous respectively. 

Through the KITTI Benchmark metrics, the depth completion results from the application of $5$ iterations of the closing morphological technique are compared to other methods in the literature in Table \ref{tab:densification}. In such a table, it is possible to note that the closing operator, with kernel $3\times 3$, is the method that best preserves the ground truth depth information. However, the learning-based techniques retrieve totally dense depth maps, whereas the closing ones can be tuned to output images with different densification levels.

\begin{table}[htb!]
\centering
\caption{Comparison of state-of-the-art depth completion methods according to the KITTI Benchmark \cite{uhrig2017sparsity} metrics ($iMAE=\frac{1}{MAE}$ and $iRMSE=\frac{1}{RMSE}$)}
\label{tab:densification}
\resizebox{1.0\linewidth}{!}{
\begin{tabular}{c|c|c|c|c}
\hline
\textbf{Method} & \textbf{MAE}$\downarrow$ & \textbf{RMSE}$\downarrow$ & \textbf{iMAE}$\downarrow$ & \textbf{iRMSE}$\downarrow$\\\hline\hline
SGDU \cite{schneider2016semantically} & 605.470 & 2312.570 & 2.050 & 7.380\\
NN+CNN \cite{uhrig2017sparsity} & 416.140 & 1419.750 & 1.290 & 3.250 \\
ADNN \cite{chodosh2018deep} & 439.480 & 1325.370 & 3.190 & 59.390 \\
IP-Basic \cite{ku2018defense} & 302.600 & 1288.460 & 1.290 & 3.780 \\
DFuseNet \cite{shivakumar2019dfusenet} & 429.930 & 1206.660 & 1.790 & 3.620\\
VOICED \cite{wong2020unsupervised} & 299.410 & 1169.970 & 1.200 & 3.560 \\
Morph-Net \cite{dimitrievski2018learning} & 310.490 & 1045.450 & 1.570 & 3.840 \\
CSPN \cite{cheng2018depth} & 279.460 & 1019.640 & 1.150 & 2.930 \\
DCrgb\textunderscore 80b\textunderscore 3coef \cite{imran2019depth} & 215.750 & 965.870 & 0.980 & 2.430 \\
Conf-Net \cite{hekmatian2019conf} & 257.540 & 962.280 & 1.090 & 3.100 \\
DFineNet \cite{zhang2019dfinenet} & 304.170 & 943.890 & 1.390 & 3.210 \\
Spade-RGBsD \cite{jaritz2018sparse} & 234.810 & 917.640 & 0.950 & 2.170 \\
IR\textunderscore L2 \cite{lu2020depth} & 292.360 & 901.430 & 1.350 & 4.920 \\
DDP \cite{yang2019dense} & 203.960 & 832.940 & 0.850 & 2.100 \\
NConv \cite{eldesokey2019confidence} & 233.260 & 829.980 & 1.030 & 2.600 \\
Sparse-to-Dense \cite{ma2019self} & 249.950 & 814.730 & 1.210 & 2.800 \\
CrossGuidance \cite{lee2020deep} & 253.980 & 807.420 & 1.330 & 2.730\\
Revisiting NN+CNN \cite{yan2020revisiting} & 225.810 & 792.800 & 0.990 & 2.420\\
PwP \cite{xu2019depth} & 235.170 & 777.050 & 1.130 & 2.420 \\
RGB\textunderscore guide\&certainty \cite{van2019sparse} & 215.020 & 772.870 & 0.930 & 2.190 \\
DSPN \cite{xu2020deformable} & 220.360 & 766.740 & 1.030 & 2.470\\
MSG-CHN \cite{li2020multi} & 220.410 & 762.190 & 0.980 & 2.300 \\
DeepLiDAR \cite{qiu2019deeplidar} & 226.500 & 758.380 & 1.150 & 2.560 \\
UberATG-FuseNet \cite{chen2019learning} & 221.190 & 752.880 & 1.140 & 2.340 \\
CSPN++ \cite{cheng2019cspn++} & 209.280 & 743.690 & 0.900 & 2.070 \\
NLSPN \cite{park2020non} & 199.590 & 741.680 & 0.840 & 1.990\\
GuideNet \cite{tang2019learning} & 218.830 & 736.240 & 0.990 & 2.250 \\
Closing ($k=3$, $it=5$) & 148.963 & 519.606 & 0.597 & 1.082 \\
Closing ($k=3$, $it=4$) & 129.353 & 457.346 & 0.533 & 0.972 \\
Closing ($k=3$, $it=3$) & 106.925 & 391.138 & 0.458 & 0.867 \\
Closing ($k=3$, $it=2$) & 82.137 & 326.323 & 0.363 & 0.747 \\
Closing ($k=3$, $it=1$) & \textbf{46.724} & \textbf{226.214} & \textbf{0.211} & \textbf{0.541} \\
\hline
\end{tabular}
} 
\end{table}

In Table \ref{tab:pixels}, we may infer that the bigger the number of iterations ($it$) the denser the KITTI Morphological dataset becomes. Moreover, the proposed KITTI Completed is the densest dataset employed once it is built with the learning-based RGB\textunderscore guide\&certainty approach. The results exposed in Table \ref{tab:tests} show that our pipeline achieves slightly better performance when jointly trained with the partially dense dataset and the BerHu loss function. 

However, the accuracy of the method decreases when it is trained with the KITTI Completed and the KITTI Continuous datasets due to the presence of greater distortions in the reference depth maps. Furthermore, although the KITTI Continuous is less dense than the KITTI Completed, the former dataset includes a higher amount of artifacts that hinder the network learning. Therefore, the KITTI Morphological presents the best balance between densification and modification of the original ground truth values, so that our CNN can better understand the structure of the scene.

\begin{table}[htb!]
\centering
\caption{Analysis of the densification degrees of the datasets used according to the average number and average percentage of valid pixels. $375\times1242$ is the typical resolution for the KITTI Depth \cite{uhrig2017sparsity} and $352\times704$ is the image size that is fed into our framework}
\label{tab:pixels}
\resizebox{1.0\linewidth}{!}{
\begin{tabular}{c|cc|cc}
\hline
\multicolumn{1}{c}{\multirow{2}{*}{\textbf{Dataset}}} \vline & \multicolumn{2}{c}{\begin{tabular}[c]{@{}c@{}}\textbf{Average number}\\\textbf{of valid pixels}\end{tabular}} \vline & \multicolumn{2}{c}{\begin{tabular}[c]{@{}c@{}}\textbf{Average percentage}\\\textbf{of valid pixels}\end{tabular}} \\
\multicolumn{1}{c}{} \vline & $375\times1242$ & \multicolumn{1}{c}{$352\times704$} \vline & $375\times1242$ & \multicolumn{1}{c}{$352\times704$} \\ \hline \hline
KITTI Depth & 75407 & 43649 & 16.190\% & 17.614\% \\
KITTI Morphological ($k=3$, $it=1$) & 142309 & 82377 & 30.555\% & 33.242\% \\
KITTI Morphological ($k=3$, $it=2$) & 166765 & 96525 & 35.806\% & 38.951\% \\
KITTI Morphological ($k=3$, $it=3$) & 182949 & 105884 & 39.280\% & 42.728\% \\
KITTI Morphological ($k=3$, $it=4$) & 195587 & 113172 & 41.994\% & 45.669\% \\
KITTI Morphological ($k=3$, $it=5$) & 205478 & 118875 & 44.118\% & 47.971\% \\
KITTI Continuous & 291489 & 158255 & 62.585\% & 63.862\% \\
KITTI Completed & 428031 & 247807 & 91.901\% & 100\% 
\\\hline
\end{tabular}
} 
\end{table}


\begin{table*}[]
\caption{Evaluation of the proposed method considering variations in the loss function and the dataset used during training. The DenseNet-121 is fixed as the feature extraction network of the DSN}
\label{tab:tests}
\resizebox{\textwidth}{!}{
\begin{tabular}{c|c|c|c|cccccc|ccc}
\hline
\textbf{Method} & \textbf{Loss} & \textbf{Dataset} & \textbf{Cap} & \textbf{Abs Rel}$\downarrow$ & \textbf{Sqr Rel}$\downarrow$ & \textbf{RMSE}$\downarrow$ & \textbf{RMSE (log)}$\downarrow$ & \textbf{SILog}$\downarrow$ & \textbf{log10}$\downarrow$ & $\bm{\delta < 1.25}$$\uparrow$ & $\bm{\delta < 1.25^2}$$\uparrow$ & $\bm{\delta < 1.25^3}$$\uparrow$\\\hline \hline
DenseNet-121 & $\mathcal{L}_2$ & KITTI Depth & $0 - 80\ m$ & 0.114 & 0.663 & 4.367 & 0.173 & 15.890 & 0.050 & 0.853 & 0.963 & 0.989 \\
DenseNet-121 & $\mathcal{L}_1$ & KITTI Depth & $0 - 80\ m$ & 0.108 & 0.624 & 4.200 & 0.166 & 15.294 & 0.047 & 0.868 & 0.965 & 0.990 \\
DenseNet-121 & $\mathcal{L}_{charbo}$ & KITTI Depth & $0 - 80\ m$ & 0.108 & 0.639 & 4.255 & 0.167 & 15.323 & 0.047 & 0.866 & 0.966 & 0.989 \\
DenseNet-121 & $\mathcal{L}_{Huber}$ & KITTI Depth & $0 - 80\ m$ & 0.099 & 0.538 & 4.076 & 0.154 & 14.308 & 0.044 & 0.883 & 0.972 & 0.992 \\
DenseNet-121 & $\mathcal{L}_{SILog}$ & KITTI Depth & $0 - 80\ m$ & 0.097 & 0.513 & 3.939 & 0.150 & 13.569 & 0.043 & 0.890 & 0.975 & 0.993 \\
DenseNet-121 & $\mathcal{L}_{cosh}$ & KITTI Depth & $0 - 80\ m$ & 0.096 & 0.518 & 3.855 & 0.150 & 13.800 & 0.042 & 0.892 & 0.974 & 0.993 \\
DenseNet-121 & $\mathcal{L}_{SILog}^{+}$ & KITTI Depth & $0 - 80\ m$ & 0.095 & 0.494 & 3.856 & 0.146 & 13.214 & 0.042 & 0.894 & 0.976 & \textbf{0.994} \\
DenseNet-121 & $\mathcal{L}_{BerHu}$  & KITTI Depth & $0 - 80\ m$ & 0.089 & 0.478 & 3.788 & 0.142 & 13.004 & \textbf{0.040} & 0.903 & 0.977 & \textbf{0.994} \\
DenseNet-121 & $\mathcal{L}_{BerHu}$  & KITTI Morphological ($k=3$, $it=1$) & $0 - 80\ m$ & \textbf{0.088} & 0.471 & 3.792 & 0.141 & 12.750 & \textbf{0.040} & 0.903 & \textbf{0.979} & \textbf{0.994} \\
DenseNet-121 & $\mathcal{L}_{BerHu}$  & KITTI Morphological ($k=3$, $it=2$) & $0 - 80\ m$ & \textbf{0.088} & 0.461 & \textbf{3.780} & \textbf{0.140} & \textbf{12.747} & \textbf{0.040} & \textbf{0.905} & 0.978 & \textbf{0.994} \\
DenseNet-121 & $\mathcal{L}_{BerHu}$  & KITTI Morphological ($k=3$, $it=3$) & $0 - 80\ m$ & \textbf{0.088} & \textbf{0.458} & 3.800 & 0.142 & 12.860 & \textbf{0.040} & 0.903 & 0.977 & \textbf{0.994} \\
DenseNet-121 & $\mathcal{L}_{BerHu}$  & KITTI Morphological ($k=3$, $it=4$) & $0 - 80\ m$ & \textbf{0.088} & 0.479 & 3.855 & 0.143 & 13.024 & \textbf{0.040} & 0.904 & 0.976 & 0.993 \\
DenseNet-121 & $\mathcal{L}_{BerHu}$  & KITTI Morphological ($k=3$, $it=5$) & $0 - 80\ m$ & \textbf{0.088} & 0.474 & 3.881 & 0.143 & 13.053 & \textbf{0.040} & 0.903 & 0.977 & \textbf{0.994} \\
DenseNet-121 & $\mathcal{L}_{BerHu}$  & KITTI Continuous & $0 - 80\ m$ & 1.640 & 36.498 & 24.577 & 0.978 & 21.627 & 0.415 & 0.013 & 0.036 & 0.072 \\
DenseNet-121 & $\mathcal{L}_{BerHu}$  & KITTI Completed & $0 - 80\ m$ & 0.220 & 1.741 & 6.269 & 0.249 & 20.577 & 0.084 & 0.698 & 0.924 & 0.980 
\\\hline\hline
DenseNet-121 & $\mathcal{L}_2$ & KITTI Depth & $0 - 50\ m$ & 0.110 & 0.509 & 3.203 & 0.161 & 14.807 & 0.047 & 0.867 & 0.970 & 0.992 \\
DenseNet-121 & $\mathcal{L}_1$ & KITTI Depth & $0 - 50\ m$ & 0.104 & 0.480 & 3.102 & 0.155 & 14.293 & 0.045 & 0.881 & 0.970 & 0.992 \\
DenseNet-121 & $\mathcal{L}_{charbo}$ & KITTI Depth & $0 - 50\ m$ & 0.103 & 0.487 & 3.122 & 0.156 & 14.277 & 0.045 & 0.879 & 0.972 & 0.991 \\
DenseNet-121 & $\mathcal{L}_{Huber}$ & KITTI Depth & $0 - 50\ m$ & 0.094 & 0.393 & 2.892 & 0.142 & 13.165 & 0.041 & 0.897 & 0.978 & 0.994 \\
DenseNet-121 & $\mathcal{L}_{SILog}$ & KITTI Depth & $0 - 50\ m$ & 0.093 & 0.377 & 2.833 & 0.139 & 12.542 & 0.041 & 0.904 & 0.980 & 0.995 \\
DenseNet-121 & $\mathcal{L}_{cosh}$ & KITTI Depth & $0 - 50\ m$ & 0.092 & 0.390 & 2.817 & 0.140 & 12.839 & 0.040 & 0.904 & 0.979 & 0.994 \\
DenseNet-121 & $\mathcal{L}_{SILog}^{+}$ & KITTI Depth & $0 - 50\ m$ & 0.090 & 0.365 & 2.771 & 0.136 & 12.229 & 0.040 & 0.907 & 0.981 & \textbf{0.996} \\
DenseNet-121 & $\mathcal{L}_{BerHu}$  & KITTI Depth & $0 - 50\ m$ & 0.085 & 0.352 & 2.721 & 0.131 & 11.981 & \textbf{0.037} & 0.916 & 0.982 & 0.995 \\
DenseNet-121 & $\mathcal{L}_{BerHu}$  & KITTI Morphological ($k=3$, $it=1$) & $0 - 50\ m$ & 0.084 & 0.342 & 2.732 & 0.130 & 11.723 & \textbf{0.037} & 0.916 & \textbf{0.983} & \textbf{0.996} \\
DenseNet-121 & $\mathcal{L}_{BerHu}$  & KITTI Morphological ($k=3$, $it=2$) & $0 - 50\ m$ & \textbf{0.083} & 0.335 & 2.701 & \textbf{0.129} & \textbf{11.722} & \textbf{0.037} & \textbf{0.919} & \textbf{0.983} & \textbf{0.996} \\
DenseNet-121 & $\mathcal{L}_{BerHu}$  & KITTI Morphological ($k=3$, $it=3$) & $0 - 50\ m$ & \textbf{0.083} & \textbf{0.331} & \textbf{2.688} & 0.130 & 11.784 & \textbf{0.037} & 0.916 & 0.982 & 0.995 \\
DenseNet-121 & $\mathcal{L}_{BerHu}$  & KITTI Morphological ($k=3$, $it=4$) & $0 - 50\ m$ & \textbf{0.083} & 0.345 & 2.729 & 0.131 & 11.906 & \textbf{0.037} & 0.917 & 0.982 & 0.995 \\
DenseNet-121 & $\mathcal{L}_{BerHu}$  & KITTI Morphological ($k=3$, $it=5$) & $0 - 50\ m$ & \textbf{0.083} & 0.338 & 2.744 & 0.131 & 11.912 & \textbf{0.037} & 0.916 & \textbf{0.983} & 0.995 \\
DenseNet-121 & $\mathcal{L}_{BerHu}$  & KITTI Continuous & $0 - 50\ m$ & 1.545 & 28.857 & 19.418 & 0.939 & 23.091 & 0.395 & 0.029 & 0.069 & 0.128 \\
DenseNet-121 & $\mathcal{L}_{BerHu}$  & KITTI Completed & $0 - 50\ m$ & 0.215 & 1.454 & 5.163 & 0.240 & 19.124 & 0.082 & 0.709 & 0.933 & 0.983 
\\\hline
\end{tabular}
} 
\end{table*}

\subsubsection{Structural Studies}

The following studies are conducted with the KITTI Morphological, with setup $k=3$ and $it=2$ since it is beneficial to the training of our CNN. To explore the response of the proposed pipeline to other encoder networks, we carried out new experiments whose results are presented in Table \ref{tab:nets}. From the metrics of Table \ref{tab:nets}, we can observe that the DSN with Model $3$ as encoder achieves the best trade-off regarding the number of trainable parameters, prediction speed and accuracy. Moreover, analyzing the methods with comparable sizes, it is possible to state that all the proposed lightweight feature extraction networks (Fig. \ref{figgrasp}, Model $1$, Model $2$ and Model $3$) surpass the state-of-the-art MobileNetV2 and DenseNet-121.

\begin{table*}[]
\centering
\caption{Comparison of the results obtained by the proposed framework with different feature extraction networks in the KITTI Depth dataset \cite{uhrig2017sparsity}. Model $1$, Model $2$ and Model $3$ refer to the DenseSIDENet with the encoder models of Table \ref{tab:netarch}. The MobileNetV2-50 and MobileNetV2-140 are versions of the MobileNetV2 with depth multipliers $1.4$ and $0.5$ respectively. The BerHu loss function is used in all experiments and the Features column indicates the number of input feature maps to the decoder stage of the DSN}
\label{tab:nets}
\resizebox{\textwidth}{!}{
\begin{tabular}{c|c|c|c|c|cccccc|ccc}
\hline
\textbf{Method} & \textbf{Features} & \textbf{Size} & \textbf{Speed} & \textbf{Cap} & \textbf{Abs Rel}$\downarrow$ & \textbf{Sqr Rel}$\downarrow$ & \textbf{RMSE}$\downarrow$ & \textbf{RMSE (log)}$\downarrow$ & \textbf{SILog}$\downarrow$ & \textbf{log10}$\downarrow$ & $\bm{\delta < 1.25}$$\uparrow$ & $\bm{\delta < 1.25^2}$$\uparrow$ & $\bm{\delta < 1.25^3}$$\uparrow$\\\hline\hline
Fig. \ref{figgrasp} CNN & 256 & 3\ M & $\boldsymbol{88}\ \boldsymbol{fps}$ & $0 - 80\ m$ & 0.102 & 0.594 & 4.094 & 0.160 & 14.797 & 0.045 & 0.876 & 0.968 & 0.991 \\
Model 1 & 256 & 3\ M & $59\ fps$ & $0 - 80\ m$ & 0.093 & 0.517 & 3.941 & 0.149 & 13.776 & 0.041 & 0.895 & 0.974 & 0.992 \\
Model 2 & 256 & 6\ M & $57\ fps$ & $0 - 80\ m$ & 0.091 & 0.501 & 3.851 & 0.146 & 13.243 & 0.041 & 0.901 & 0.977 & 0.993 \\
Model 3 & 256 & 12\ M & $50\ fps$ & $0 - 80\ m$ & \textbf{0.083} & \textbf{0.434} & \textbf{3.646} & \textbf{0.135} & \textbf{12.326} & \textbf{0.037} & \textbf{0.914} & \textbf{0.980} & \textbf{0.994} \\
MobileNetV2-50 & 128 & \textbf{2\ M} & $87\ fps$ & $0 - 80\ m$ & 0.122 & 0.760 & 4.612 & 0.184 & 17.074 & 0.052 & 0.842 & 0.955 & 0.986 \\
MobileNetV2-50 & 256 & 5\ M & $81\ fps$ & $0 - 80\ m$ & 0.116 & 0.729 & 4.508 & 0.178 & 16.326 & 0.051 & 0.853 & 0.959 & 0.987 \\
MobileNetV2-140 & 256 & 10\ M & $62\ fps$ & $0 - 80\ m$ & 0.104 & 0.596 & 4.084 & 0.161 & 14.799 & 0.045 & 0.877 & 0.969 & 0.991 \\
MobileNetV2-140 & 512 & 20\ M & $39\ fps$ & $0 - 80\ m$ & 0.100 & 0.575 & 4.065 & 0.158 & 14.206 & 0.044 & 0.883 & 0.970 & 0.991 \\
DenseNet-121 & 256 & 12\ M & $41\ fps$ & $0 - 80\ m$ & 0.088 & 0.461 & 3.780 & 0.140 & 12.747 & 0.040 & 0.905 & 0.978 & \textbf{0.994} \\\hline\hline
Fig. \ref{figgrasp} CNN & 256 & 3\ M & $\boldsymbol{88}\ \boldsymbol{fps}$ & $0 - 50\ m$ & 0.098 & 0.456 & 3.030 & 0.150 & 13.796 & 0.043 & 0.888 & 0.974 & 0.993 \\
Model 1 & 256 & 3\ M & $59\ fps$ & $0 - 50\ m$ & 0.089 & 0.380 & 2.843 & 0.138 & 12.712 & 0.039 & 0.907 & 0.980 & 0.994 \\
Model 2 & 256 & 6\ M & $57\ fps$ & $0 - 50\ m$ & 0.086 & 0.370 & 2.789 & 0.134 & 12.186 & 0.038 & 0.914 & 0.981 & 0.994 \\
Model 3 & 256 & 12\ M & $50\ fps$ & $0 - 50\ m$ & \textbf{0.079} & \textbf{0.314} & \textbf{2.592} & \textbf{0.124} & \textbf{11.315} & \textbf{0.035} & \textbf{0.926} & \textbf{0.984} & \textbf{0.996}\\
MobileNetV2-50 & 128 &\textbf{2\ M} & $87\ fps$ & $0 - 50\ m$ & 0.117 & 0.592 & 3.431 & 0.172 & 15.892 & 0.049 & 0.856 & 0.961 & 0.989 \\
MobileNetV2-50 & 256 & 5\ M & $81\ fps$ & $0 - 50\ m$ & 0.111 & 0.560 & 3.339 & 0.166 & 15.139 & 0.048 & 0.866 & 0.965 & 0.989 \\
MobileNetV2-140 & 256 & 10\ M & $62\ fps$ & $0 - 50\ m$ & 0.100 & 0.463 & 3.044 & 0.151 & 13.810 & 0.043 & 0.889 & 0.974 & 0.993 \\
MobileNetV2-140 & 512 & 20\ M & $39\ fps$ & $0 - 50\ m$ & 0.095 & 0.434 & 2.973 & 0.146 & 13.172 & 0.042 & 0.896 & 0.975 & 0.993 \\
DenseNet-121 & 256 & 12\ M & $41\ fps$ & $0 - 50\ m$ & 0.083 & 0.335 & 2.701 & 0.129 & 11.722 & 0.037 & 0.919 & 0.983 & \textbf{0.996} \\\hline
\end{tabular}
}
\end{table*}

To improve the performance of the DSN with Model $3$ as its feature extraction network (DSN $3$), we impose modifications in the structure of the CNN decoder and in its training procedure. Through Table \ref{tab:kittinets}, we may verify that the use of the attention loss, combined with the $\mathcal{L}_{BerHu}$ expression, decreases the framework accuracy, whereas the application of the $\mathcal{L}_{BerHu}^{+}$ function enhances the results. This indicates that the BerHu loss function is already capable of balancing the penalties for residuals generated in regions closer and more distant to the scene, without the need for an attention term. Furthermore, the network learning is boosted when the term $\mathcal{L}_1$ of the BerHu penalty acts on a wider range of residuals, considering the KITTI Morphological dataset.

The pyramid modules are simple blocks that concatenate multi-scale feature maps of the DSN $3$. In the decoder, the Pyramid$^1$ module concatenates the upsampled feature maps from the last $3$ convolution blocks with the output from the final upconvolution, while the Pyramid$^2$ adds the feature maps from the first convolution to this concatenation. Regarding the metrics presented in Table \ref{tab:kittinets}, both the pyramid modules are beneficial to the proposed framework, yet they slow down the network's processing speed.

Also, the results obtained when the DSN $3$ is pre-trained with semantic and disparity maps show that both pre-trainings improve the metrics values. Although our method acquires important cues, like the object edges, from the semantic maps with $19$ different categories, the monocular features learned from the same depth estimation task show better improvements. All the structural changes in the decoder of the proposed pipeline reduce its inference speed, yet the number of trainable parameters is not significantly increased.      

\begin{table*}[]
\centering
\caption{Behavior of the DSN when subjected to changes in its structure and training. The DSN $3$ is the baseline CNN with the Model $3$ as encoder. The Pyramid$^1$ and Pyramid$^2$ are pyramid-shaped modules implemented in the decoder. CS and Sem refer to pre-training with the disparities of the Cityscapes dataset \cite{cordts2016cityscapes} and the semantic maps of the KITTI \cite{Alhaija2018IJCV} and Cityscapes \cite{cordts2016cityscapes} respectively. The $\mathcal{L}_{BerHu}^{+}$ is equal to the $\mathcal{L}_{BerHu}$ expression, yet with $\kappa=4$}
\label{tab:kittinets}
\resizebox{\textwidth}{!}{
\begin{tabular}{c|c|c|c|c|cccccc|ccc}
\hline
\textbf{Method} & \textbf{Loss} & \textbf{Size} & \textbf{Speed} & \textbf{Cap} & \textbf{Abs Rel}$\downarrow$ & \textbf{Sqr Rel}$\downarrow$ & \textbf{RMSE}$\downarrow$ & \textbf{RMSE (log)}$\downarrow$ & \textbf{SILog}$\downarrow$ & \textbf{log10}$\downarrow$ & $\bm{\delta < 1.25}$$\uparrow$ & $\bm{\delta < 1.25^2}$$\uparrow$ & $\bm{\delta < 1.25^3}$$\uparrow$\\\hline\hline
DSN 3 & $\mathcal{L}_{BerHu}$ & \textbf{12\ M} & $\boldsymbol{50}\ \boldsymbol{fps}$ & $0 - 80\ m$ & 0.083 & 0.434 & 3.646 & 0.135 & 12.326 & 0.037 & 0.914 & 0.980 & 0.994 \\
DSN 3 & $\mathcal{L}_a$ $+$ $\mathcal{L}_{BerHu}$ & \textbf{12\ M} & $\boldsymbol{50}\ \boldsymbol{fps}$ & $0 - 80\ m$ & 0.084 & 0.437 & 3.621 & 0.133 & 12.259 & 0.037 & 0.914 & 0.981 & 0.994\\
DSN 3 & $\mathcal{L}_{BerHu}^{+}$ & \textbf{12\ M} & $\boldsymbol{50}\ \boldsymbol{fps}$ & $0 - 80\ m$ & 0.083 & 0.431 & 3.628 & 0.133 & 12.146 & 0.037 & 0.915 & 0.982 & 0.995\\ 
DSN 3 $+$ Pyramid$^1$ & $\mathcal{L}_{BerHu}^{+}$ & \textbf{12\ M} & $39\ fps$ & $0 - 80\ m$ & 0.082 & 0.418 & 3.593 & 0.132 & 12.122 & 0.037 & 0.919 & 0.981 & 0.995 \\
DSN 3 $+$ Pyramid$^1$ $+$ Sem & $\mathcal{L}_{BerHu}^{+}$ & \textbf{12\ M} & $39\ fps$ & $0 - 80\ m$ & 0.081 & 0.414 & 3.557 & 0.133 & 12.287 & 0.036 & 0.916 & 0.980 & 0.995 \\
DSN 3 $+$ Pyramid$^1$ $+$ CS & $\mathcal{L}_{BerHu}^{+}$ & \textbf{12\ M} & $39\ fps$ & $0 - 80\ m$ & \textbf{0.078} & 0.405 & 3.399 & 0.125 & 11.490 & \textbf{0.034} & \textbf{0.926} & \textbf{0.985} & 0.995\\
DSN 3 $+$ Pyramid$^2$ $+$ CS & $\mathcal{L}_{BerHu}^{+}$ & \textbf{12\ M} & $32\ fps$ & $0 - 80\ m$ & \textbf{0.078} & \textbf{0.376} & \textbf{3.344} & \textbf{0.124} & \textbf{11.323} & \textbf{0.034} & \textbf{0.926} & \textbf{0.985} & \textbf{0.996}\\\hline \hline
DSN 3 & $\mathcal{L}_{BerHu}$ & \textbf{12\ M} & $\boldsymbol{50}\ \boldsymbol{fps}$ & $0 - 50\ m$ & 0.079 & 0.314 & 2.592 & 0.124 & 11.315 & 0.035 & 0.926 & 0.984 & 0.996 \\
DSN 3 & $\mathcal{L}_{BerHu}$ $+$ $\mathcal{L}_a$ & \textbf{12\ M} & $\boldsymbol{50}\ \boldsymbol{fps}$ & $0 - 50\ m$ & 0.080 & 0.319 & 2.594 & 0.123 & 11.267 & 0.035 & 0.927 & 0.985 & 0.996\\
DSN 3 & $\mathcal{L}_{BerHu}^{+}$ & \textbf{12\ M} & $\boldsymbol{50}\ \boldsymbol{fps}$ & $0 - 50\ m$ & 0.078 & 0.312 & 2.596 & 0.122 & 11.143 & 0.035 & 0.927 & 0.986 & 0.996 \\
DSN 3 $+$ Pyramid$^1$ & $\mathcal{L}_{BerHu}^{+}$ & \textbf{12\ M} & $39\ fps$ & $0 - 50\ m$ & 0.078 & 0.301 & 2.553 & 0.121 & 11.093 & 0.034 & 0.931 & 0.985 & 0.996 \\
DSN 3 $+$ Pyramid$^1$ $+$ Sem & $\mathcal{L}_{BerHu}^{+}$ & \textbf{12\ M} & $39\ fps$ & $0 - 50\ m$ & 0.077 & 0.301 & 3.539 & 0.123 & 11.309 & 0.034 & 0.927 & 0.984 & 0.996 \\
DSN 3 $+$ Pyramid$^1$ $+$ CS & $\mathcal{L}_{BerHu}^{+}$ & \textbf{12\ M} & $39\ fps$ & $0 - 50\ m$ & \textbf{0.074} & 0.302 & 2.437 & 0.116 & 10.579 & \textbf{0.032} & \textbf{0.937} & \textbf{0.988} & 0.996 \\
DSN 3 $+$ Pyramid$^2$ $+$ CS & $\mathcal{L}_{BerHu}^{+}$ & \textbf{12\ M} & $32\ fps$ & $0 - 50\ m$ & \textbf{0.074} & \textbf{0.274} & \textbf{2.393} & \textbf{0.114} & \textbf{10.413} & \textbf{0.032} & \textbf{0.937} & \textbf{0.988} & \textbf{0.997} \\\hline
\end{tabular}
}
\end{table*}

\subsubsection{Surface Normals Analysis}

The surface normals module is tested in the DSN $3$ framework, along with the Pyramid$^1$ module, the edge detection phase, the 2.5D loss function and the weights pre-trained on the reference depth maps of the Cityscapes \cite{cordts2016cityscapes} which are produced from their respective disparity maps. Based on the Abs Rel metric in Table \ref{tab:kittisn}, we may see that the network configuration with $\psi_1=10^6$, $\psi_2=10^6$ and without the presence of the Gaussian filter is the one that generates the smallest errors, implying that the edge detector of the surface normals module is robust to noise in the input image. 

Moreover, considering the errors and the accuracy of the methods exposed in Table \ref{tab:kittisn}, we have that the geometric cues, provided by the surface normals ground truth and computed by the geometric loss function, support the DSN to retrieve 2.5D maps more consistent with the scene. The surface normals module has little influence on the total size of the framework and can be used in other methods that tackle SIDE problems. Only the prediction speed is negatively affected by the use of the aforementioned strategy.

An overview of the works that address the SIDE task and that evaluate its approaches in the KITTI Depth \cite{uhrig2017sparsity} can be visualized in Table \ref{kitti_tab:comparison}. The results obtained by our pipeline are comparable to those generated by state-of-the-art methods with different types of training procedures. Furthermore, based on Table \ref{tab:litside}, the proposed CNN has fewer trainable parameters and, due to its prediction speed, it may be applied to real self-driving scenarios.

\begin{table*}[]
\centering
\caption{Evaluation of the DSN with the surface normals module in the KITTI Depth dataset \cite{uhrig2017sparsity}. The DSN $3$ $+$ Pyramid$^1$ $+$ CS $+$ SN setup, as well as the Sobel edge detector and the $\mathcal{L}_{BerHu}^{+}$ loss, is standard for all experiments. SN means that the training is carried out with additional surface normals ground truth. The Blur column indicates the presence or absence of a Convolution with Gaussian kernel. $\psi_1$ and $\psi_2$ are the constants of the 2.5D loss function employed in the pre-training and training steps respectively}
\label{tab:kittisn}
\resizebox{\textwidth}{!}{
\begin{tabular}{c|c|c|c|c|c|c|cccccc|ccc}
\hline
\textbf{Method} & \textbf{Size} & \textbf{Speed} & \textbf{Blur} & \textbf{$\psi_1$} & \textbf{$\psi_2$} & \textbf{Cap} & \textbf{Abs Rel}$\downarrow$ & \textbf{Sqr Rel}$\downarrow$ & \textbf{RMSE}$\downarrow$ & \textbf{RMSE (log)}$\downarrow$ & \textbf{SILog}$\downarrow$ & \textbf{log10}$\downarrow$ & $\bm{\delta < 1.25}$$\uparrow$ & $\bm{\delta < 1.25^2}$$\uparrow$ & $\bm{\delta < 1.25^3}$$\uparrow$ \\\hline \hline
DSN 3 $+$ Pyramid$^1$ $+$ CS $+$ SN & \textbf{12\ M} & $\boldsymbol{33}\ \boldsymbol{fps}$ & \xmark & $10^6$ & $10^6$ & $0 - 80\ m$ & \textbf{0.075} & 0.363 & \textbf{3.253} & \textbf{0.119} & \textbf{10.886} & \textbf{0.033} & \textbf{0.934} & \textbf{0.986} & \textbf{0.996}\\
DSN 3 $+$ Pyramid$^1$ $+$ CS $+$ SN & \textbf{12\ M} & $\boldsymbol{33}\ \boldsymbol{fps}$ & \xmark & $10^6$ & $10^7$ & $0 - 80\ m$ & 0.076 & \textbf{0.360} & 3.259 & 0.120 & 10.950 & 0.034 & 0.932 & \textbf{0.986} & \textbf{0.996}\\
DSN 3 $+$ Pyramid$^1$ $+$ CS $+$ SN & \textbf{12\ M} & $32\ fps$ & \cmark & $10^6$ & $10^6$ & $0 - 80\ m$ & \textbf{0.075} & 0.365 & 3.264 & 0.120 & 10.947 & \textbf{0.033} & \textbf{0.934} & \textbf{0.986} & \textbf{0.996}\\
DSN 3 $+$ Pyramid$^1$ $+$ CS $+$ SN & \textbf{12\ M} & $32\ fps$ & \cmark & $10^6$ & $10^7$ & $0 - 80\ m$ & 0.081 & 0.391 & 3.347 & 0.126 & 11.591 & 0.036 & 0.924 & 0.985 & \textbf{0.996}\\\hline\hline
DSN 3 $+$ Pyramid$^1$ $+$ CS $+$ SN & \textbf{12\ M} & $\boldsymbol{33}\ \boldsymbol{fps}$ & \xmark & $10^6$ & $10^6$ & $0 - 50\ m$ & \textbf{0.071} & 0.267 & \textbf{2.351} & \textbf{0.111} & \textbf{10.053} & \textbf{0.031} & \textbf{0.944} & \textbf{0.989} & \textbf{0.997}\\
DSN 3 $+$ Pyramid$^1$ $+$ CS $+$ SN & \textbf{12\ M} & $\boldsymbol{33}\ \boldsymbol{fps}$ & \xmark & $10^6$ & $10^7$ & $0 - 50\ m$ & 0.072 & \textbf{0.263} & 2.365 & \textbf{0.111} & 10.129 & 0.032 & 0.942 & \textbf{0.989} & \textbf{0.997}\\
DSN 3 $+$ Pyramid$^1$ $+$ CS $+$ SN & \textbf{12\ M} & $32\ fps$ & \cmark & $10^6$ & $10^6$ & $0 - 50\ m$ & 0.072 & 0.268 & 2.355 & \textbf{0.111} & 10.113 & \textbf{0.031} & \textbf{0.944} & \textbf{0.989} & \textbf{0.997}\\
DSN 3 $+$ Pyramid$^1$ $+$ CS $+$ SN & \textbf{12\ M} & $32\ fps$ & \cmark & $10^6$ & $10^7$ & $0 - 50\ m$ & 0.077 & 0.296 & 2.473 & 0.118 & 10.827 & 0.034 & 0.934 & 0.987 & 0.996\\
\hline
\end{tabular}
}
\end{table*}

\begin{table*}[]
\centering
\caption{Performances of our framework and other works in the SIDE literature according to the KITTI Depth \cite{uhrig2017sparsity} metrics. Legend: DSN$\dagger$ - DSN setup that yields the most accurate overall results; MV - monocular video; SP - stereo pair; CM - camera motion; SI - single image; GT - ground truth; SL - semantic labels; SN - surface normals; DSO - direct sparse odometry; SGM - semi-global matching; DM - disparity map; Cal. Cam. - calibrated camera}
\label{kitti_tab:comparison}
\resizebox{\textwidth}{!}{%
\begin{tabular}{c|c|c|cccc|ccc}
\hline
\textbf{Method} & Training Input/Auxiliar Info & \textbf{Cap} & \textbf{Abs Rel}$\downarrow$ & \textbf{Sqr Rel}$\downarrow$ & \textbf{RMSE}$\downarrow$ & \textbf{RMSE \textit{log}}$\downarrow$ & $\bm{\delta < 1.25}$$\uparrow$ & $\bm{\delta < 1.25^2}$$\uparrow$ & $\bm{\delta < 1.25^3}$$\uparrow$\\
\hline
\hline
Saxena \etal \cite{saxena2008make3d} & SI/GT Depth & $0 - 80\ m$ & 0.280 & 3.012 & 8.734 & 0.361 & 0.601 & 0.820 & 0.926\\
Eigen \etal \cite{eigen2014depth} & SI/GT Depth & $0 - 80\ m$ & 0.203 & 1.548 & 6.307 & 0.282 & 0.702 & 0.898 & 0.967\\
Liu \etal \cite{liu2015learning} & SI/GT Depth & $0 - 80\ m$ & 0.201 & 1.584 & 6.471 & 0.273 & 0.680 & 0.898 & 0.967\\
Zhou \etal \cite{zhou2017unsupervised} & MV/- & $0 - 80\ m$ & 0.198 & 1.836 & 6.565 & 0.275 & 0.718 & 0.901 & 0.960\\
UnDeepVO \cite{li2018undeepvo} & SP/Cal. Cam. & $0 - 80\ m$ & 0.183 & 1.730 & 6.570 & 0.268 & - & - & -\\
Yang \etal \cite{yang2017unsupervised} & MV/- & $0 - 80\ m$ & 0.182 & 1.481 & 6.501 & 0.267 & 0.725 & 0.906 & 0.963\\
AdaDepth \cite{nath2018adadepth} & SI and Synt. SI/Synt. GT Depth (self) & $0 - 80\ m$ & 0.167 & 1.257 & 5.578 & 0.237 & 0.771 & 0.922 & 0.971\\
Klodt \etal \cite{klodt2018supervising} & MV/- & $0 - 80\ m$ & 0.166 & 1.490 & 5.988 & - & 0.778 & 0.919 & 0.966\\
Mahjourian \etal \cite{mahjourian2018unsupervised} & MV/- & $0 - 80\ m$ & 0.163 & 1.240 & 6.220 & 0.250 & 0.762 & 0.916 & 0.968\\
LEGO \cite{yang2018lego} & MV/- & $0 - 80\ m$ & 0.162 & 1.352 & 6.276 & 0.252 & - & - & -\\
Poggi \etal \cite{poggi2018towards} & SP/- & $0 - 80\ m$ & 0.153 & 1.363 & 6.030 & 0.252 & 0.789 & 0.918 & 0.963\\
GeoNet \cite{yin2018geonet} & MV/- & $0 - 80\ m$ & 0.153 & 1.328 & 5.737 & 0.232 & 0.802 & 0.934 & 0.972\\
Pilzer \etal \cite{pilzer2018unsupervised} & SP/Cal. Cam. & $0 - 80\ m$ & 0.152 & 1.388 & 6.016 & 0.247 & 0.789 & 0.918 & 0.965\\
DDVO \cite{wang2018learning} & MV/- & $0 - 80\ m$ & 0.151 & 1.257 & 5.583 & 0.228 & 0.810 & 0.936 & 0.974\\
DF-Net \cite{zou2018df} & MV/- & $0 - 80\ m$ & 0.150 & 1.124 & 5.507 & 0.223 & 0.806 & 0.933 & 0.973\\
Ranjan \etal \cite{ranjan2019competitive} & MV/- & $0 - 80\ m$ & 0.148 & 1.149 & 5.464 & 0.226 & 0.815 & 0.935 & 0.973\\
MiniNet \etal \cite{liu2020mininet} & MV/- & $0 - 50\ m$ & 0.141 & 1.080 & 5.264 & 0.216 & 0.825 & 0.941 & 0.976\\
Struct2depth \cite{casser2019depth} & MV/Cal. Cam. (Only intrinsics matrix) & $0 - 80\ m$ & 0.141 & 1.026 & 5.291 & 0.215 & 0.816 & 0.945 & 0.979\\
Elkerdawy \etal \cite{elkerdawy2019lightweight} & SP/Cal. Cam. & $0 - 80\ m$ & 0.136 & - & 5.891 & - & 0.827 & - & -\\
PHN \etal\cite{zhang2018progressive} & SI/GT Depth & $0 - 80\ m$ & 0.136 & - & 4.082 & 0.164 & 0.864 & 0.966 & 0.989\\
Zhan FullNYU \cite{zhan2018unsupervised} & SP/CM & $0 - 80\ m$ & 0.135 & 1.132 & 5.585 & 0.229 & 0.820 & 0.933 & 0.971\\
Wong \etal \cite{wong2019bilateral} & SP/- (Cal. needed in test) & $0 - 80\ m$ & 0.133 & 1.126 & 5.515 & 0.231 & 0.826 & 0.934 & 0.969\\
3Net (ResNet-50) \cite{poggi2018learning} & Trinocular setup/- & $0 - 80\ m$ & 0.129 & 0.996 & 5.281 & 0.223 & 0.831 & 0.939 & 0.974\\
StrAT \cite{mehta2018structured} & SI/Cal. Cam. & $0 - 80\ m$ & 0.128 & 1.019 & 5.403 & 0.227 & 0.827 & 0.935 & 0.971\\
EPC++ \cite{luo2018every} & MV/Cal. Cam. (Only intrinsics matrix) & $0 - 80\ m$ & 0.128 & 0.935 & 5.011 & 0.209 & 0.831 & 0.945 & 0.979\\
Gordon \etal \cite{gordon2019depth} & MV/- & $0 - 80\ m$ & 0.124 & 0.930 & 5.120 & 0.206 & 0.851 & 0.950 & 0.978\\
Sparse-to-Continuous \cite{dos2019sparse} & SI/GT Depth & $0 - 80\ m$ & 0.123 & 0.641 & 4.524 & 0.199 & 0.881 & 0.966 & 0.986\\
SA-Attention \cite{zhou2019unsupervised} &  MV/Cal. Cam. (Only intrinsics matrix) & $0 - 80\ m$ & 0.121 & 0.837 & 4.945 & 0.197 & 0.853 & 0.955 & 0.982\\
3Net (VGG) \cite{poggi2018learning} & Trinocular setup/- & $0 - 80\ m$ & 0.119 & 1.201 & 5.888 & 0.208 & 0.844 & 0.941 & 0.978\\
SceneNet \cite{chen2019towards} & SI and SP/SL & $0 - 80\ m$ & 0.118 & 0.905 & 5.096 & 0.211 & 0.839 & 0.945 & 0.977\\
Su \etal\cite{su2020monocular} & SI/GT Depth & $0 - 80\ m$ & 0.117 & - & 4.251 & 0.174 & 0.894 & 0.971 & 0.984\\
SDNet \cite{ochs2019sdnet} & SI/GT Depth and SL & $0 - 80\ m$ & 0.116 & 0.945 & 4.916 & 0.208 & 0.861 & 0.952 & 0.968\\
Cao \etal\cite{cao2017estimating} & SI/GT Depth & $0 - 80\ m$ & 0.115 & - & 4.712 & 0.198 & 0.887 & 0.963 & 0.982\\
Monodepth \cite{godard2017unsupervised} & SP/Cal. Cam. & $0 - 80\ m$ & 0.114 & 0.898 & 4.935 & 0.206 & 0.861 & 0.949 & 0.976\\
LSIM \cite{goldman2019learn} & SP/Cal. Cam. & $0 - 80\ m$ & 0.113 & 0.898 & 5.048 & 0.208 & 0.853 & 0.948 & 0.976\\
Kuznietsov \etal \cite{kuznietsov2017semi} & SP/Cal. Cam. and GT Depth & $0 - 80\ m$ & 0.113 & 0.741 & 4.621 & 0.189 & 0.862 & 0.960 & 0.986\\
SuperDepth \cite{pillai2019superdepth} & SP/Cal. Cam. & $0 - 80\ m$ & 0.112 & 0.875 & 4.958 & 0.207 & 0.852 & 0.947 & 0.977\\
DeepLabV3+ (F10) \cite{gur2019single} & SI/GT Depth and Cal. Cam. & $0 - 80\ m$ & 0.110 & 0.666 & 4.186 & 0.168 & 0.880 & 0.966 & 0.988\\
Monodepth2 \cite{godard2019digging} & SP or MV or both/Cal. Cam. & $0 - 80\ m$ & 0.106 & 0.806 & 4.630 & 0.193 & 0.876 & 0.958 & 0.980\\
PackNet-SfM \cite{guizilini2019packnet} & MV/Velocity term (optional) & $0 - 80\ m$ & 0.104 & 0.758 & 4.386 & 0.182 & 0.895 & 0.964 & 0.982\\
Guizilini \etal \cite{guizilini2020semantically} & MV/Pre-tained network with SL & $0 - 80\ m$ & 0.100 & 0.761 & 4.270 & 0.175 & 0.902 & 0.965 & 0.982\\
CFA \cite{gurram2018monocular} & SI/GT Depth and SL & $0 - 80\ m$ & 0.100 & 0.601 & 4.298 & 0.174 & 0.874 & 0.966 & 0.989\\
Refine\&Distill \cite{pilzer2019refine} & SP/Cal. Cam. & $0 - 80\ m$ & 0.098 & 0.831 & 4.656 & 0.202 & 0.882 & 0.948 & 0.973\\
DVSO \cite{yang2018deep} & SP/GT Stereo DSO Depth & $0 - 80\ m$ & 0.097 & 0.734 & 4.442 & 0.187 & 0.888 & 0.958 & 0.980\\
SOM \cite{zhu2019structure} & SI/GT Depth & $0 - 80\ m$ & 0.097 & 0.398 & 3.007 & 0.133 & 0.913 & 0.985 & 0.997\\
Depth Hints \cite{watson2019self} & SP/Cal. Cam. & $0 - 80\ m$ & 0.096 & 0.710 & 4.393 & 0.185 & 0.890 & 0.962 & 0.981\\
monoResMatch \cite{tosi2019learning} & SP/Proxy GT with SGM & $0 - 80\ m$ & 0.096 & 0.673 & 4.351 & 0.184 & 0.890 & 0.961 & 0.981\\
VGG16-UNet \cite{guo2018learning} & SI/Synthetic SP-DM and Cal. Cam.  & $0 - 80\ m$ & 0.096 & 0.641 & 4.095 & 0.168 & 0.892 & 0.967 & 0.986\\
SemiDepth \cite{amiri2019semi} & SP/Cal. Cam. and GT Depth & $0 - 80\ m$ & 0.096 & 0.552 & 3.995 & 0.152 & 0.892 & 0.972 & 0.992\\
SVS \cite{luo2018single} & SP/Cal. Cam. & $0 - 80\ m$ & 0.094 & 0.626 & 4.252 & 0.177 & 0.891 & 0.965 & 0.984\\
DenseDepth \cite{alhashim2018high} & SI/GT Depth & $0 - 80\ m$ & 0.093 & 0.589 & 4.170 & 0.171 & 0.886 & 0.965 & 0.986\\
VOMonodepth \cite{andraghetti2019enhancing} & SP/Sparse Depths and Cal. Cam. & $0 - 80\ m$ & 0.091 & 0.548 & 3.690 & 0.181 & 0.892 & 0.956 & 0.979\\
FAL-netB49 \cite{gonzalezbello2020forget} & SP/Cal. Cam. & $0 - 80\ m$ & 0.088 & 0.547 & 4.004 & 0.175 & 0.898 & 0.966 & 0.984\\
Monodepth2-Self \cite{poggi2020uncertainty} & SP or MV or both/Cal. Cam. & $0 - 80\ m$ & 0.083 & - & 3.682 & - & 0.919 & - & -\\
BA-Net \cite{tang2018ba} & MV/Cal. Cam. & $0 - 80\ m$ & 0.083 & - & 3.640 & 0.134 & - & - & -\\
\textbf{DSN$\dagger$} & SI/GT and SN & $0 - 80\ m$ & 0.075 & 0.363 & 3.253 & 0.119 & 0.934 & 0.986 & 0.996\\
VNL \cite{yin2019enforcing} & SI/GT Depth and VN & $0 - 80\ m$ & 0.072 & - & 3.258 & 0.117 & 0.938 & 0.990 & \textbf{0.998}\\
DORN \cite{fu2018deep} & SI/GT Depth & $0 - 80\ m$ & 0.072 & 0.307 & \textbf{2.727} & 0.120 & 0.932 & 0.984 & 0.994\\
BTS \cite{lee2019big} & SI/GT Depth & $0 - 80\ m$ & \textbf{0.059} & \textbf{0.245} & 2.756 & \textbf{0.096} & \textbf{0.956} & \textbf{0.993} & \textbf{0.998}\\
\hline\hline
Zhou \etal \cite{zhou2017unsupervised} & MV/- & $0 - 50\ m$ & 0.190 & 1.436 & 4.975 & 0.258 & 0.735 & 0.915 & 0.968\\
Kim \etal\cite{kim2018deep} & SI/GT Depth & $0 - 50\ m$ & 0.177 & - & 6.570 & 0.254 & - & - & -\\
Garg \etal \cite{garg2016unsupervised} & SP/CM & $1 - 50\ m$ & 0.169 & 1.080 & 5.104 & 0.273 & 0.740 & 0.904 & 0.962\\
Mahjourian \etal \cite{mahjourian2018unsupervised} & MV/- & $0 - 50\ m$ & 0.155 & 0.927 & 4.549 & 0.231 & 0.781 & 0.931 & 0.975\\
GeoNet \cite{yin2018geonet} & MV/- & $0 - 50\ m$ & 0.147 & 0.936 & 4.348 & 0.218 & 0.810 & 0.941 & 0.977\\
Poggi \etal \cite{poggi2018towards} & SP/- & $0 - 50\ m$ & 0.145 & 1.014 & 4.608 & 0.227 & 0.813 & 0.934 & 0.972\\
Pilzer \etal \cite{pilzer2018unsupervised} & SP/Cal. Cam. & $0 - 50\ m$ & 0.144 & 1.007 & 4.660 & 0.240 & 0.793 & 0.923 & 0.968\\
MiniNet \etal \cite{liu2020mininet} & MV/- & $0 - 50\ m$ & 0.135 & 0.839 & 4.067 & 0.205 & 0.838 & 0.947 & 0.978\\
Wong \etal \cite{wong2019bilateral} & SP/- (Cal. needed in test) & $0 - 50\ m$ & 0.126 & 1.832 & 4.172 & 0.217 & 0.840 & 0.941 & 0.973\\
Sparse-to-Continuous \cite{dos2019sparse} & SI/GT Depth & $0 - 50\ m$ & 0.119 & 0.444 & 3.097 & 0.180 & 0.893 & 0.974 & 0.990\\
Monodepth \cite{godard2017unsupervised} & SP/Cal. Cam. & $0 - 50\ m$ & 0.108 & 0.657 & 3.729 & 0.194 & 0.873 & 0.954 & 0.979\\
Kuznietsov \etal \cite{kuznietsov2017semi} & SP/Cal. Cam. and GT Depth & $1 - 50\ m$ & 0.108 & 0.595 & 3.518 & 0.179 & 0.875 & 0.964 & 0.988\\
Cao \etal\cite{cao2017estimating} & SI/GT Depth & $0 - 50\ m$ & 0.107 & - & 3.605 & 0.187 & 0.898 & 0.966 & 0.984\\
Su \etal\cite{su2020monocular} & SI/GT Depth & $0 - 50\ m$ & 0.107 & - & 3.440 & 0.172 & 0.900 & 0.976 & 0.990\\
LSIM \etal \cite{goldman2019learn} & SP/Cal. Cam. & $0 - 50\ m$ & 0.106 & 0.653 & 3.790 & 0.195 & 0.867 & 0.954 & 0.979\\
CFA \cite{gurram2018monocular} & SI/GT Depth and SL & $0 - 50\ m$ & 0.096 & 0.482 & 3.338 & 0.166 & 0.886 & 0.980 & 0.995\\
Gan \etal \cite{gan2018monocular} & SI/GT Depth & $0 - 50\ m$ & 0.094 & 0.552 & 3.133 & 0.165 & 0.898 & 0.967 & 0.986\\
\textbf{DSN$\dagger$} & SI/GT and SN & $0 - 50\ m$ & 0.071 & 0.267 & 2.351 & 0.111 & 0.944 & 0.989 & 0.997\\
DORN \cite{fu2018deep} & SI/GT Depth & $0 - 50\ m$ & 0.071 & 0.268 & 2.271 & 0.116 & 0.936 & 0.985 & 0.995\\
BTS \cite{lee2019big} & SI/GT Depth & $0 - 50\ m$ & \textbf{0.056} & \textbf{0.169} & \textbf{1.925} & \textbf{0.087} & \textbf{0.964} & \textbf{0.994} & \textbf{0.999}\\\hline
\end{tabular}%
} 
\end{table*}

\begin{figure*}[t!]
     \centering
          \begin{subfigure}[t]{0.16\textwidth}
         \centering
         \caption{RGB Input}
         \includegraphics[width=\textwidth]{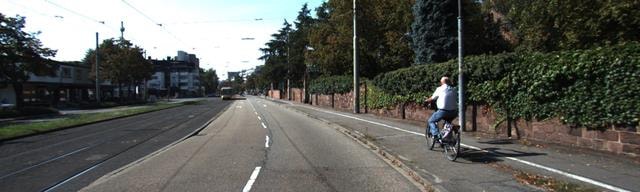}
         \hspace{1em}
         \includegraphics[width=\textwidth]{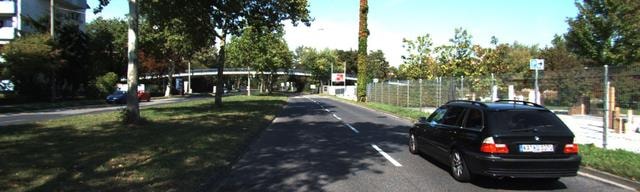}
         \hspace{1em}
         \includegraphics[width=\textwidth]{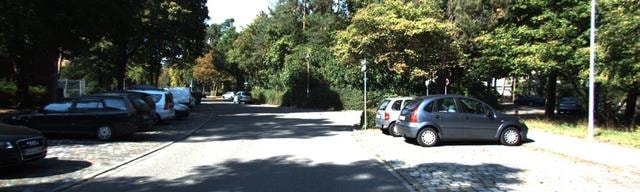}
         \hspace{1em}
         \includegraphics[width=\textwidth]{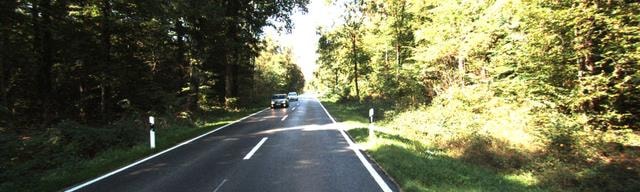}
         \hspace{1em}
         \includegraphics[width=\textwidth]{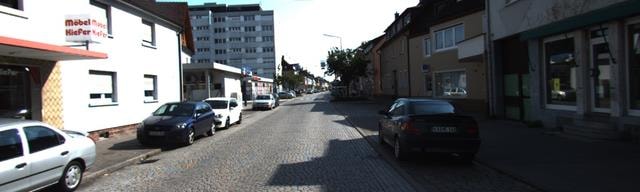}
         \hspace{1em}
         \includegraphics[width=\textwidth]{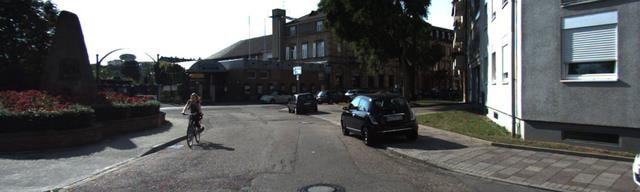}
         \hspace{1em}
         \includegraphics[width=\textwidth]{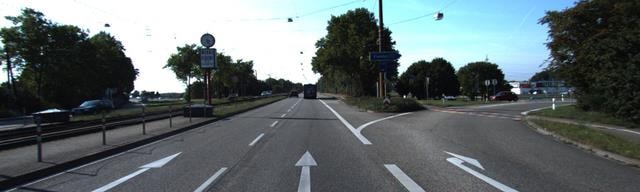}
         \hspace{1em}
         \includegraphics[width=\textwidth]{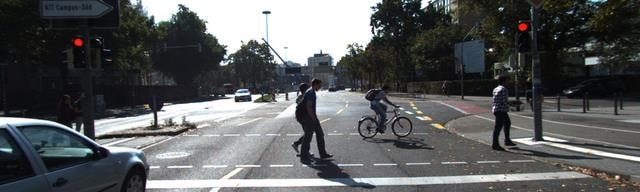}
         \hspace{1em}
         \includegraphics[width=\textwidth]{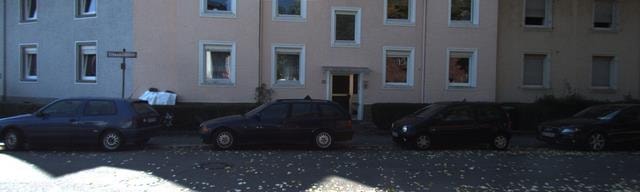}
         \label{kittisubfig0:rgb}
     \end{subfigure}
     \begin{subfigure}[t]{0.16\textwidth}
         \centering
         \caption{Ground Truth}
         \includegraphics[width=\textwidth]{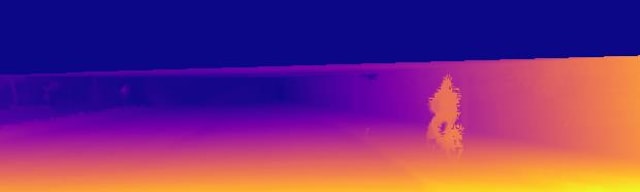}
         \hspace{1em}
         \includegraphics[width=\textwidth]{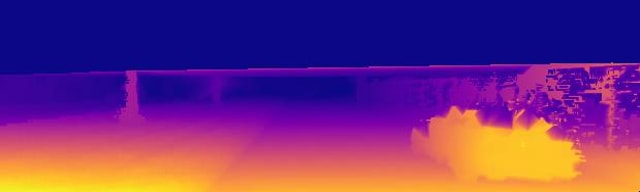}
         \hspace{1em}
         \includegraphics[width=\textwidth]{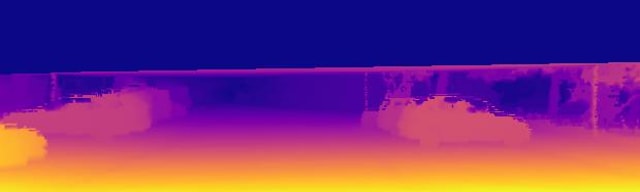}
         \hspace{1em}
         \includegraphics[width=\textwidth]{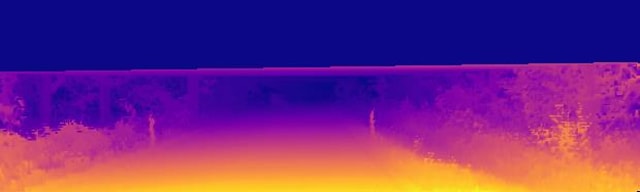}
         \hspace{1em}
         \includegraphics[width=\textwidth]{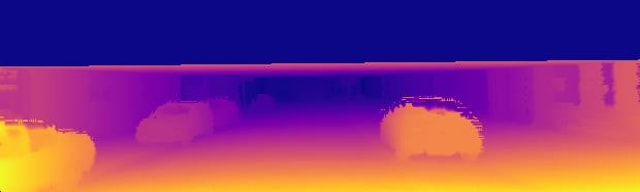}
         \hspace{1em}
         \includegraphics[width=\textwidth]{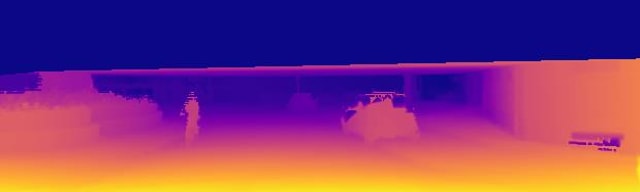}
         \hspace{1em}
         \includegraphics[width=\textwidth]{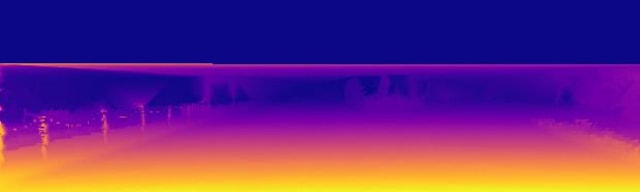}
         \hspace{1em}
         \includegraphics[width=\textwidth]{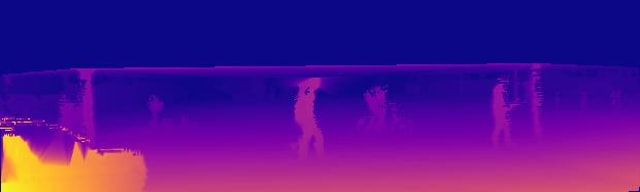}
         \hspace{1em}
         \includegraphics[width=\textwidth]{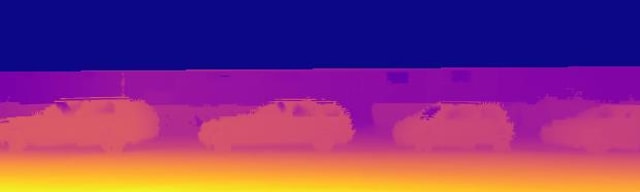}
         \label{kittisubfig0:gt}
     \end{subfigure}
         \begin{subfigure}[t]{0.16\textwidth}
         \centering
         \caption{GeoNet \cite{yin2018geonet}}
         \includegraphics[width=\textwidth]{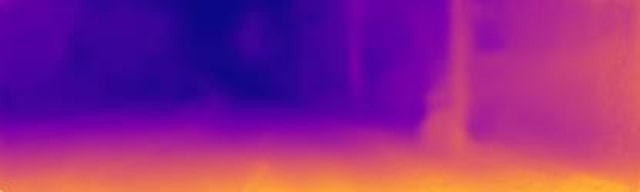}
         \hspace{1em}
         \includegraphics[width=\textwidth]{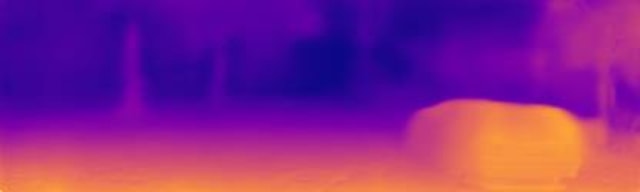}
         \hspace{1em}
         \includegraphics[width=\textwidth]{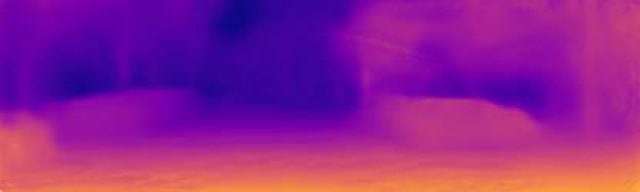}
         \hspace{1em}
         \includegraphics[width=\textwidth]{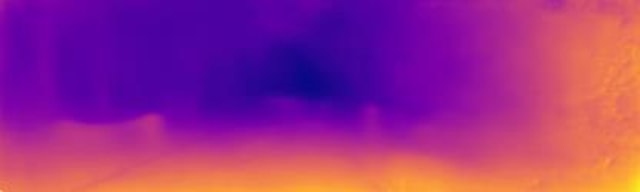}
         \hspace{1em}
         \includegraphics[width=\textwidth]{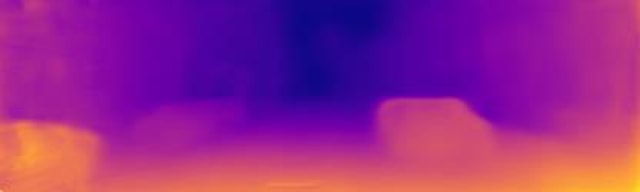}
         \hspace{1em}
         \includegraphics[width=\textwidth]{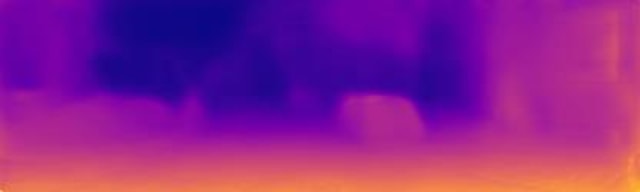}
         \hspace{1em}
         \includegraphics[width=\textwidth]{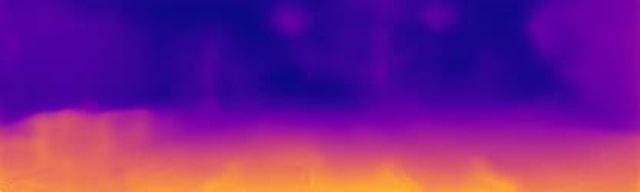}
         \hspace{1em}
         \includegraphics[width=\textwidth]{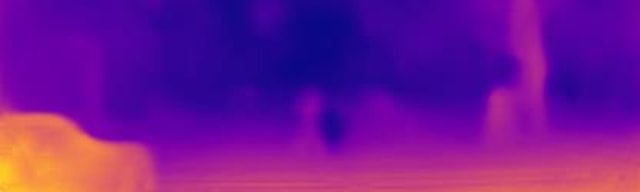}
         \hspace{1em}
         \includegraphics[width=\textwidth]{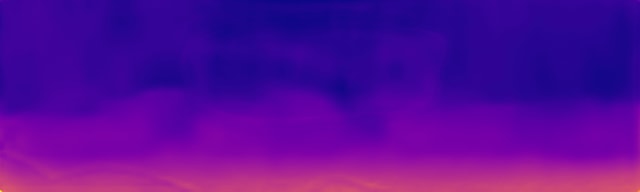}
         \label{kittisubfig0:geonet}
     \end{subfigure}
     \begin{subfigure}[t]{0.16\textwidth}
         \centering
         \caption{Poggi \etal \cite{poggi2018towards}}
         \includegraphics[width=\textwidth]{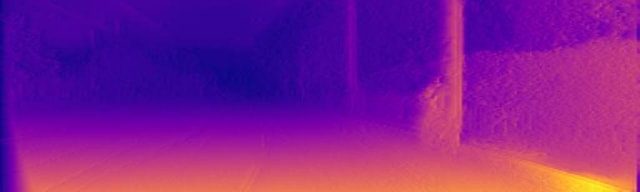}
         \hspace{1em}
         \includegraphics[width=\textwidth]{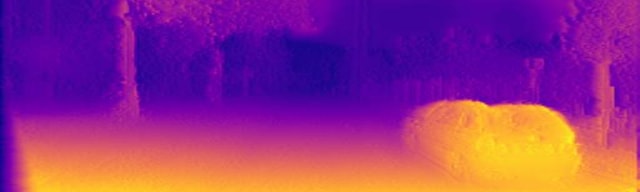}
         \hspace{1em}
         \includegraphics[width=\textwidth]{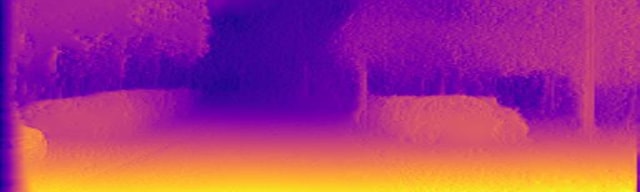}
         \hspace{1em}
         \includegraphics[width=\textwidth]{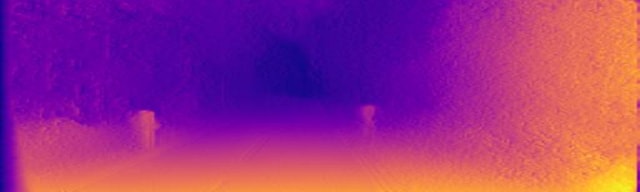}
         \hspace{1em}
         \includegraphics[width=\textwidth]{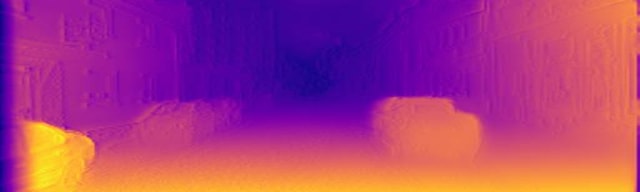}
         \hspace{1em}
         \includegraphics[width=\textwidth]{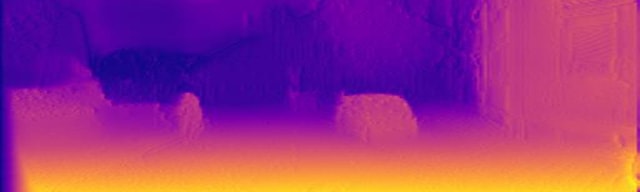}
         \hspace{1em}
         \includegraphics[width=\textwidth]{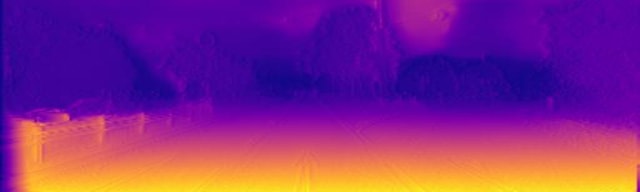}
         \hspace{1em}
         \includegraphics[width=\textwidth]{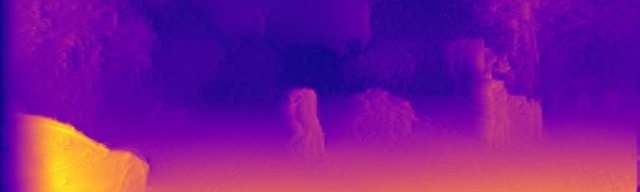}
         \hspace{1em}
         \includegraphics[width=\textwidth]{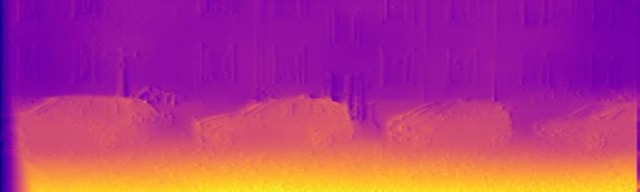}
         \label{kittisubfig0:poggi}
     \end{subfigure}
         \begin{subfigure}[t]{0.16\textwidth}
         \centering
         \caption{MiniNet \cite{liu2020mininet}}
         \includegraphics[width=\textwidth]{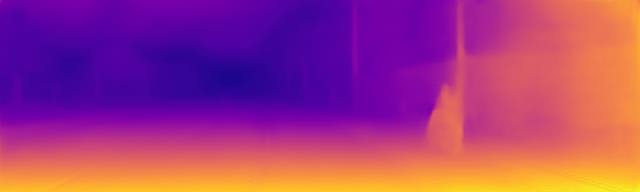}
         \hspace{1em}
         \includegraphics[width=\textwidth]{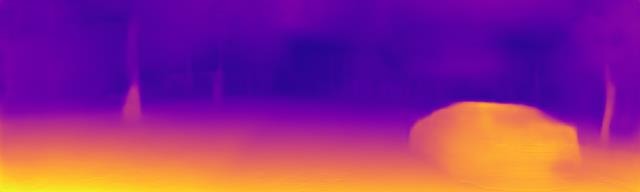}
         \hspace{1em}
         \includegraphics[width=\textwidth]{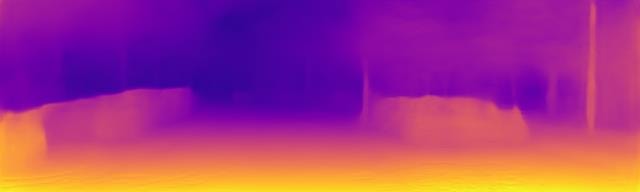}
         \hspace{1em}
         \includegraphics[width=\textwidth]{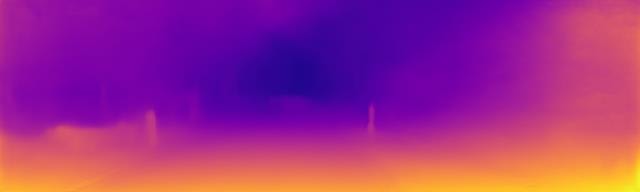}
         \hspace{1em}
         \includegraphics[width=\textwidth]{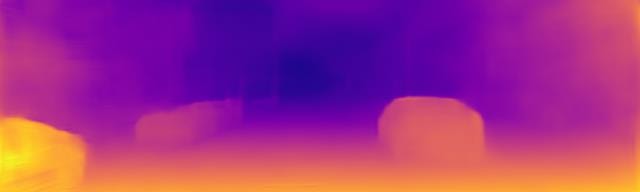}
         \hspace{1em}
         \includegraphics[width=\textwidth]{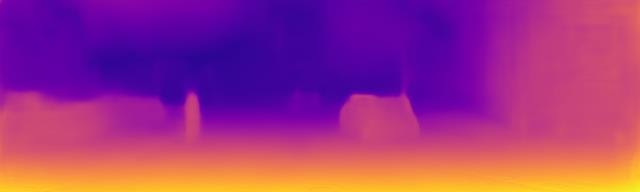}
         \hspace{1em}
         \includegraphics[width=\textwidth]{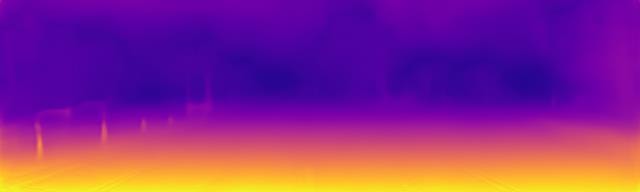}
         \hspace{1em}
         \includegraphics[width=\textwidth]{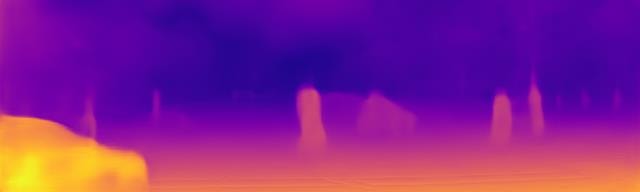}
         \hspace{1em}
         \includegraphics[width=\textwidth]{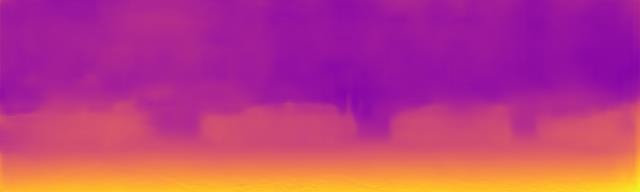}
         \label{kittisubfig0:mininet}
     \end{subfigure}
     \begin{subfigure}[t]{0.16\textwidth}
         \centering
         \caption{DSN$\dagger$}
         \includegraphics[width=\textwidth]{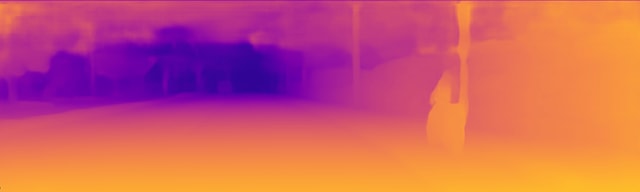}
         \hspace{1em}
         \includegraphics[width=\textwidth]{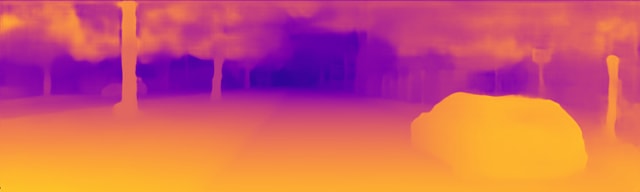}
         \hspace{1em}
         \includegraphics[width=\textwidth]{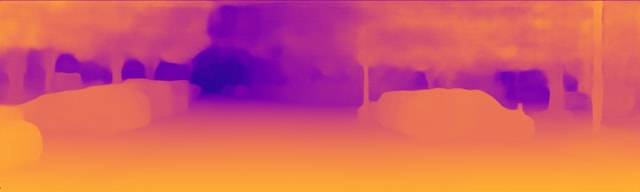}
         \hspace{1em}
         \includegraphics[width=\textwidth]{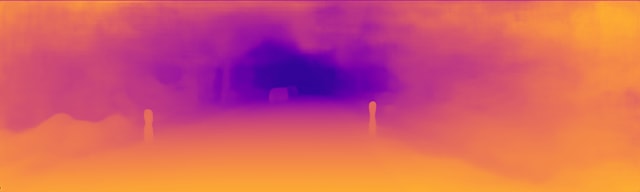}
         \hspace{1em}
         \includegraphics[width=\textwidth]{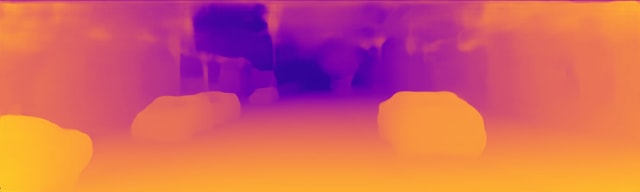}
         \hspace{1em}
         \includegraphics[width=\textwidth]{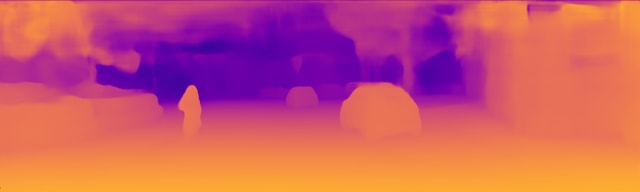}
         \hspace{1em}
         \includegraphics[width=\textwidth]{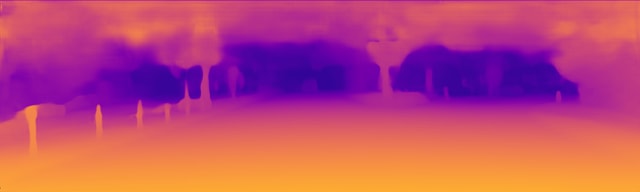}
         \hspace{1em}
         \includegraphics[width=\textwidth]{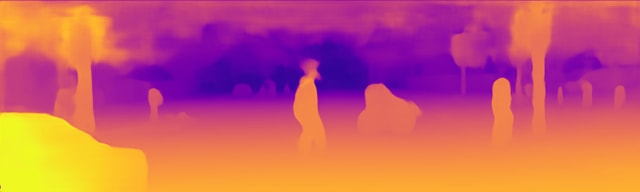}
         \hspace{1em}
         \includegraphics[width=\textwidth]{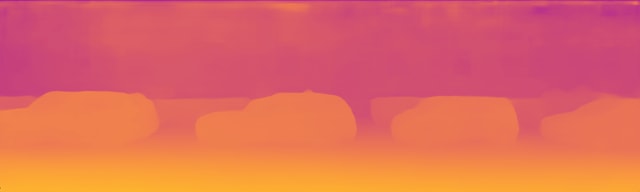}
         \label{kittisubfig0:DSN}
     \end{subfigure}
     \caption{Qualitative comparison between the predictions from our best approach (DSN$\dagger$) and from recent self-supervised methods that address the SIDE task. The input images and the ground truth belong to the KITTI Depth dataset \cite{uhrig2017sparsity}. The reference depth maps are interpolated for visualization purposes} 
     \label{kittisubfig0:all}
\end{figure*}

The depth maps retrieved by the DSN configuration of Table \ref{kitti_tab:comparison} are depicted in Fig. \ref{kittisubfig0:DSN}. Compared to the methods of Figs. \ref{kittisubfig0:geonet}, \ref{kittisubfig0:poggi} and \ref{kittisubfig0:mininet}, ours is able to better identify the shape of the objects that compose the scene, outputting more smoothly and well-outlined contours, even in regions close to occlusions. Also, in the depth maps of Fig. \ref{kittisubfig0:DSN}, we may notice a higher amount of inferred artifacts, mainly in the more distant regions of the scene, which are more difficult to be comprehended by the other methods.

\begin{figure*}[t!]
     \centering
     \begin{subfigure}[t]{0.2\textwidth}
         \centering
         \caption{RGB Input}
         \includegraphics[width=\textwidth]{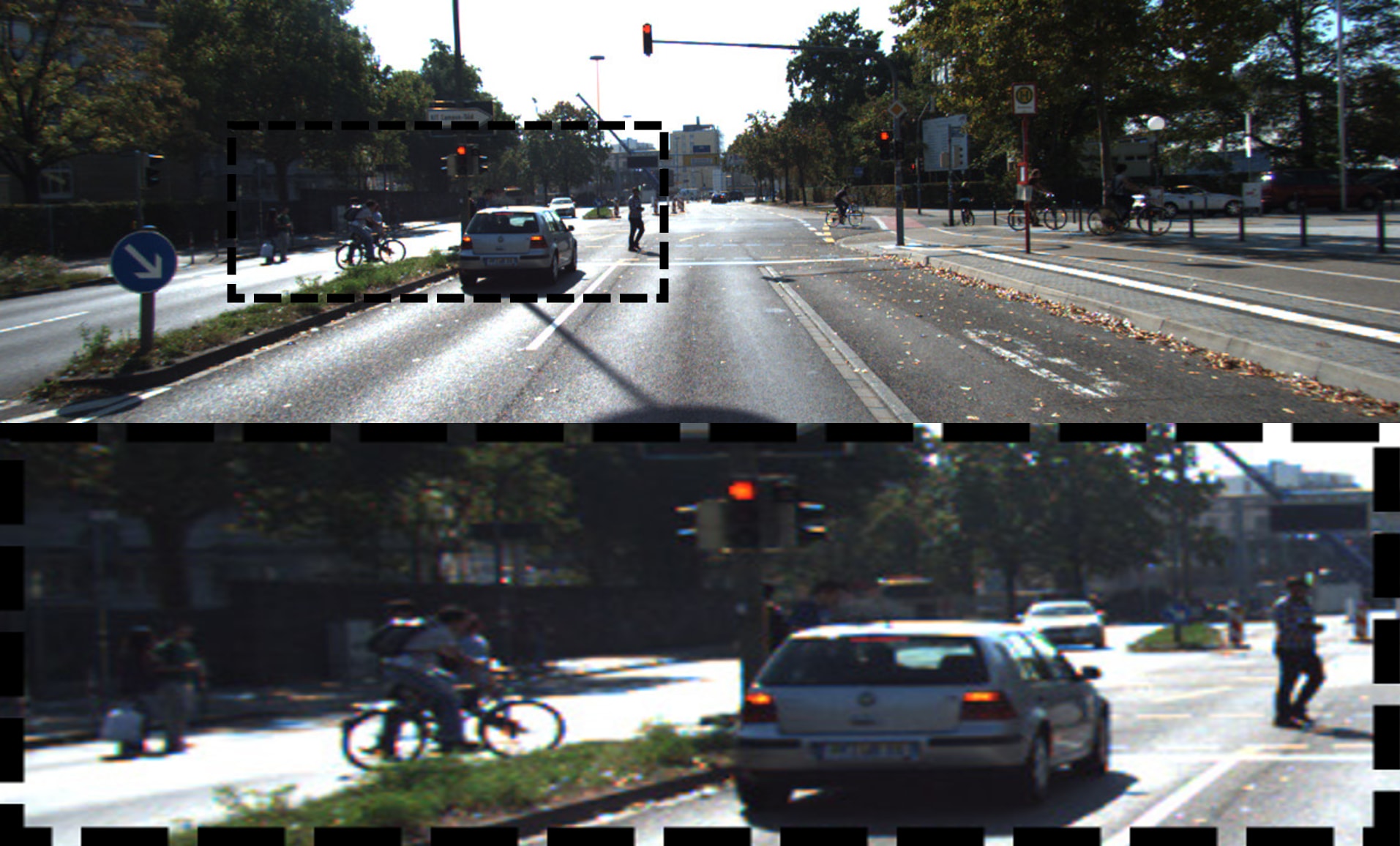}
         \hspace{1em}
         \includegraphics[width=\textwidth]{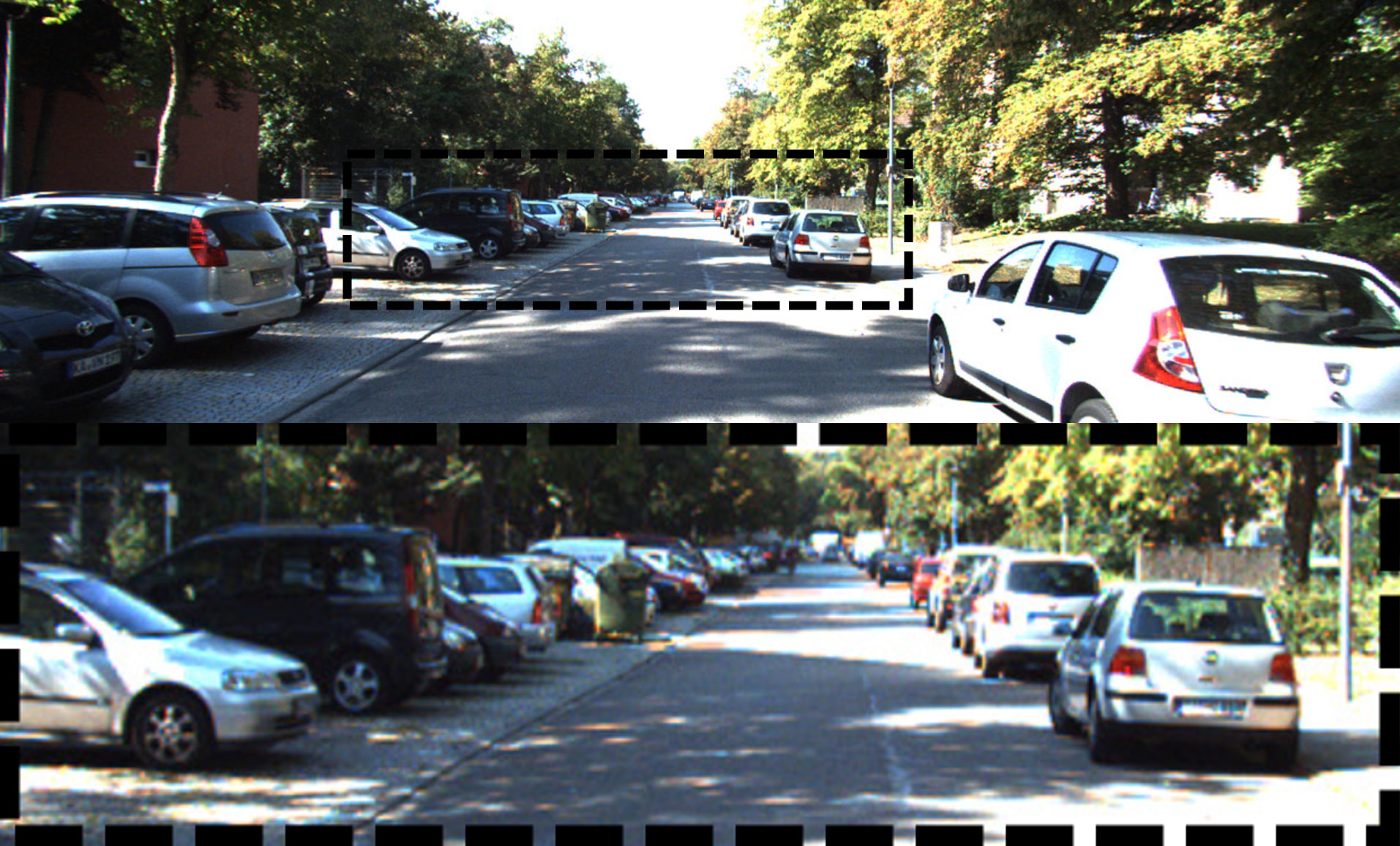}
         \hspace{1em}
         \includegraphics[width=\textwidth]{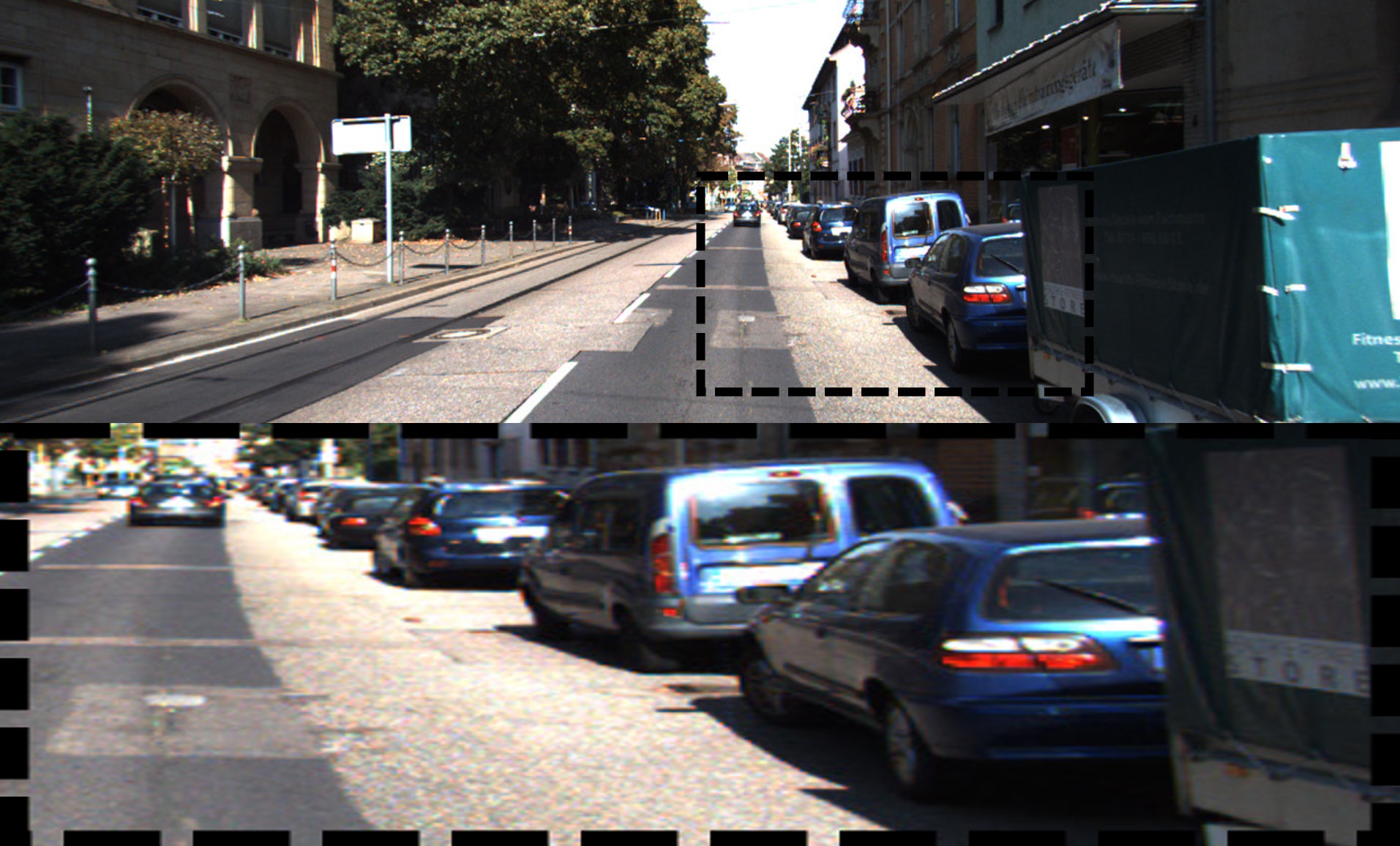}
         \label{kittisub1:rgb}
     \end{subfigure}
     \begin{subfigure}[t]{0.2\textwidth}
         \centering
         \caption{DORN \cite{fu2018deep}}
         \includegraphics[width=\textwidth]{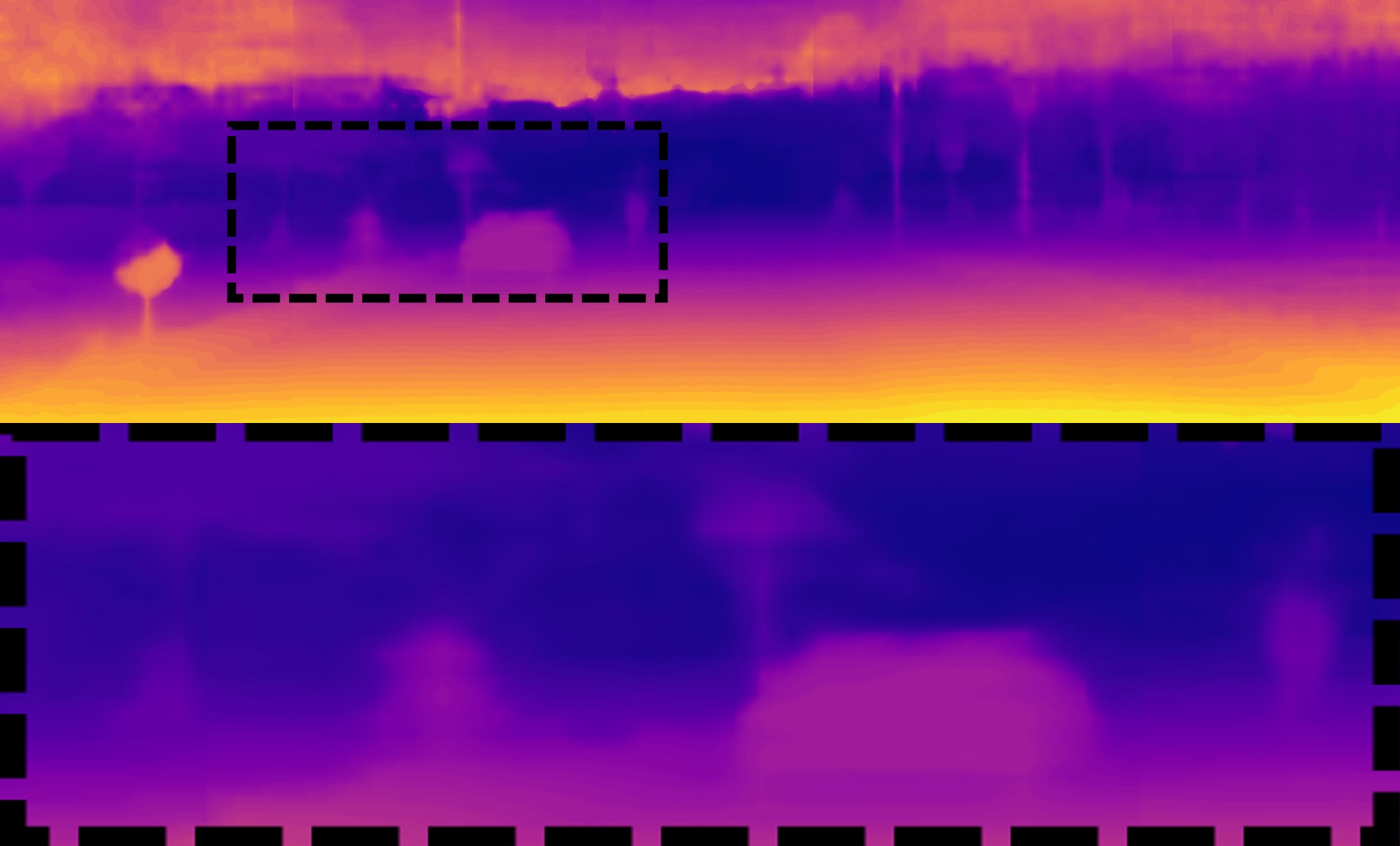}
         \hspace{1em}
         \includegraphics[width=\textwidth]{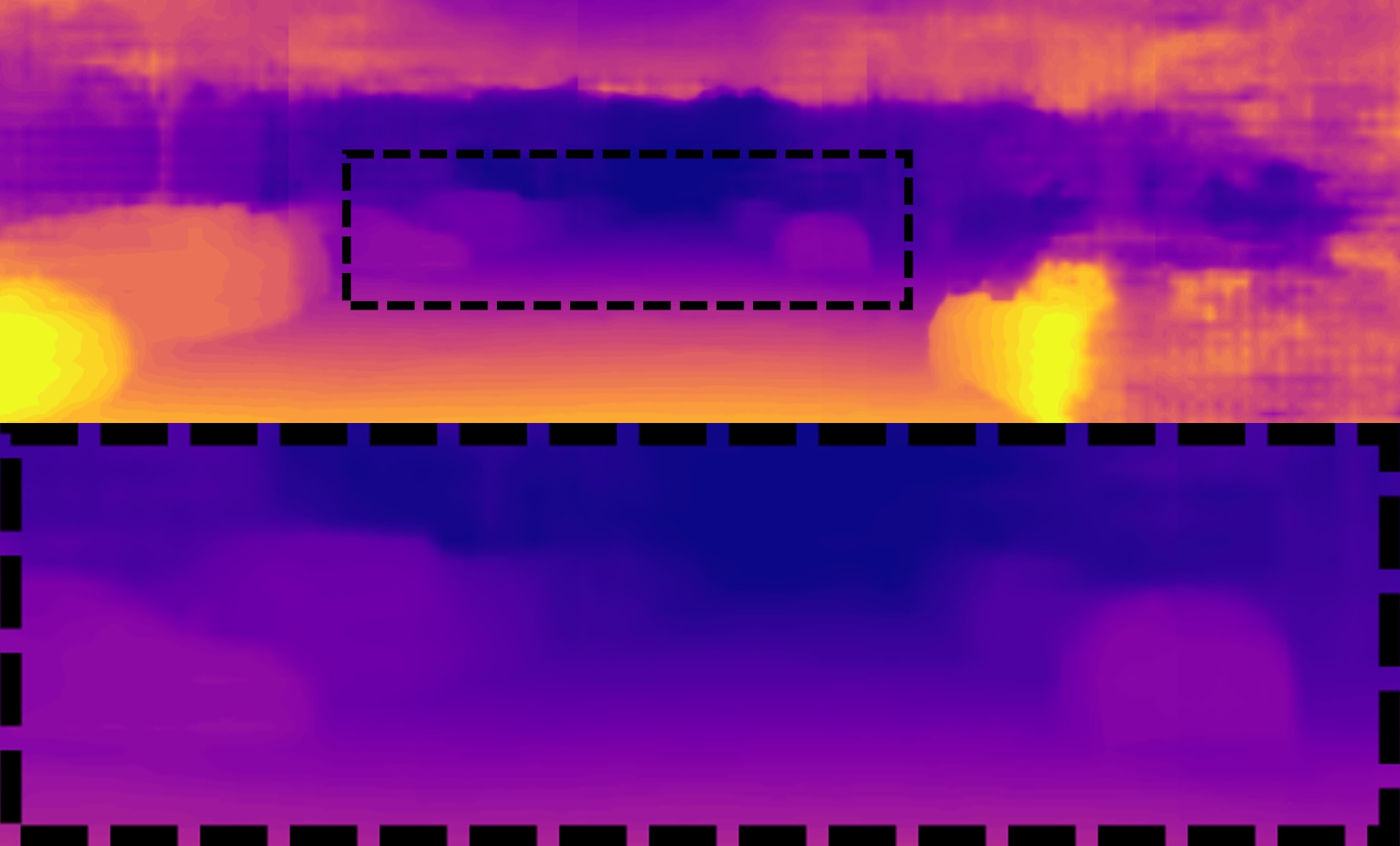}
         \hspace{1em}
         \includegraphics[width=\textwidth]{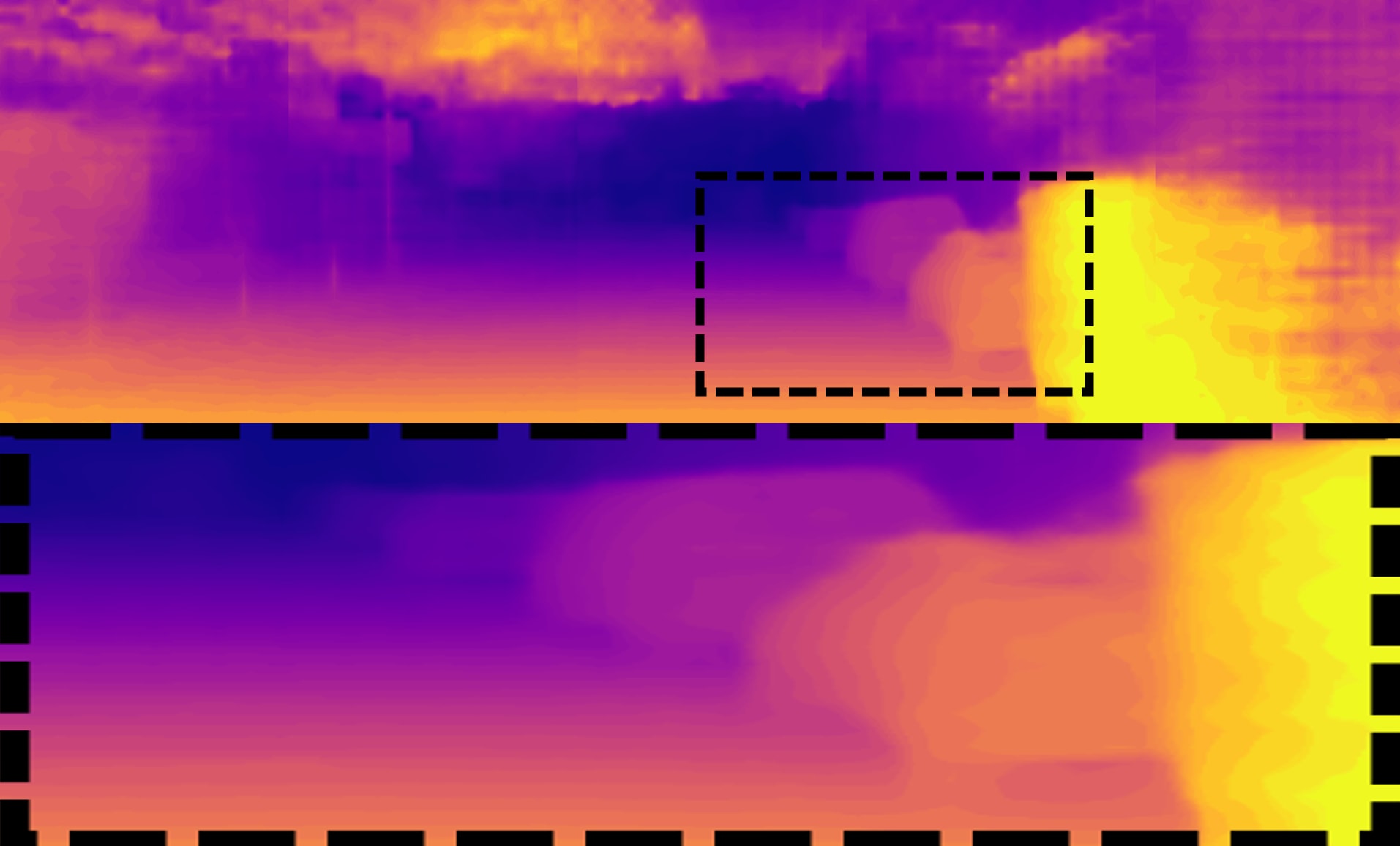}
         \label{kittisub1:dorn}
     \end{subfigure}
          \begin{subfigure}[t]{0.2\textwidth}
         \centering
         \caption{DenseDepth \cite{alhashim2018high}}
         \includegraphics[width=\textwidth]{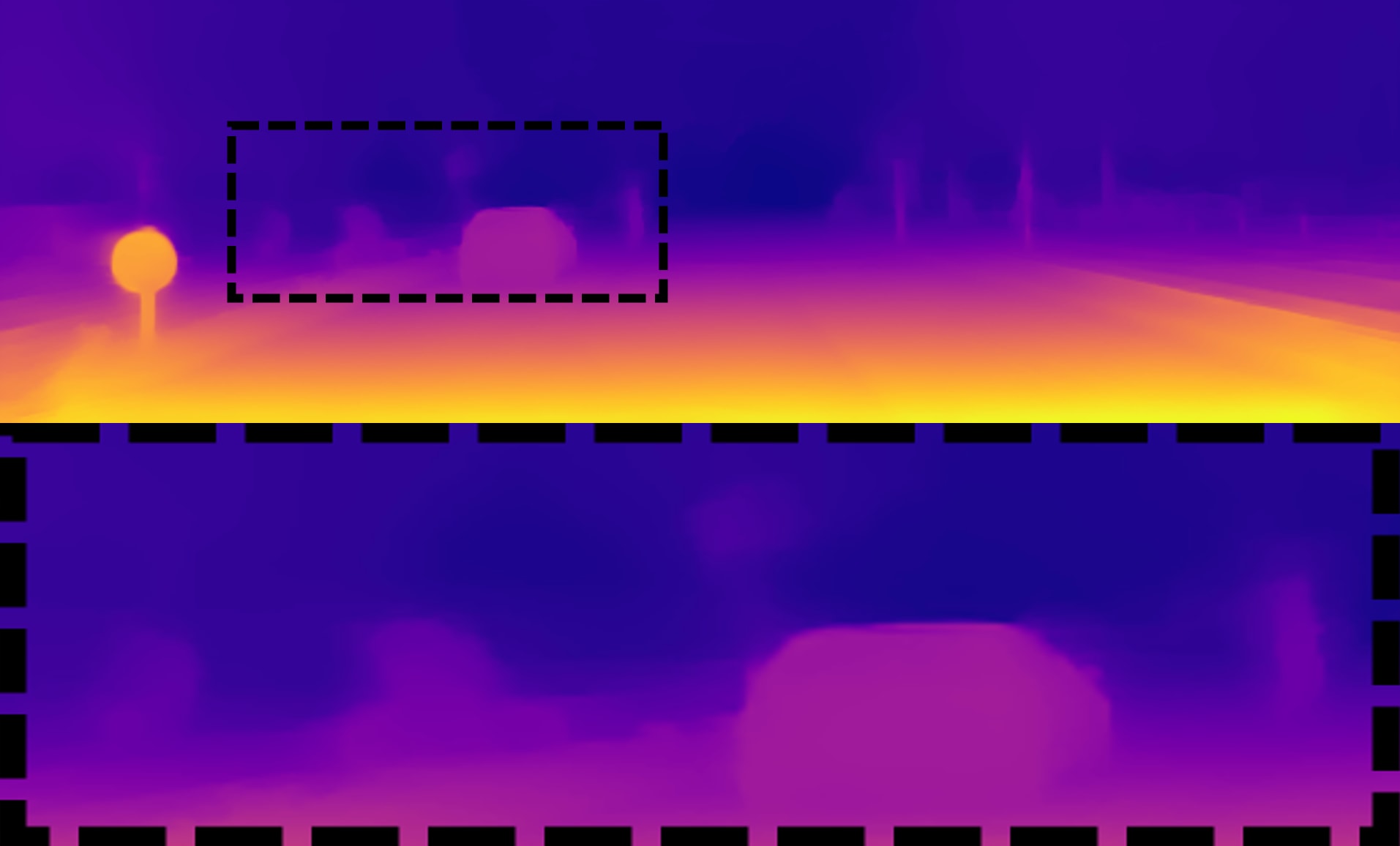}
         \hspace{1em}
         \includegraphics[width=\textwidth]{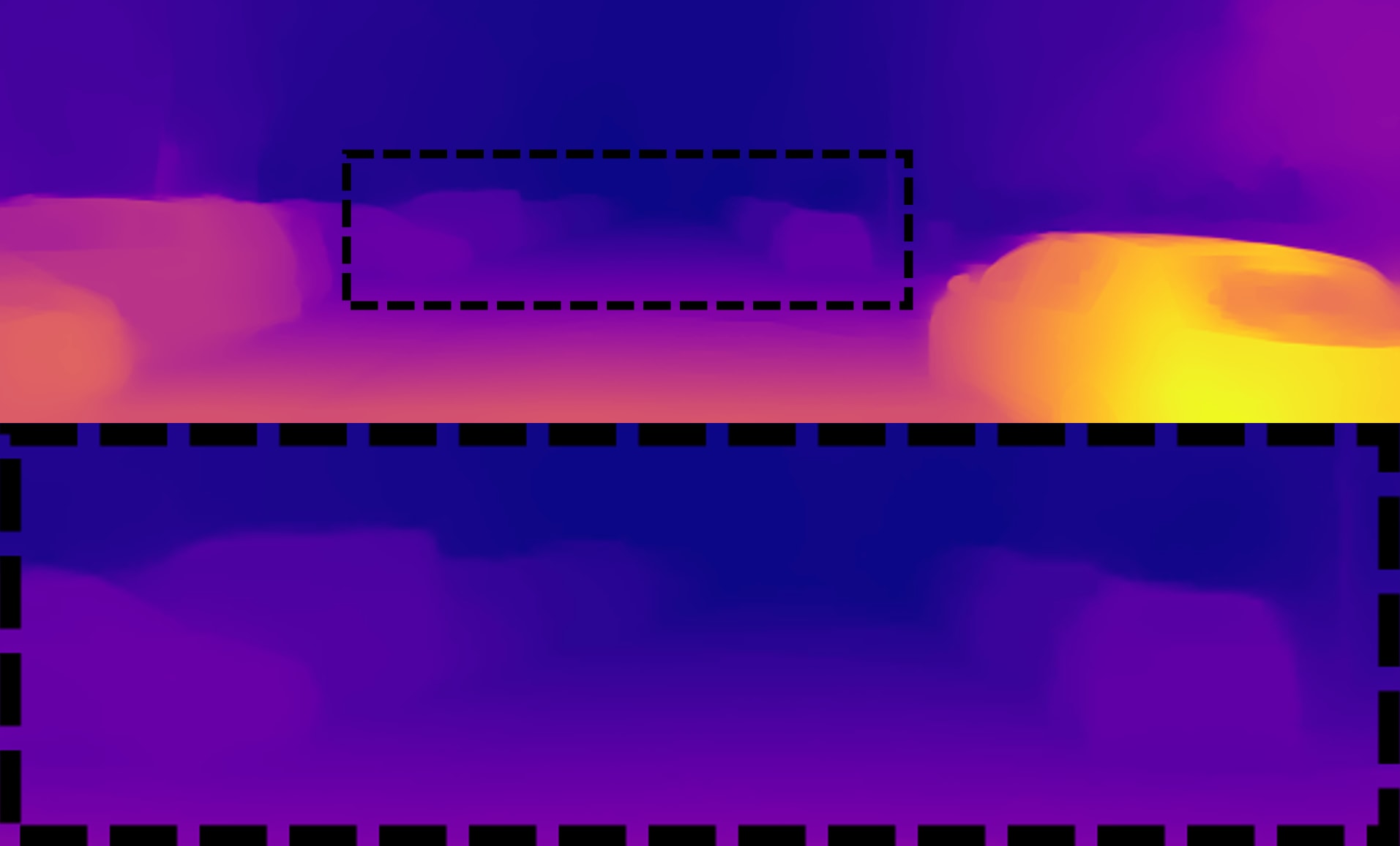}
         \hspace{1em}
         \includegraphics[width=\textwidth]{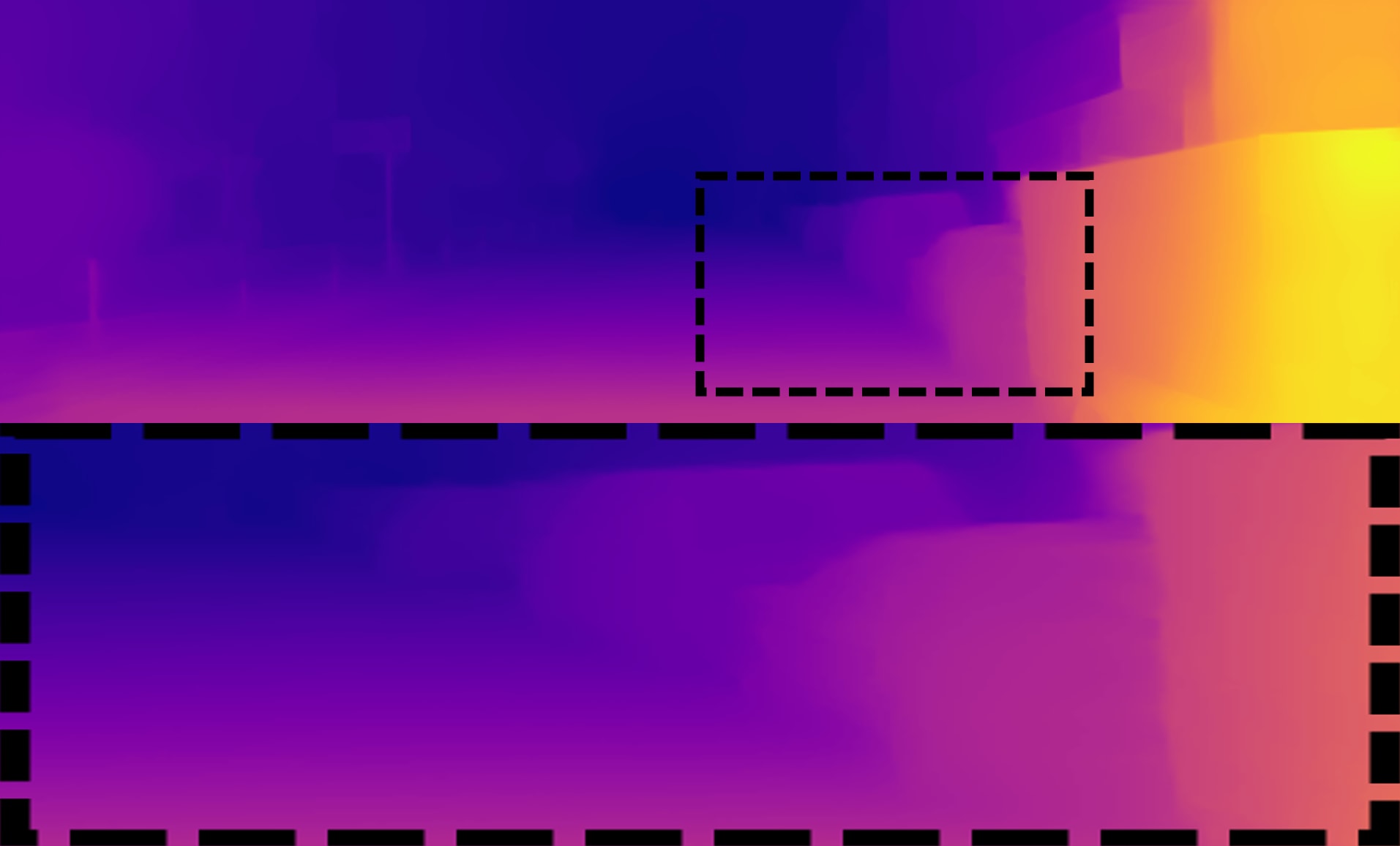}
         \label{kittisub1:densed}
     \end{subfigure}
          \begin{subfigure}[t]{0.2\textwidth}
         \centering
         \caption{DSN$\dagger$}
         \includegraphics[width=\textwidth]{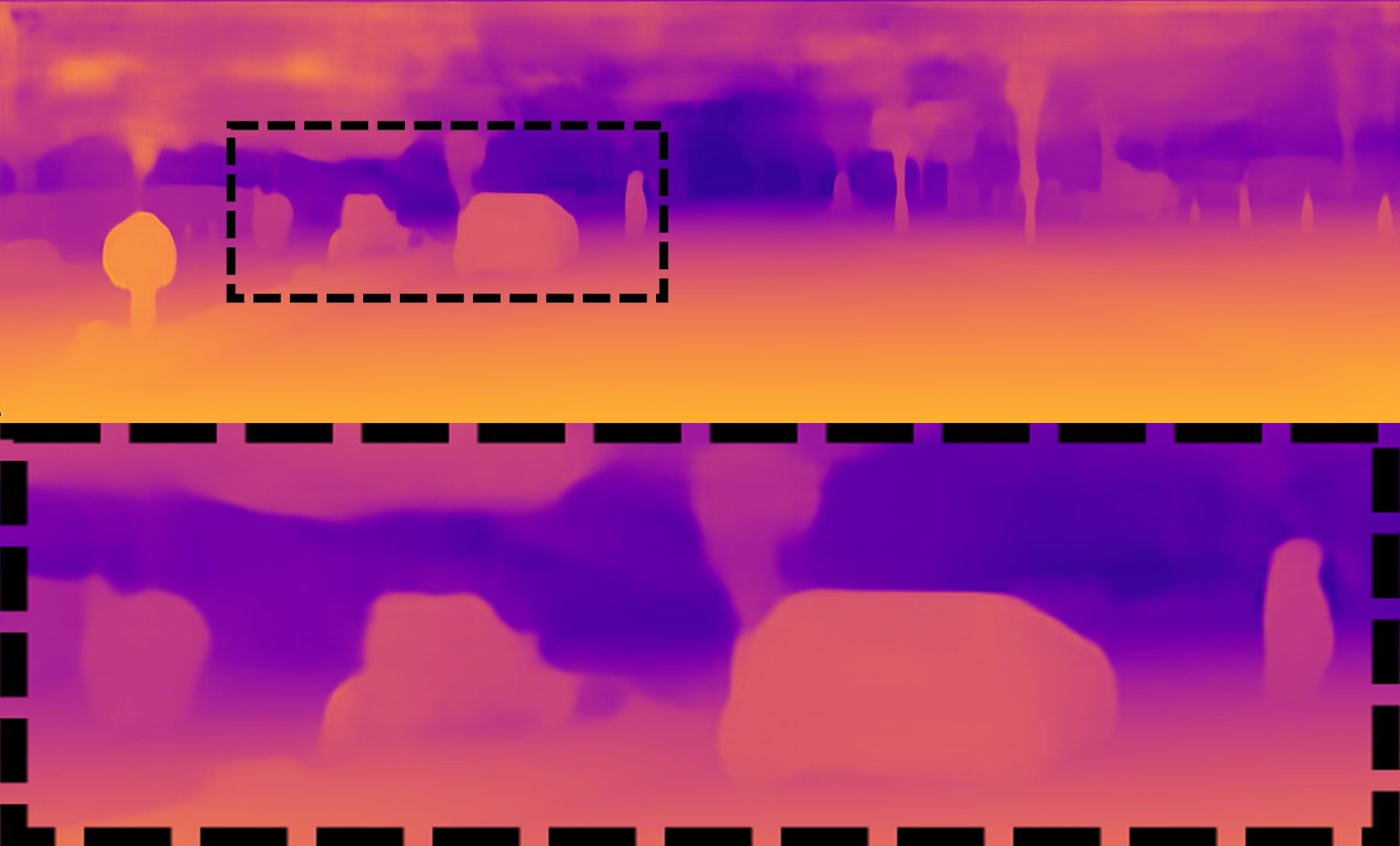}
         \hspace{1em}
         \includegraphics[width=\textwidth]{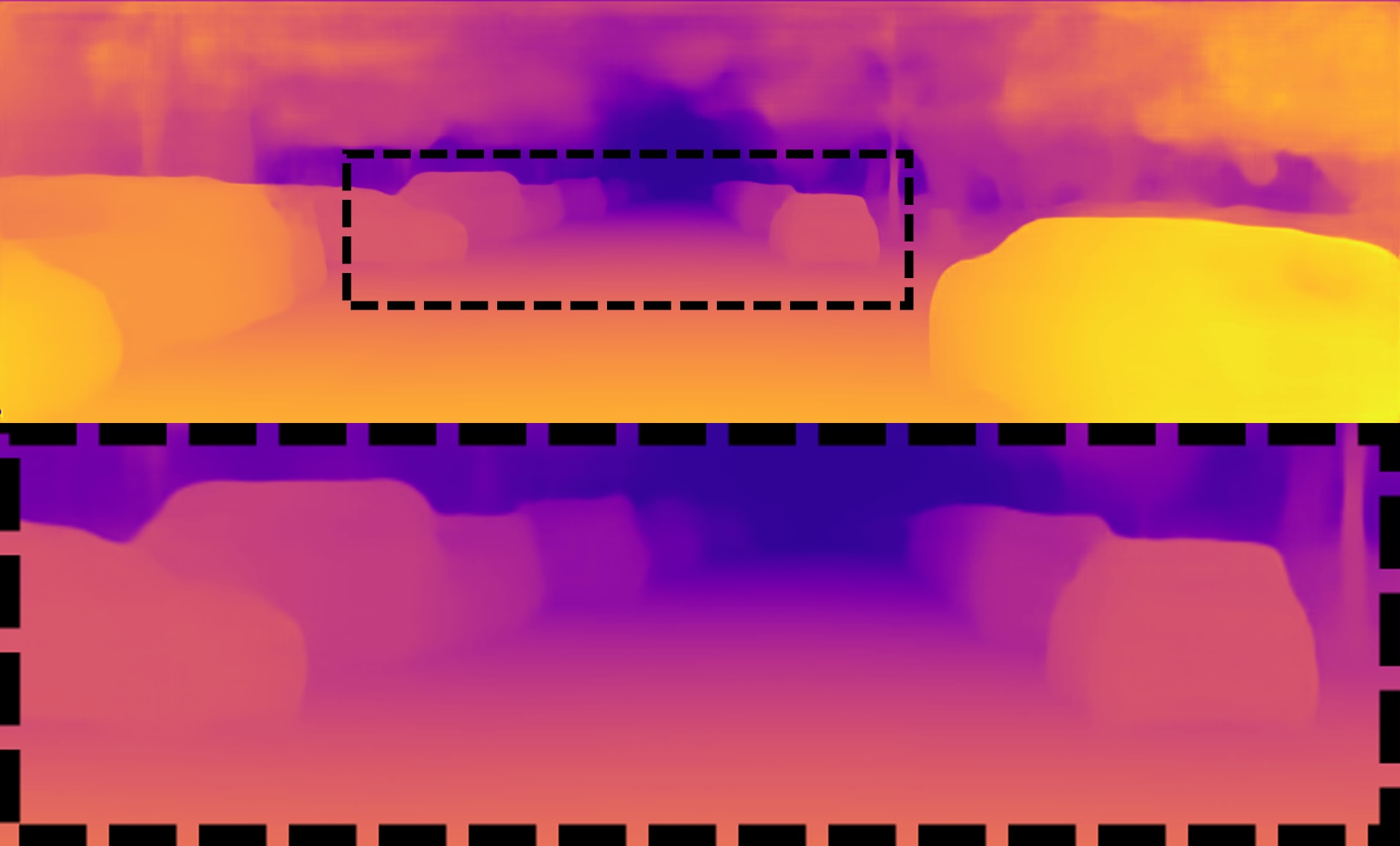}
         \hspace{1em}
         \includegraphics[width=\textwidth]{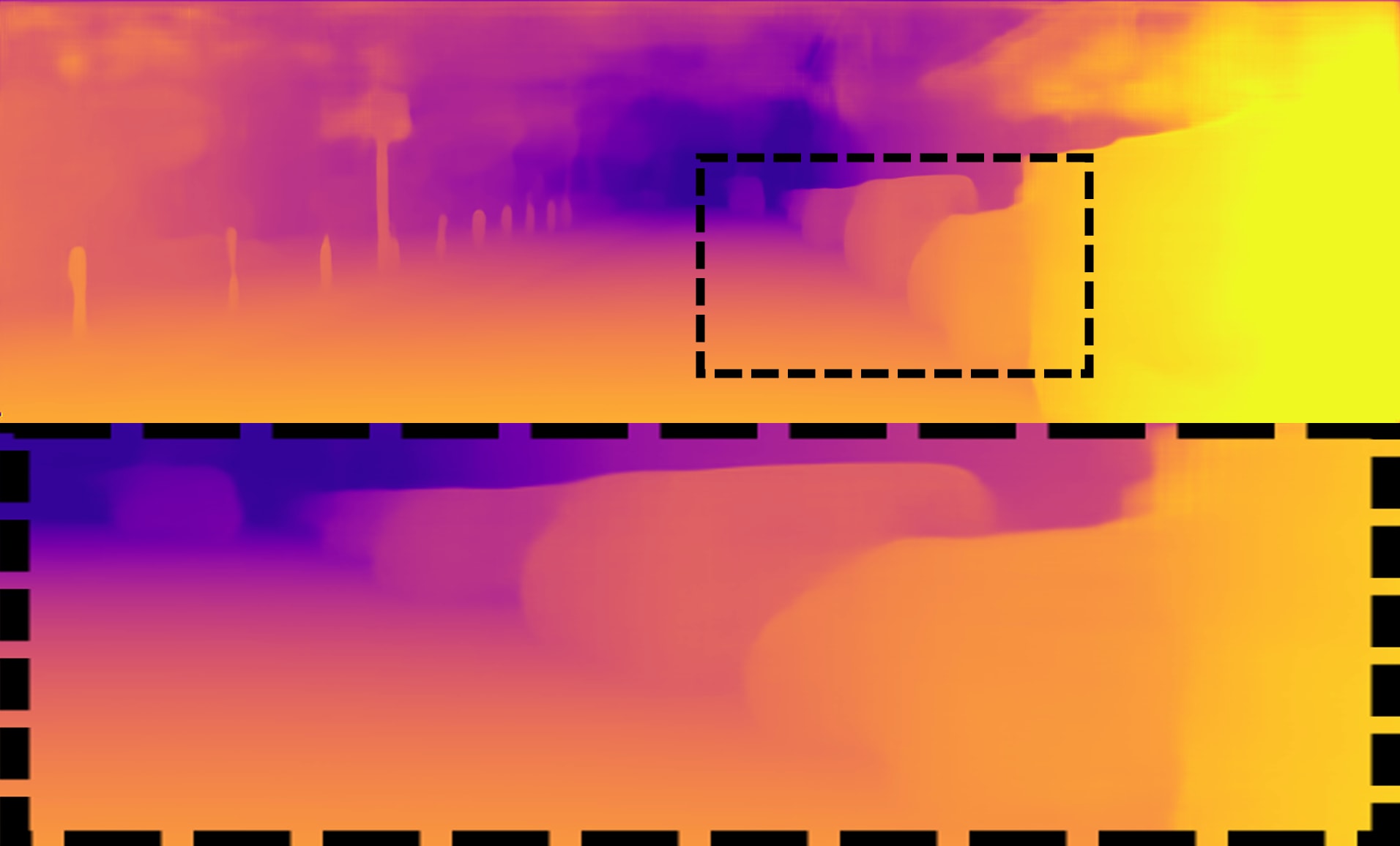}
         \label{kittisub1:dsn}
     \end{subfigure}
     \caption{Differences between the depth maps retrieved by the best DSN configuration (DSN$\dagger$) and by SIDE approaches in the state-of-the-art with a supervised training pipeline. The input RGB images are part of the KITTI Depth dataset \cite{uhrig2017sparsity}}
     \label{kittisub1:all}
\end{figure*}

\begin{figure*}[t!]
\centering
     \begin{subfigure}[t]{0.19\textwidth}
         \centering
         \caption{RGB Input}
         \includegraphics[width=\textwidth]{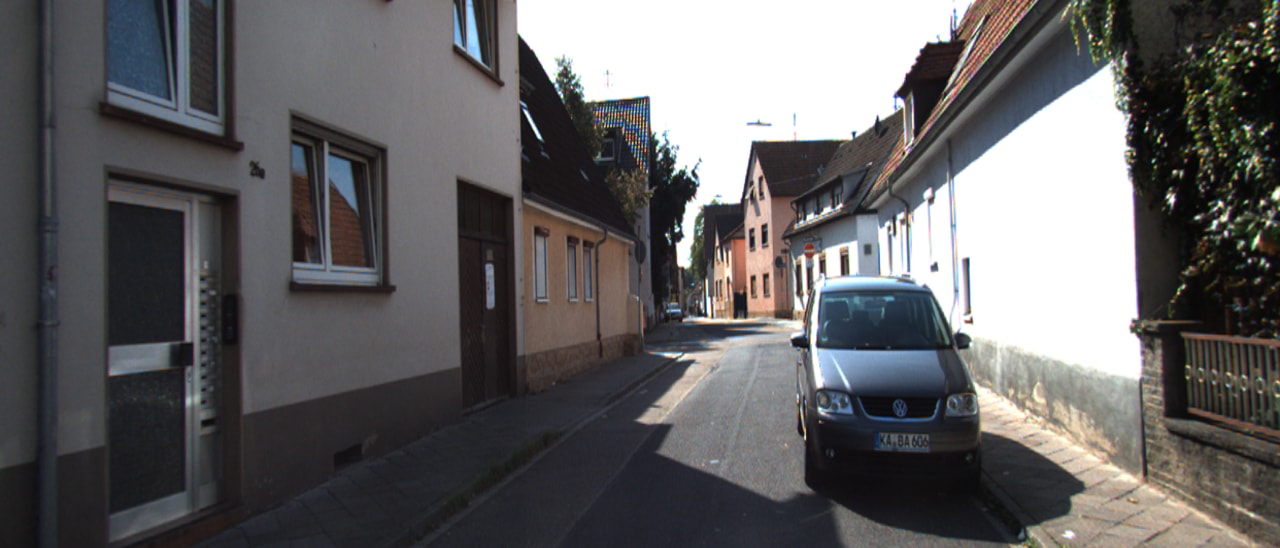}
         \hspace{1em}
         \includegraphics[width=\textwidth]{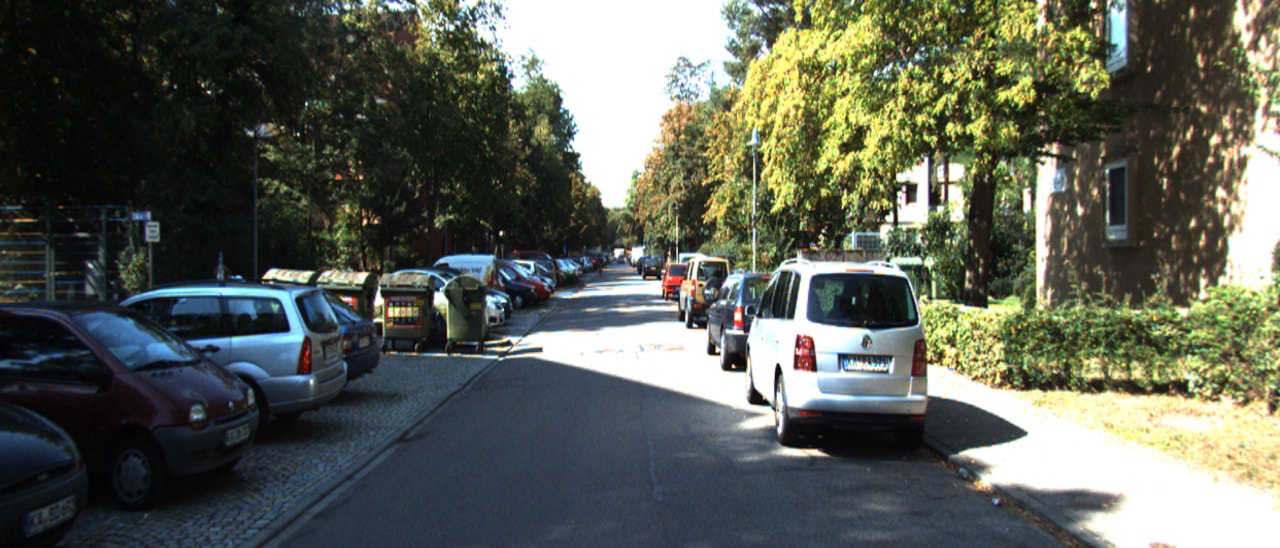}
         \hspace{1em}
         \includegraphics[width=\textwidth]{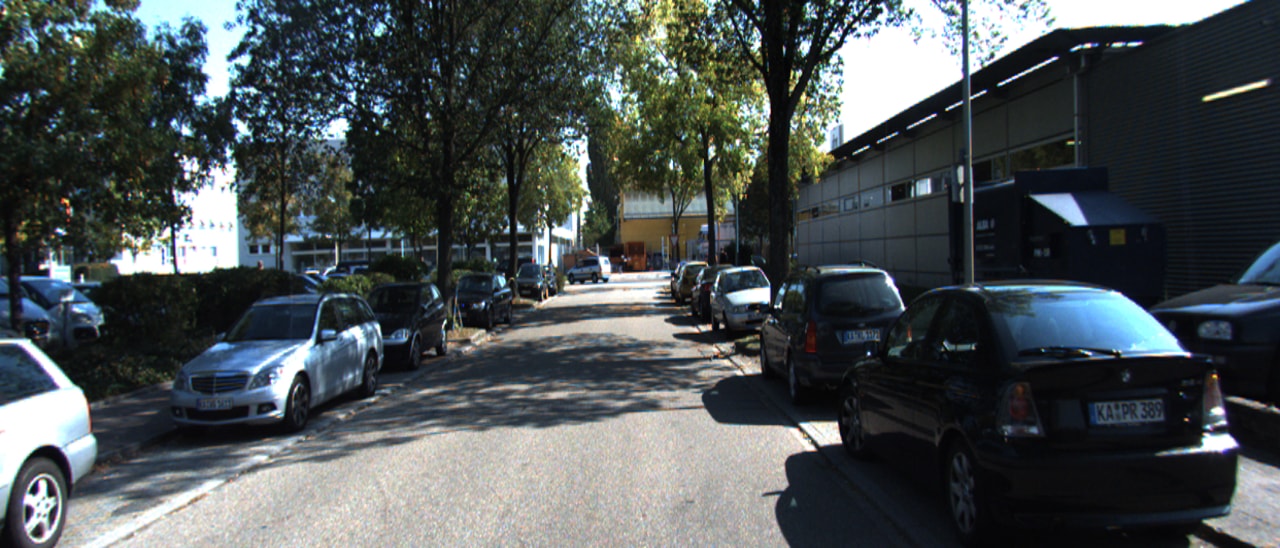}
         \label{kittisubfig3d:0}
     \end{subfigure}
     \begin{subfigure}[t]{0.19\textwidth}
         \centering
         \caption{Ground Truth}
         \includegraphics[width=\textwidth]{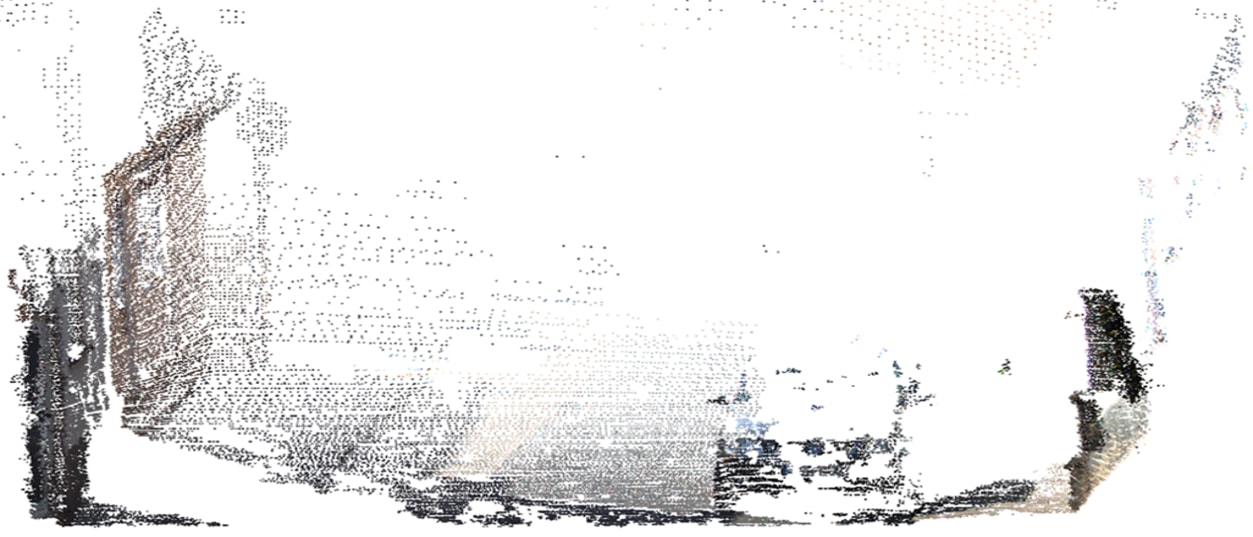}
         \hspace{1em}
         \includegraphics[width=\textwidth]{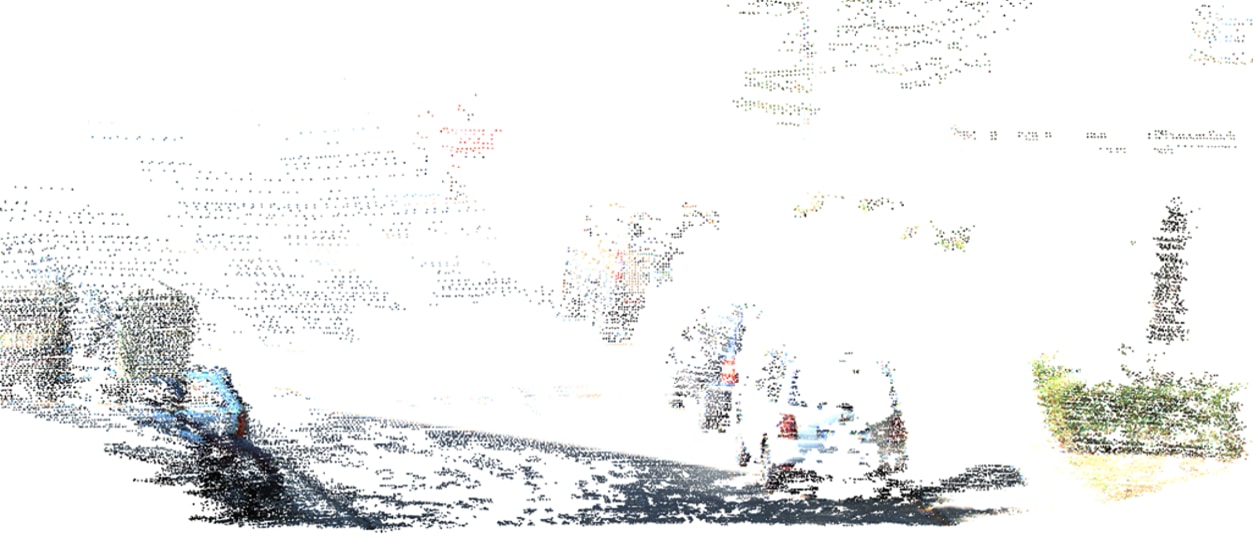}
         \hspace{1em}
         \includegraphics[width=\textwidth]{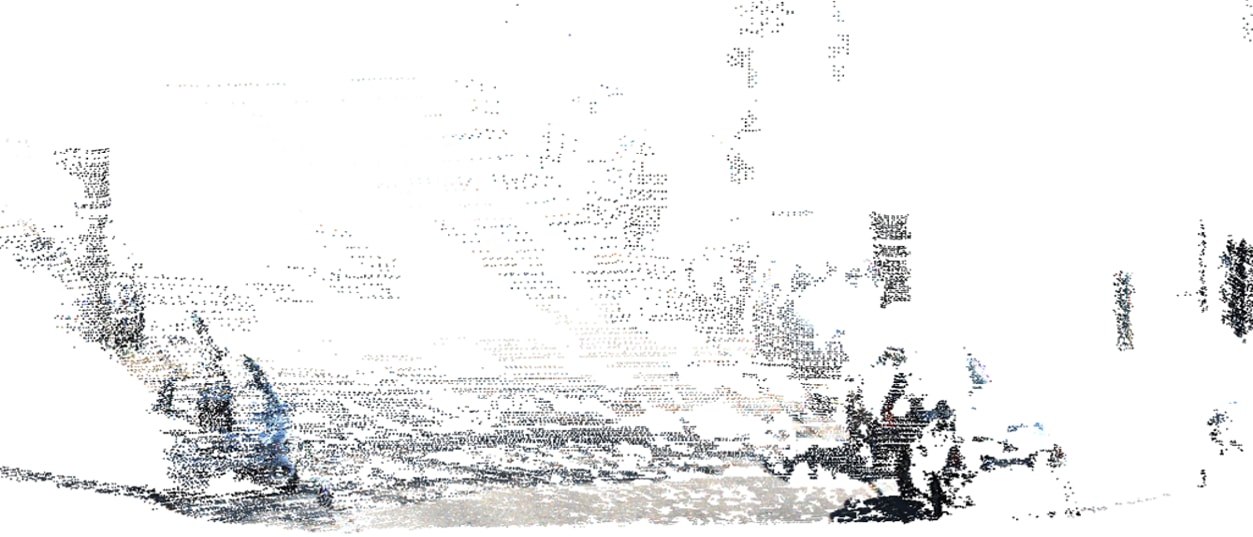}
         \label{kittisubfig3d:1}
     \end{subfigure}
         \begin{subfigure}[t]{0.19\textwidth}
         \centering
         \caption{DSN$\dagger$}
         \includegraphics[width=\textwidth]{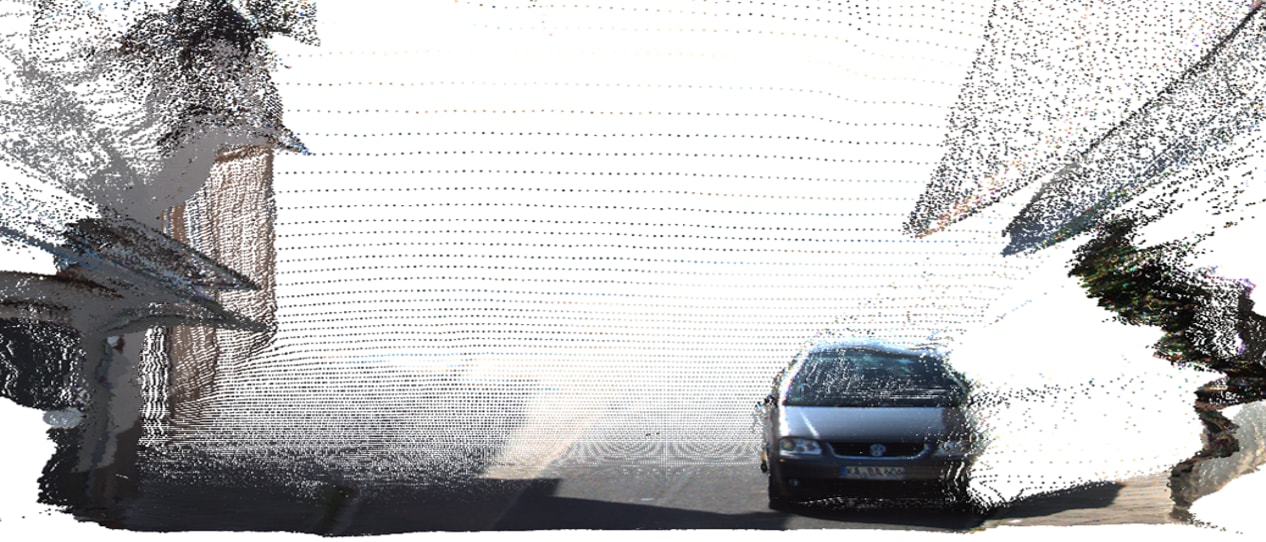}
         \hspace{1em}
         \includegraphics[width=\textwidth]{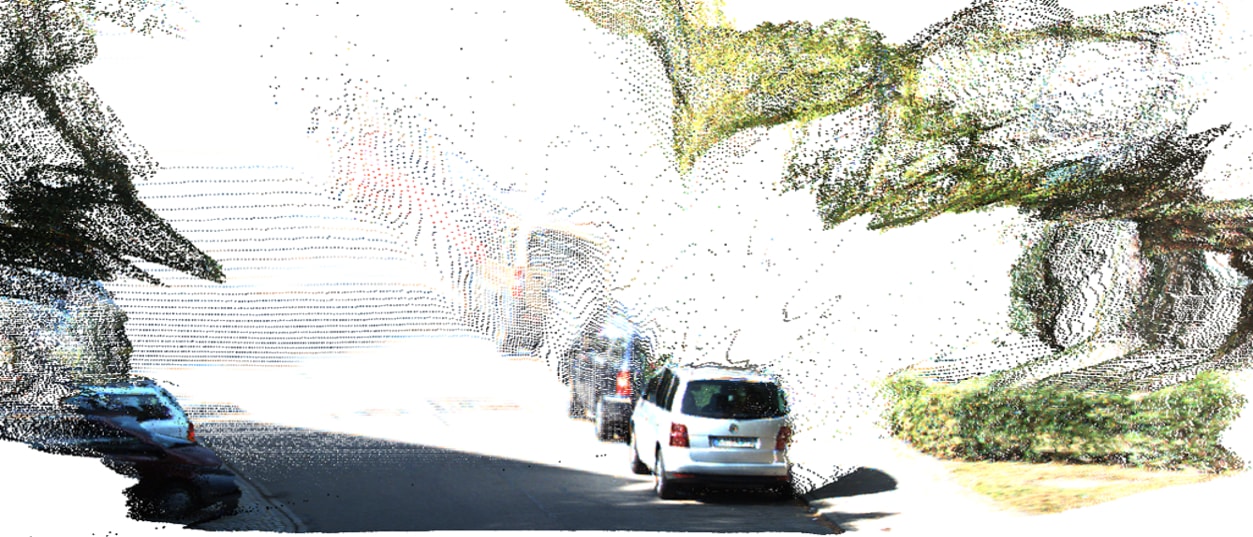}
         \hspace{1em}
         \includegraphics[width=\textwidth]{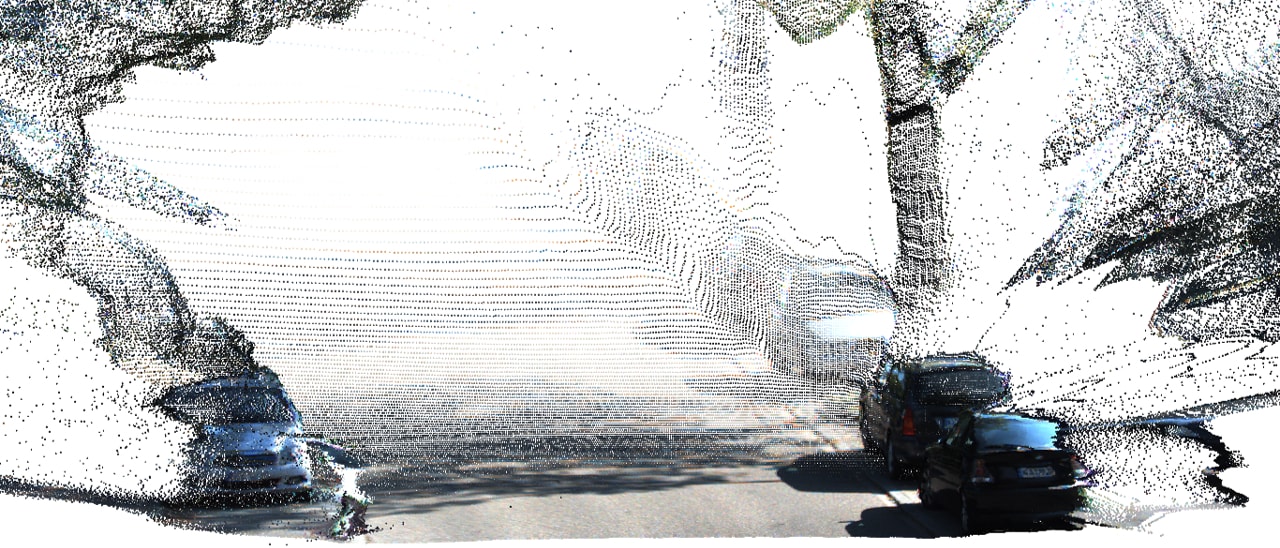}
         \label{kittisubfig3d:2}
     \end{subfigure}
     \begin{subfigure}[t]{0.19\textwidth}
         \centering
         \caption{Ground Truth}
         \includegraphics[width=\textwidth]{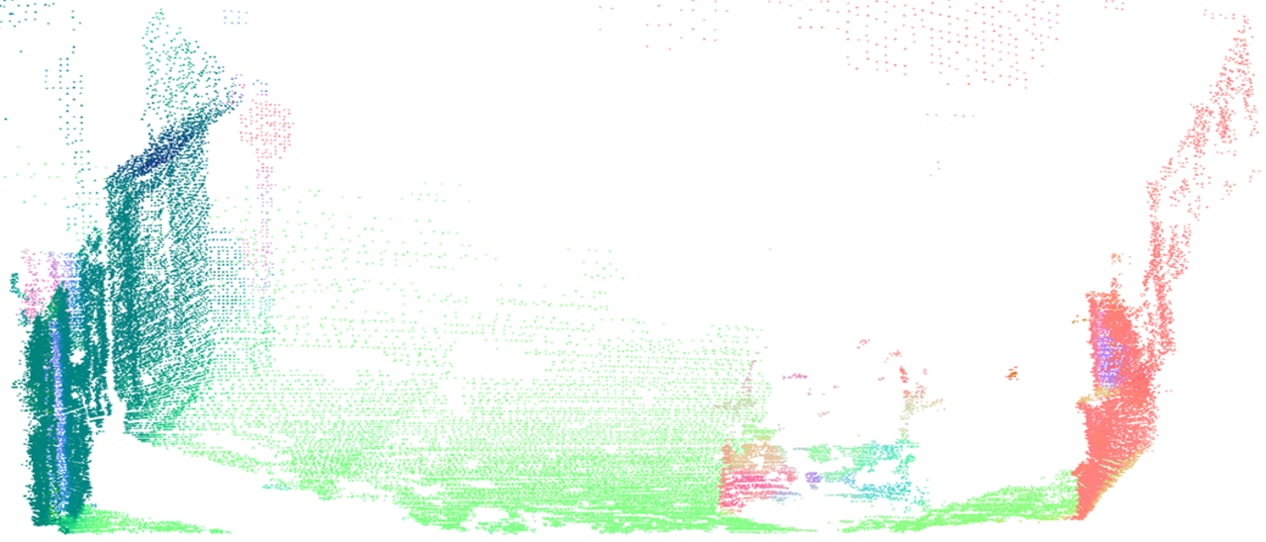}
         \hspace{1em}
         \includegraphics[width=\textwidth]{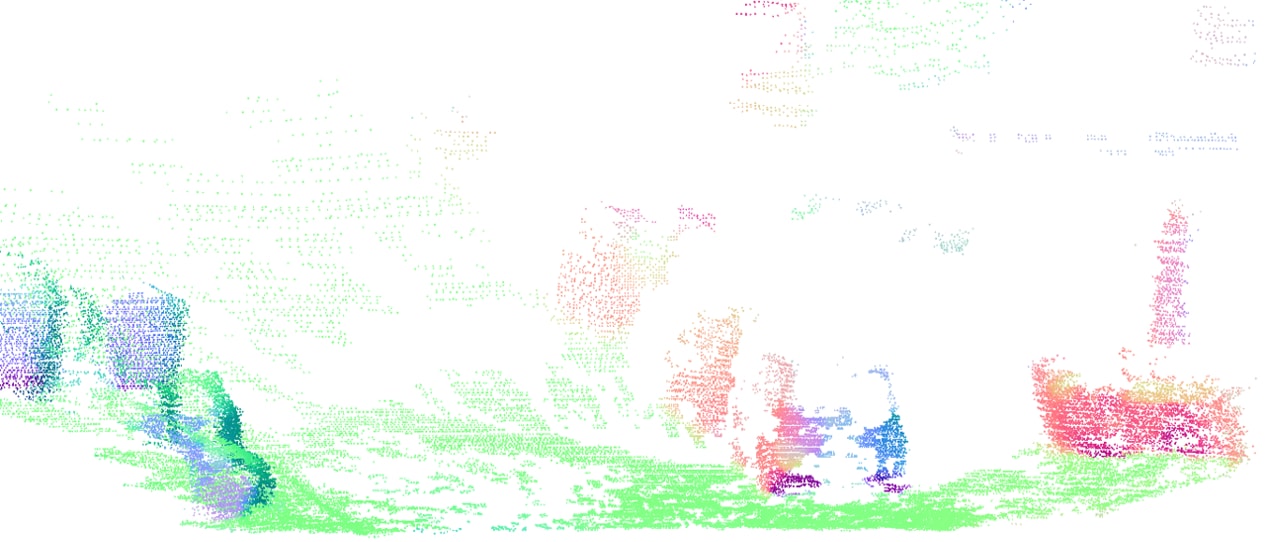}
         \hspace{1em}
         \includegraphics[width=\textwidth]{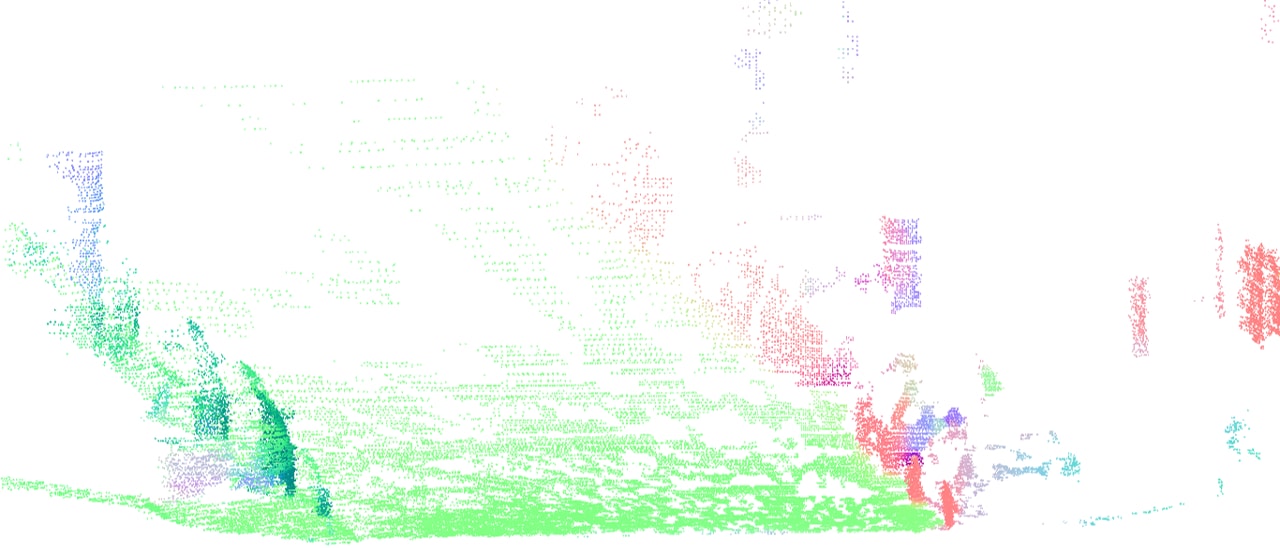}
         \label{kittisubfig3d:3}
     \end{subfigure}
          \begin{subfigure}[t]{0.19\textwidth}
         \centering
         \caption{DSN$\dagger$}
         \includegraphics[width=\textwidth]{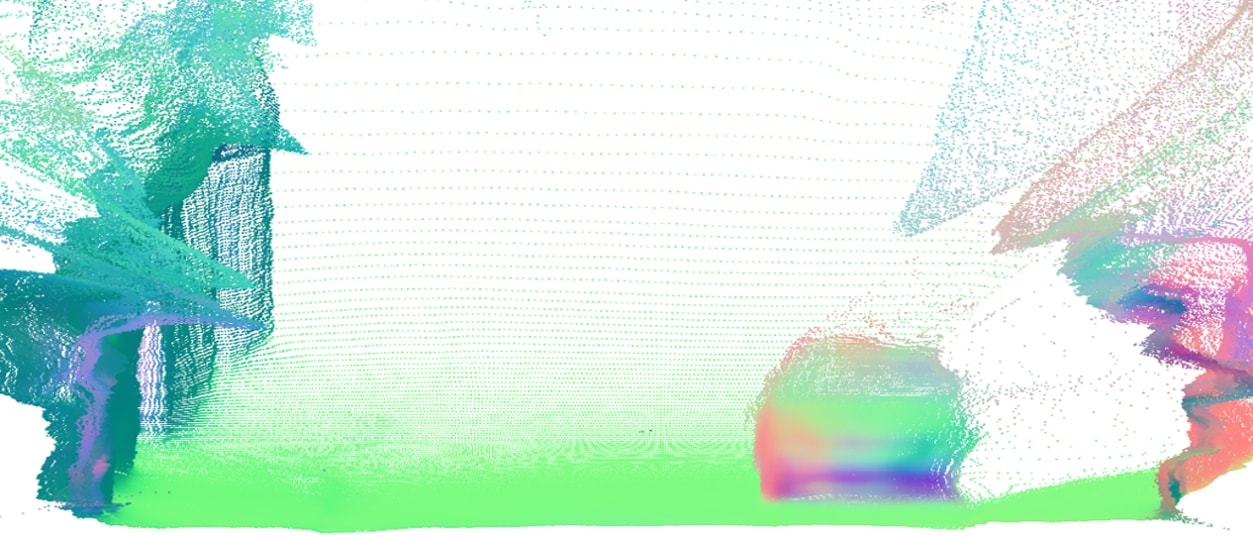}
         \hspace{1em}
         \includegraphics[width=\textwidth]{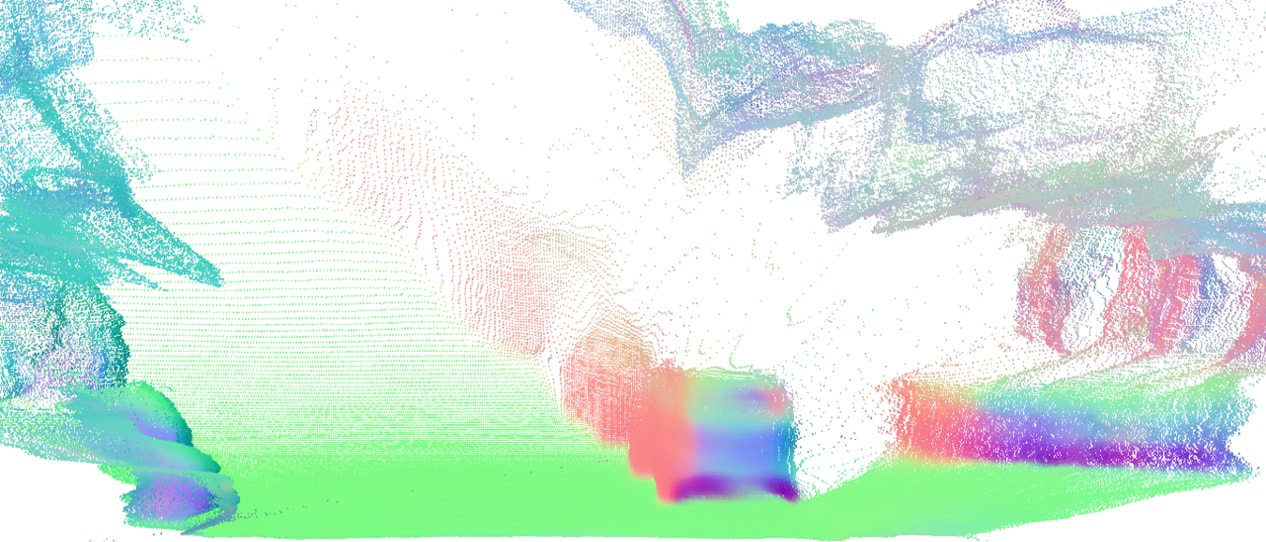}
         \hspace{1em}
         \includegraphics[width=\textwidth]{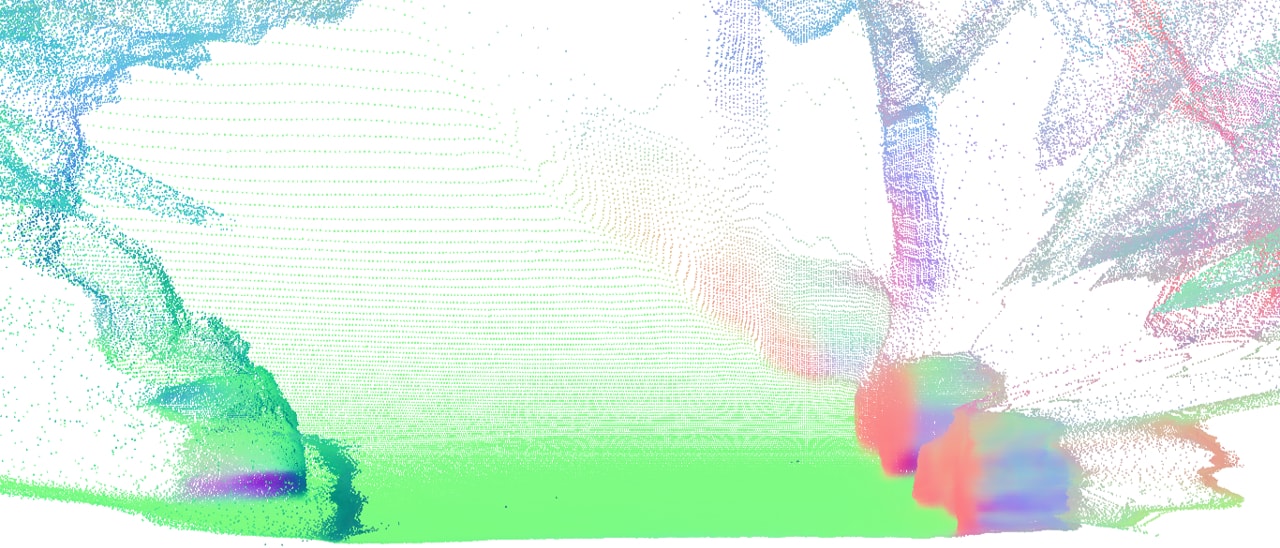}
         \label{kittisubfig3d:4}
     \end{subfigure}
     \caption{Qualitative analysis of the point clouds reconstructed from the DSN predictions. DSN$\dagger$ indicates the DSN setup that leads to the most accurate overall metrics. The RGB input and the reference point clouds with depth data are taken from the KITTI Depth dataset \cite{uhrig2017sparsity}. However, the reference surface normals information are generated by the authors with the aid of the algorithm introduced by Fouhey \etal\cite{fouhey2013data}}
     \label{kittisubfig3d:all}
\end{figure*}

On the other hand, Fig. \ref{kittisub1:all} compares the predictions from our approach with the ones produced by recent supervised frameworks. As can be seen in such a figure, the DSN outputs high-definition depth maps with fewer blurred regions under the sky limit. Through Fig. \ref{kittisubfig3d:all}, it is possible to verify that the 3D point clouds reconstructed from the predictions of the DSN have a denser distribution than the point clouds created with the sparse and noisy ground truth. The 3D maps from Fig. \ref{kittisubfig3d:all} also show that our framework is robust to data noise and it is capable of modeling the shape of the objects with few errors.

\subsubsection{Depth Completion Studies}

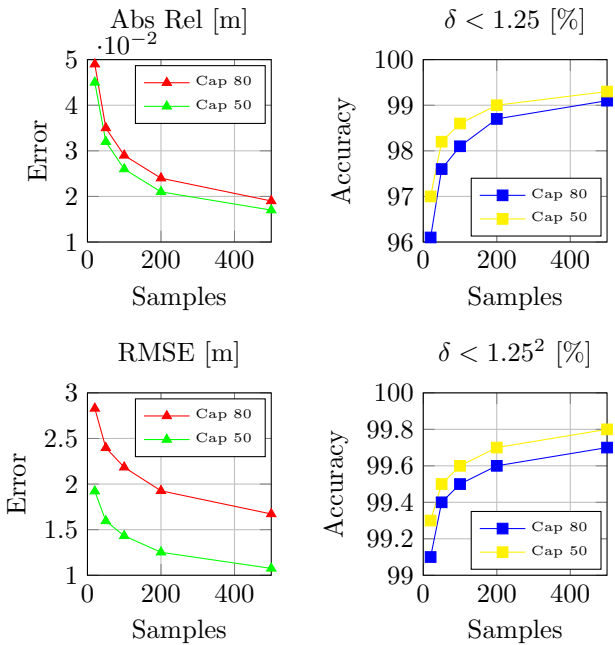
\begin{figure}[!htb]
\centering
\begin{tikzpicture}
  \begin{axis}[title={Abs Rel [m]},
    ylabel near ticks,
    xlabel near ticks,
    xlabel={Samples},
    ylabel={Error},
    xmin=0, xmax=500,
    ymin=0.01, ymax=0.05,
    legend pos=north east,
    ymajorgrids=true,
    ymajorgrids=true,
    grid=major,
    name=absrel,
    height=4cm,
    width=4cm]
    \addplot[color=red,mark=triangle*] coordinates{(20,0.049)(50,0.035) (100,0.029)(200,0.024)(500,0.019)};
    \addlegendentry{\tiny Cap 80}
    \addplot[color=green,mark=triangle*] coordinates{(20,0.045)(50,0.032) (100,0.026)(200,0.021)(500,0.017)};
    \addlegendentry{\tiny Cap 50}
  \end{axis}
  \begin{axis}[title={$\delta<1.25$ $[\%]$},
    ylabel near ticks,
    xlabel near ticks,
    xlabel={Samples},
    ylabel={Accuracy},
    xmin=0, xmax=500,
    ymin=96.0, ymax=100.0,
    legend pos=south east,
    ymajorgrids=true,
    ymajorgrids=true,
    grid=major,
    name=sigma1,
    at={($(absrel.east)+(2.0cm,0)$)},
    anchor=west,
    height=4cm,
    width=4cm]
    \addplot[color=blue,mark=square*] coordinates{(20,96.1)(50,97.6) (100,98.1)(200,98.7)(500,99.1)};
    \addlegendentry{\tiny Cap 80}
    \addplot[color=yellow,mark=square*] coordinates{(20,97.0)(50,98.2) (100,98.6)(200,99.0)(500,99.3)};
    \addlegendentry{\tiny Cap 50}
  \end{axis}
  \begin{axis}[title={RMSE [m]},
    ylabel near ticks,
    xlabel near ticks,
    xlabel={Samples},
    ylabel={Error},
    xmin=0, xmax=500,
    ymin=1.0, ymax=3.0,
    legend pos=north east,
    ymajorgrids=true,
    ymajorgrids=true,
    grid=major,
    name=rmse,
    at={($(absrel.south)-(0,2cm)$)},
    anchor=north,
    height=4cm,
    width=4cm]
    \addplot[color=red,mark=triangle*] coordinates{(20,2.829)(50,2.397) (100,2.184)(200,1.926)(500,1.672)};
    \addlegendentry{\tiny Cap 80}
    \addplot[color=green,mark=triangle*] coordinates{(20,1.922)(50,1.595) (100,1.432)(200,1.252)(500,1.075)};
    \addlegendentry{\tiny Cap 50}
  \end{axis}
  \begin{axis}[title={$\delta<1.25^2$ $[\%]$},
    ylabel near ticks,
    xlabel near ticks,
    xlabel={Samples},
    ylabel={Accuracy},
    xmin=0, xmax=500,
    ymin=99.0, ymax=100,
    legend pos=south east,
    ymajorgrids=true,
    ymajorgrids=true,
    grid=major,
    name=sigma2,
    at={($(sigma1.south)-(0,2cm)$)},
    anchor=north,
    height=4cm,
    width=4cm]
    \addplot[color=blue,mark=square*] coordinates{(20,99.1)(50,99.4) (100,99.5)(200,99.6)(500,99.7)};
    \addlegendentry{\tiny Cap 80}
    \addplot[color=yellow,mark=square*] coordinates{(20,99.3)(50,99.5) (100,99.6)(200,99.7)(500,99.8)};
    \addlegendentry{\tiny Cap 50}
  \end{axis}
\end{tikzpicture}
\caption{Errors and accuracy evolution of the DSN predictions according to the number of input sparse samples. We used the ground truth of the KITTI Depth dataset \cite{uhrig2017sparsity} to present the results}
\label{graph:kittilidar}
\end{figure}

To study the performance of the DSN on depth completion applications, we employed the sampling technique based on the categorical distribution of Eq. \ref{prob1} to feed the CNN with RGBD maps. The results of such experiments, exposed in Table \ref{tab:kittilidar}, enforce the theory that the more LiDAR points are presented to the network input, the more accurate the retrieved depth maps are. This can be visualized in the graphs of Fig. \ref{graph:kittilidar}, in which the error metrics decrease and the precision metrics increase as more sparse depth data is added to the network input.

Based on Table \ref{tab:kittilidar} metrics, we can conclude that the 3D space cues yielded by the surface normals reference maps cause a slightly positive influence in the densification process. Moreover, the best performance of our depth completion pipeline is compared to the results from other approaches in the literature in Table \ref{soa_tab:kittilidar}. 

\begin{figure}[]
     \centering
         \begin{subfigure}[t]{0.11\textwidth}
         \centering
         \caption{Input}
         \includegraphics[width=\textwidth]{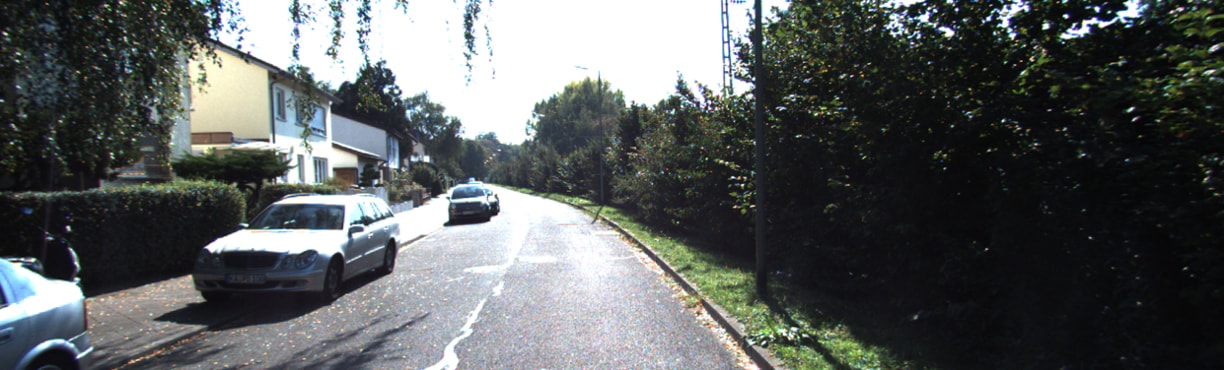}
         \hspace{1em}
         \includegraphics[width=\textwidth]{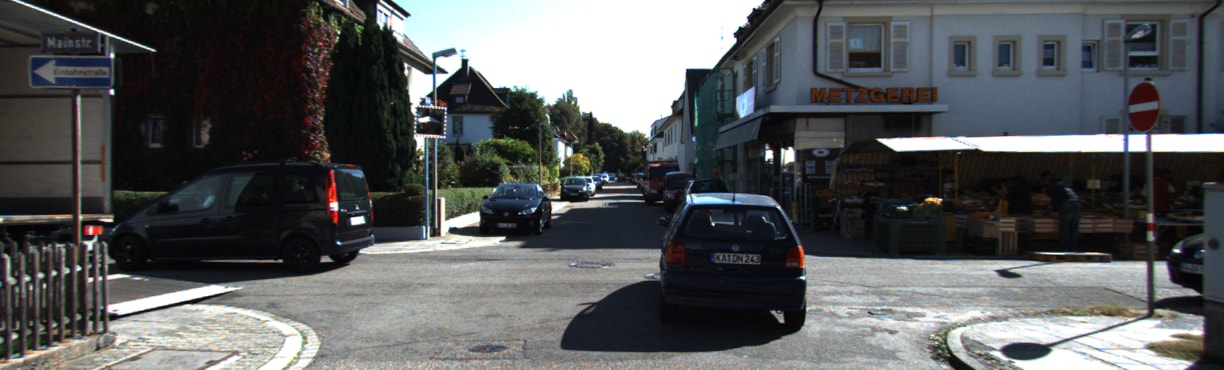}
         \hspace{1em}
         \includegraphics[width=\textwidth]{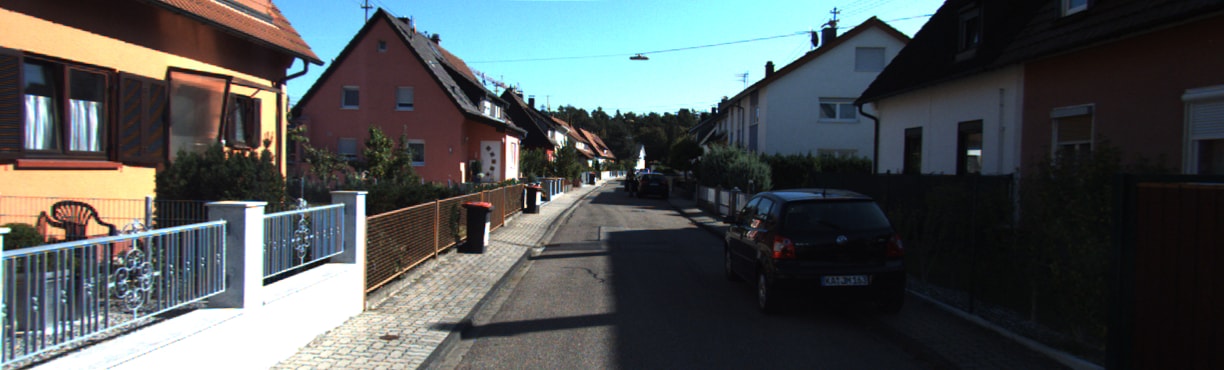}
         \hspace{1em}
         \includegraphics[width=\textwidth]{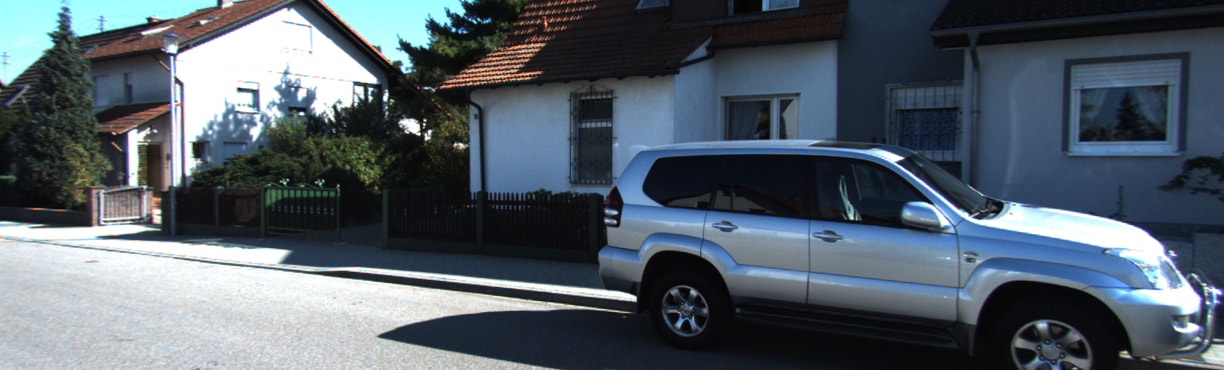}
         \hspace{1em}
         \includegraphics[width=\textwidth]{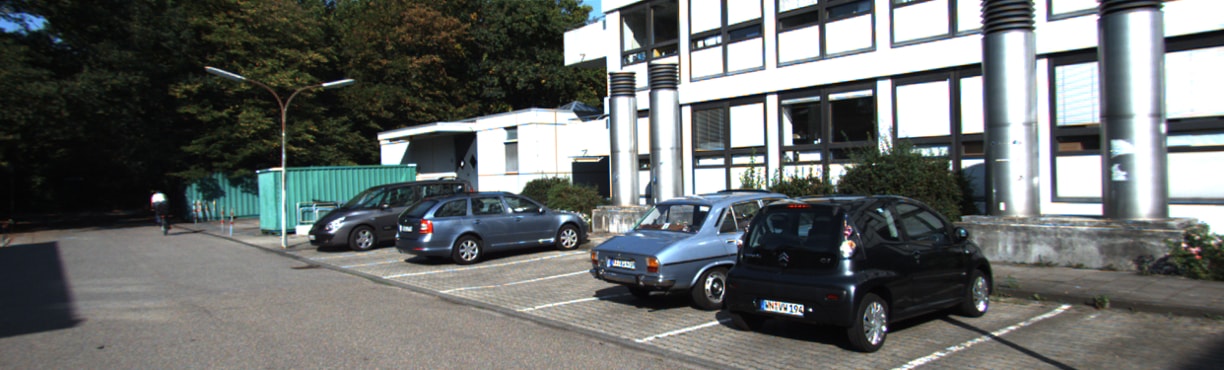}
         \label{kittilidarsubfig0:rgb}
     \end{subfigure}
          \begin{subfigure}[t]{0.11\textwidth}
         \centering
         \caption{GT}
         \includegraphics[width=\textwidth]{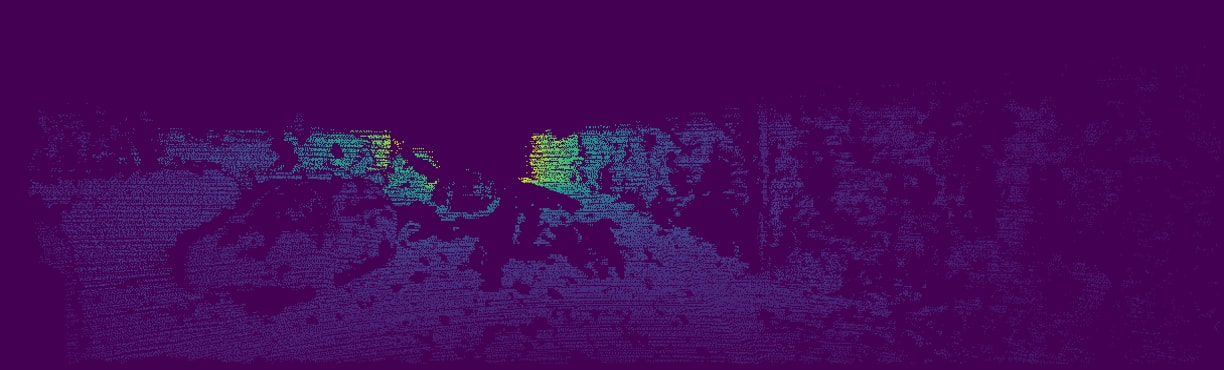}
         \hspace{1em}
         \includegraphics[width=\textwidth]{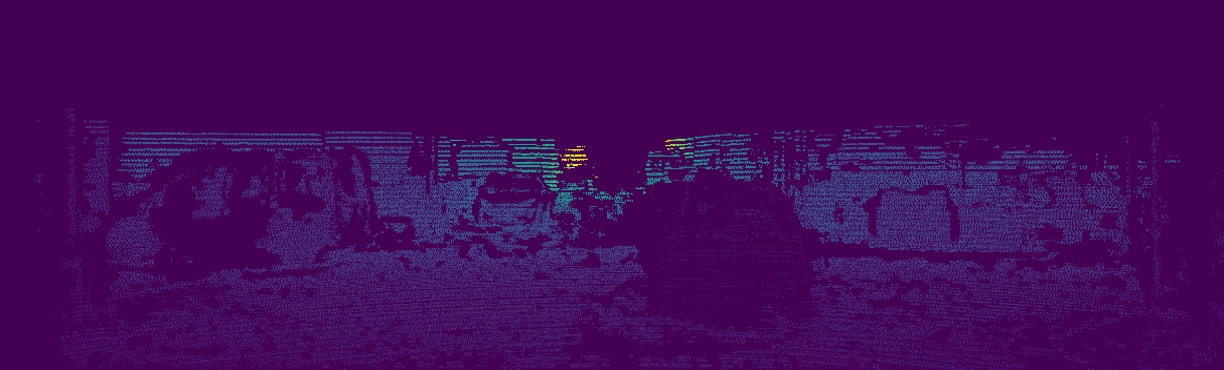}
         \hspace{1em}
         \includegraphics[width=\textwidth]{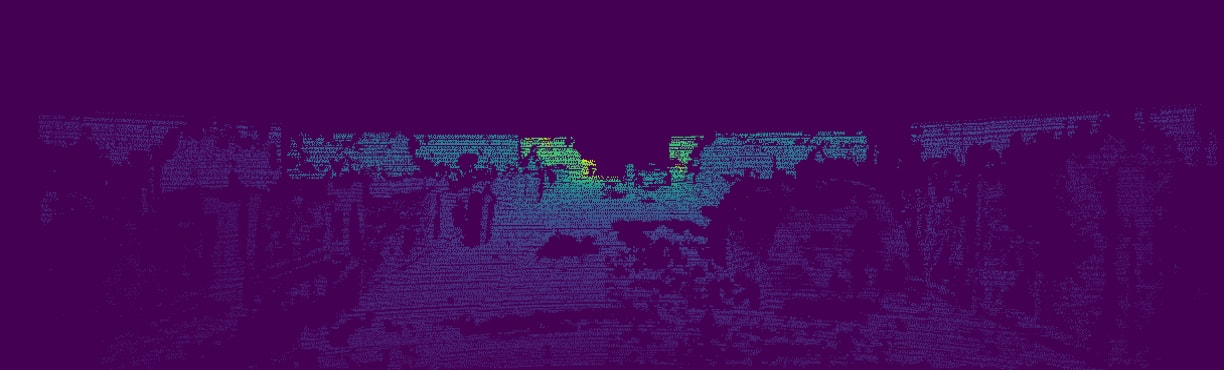}
         \hspace{1em}
         \includegraphics[width=\textwidth]{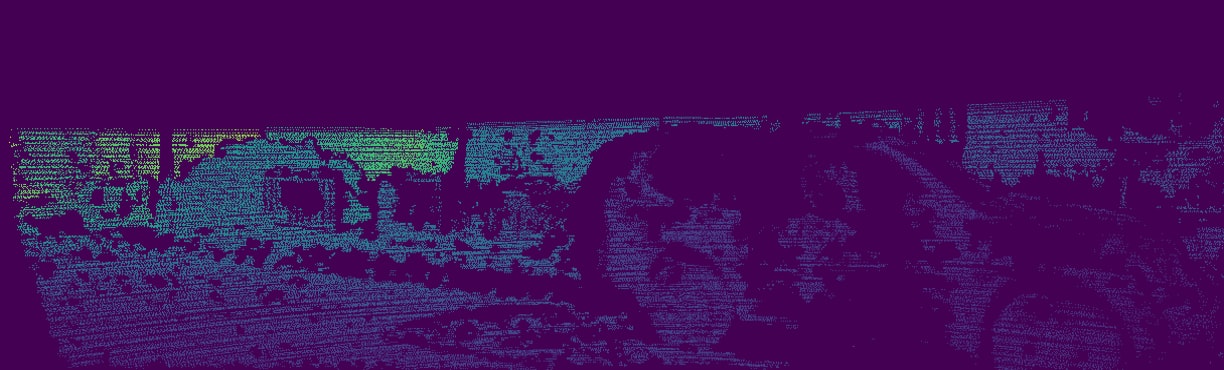}
         \hspace{1em}
         \includegraphics[width=\textwidth]{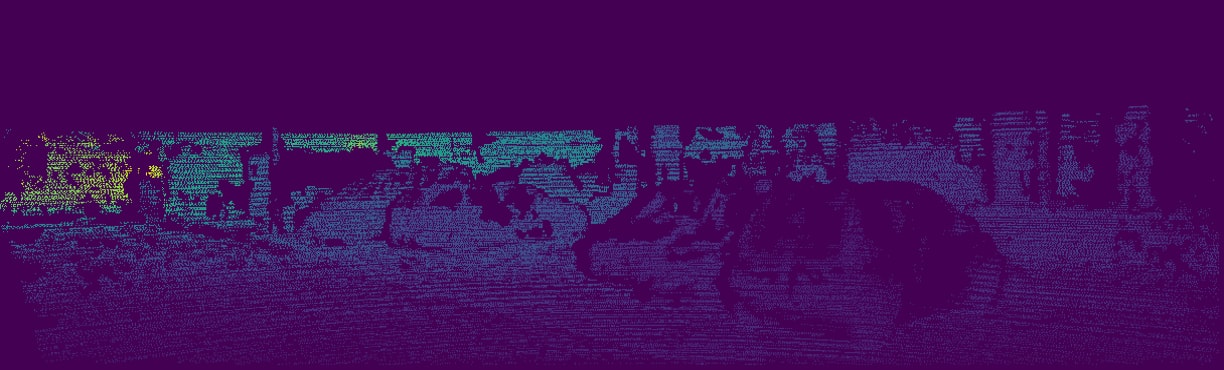}
         \label{kittilidarsubfig0:gt}
     \end{subfigure}
     \begin{subfigure}[t]{0.11\textwidth}
         \centering
         \caption{200d}
         \includegraphics[width=\textwidth]{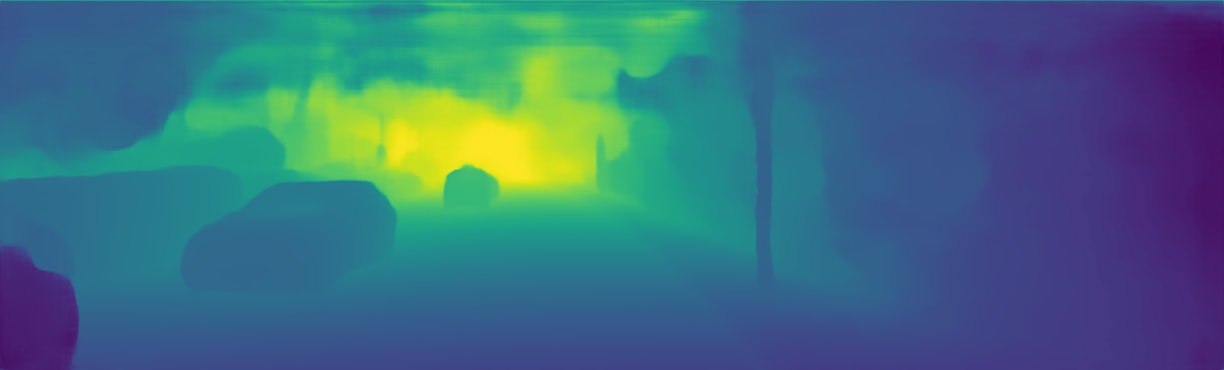}
         \hspace{1em}
         \includegraphics[width=\textwidth]{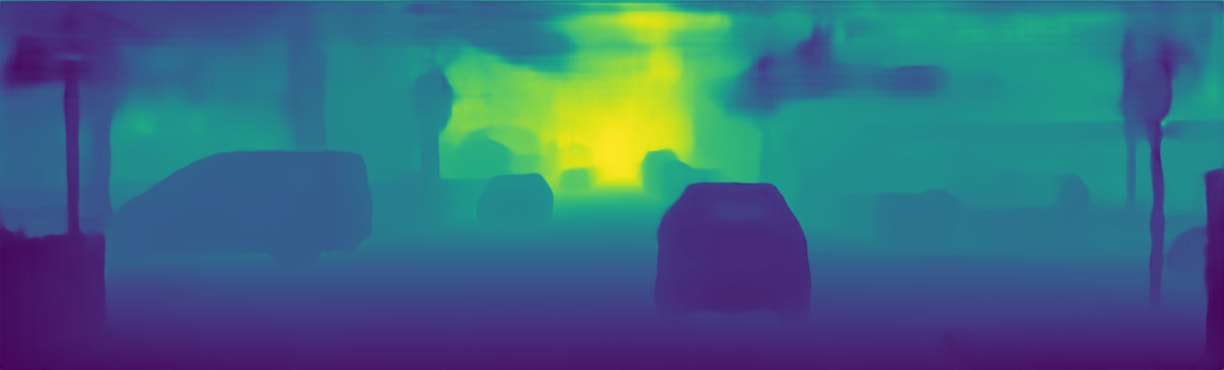}
         \hspace{1em}
         \includegraphics[width=\textwidth]{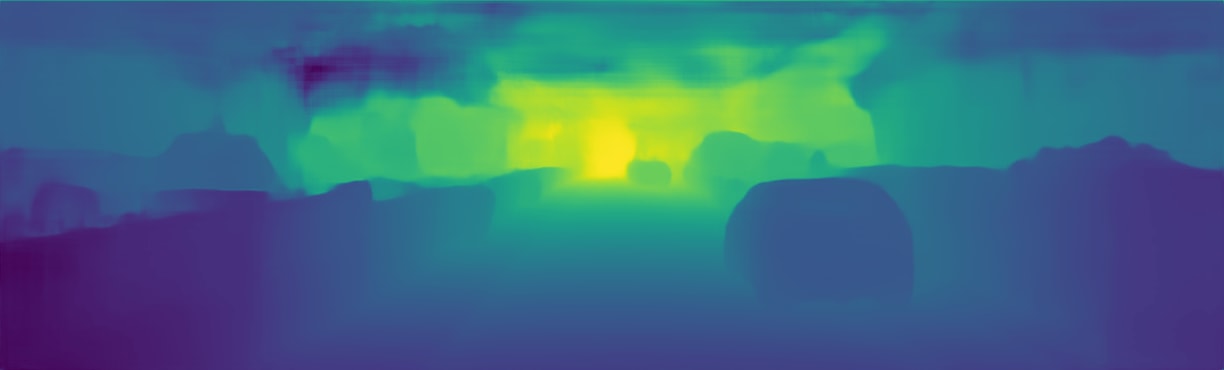}
         \hspace{1em}
         \includegraphics[width=\textwidth]{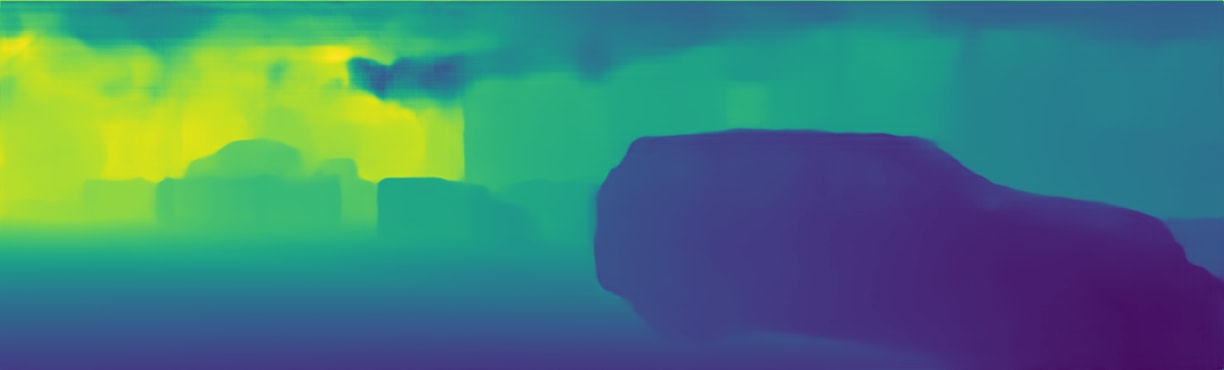}
         \hspace{1em}
         \includegraphics[width=\textwidth]{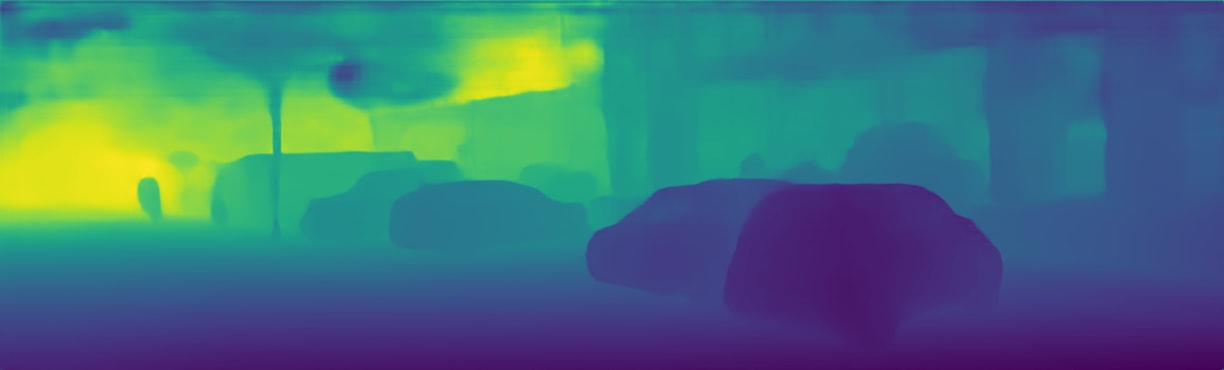}
         \label{kittilidarsubfig0:200}
     \end{subfigure}
         \begin{subfigure}[t]{0.11\textwidth}
         \centering
         \caption{500d}
         \includegraphics[width=\textwidth]{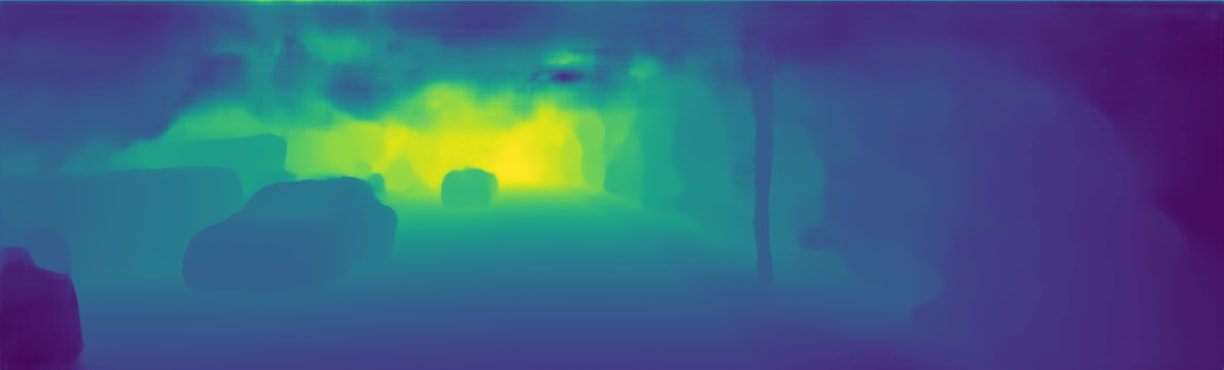}
         \hspace{1em}
         \includegraphics[width=\textwidth]{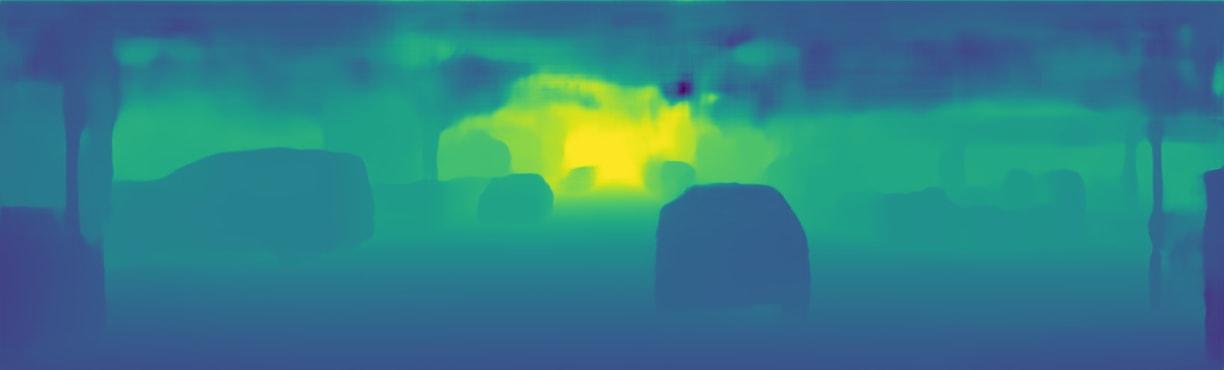}
         \hspace{1em}
         \includegraphics[width=\textwidth]{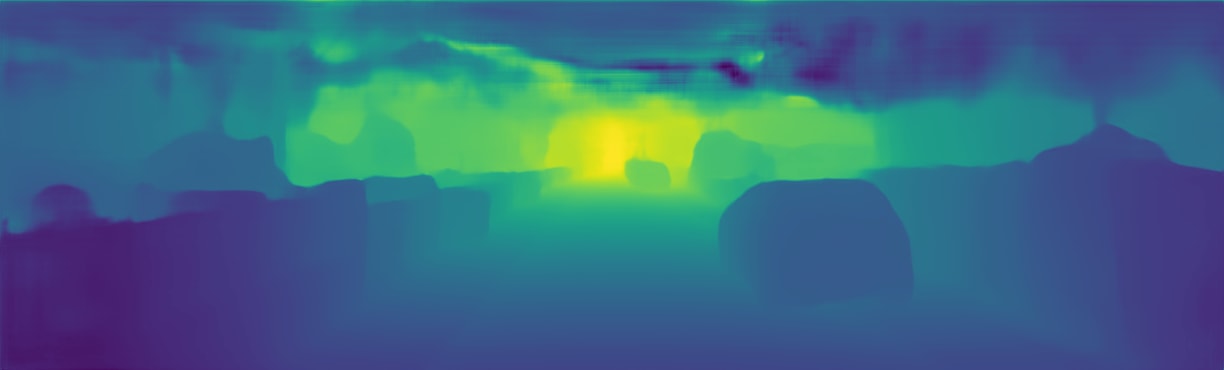}
         \hspace{1em}
         \includegraphics[width=\textwidth]{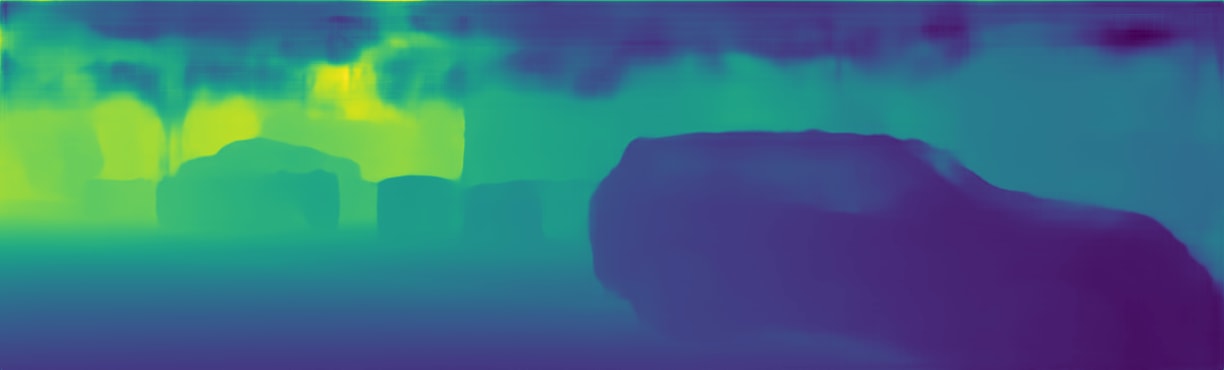}
         \hspace{1em}
         \includegraphics[width=\textwidth]{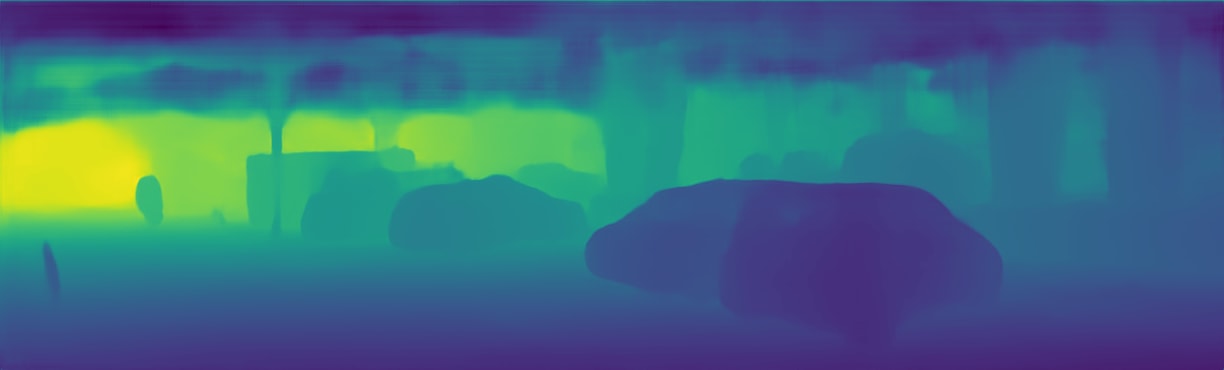}
         \label{kittilidarsubfig0:500}
     \end{subfigure}
     \caption{Qualitative evaluation of our depth completion framework based on the noisy and sparse ground truth of the KITTI Depth dataset \cite{uhrig2017sparsity}. Legend: Input - input RGB image; GT - sparse ground truth; d - sparse input data} 
     \label{kittilidarsubfig0:all}
\end{figure}

Although our CNN is capable of densifying depth maps at a high frame rate, it also generates dense maps with a quality that surpasses that of other state-of-the-art works. Our most accurate depth completion results are also depicted in Fig. \ref{kittilidarsubfig0:all}, in which both the experiments regarding $200$ and $500$ sparse depth data, fed into the DSN, achieve high-quality dense depth maps.

\begin{table}[]
\centering
\caption{Evaluation of our depth completion pipeline according to Eigen Split applied to the KITTI Depth dataset \cite{uhrig2017sparsity}. The DSN $3$ $+$ Pyramid$^1$ $+$ CS configuration, besides the $\mathcal{L}_{BerHu}^{+}$ function, is taken as default for the first experiments. The DSN $3$ $+$ Pyramid$^1$ $+$ CS $+$ SN setup, along with the edge detector, is analyzed afterward}
\label{tab:kittilidar}
\resizebox{\linewidth}{!}{
\begin{tabular}{c|c|c|ccc|ccc}
\hline
\textbf{Method} & \textbf{Cap} & \textbf{Samples} & \textbf{Abs Rel}$\downarrow$ & \textbf{RMSE}$\downarrow$ & \textbf{log10}$\downarrow$ & $\bm{\delta < 1.25}$$\uparrow$ & $\bm{\delta < 1.25^2}$$\uparrow$ & $\bm{\delta < 1.25^3}$$\uparrow$ \\\hline \hline
DSN 3 $+$ Pyramid$^1$ $+$ CS & 80 & 20 & 0.049 & 2.829 & 0.021 & 0.961 & 0.991 & 0.997 \\
DSN 3 $+$ Pyramid$^1$ $+$ CS & 80 & 50 & 0.035 & 2.397 & 0.015 & 0.976 & 0.994 & 0.998 \\
DSN 3 $+$ Pyramid$^1$ $+$ CS & 80 & 100 & 0.029 & 2.184 & 0.013 & 0.981 & 0.995 & 0.998 \\
DSN 3 $+$ Pyramid$^1$ $+$ CS & 80 & 200 & 0.024 & 1.926 & 0.010 & 0.987 & 0.996 & 0.999 \\
DSN 3 $+$ Pyramid$^1$ $+$ CS & 80 & 500 & \textbf{0.019} & 1.672 & \textbf{0.008} & \textbf{0.991} & \textbf{0.997} & \textbf{0.999} \\
DSN 3 $+$ Pyramid$^1$ $+$ CS $+$ SN & 80 & 200 & 0.024 & 1.863 & 0.010 & 0.987 & 0.996 & 0.999\\
DSN 3 $+$ Pyramid$^1$ $+$ CS $+$ SN & 80 & 500 & \textbf{0.019} & \textbf{1.588} & \textbf{0.008} & \textbf{0.991} & \textbf{0.997} & \textbf{0.999}\\\hline\hline
DSN 3 $+$ Pyramid$^1$ $+$ CS & 50 & 20 & 0.045 & 1.922 & 0.019 & 0.970 & 0.993 & 0.998 \\
DSN 3 $+$ Pyramid$^1$ $+$ CS & 50 & 50 & 0.032 & 1.595 & 0.014 & 0.982 & 0.995 & 0.999 \\
DSN 3 $+$ Pyramid$^1$ $+$ CS & 50 & 100 & 0.026 & 1.432 & 0.011 & 0.986 & 0.996 & 0.999 \\
DSN 3 $+$ Pyramid$^1$ $+$ CS & 50 & 200 & 0.021 & 1.252 & 0.009 & 0.990 & 0.997 & 0.999 \\
DSN 3 $+$ Pyramid$^1$ $+$ CS & 50 & 500 & \textbf{0.017} & 1.075 & \textbf{0.007} & \textbf{0.993} & \textbf{0.998} & \textbf{0.999}\\
DSN 3 $+$ Pyramid$^1$ $+$ CS $+$ SN & 50 & 200 & 0.022 & 1.247 & 0.09 & 0.990 & 0.997 & 0.999\\
DSN 3 $+$ Pyramid$^1$ $+$ CS $+$ SN & 50 & 500 & 0.018 & \textbf{1.051} & 0.008 & \textbf{0.993} & \textbf{0.998} & \textbf{0.999}\\\hline
\end{tabular}
}
\end{table}

\begin{table}[]
\centering
\caption{Quantitative comparison of the results from our framework and the results from other depth completion techniques. The metrics are computed considering the Eigen Split and the KITTI Depth dataset \cite{uhrig2017sparsity}. We apply cap $80$ in all the analysis}
\label{soa_tab:kittilidar}
\resizebox{\linewidth}{!}{
\begin{tabular}{c|c|cc|ccc}
\hline
\textbf{Method} & \textbf{Samples} & \textbf{Abs Rel}$\downarrow$ & \textbf{RMSE}$\downarrow$ & $\bm{\delta < 1.25}$$\uparrow$ & $\bm{\delta < 1.25^2}$$\uparrow$ & $\bm{\delta < 1.25^3}$$\uparrow$ \\\hline \hline
full-MAE \cite{cadena2016multi} & $\sim$650 & 0.179 & 7.14 & 0.709 & 0.888 & 0.956\\
Liao \etal \cite{liao2017parse} & 225 & 0.113 & 4.50 & 0.874 & 0.960 & 0.984 \\
Ma \etal \cite{ma2018sparse} & 200 & 0.083 & 3.851 & 0.919 & 0.970 & 0.986\\
Ma \etal \cite{ma2018sparse} & 500 & 0.073 & 3.378 & 0.935 & 0.976 & 0.989\\
MC $+$ SPN \cite{cheng2018depth} & 500 & 0.059 & 3.248 & 0.944 & 0.977 & 0.989\\
SPN \cite{liu2017learning} & 500 & 0.063 & 3.243 & 0.943 & 0.978 & 0.991\\
Mirror connection (MC) \cite{cheng2018depth} & 500 & 0.051 & 3.049 & 0.953 & 0.979 & 0.990\\
CSPN \cite{cheng2018depth} & 500 & 0.049 & 3.029 & 0.955 & 0.980 & 0.990\\
MC $+$ CSPN \cite{cheng2018depth} & 500 & 0.044 & 2.977 & 0.957 & 0.980 & 0.991\\
MC $+$ CSPN $+$ ACSPF \cite{cheng2018depth} & 500 & 0.042 & 2.843 & 0.961 & 0.982 & 0.992\\
DSN 3 $+$ Pyramid$^1$ $+$ CS $+$ SN & 200 & 0.024 & 1.863 & 0.987 & 0.996 & \textbf{0.999}\\
DSN 3 $+$ Pyramid$^1$ $+$ CS $+$ SN & 500 & \textbf{0.019} & \textbf{1.588} & \textbf{0.991} & \textbf{0.997} & \textbf{0.999}\\\hline
\end{tabular}
}
\end{table}

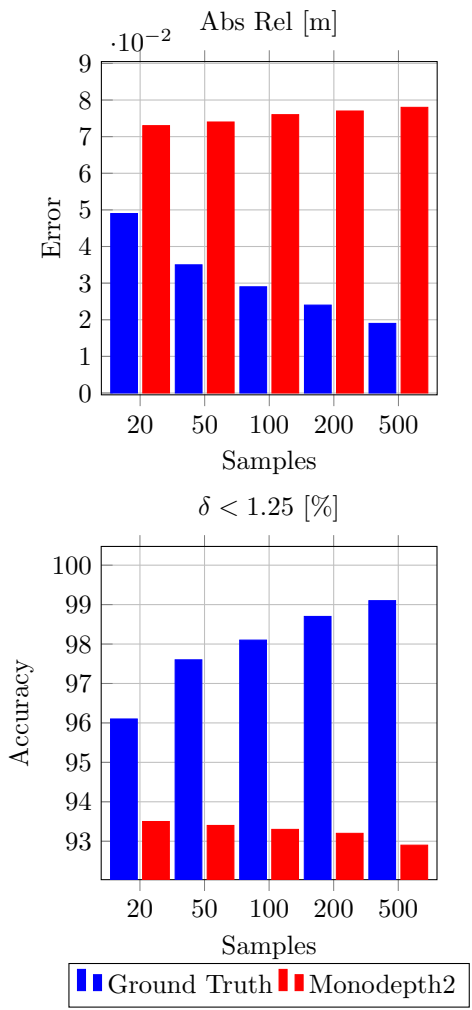
\begin{figure}[!htb]
\centering
\begin{tikzpicture}
\begin{axis}[title={Abs Rel [m]},
    ybar,
    enlargelimits=0.15,
    symbolic x coords={20,50,100,200,500},
	xtick=data,
	ylabel near ticks,
	xlabel near ticks,
	ylabel=Error,
	xlabel=Samples,
	grid=major,
	legend style={at={(0.5,-0.25)},
    anchor=north,legend columns=-1},
	ytick distance = 0.01,
    ytick pos = left,
    ymin=0.01, ymax=0.08,
    height=6cm,
    width=6cm,
]
\addplot[color=blue,fill=blue] 
	coordinates{(20,0.049)(50,0.035)(100,0.029)(200,0.024)(500,0.019)};
\addplot[color=red,fill=red] 
	coordinates{(20,0.073)(50,0.074)(100,0.076)(200,0.077)(500,0.078)};
\end{axis}
\end{tikzpicture}
\begin{tikzpicture}
\begin{axis}[title={$\delta<1.25$ $[\%]$},
    ybar,
    enlargelimits=0.15,
    symbolic x coords={20,50,100,200,500},
	xtick=data,
	ylabel near ticks,
	xlabel near ticks,
	ylabel=Accuracy,
	xlabel=Samples,
	grid=major,
	legend style={at={(0.5,-0.25)},
    anchor=north,legend columns=-1},
	ytick distance = 1.0,
    ytick pos = left,
    ymin=93.0, ymax=99.5,
    height=6cm,
    width=6cm,
]
\addplot[color=blue,fill=blue] 
	coordinates{(20,96.1)(50,97.6)(100,98.1)(200,98.7)(500,99.1)};
\addplot[color=red,fill=red] 
	coordinates{(20,93.5)(50,93.4)(100,93.3)(200,93.2)(500,92.9)};
\legend{Ground Truth, Monodepth2}
\end{axis}
\end{tikzpicture}
\caption{Comparison between the errors and precision obtained by our depth completion framework (without the surface normals module) when it is introduced to sparse data from the ground truth and the depth maps generated by the Monodepth2 \cite{godard2019digging}}
\label{graph:kittilidar_monodepth2}
\end{figure}

From the graphs in Fig. \ref{graph:kittilidar_monodepth2}, we analyze the effects caused on our depth completion CNN when it is fed with sparse data generated by the Monodepth2 pre-trained with the mono $+$ stereo configuration. The behavior of the curve related to the Monodepth2 approach is increasing for the Abs Rel metric and decreasing for the $\delta<1.25$ $[\%]$ since the more sparse data are sampled from the depth maps produced by Monodepth2, the more the errors of these maps influence the DSN results. 

However, the graphs of Fig. \ref{graph:kittilidar_monodepth2} also indicate that the performance of our network (without the surface normals module) is improved by using fewer amounts of depth data sampled from the Monodepth2 results. Therefore, the DSN may be successfully applied in real-world situations, where there is no ground truth available. 

\subsubsection{Visual Odometry Results}

Once depth data are estimated, they can be incorporated into VO/SLAM systems to improve the quality of visual odometry and the reconstructed map. With respect to Semi-Direct Visual Odometry (SVO) methods, the accurate depth information obtained from SIDE approaches can be added to enhance the invariance of the tracking algorithm to illumination changes in the input overexposed or underexposed images \cite{loo2019cnn}. Moreover, with the aid of such additional cues, the mapping points step of the SVO framework is performed with a smaller uncertainty, leading to an increase in the matching features accuracy and to a decrease in the convergence time of this feature correspondence mechanism \cite{loo2019cnn}. 

CNN-SLAM and CNN-SVO \cite{loo2019cnn} are examples of works in which the incorporation of the depth maps, predicted by a deep neural network, into the processing pipeline is done successfully. More specifically, the former work used a ResNet-based FCN \cite{laina2016deeper}, whereas the latter one used the Monodepth \cite{godard2017unsupervised} to generate the depth estimates. Since the predictions from the DSN are better than those produced by the cited methods, we fed the CNN-SVO framework with our depth estimations to better determine corresponding features. Using different setups of our CNN, we evaluated the improvements that this modification in mapping the points of consecutive keyframes brings to the SVO task. 

Particularly, we compared the results obtained by the joint application of the CNN-SVO and variations of the DSN with the CNN-SVO alone, the Direct Sparse Odometry (DSO) \cite{engel2017direct}, the SVO \citep{forster2014svo} and the ORB-SLAM without loop closure \cite{mur2017orb} through Table \ref{tab:slam}. Such experiments are conducted with the help of the $11$ sequences with ground truth trajectories provided by the KITTI Odometry dataset \cite{Geiger2012CVPR}. Also, we employed the Absolute Trajectory Error (ATE) and the RMSE metric to expose the aforementioned results. In Table \ref{tab:slam}, the spaces that are not filled with metric values in meters are related to the techniques that are not able to compute the entire sequence due to failures in the tracking algorithm. As it is proposed in the work of Loo \etal\cite{loo2019cnn}, we stored the ATEs with a median of $5$ runs.

\begin{table*}[]
\centering
\caption{Evaluation of DSN predictions applied to the CNN-SVO framework according to the RMSE metric. We used the first $11$ sequences of the KITTI Odometry dataset \cite{Geiger2012CVPR} to generate the results. Legend: SVO - Semi-Direct Visual Odometry; DSO - Direct Sparse Odometry; DSN$\dagger$ - DSN setup, without the surface normals module, that produces the best results; BA - Bundle Adjustment; DSN$\ddagger$ - DSN setup, with the surface normals module, that produces the most accurate results}
\label{tab:slam}
\resizebox{\textwidth}{!}{
\begin{tabular}{c|c|c|c|c|c|c|c}
\hline
\textbf{Sequence} & \textbf{SVO \cite{forster2014svo}} & \textbf{CNN-SVO $+$ BA \cite{loo2019cnn}} & \textbf{\makecell{CNN-SVO \cite{loo2019cnn} \\ $+$ BA $+$ DSN$\ddagger$}} & \textbf{\makecell{CNN-SVO \cite{loo2019cnn} \\ $+$ BA $+$ DSN$\dagger$}} & \textbf{\makecell{CNN-SVO \cite{loo2019cnn} \\ $+$ DSN$\dagger$}} & \textbf{DSO \cite{engel2017direct}} & \makecell{\textbf{ORB-SLAM \cite{mur2017orb}} \\ \textbf{(without loop closure)}} \\\hline \hline
00 & - & 17.5269 & \textbf{16.9438} & 19.7020 & 42.9432 & 113.1838 & 77.9502 \\
01 & - & - & - & - & - & - & - \\
02 & - & 50.5119 & \textbf{14.9513} & 15.7180 & 45.0136 & 116.8108 & 41.0064 \\
03 & - & 3.4588 & 1.9309 & 3.4340 & 12.4293 & 1.3943 & \textbf{1.0182} \\
04 & 58.3970 & 2.4414 & 2.3090 & 4.1912 & 5.6721 & 0.422 & \textbf{0.9302} \\
05 & - & 8.1513 & \textbf{7.1224} & 9.4541 & 30.2511 & 47.4605 & 40.3542 \\
06 & - & 11.5091 & \textbf{8.7421} & 11.4059 & 22.5910 & 55.6173 & 52.2282 \\
07 & - & 6.5141 & \textbf{1.9292} & 2.3099 & 8.0324 & 16.7192 & 16.546 \\
08 & - & 10.9755 & \textbf{8.3792} & 9.4699 & 13.3337 & 111.0832 & 51.6215 \\
09 & - & \textbf{10.6873} & 10.9941 & 15.0341 & 28.6044 & 52.2251 & 58.1742 \\
10 & - & 4.8354 & \textbf{2.3163} & 2.8371 & 4.1040 & 11.090 & 18.4765 \\\hline
\end{tabular}
}
\end{table*}

Analyzing the metrics from Table \ref{tab:slam}, it is plausible to conclude that when the CNN-SVO technique, along with the Bundle Adjustment algorithm (BA), employs the depth maps retrieved by the DSN, it is capable of tracking most of the addressed sequences in a more precise way. This occurs due to the initialization of new map points in the CNN-SVO pipeline with lower depth uncertainties. Furthermore, the results from Table \ref{tab:slam} also indicate that when the CNN-SVO is combined with the DSN configuration that leverages surface normals cues, it provides the best overall metric values. 

\begin{figure}[t!]
     \centering
          \begin{subfigure}[t]{0.15\textwidth}
         \centering
         \caption{Seq00}
         \includegraphics[width=\textwidth]{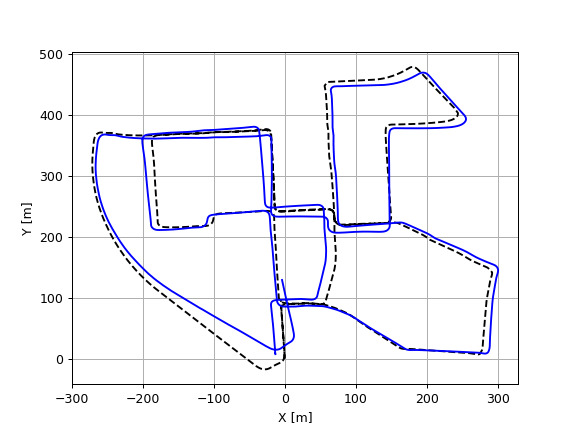}
         \hspace{1em}
         \caption{Seq07}
         \includegraphics[width=\textwidth]{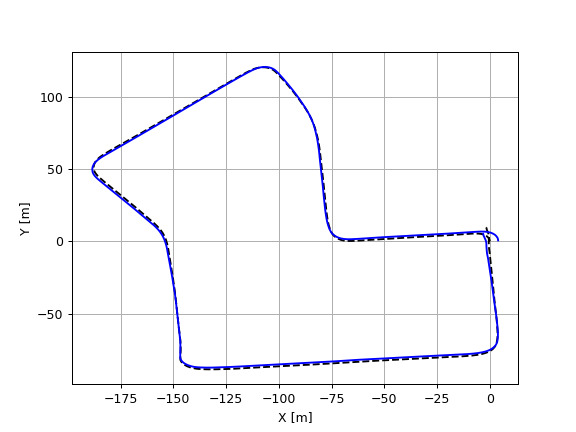}
         \label{subfig:traj1}
     \end{subfigure}
     \begin{subfigure}[t]{0.15\textwidth}
         \centering
         \caption{Seq02}
         \includegraphics[width=\textwidth]{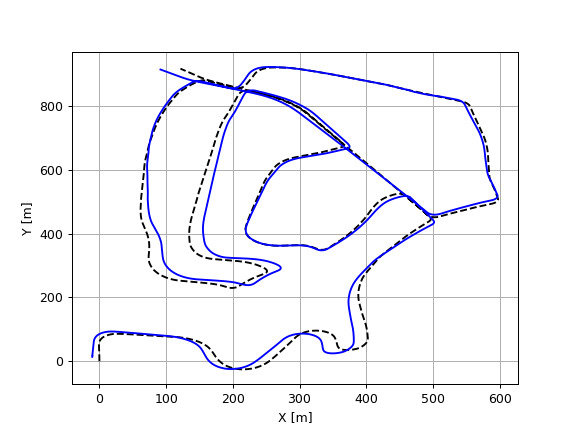}
         \hspace{1em}
         \caption{Seq08}
         \includegraphics[width=\textwidth]{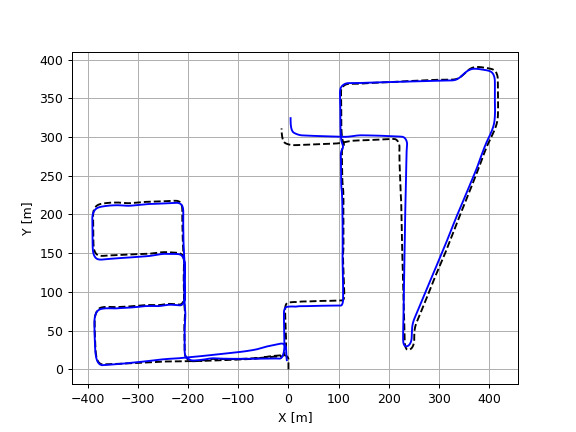}
         \label{subfig:traj2}
     \end{subfigure}
         \begin{subfigure}[t]{0.15\textwidth}
         \centering
         \caption{Seq05}
         \includegraphics[width=\textwidth]{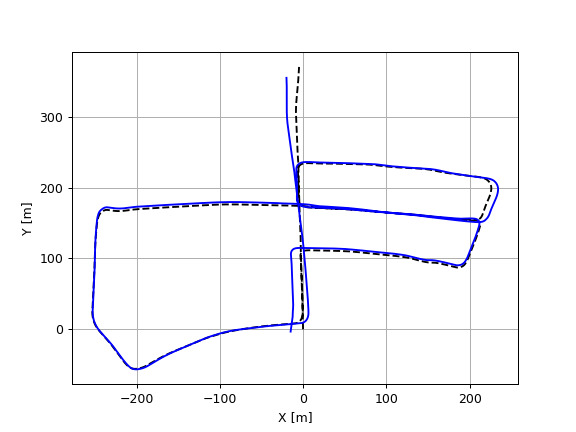}
         \hspace{1em}
         \caption{Seq10}
         \includegraphics[width=\textwidth]{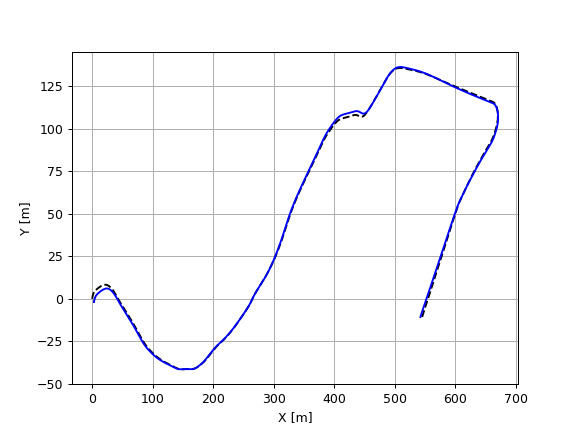}
         \label{subfig:traj3}
     \end{subfigure}
     \caption{Qualitative results of the camera trajectory estimations obtained on the KITTI Odometry \cite{Geiger2012CVPR} sequences $00$, $02$, $05$, $07$, $08$ and $10$ by the CNN-SVO $+$ BA jointly with the DSN and the surface normals module. The black dotted and solid blue trajectories represent the ground truth and the predictions respectively}
     \label{subfig:trajall}
\end{figure}

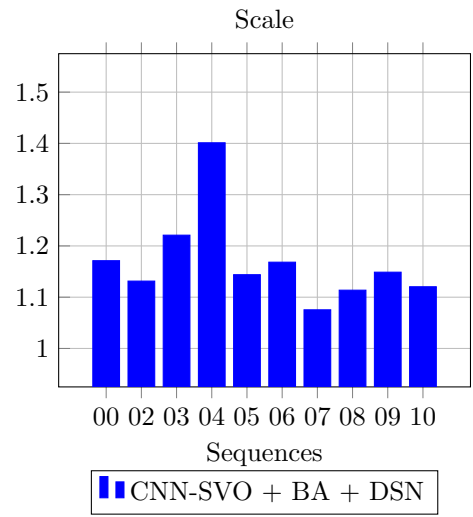
\begin{figure}[!htb]
\centering
\begin{tikzpicture}
\begin{axis}[title={Scale},
    ybar,
    enlargelimits=0.15,
    symbolic x coords={00,02,03,04,05,06,07,08,09,10},
	xtick=data,
	ylabel near ticks,
	xlabel near ticks,
	xlabel=Sequences,
	grid=major,
	legend style={at={(0.5,-0.25)},
    anchor=north,legend columns=-1},
	ytick distance = 0.1,
    ytick pos = left,
    ymin=1.0, ymax=1.5,
    height=6cm,
    width=7cm,
]
\addplot[color=blue,fill=blue] 
	coordinates{(00,1.1712)(02,1.1314)(03,1.2210)(04,1.4012)(05,1.1439)(06,1.1683)(07,1.0754)(08,1.1137)(09,1.1488)(10,1.1204)};
\legend{CNN-SVO $+$ BA $+$ DSN}
\end{axis}
\end{tikzpicture}
\caption{Scale drift profile of the tracking results generated by the CNN-SVO $+$ BA $+$ DSN (with the surface normals module). The sequences are taken from the KITTI Odometry dataset \cite{Geiger2012CVPR}}
\label{graph:kittiodometry_scale}
\end{figure}

The trajectories illustrated in Fig. \ref{subfig:trajall} are the aligned qualitative results of our best visual odometry approach in the KITTI Odometry sequences. Such trajectories confirm the metrics exposed in Table \ref{tab:slam}. The scale drift values from Fig. \ref{graph:kittiodometry_scale} show that our tracking results have a scale close to the ground truth. However, the small lags of scale occur due to the distortions suffered by the output depth maps of the DSN, which are resized to be applied in the CNN-SVO framework.

\subsection{Indoor Ablation Studies}

\subsubsection{Structural Studies}

To analyze the behavior of the DSN $3$ when it is exposed to other types of scenes, we also conducted ablation studies with indoor datasets, which are composed of images with different scales than those present in the outdoor datasets. At first, we modified the loss function and the decoder structure of our framework to determine an optimized combination. Regarding the RMSE metric, Table \ref{tab:nyunets} indicates that both the attention term and the modification in the constant $\kappa$ of the BerHu function do not bring benefits to the network predictions since, in comparison to the KITTI Depth \cite{uhrig2017sparsity}, the NYU Depth V2 dataset \cite{silberman2012indoor} is denser both in regions closer and further from the scene. Due to the denser profile of the ground truth of indoor datasets, simple loss functions, such as $\mathcal{L}_1^+$ and $\mathcal{L}_2^+$, are capable of handling the pixel distribution of the reference maps and update the network weights in a faster and more efficient way.

As also can be inferred from Table \ref{tab:nyunets}, the Pyramid$^1$ module enhances the value of most of the metrics. However, the block Pyramid$^2$ is not advantageous for the DSN $3$ predictions once, for the NYU Depth V2 \cite{silberman2012indoor}, the upsampled feature maps from the initial convolution of the network decoder do not present the necessary monocular cues to support the estimation of the output depth map. Moreover, we utilized the semantic maps, with $13$ fine annotated labels, of the NYU Depth V2 \cite{silberman2012indoor} and SUNRGBD \cite{song2015sun} datasets, as well as the depth ground truth of all indoor datasets covered in this work to pre-train the proposed framework. As in the experiments addressing outdoor data, we conclude that the pre-trained weights for semantic segmentation help our approach to reason about high-frequency regions of the image, yet the features learned from the pre-training of the DSN $3$ on the same task leads to more accurate results.

\begin{table*}[]
\centering
\caption{Performance of the proposed SIDE method for different architecture configurations and training strategies. The feature extraction network of the DSN $3$ is the developed Model $3$. Both pyramid modules, besides the loss function $\mathcal{L}_{BerHu}^{+}$, are implemented as in the analyzes involving the outdoor datasets. The terms Sem and Indoor indicate the pre-training of the DSN $3$ with semantic maps and extra depth maps, respectively, provided by the indoor datasets}
\label{tab:nyunets}
\resizebox{\textwidth}{!}{
\begin{tabular}{c|c|c|c|cccccc|ccc}
\hline
\textbf{Method} & \textbf{Loss} & \textbf{Size} & \textbf{Speed} & \textbf{Abs Rel}$\downarrow$ & \textbf{Sqr Rel}$\downarrow$ & \textbf{RMSE}$\downarrow$ & \textbf{RMSE (log)}$\downarrow$ & \textbf{SILog}$\downarrow$ & \textbf{log10}$\downarrow$ & $\bm{\delta < 1.25}$$\uparrow$ & $\bm{\delta < 1.25^2}$$\uparrow$ & $\bm{\delta < 1.25^3}$$\uparrow$\\\hline\hline
DSN 3 & $\mathcal{L}_{BerHu}$ & \textbf{12\ M} & $\boldsymbol{51}\ \boldsymbol{fps}$ & 0.169 & 0.136 & 0.523 & 0.208 & 17.143 & 0.070 & 0.768 & 0.934 & 0.978\\
DSN 3 & $\mathcal{L}_a$ $+$ $\mathcal{L}_{BerHu}$ & \textbf{12\ M} & $\boldsymbol{51}\ \boldsymbol{fps}$ & 0.179 & 0.144 & 0.540 & 0.216 & 17.315 & 0.073 & 0.746 & 0.926 & 0.977\\
DSN 3 & $\mathcal{L}_{BerHu}^{+}$ & \textbf{12\ M} & $\boldsymbol{51}\ \boldsymbol{fps}$ & 0.179 & 0.151 & 0.538 & 0.215 & 17.277 & 0.073 & 0.750 & 0.924 & 0.976\\
DSN 3 & $\mathcal{L}_1$ & \textbf{12\ M} & $\boldsymbol{51}\ \boldsymbol{fps}$ & 0.180 & 0.151 & 0.548 & 0.217 & 17.588 & 0.073 & 0.746 & 0.927 & 0.978\\
DSN 3 $+$ Pyramid$^1$ & $\mathcal{L}_{BerHu}$ & \textbf{12\ M} & $41\ fps$ & 0.171 & 0.133 & 0.521 & 0.207 & 16.605 & 0.070 & 0.758 & 0.936 & 0.982\\
DSN 3 $+$ Pyramid$^1$ & $\mathcal{L}_2^+$ & \textbf{12\ M} & $41\ fps$ & 0.175 & 0.138 & 0.520 & 0.207 & 16.462 & 0.071 & 0.757 & 0.934 & 0.981\\
DSN 3 $+$ Pyramid$^1$ & $\mathcal{L}_1^+$ & \textbf{12\ M} & $41\ fps$ & 0.167 & 0.130 & 0.519 & 0.207 & 16.859 & 0.069 & 0.763 & 0.933 & 0.981\\
DSN 3 $+$ Pyramid$^2$ & $\mathcal{L}_1^+$ & \textbf{12\ M} & $32\ fps$ & 0.174 & 0.143 & 0.529 & 0.210 & 16.896 & 0.071 & 0.756 & 0.933 & 0.979\\
DSN 3 $+$ Pyramid$^2$ $+$ Sem & $\mathcal{L}_1^+$ & \textbf{12\ M} & $32\ fps$ & 0.174 & 0.139 & 0.522 & 0.211 & 17.231 & 0.071 & 0.760 & 0.932 & 0.977\\
DSN 3 $+$ Pyramid$^1$ $+$ Indoor & $\mathcal{L}_1^+$ & \textbf{12\ M} & $41\ fps$ & \textbf{0.142} & \textbf{0.098} & \textbf{0.445} & \textbf{0.177} & \textbf{14.197} & \textbf{0.059} & \textbf{0.823} & \textbf{0.954} & \textbf{0.987}\\\hline
\end{tabular}
}
\end{table*}

\subsubsection{Surface Normals Analysis}

Considering indoor scenes, the results obtained from the experiments involving our approach that is trained with both depth and surface normals reference maps are presented in Table \ref{tab:nyusn}. From the RMSE metric, it is possible to evidence that the application of the edge detector, implemented in the surface normals module, improves the results once it supports the proposed CNN in the process of predicting more accurate boundaries. The edge detection operation reduces the network's prediction speed by $1$ fps and does not change its number of trainable parameters.

When the DSN $3$ is pre-trained on the additional surface normals ground truth, with the help of the geometric 2.5D penalty, the results are also enhanced. This occurs since our learning-based method is introduced to a greater amount of 3D geometric features which help it to better generalize. Although the presence of the Gaussian filter leads to better values for the Sqr Rel and $\delta<1.25^3$ metrics, its absence is related to the configuration of the proposed framework that generated the most accurate overall results. This proves the efficiency and robustness of the surface normals module, which can be successfully applied to our SIDE approach at a high frame rate. 

\begin{table*}[]
\centering
\caption{Analysis of our pipeline that leverages surface normals cues. The configuration DSN $3$ $+$ Pyramid$^1$ $+$ Indoor $+$ SN, along with the $\mathcal{L}_1^+$ loss, is maintained in all experiments. Legend: DSN $3$ - DSN with Model $3$ as encoder; Pyramid$^1$ - decoder's pyramid-shaped module; Indoor - pre-training with indoor datasets; SN - surface normals ground truth; Edge - Sobel edge detector; Blur - convolution with Gaussian kernel; $\psi_1$ and $\psi_2$ - constants of the 2.5D loss function used in the pre-training and training steps respectively}
\label{tab:nyusn}
\resizebox{\textwidth}{!}{
\begin{tabular}{c|c|c|c|c|c|c|cccccc|ccc}
\hline
\textbf{Method} & \textbf{Size} & \textbf{Speed} & \textbf{Edge} & \textbf{Blur} & \textbf{$\psi_1$} & \textbf{$\psi_2$} & \textbf{Abs Rel} & \textbf{Sqr Rel} & \textbf{RMSE} & \textbf{RMSE (log)} & \textbf{SILog} & \textbf{log10} & $\bm{\delta < 1.25}$ & $\bm{\delta < 1.25^2}$ & $\bm{\delta < 1.25^3}$ \\\hline\hline
DSN 3 $+$ Pyramid$^1$ $+$ Indoor $+$ SN & \textbf{12\ M} & $\boldsymbol{35}\ \boldsymbol{fps}$ & \xmark & \xmark & \xmark & $10^0$ & 0.138 & 0.096 & 0.444 & 0.177 & 14.227 & 0.058 & 0.825 & 0.954 & 0.986\\
DSN 3 $+$ Pyramid$^1$ $+$ Indoor $+$ SN & \textbf{12\ M} & $\boldsymbol{35}\ \boldsymbol{fps}$ & \xmark & \xmark & \xmark & $10^6$ & 0.137 & 0.096 & 0.441 & 0.176 & 14.171 & 0.058 & 0.825 & 0.954 & 0.986\\
DSN 3 $+$ Pyramid$^1$ $+$ Indoor $+$ SN & \textbf{12\ M} & $34\ fps$ & \cmark & \xmark & \xmark & $10^6$ & 0.137 & 0.094 & 0.439 & 0.175 & 14.134 & 0.058 & 0.826 & 0.954 & 0.986\\
DSN 3 $+$ Pyramid$^1$ $+$ Indoor $+$ SN & \textbf{12\ M} & $34\ fps$ & \cmark & \xmark & \xmark & $10^7$ & 0.138 & 0.095 & 0.442 & 0.176 & 14.171 & 0.058 & 0.825 & 0.956 & 0.986\\
DSN 3 $+$ Pyramid$^1$ $+$ Indoor $+$ SN & \textbf{12\ M} & $34\ fps$ & \cmark & \xmark & \xmark & $10^5$ & 0.139 & 0.096 & 0.443 & 0.177 & 14.289 & 0.059 & 0.822 & 0.953 & 0.986\\
DSN 3 $+$ Pyramid$^1$ $+$ Indoor $+$ SN & \textbf{12\ M} & $34\ fps$ & \cmark & \xmark & $10^6$ & $10^6$ & 0.134 & 0.091 & 0.435 & 0.173 & 13.917 & 0.057 & 0.831 & 0.958 & 0.986\\
DSN 3 $+$ Pyramid$^1$ $+$ Indoor $+$ SN & \textbf{12\ M} & $34\ fps$ & \cmark & \xmark & $10^7$ & $10^7$ & 0.134 & 0.089 & 0.434 & 0.172 & 13.841 & 0.057 & 0.831 & 0.957 & 0.987\\
DSN 3 $+$ Pyramid$^1$ $+$ Indoor $+$ SN & \textbf{12\ M} & $34\ fps$ & \cmark & \xmark & $10^6$ & $10^7$ & \textbf{0.132} & 0.088 & \textbf{0.429} & \textbf{0.171} & \textbf{13.732} & \textbf{0.056} & \textbf{0.834} & \textbf{0.959} & 0.987\\
DSN 3 $+$ Pyramid$^1$ $+$ Indoor $+$ SN & \textbf{12\ M} & $33\ fps$ & \cmark & \cmark & $10^6$ & $10^7$ & \textbf{0.132} & \textbf{0.086} & 0.431 & \textbf{0.171} & 13.793 & \textbf{0.056} & 0.833 & \textbf{0.959} & \textbf{0.988}\\
\hline
\end{tabular}
}
\end{table*}

Comparing the most accurate results of the DSN, on the NYU Depth V2 dataset \cite{silberman2012indoor}, with the metrics obtained by different works in the state-of-the-art, Table \ref{tab:comparison} shows that our method reaches the top-3 of the benchmark ranking. However, among the approaches in the top-3, ours has the lowest runtime and total size, as shown in Table \ref{tab:litside}. 

Also, we evaluated the predictions from the surface normals module of the DSN and compared the acquired results with the ones generated by recent works through Table \ref{soa_tab:nyusn}. Masking out the invalid pixels of the reference surface normals maps, we achieve the most accurate results presented in Table \ref{soa_tab:nyusn}. 

On the other hand, considering all the pixels of the surface normals ground truth, we still get the best results of the Median and the accuracy metrics. Therefore, this demonstrates that our surface normals module is more advantageous than the other techniques since, besides not consisting of an entire branch or another model, it works mutually with the monocular depth estimation step.

\begin{table}[]
\centering
\caption{Quantitative comparison between our method and others that address the single-view depth estimation task. The results are evaluated on the NYU Depth V2 dataset \cite{silberman2012indoor}. DSN$\dagger$ refers to the DSN setup that provides the most accurate overall results}
\label{tab:comparison}
\resizebox{\linewidth}{!}{%
\begin{tabular}{c|ccc|ccc}
\hline
\textbf{Method} & \textbf{Abs Rel}$\downarrow$ & \textbf{RMSE}$\downarrow$ & \textbf{log10}$\downarrow$ & $\bm{\delta < 1.25}$$\uparrow$ & $\bm{\delta < 1.25^2}$$\uparrow$ & $\bm{\delta < 1.25^3}$$\uparrow$\\\hline\hline
Saxena \etal \cite{saxena2008make3d} & 0.349 & 1.214 & - & 0.447 & 0.745 & 0.897\\
Karsch \etal \cite{karsch2014depth} & 0.349 & 1.210 & 1.131 & - & - & -\\
Liu \etal \cite{liu2014discrete} & 0.335 & 1.060 & 1.127 & - & - & -\\
Ladicky \etal \cite{ladicky2014pulling} & - & - & - & 0.542 & 0.829 & 0.941\\
Eigen \etal \cite{eigen2014depth} & 0.214 & 0.877 & 0.285 & 0.611 & 0.887 & 0.971\\
Wang \etal \cite{wang2015towards} & 0.220 & 0.824 & - & 0.605 & 0.890 & 0.970\\
HCRF \cite{li2015depth} & 0.232 & 0.821 & 0.094 & 0.621 & 0.886 & 0.968\\
DCNF \cite{liu2015learning} & 0.213 & 0.759 & 0.087 & 0.650 & 0.906 & 0.976\\
NR forest \cite{roy2016monocular} & 0.187 & 0.744 & 0.078 & - & - & -\\
MS-CNN \cite{eigen2015predicting} & 0.158 & 0.641 & - & 0.769 & 0.950 & 0.988\\
Chakrabarti \etal \cite{chakrabarti2016depth} & 0.149 & 0.620 & - & 0.806 & 0.958 & 0.987\\
Li \etal \cite{li2017two} & 0.152 & 0.611 & 0.064 & 0.789 & 0.955 & 988\\
SOM \cite{zhu2019structure} & 0.136 & 0.604 & 0.067 & 0.814 & 0.959 & 0.990\\
Xu \etal \cite{xu2018structured} & 0.125 & 0.593 & - & 0.806 & 0.952 & 0.986\\
MS-CRF \cite{xu2017multi} & 0.121 & 0.586 & 0.052 & 0.811 & 0.954 & 0.987\\
PAD-Net \cite{xu2018pad} & 0.120 & 0.582 & - & 0.817 & 0.954 & 0.987\\
DeepLabV3+ (F10) \cite{gur2019single} & 0.162 & 0.575 & - & 0.772 & 0.942 & 0.984\\
FCRN \cite{laina2016deeper} & 0.127 & 0.573 & 0.055 & 0.811 & 0.953 & 0.988\\
Lee \etal \cite{lee2018single} & 0.139 & 0.572 & - & 0.815 & 0.963 & 0.991\\
Qi \etal \cite{qi2018geonet} & 0.128 & 0.569 & 0.057 & 0.834 & 0.960 & 0.990\\
Index Network \cite{lu2019index} & - & 0.565 & - & 0.786 & - & -\\
RF-LW \cite{nekrasov2019real} & 0.149 & 0.565 & 0.105 & 0.790 & 0.955 & 0.990\\
Cao \etal\cite{cao2017estimating} & 0.141 & 0.540 & 0.060 & 0.790 & 0.955 & 0.990\\
RelativeDepth \cite{lee2019monocular} & 0.131 & 0.538 & 0.087 & 0.837 & 0.971 & 0.994\\
SC-SfMLearner-ResNet18 \cite{bian2020unsupervised} & 0.147 & 0.536 & 0.062 & 0.804 & 0.950 & 0.986\\
SENet-154 \cite{hu2019revisiting} & 0.115 & 0.530 & 0.050 & 0.866 & 0.975 & 0.993\\
Kim \etal\cite{kim2018deep} & 0.117 & 0.525 & - & 0.825 & 0.976 & 0.993\\
SARPN \cite{chen2019structure} & 0.111 & 0.514 & 0.048 & 0.846 & 0.968 & 0.994\\
DORN \cite{fu2018deep} & 0.115 & 0.509 & 0.051 & 0.828 & 0.965 & 0.992\\
AdaDepth \cite{nath2018adadepth} & 0.114 & 0.506 & - & 0.856 & 0.966 & 0.991\\
TRL \cite{zhang2018joint} & 0.144 & 0.501 & 0.181 & 0.815 & 0.962 & 0.992\\
Zhang \etal\cite{zhang2018joint} & 0.144 & 0.501 & 0.181 & 0.815 & 0.962 & 0.992\\
PHN \etal\cite{zhang2018progressive} & 0.144 & 0.501 & - & 0.835 & 0.962 & 0.992\\
Su \etal\cite{su2020monocular} & 0.137 & 0.498 & 0.058 & 0.826 & 0.967 & 0.995\\
SDC-Depth \cite{wang2020sdc} & 0.128 & 0.497 & 0.174 & 0.845 & 0.966 & 0.990\\
PAP \cite{zhang2019pattern} & 0.121 & 0.497 & 0.175 & 0.846 & 0.968 & 0.994\\
ACAN \cite{chen2019attention} & 0.138 & 0.496 & 0.101 & 0.826 & 0.964 & 0.990\\
SharpNet \cite{ramamonjisoa2019sharpnet} & 0.139 & 0.495 & \textbf{0.047} & \textbf{0.888} & \textbf{0.979} & \textbf{0.995}\\
DenseDepth \cite{alhashim2018high} & 0.123 & 0.465 & 0.053 & 0.846 & 0.974 & 0.994\\
\textbf{DSN$\dagger$} & 0.132 & 0.429 & 0.056 & 0.834 & 0.959 & 0.987\\
VNL \cite{yin2019enforcing} & \textbf{0.108} & 0.416 & 0.048 & 0.875 & 0.976 & 0.994\\
BTS \cite{lee2019big} & 0.110 & \textbf{0.392} & \textbf{0.047} & 0.885 & 0.978 & 0.994\\
\hline
\end{tabular}%
} 
\end{table}

\begin{table}[]
\centering
\caption{Performance of the proposed CNN and other techniques on the surface normals ground truth of the NYU Depth V2 \cite{silberman2012indoor}. Legend: DSN$\dagger$ - DSN configuration that provides the most accurate overall results; DSN$\ddagger$ - same DSN configuration as the DSN$\dagger$, yet with the Gaussian filter; Mask - mask out the invalid pixels related to the surface normals ground truth}
\label{soa_tab:nyusn}
\resizebox{\linewidth}{!}{
\begin{tabular}{c|ccc|ccc}
\hline
\textbf{Method} & \textbf{Mean}$\downarrow$ & \textbf{Median}$\downarrow$ & \textbf{RMSE}$\downarrow$ & $\bm{11.25^{\circ}}$$\uparrow$ & $\bm{22.50^{\circ}}$$\uparrow$ & $\bm{30^{\circ}}$$\uparrow$\\\hline \hline
Fouhey \etal \cite{fouhey2013data} & 36.3 & 19.2 & - & 16.4 & 36.6 & 48.2\\
UIOW \cite{fouhey2014unfolding} & 35.2 & 17.9 & - & 40.5 & 54.1 & 58.9\\
Ladicky \etal \cite{ladicky2014discriminatively} & 33.5 & 23.1 & - & 27.7 & 49.0 & 58.7\\
MS-CNN \cite{eigen2015predicting} & 23.7 & 15.5 & - & 39.2 & 62.0 & 71.1\\
Deep3D \cite{wang2015designing} & 26.9 & 14.8 & - & 42.0 & 61.2 & 68.2\\
SkipNet \cite{bansal2016marr} & 19.8 & 12.0 & 28.2 & 47.9 & 70.0 & 77.8\\
SURGE \cite{wang2016surge} & 20.6 & 12.2 & - & 47.3 & 68.9 & 76.6\\
Qi \etal \cite{qi2018geonet} & 19.0 & 11.8 & 26.9 & 48.4 & 71.5 & 79.5\\
PAP \cite{zhang2019pattern} & 18.6 & 11.7 & 25.5 & 48.8 & 72.2 & 79.8\\
DSN$\dagger$ $+$ Mask & \textbf{8.6} & \textbf{5.5} & \textbf{12.2} & 72.2 & \textbf{91.0} & 96.8\\
DSN$\dagger$ & 18.7 & 6.9 & 33.7 & 63.2 & 79.7 & 84.8\\
DSN$\ddagger$ $+$ Mask & \textbf{8.6} & \textbf{5.5} & \textbf{12.2} & \textbf{72.3} & \textbf{91.0} & \textbf{96.9}\\
DSN$\ddagger$ & 18.7 & 6.8 & 33.7 & 63.3 & 79.7 & 84.8\\
\hline
\end{tabular}
}%
\end{table}

\begin{figure*}[t!]
     \centering
          \begin{subfigure}[t]{0.16\textwidth}
         \centering
         \caption{RGB Input}
         \includegraphics[width=\textwidth]{images/sharprgb1.jpg}
         \hspace{1em}
         \includegraphics[width=\textwidth]{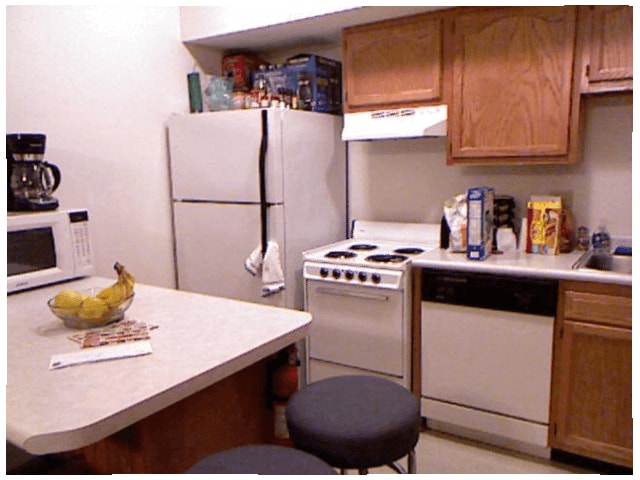}
         \hspace{1em}
         \includegraphics[width=\textwidth]{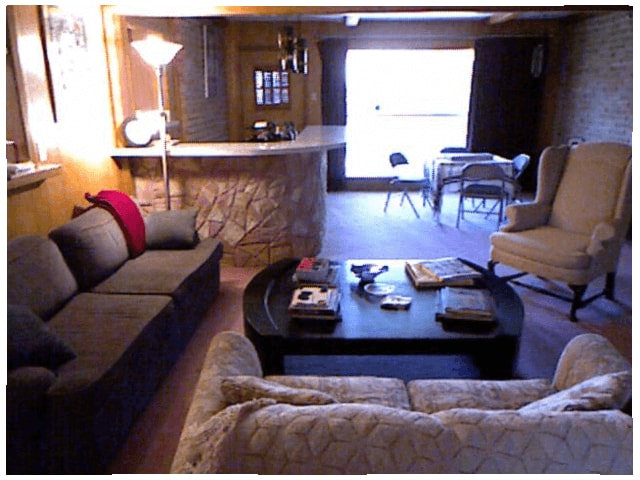}
         \label{subfig1:rgb}
     \end{subfigure}
     \begin{subfigure}[t]{0.16\textwidth}
         \centering
         \caption{Ground Truth}
         \includegraphics[width=\textwidth]{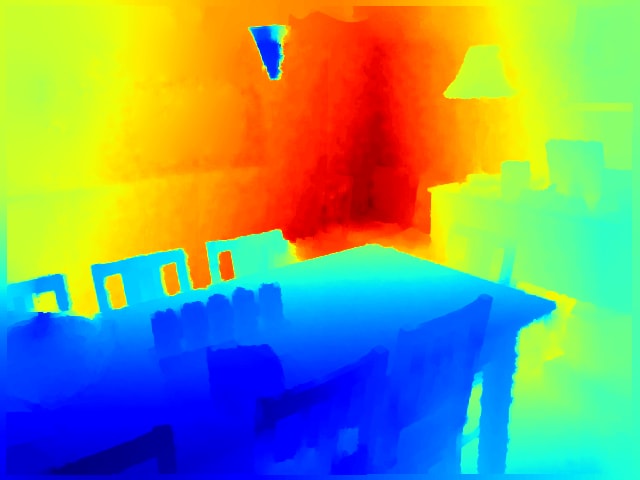}
         \hspace{1em}
         \includegraphics[width=\textwidth]{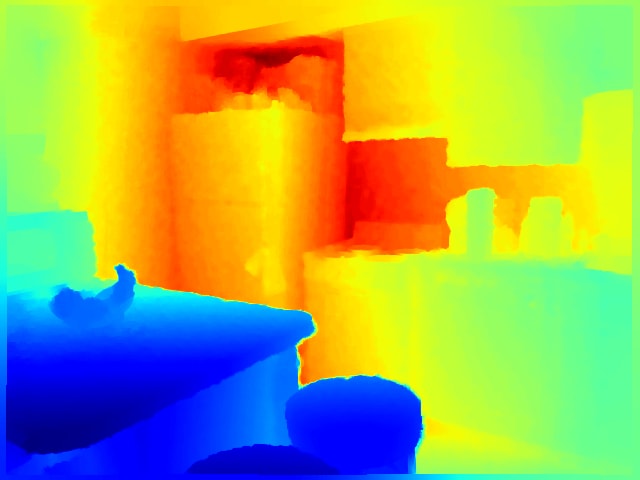}
         \hspace{1em}
         \includegraphics[width=\textwidth]{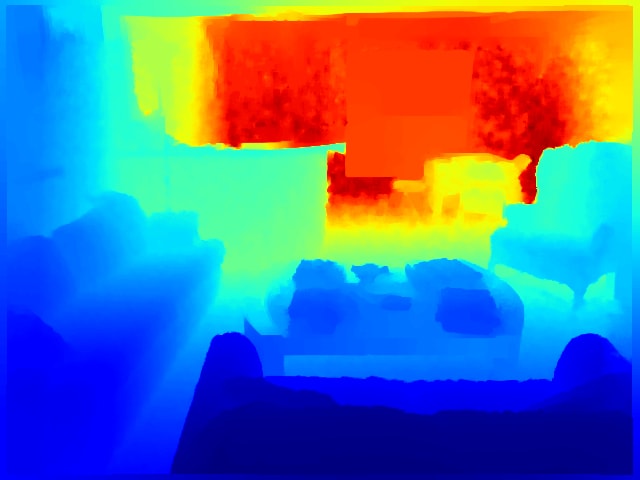}
         \label{subfig1:gt}
     \end{subfigure}
         \begin{subfigure}[t]{0.16\textwidth}
         \centering
         \caption{Laina \etal \cite{laina2016deeper}}
         \includegraphics[width=\textwidth]{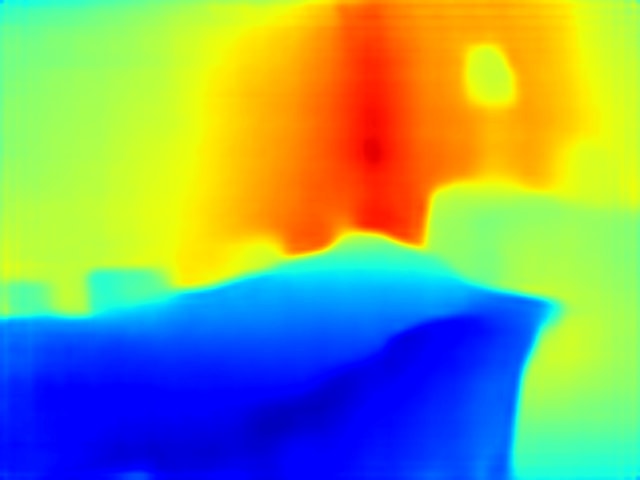}
         \hspace{1em}
         \includegraphics[width=\textwidth]{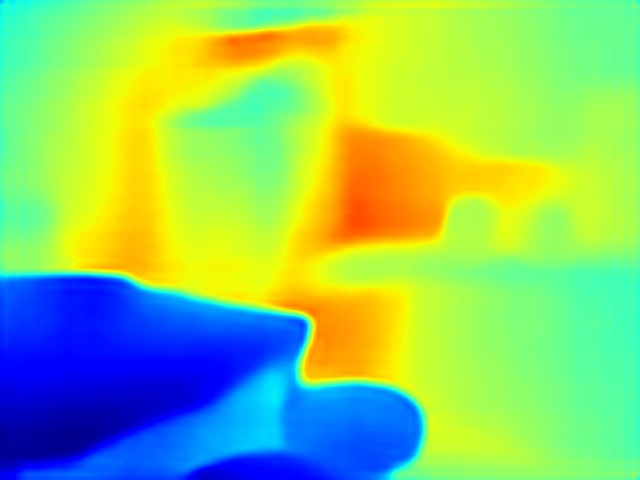}
         \hspace{1em}
         \includegraphics[width=\textwidth]{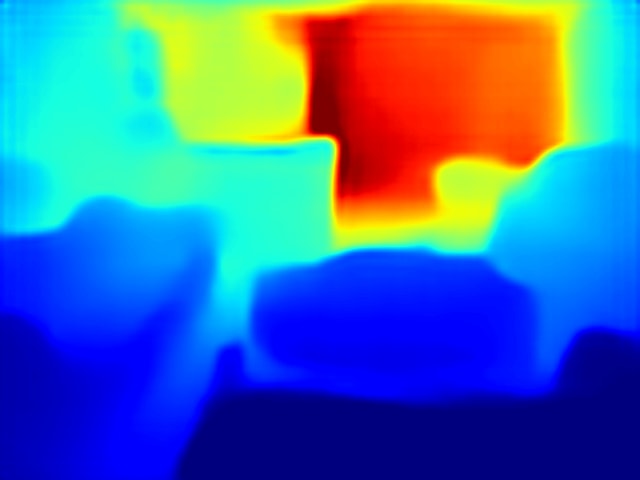}
         \label{subfig1:laina}
     \end{subfigure}
     \begin{subfigure}[t]{0.16\textwidth}
         \centering
         \caption{DORN \cite{fu2018deep}}
         \includegraphics[width=\textwidth]{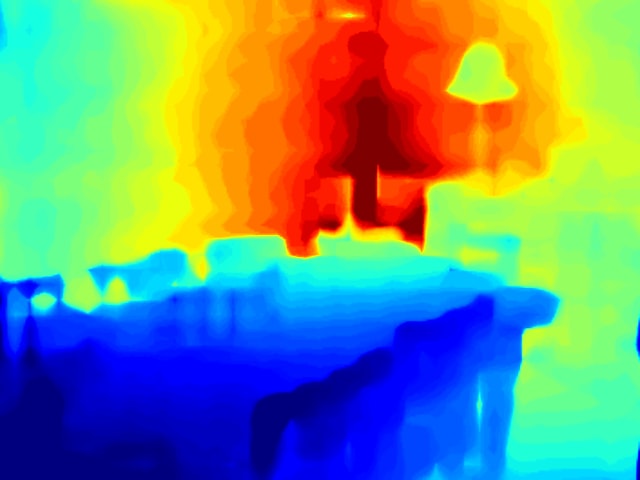}
         \hspace{1em}
         \includegraphics[width=\textwidth]{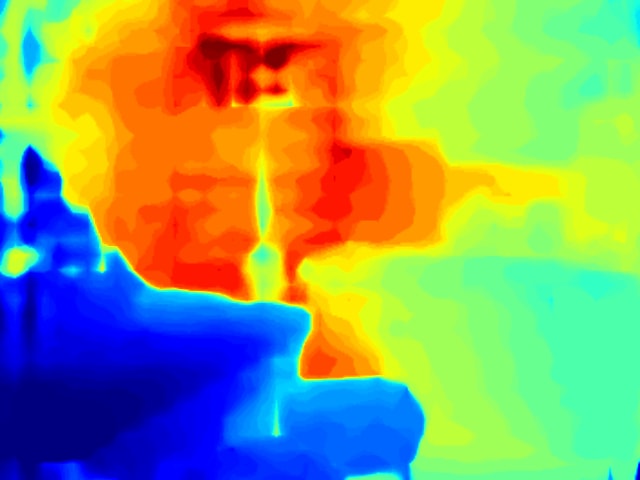}
         \hspace{1em}
         \includegraphics[width=\textwidth]{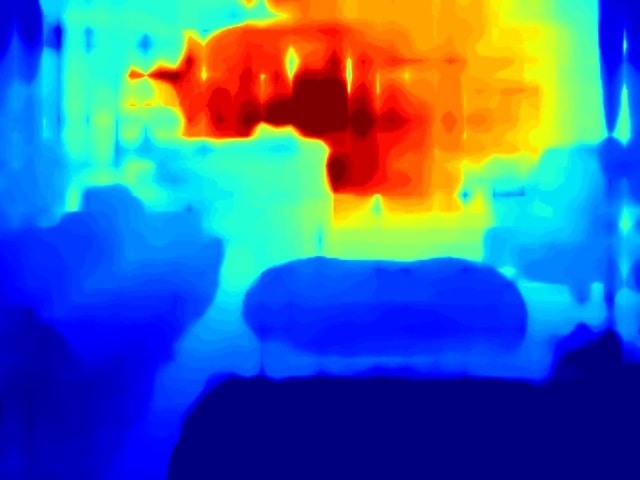}
         \label{subfig1:dorn}
     \end{subfigure}
         \begin{subfigure}[t]{0.16\textwidth}
         \centering
         \caption{SharpNet \cite{ramamonjisoa2019sharpnet}}
         \includegraphics[width=\textwidth]{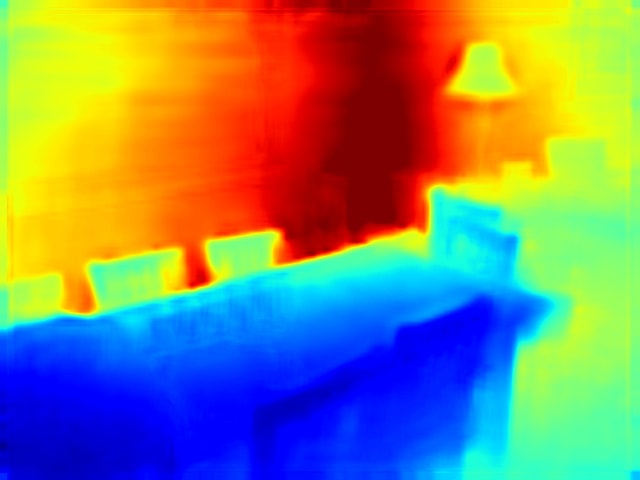}
         \hspace{1em}
         \includegraphics[width=\textwidth]{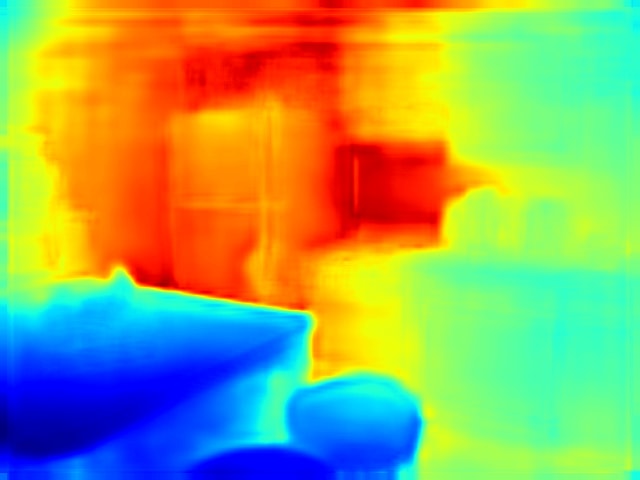}
         \hspace{1em} 
         \includegraphics[width=\textwidth]{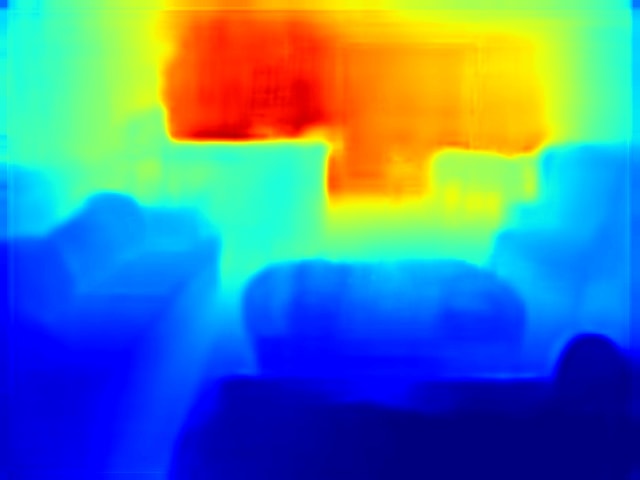}
         \label{subfig1:sharpnet}
     \end{subfigure}
     \begin{subfigure}[t]{0.16\textwidth}
         \centering
         \caption{DSN$\dagger$}
         \includegraphics[width=\textwidth]{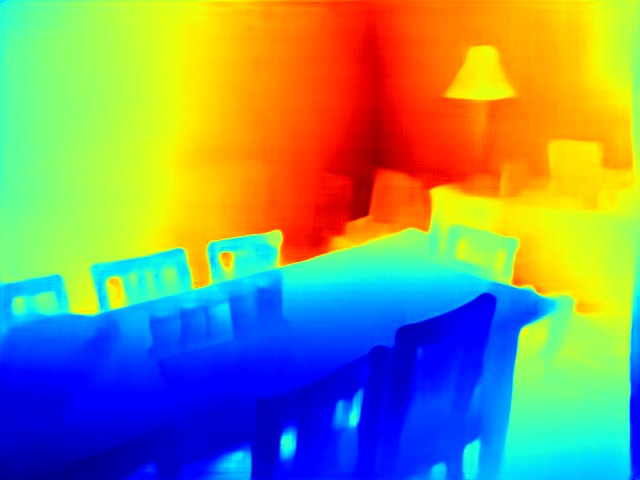}
         \hspace{1em}
         \includegraphics[width=\textwidth]{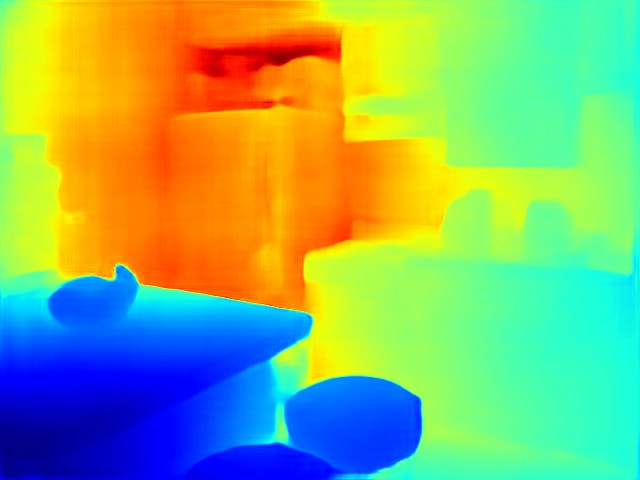}
         \hspace{1em}
         \includegraphics[width=\textwidth]{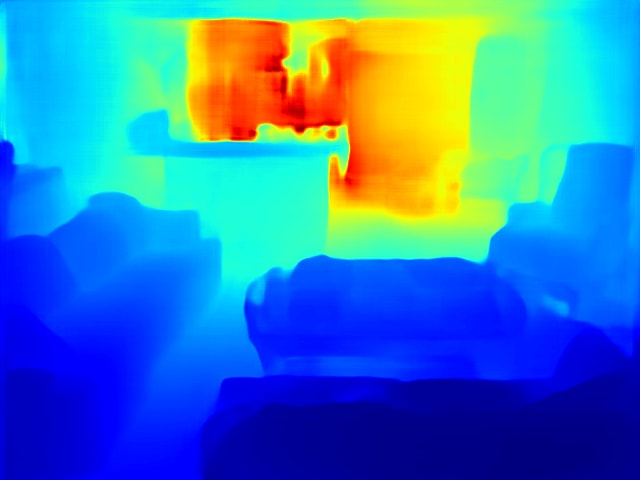}
         \label{subfig1:DSN}
     \end{subfigure}
     \caption{Comparison between the depth predictions of indoor scenes from our SIDE pipeline and other supervised methods in the state-of-the-art. The DSN$\dagger$ approach is related to the DSN configuration that produces the most accurate overall results. The input images and the ground truth come from the NYU Depth V2 dataset \cite{silberman2012indoor}. The ground truth is interpolated for visualization purposes}
     \label{subfig1NYU:all}
\end{figure*}

\begin{figure*}[t!]
     \centering
     \begin{subfigure}[t]{0.16\textwidth}
         \centering
         \caption{RGB Input}
         \includegraphics[width=\textwidth]{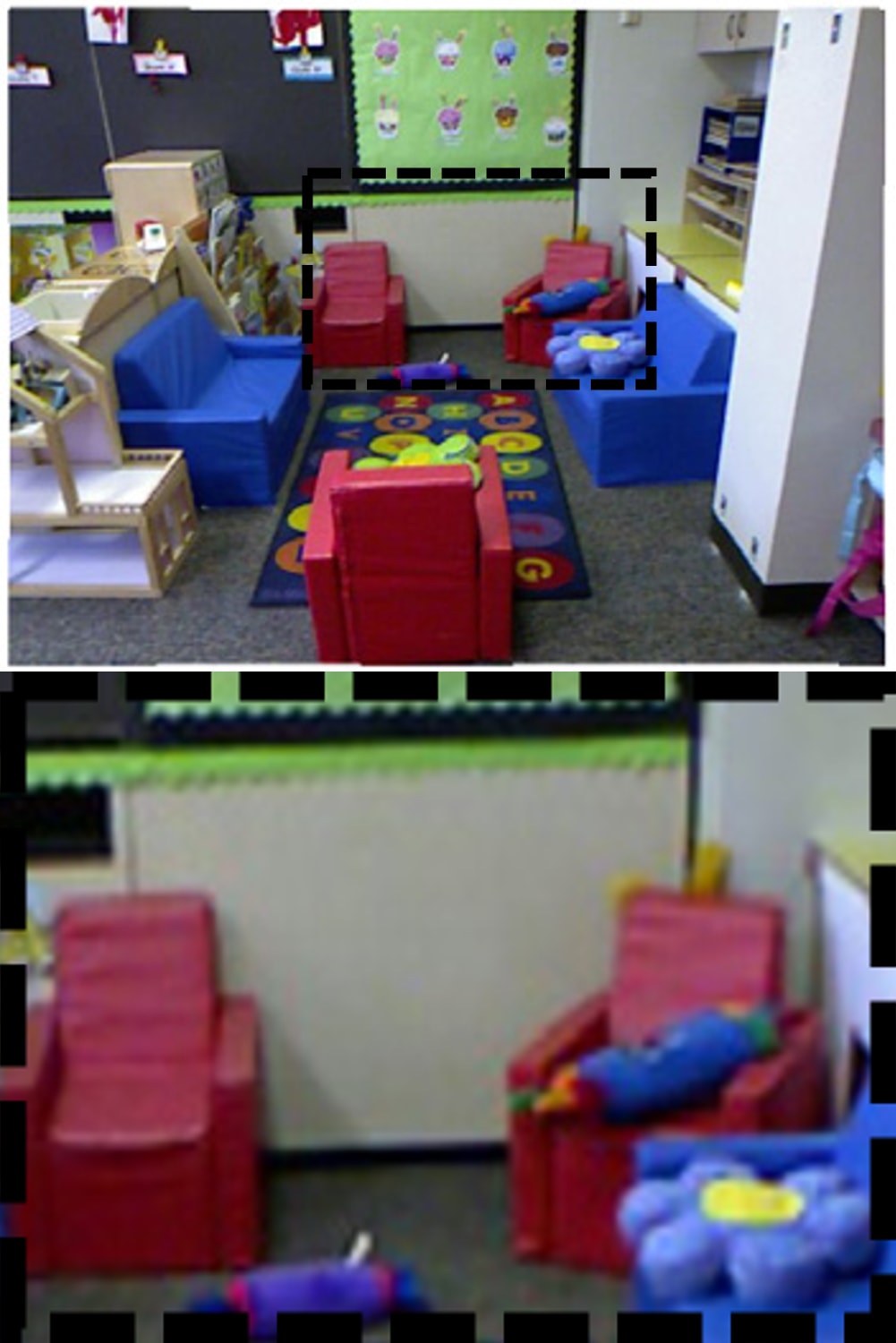}
         \hspace{1em}
         \includegraphics[width=\textwidth]{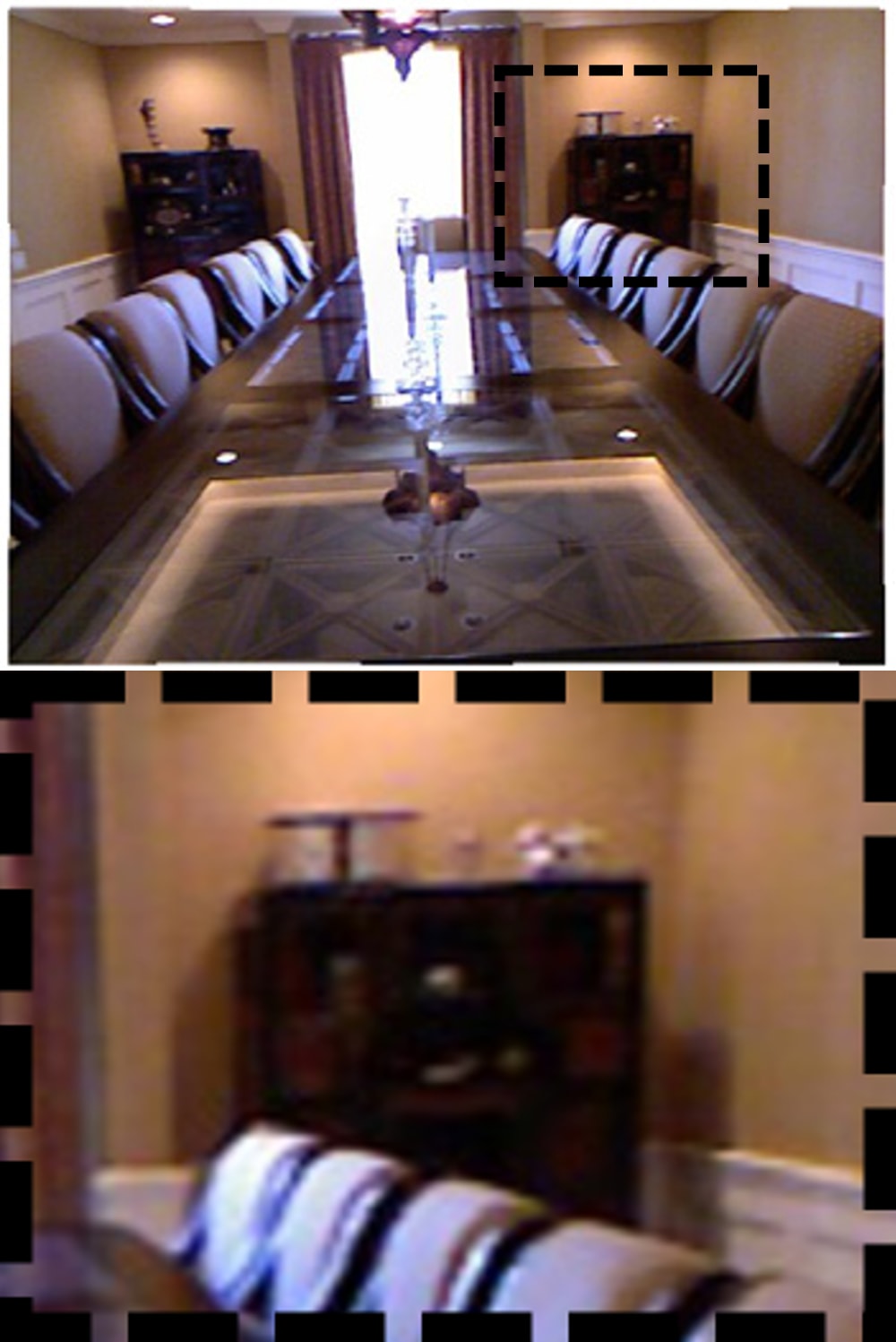}
         \hspace{1em}
         \includegraphics[width=\textwidth]{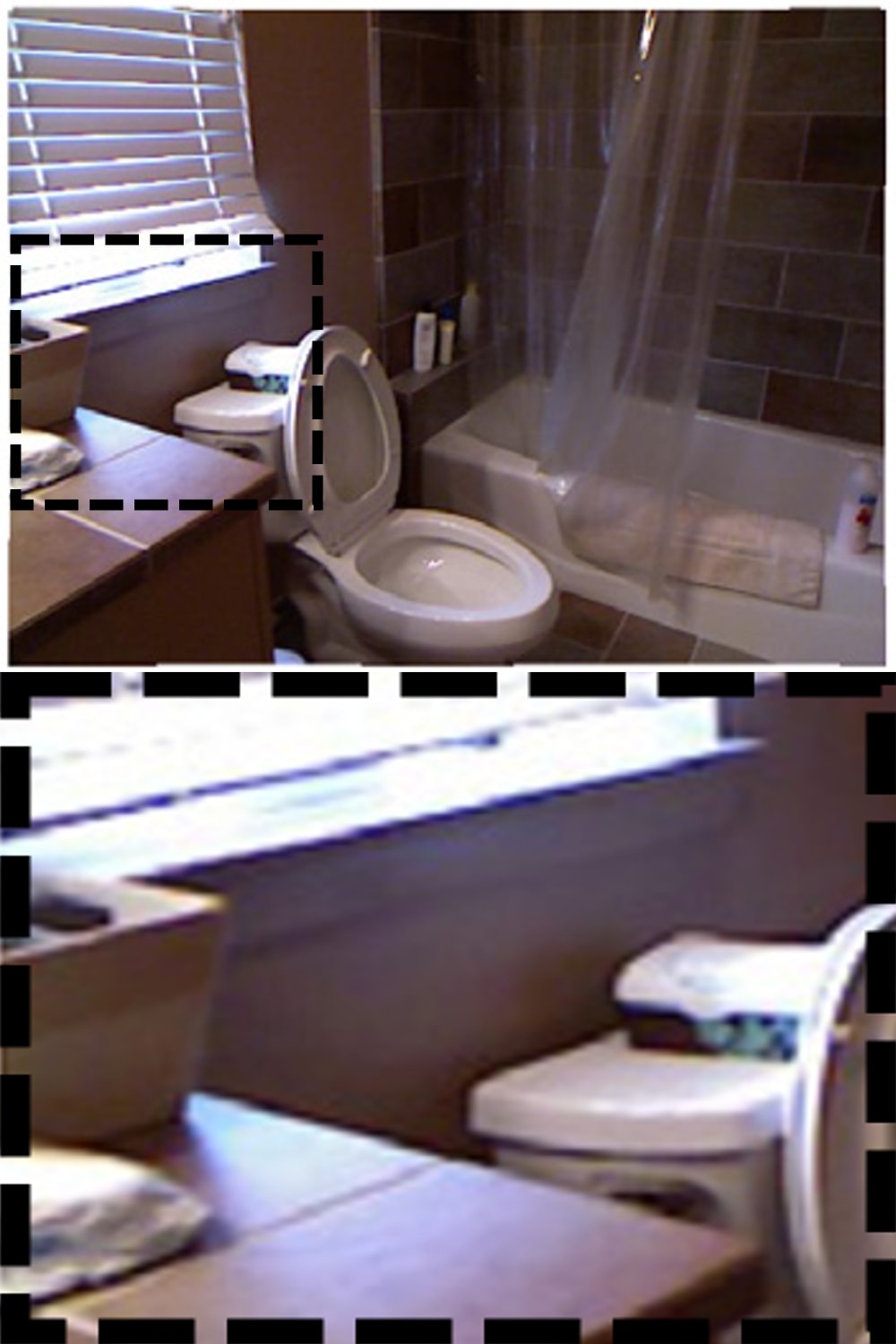}
         \label{subfig2:rgb}
     \end{subfigure}
     \begin{subfigure}[t]{0.16\textwidth}
         \centering
         \caption{DORN \cite{fu2018deep}}
         \includegraphics[width=\textwidth]{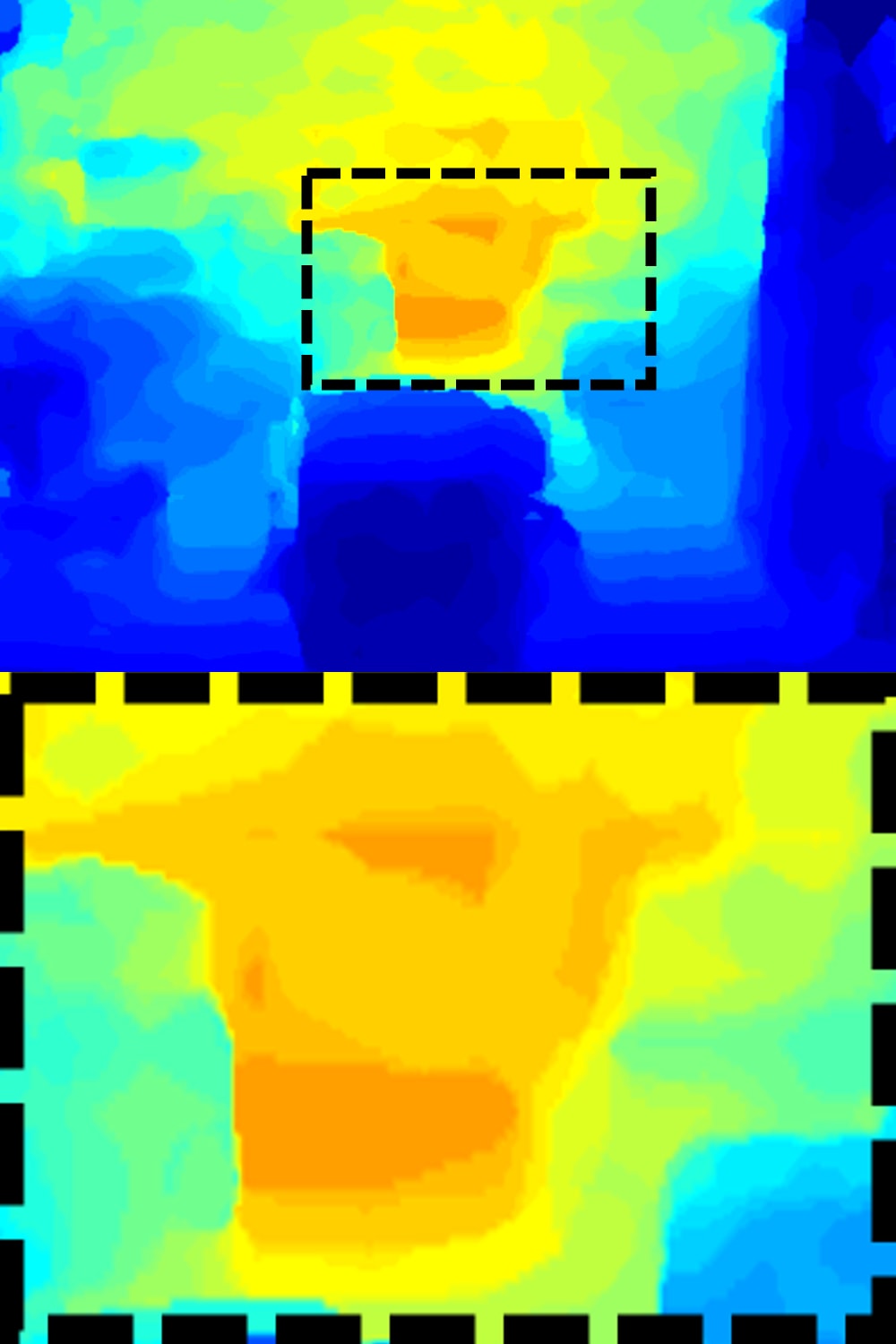}
         \hspace{1em}
         \includegraphics[width=\textwidth]{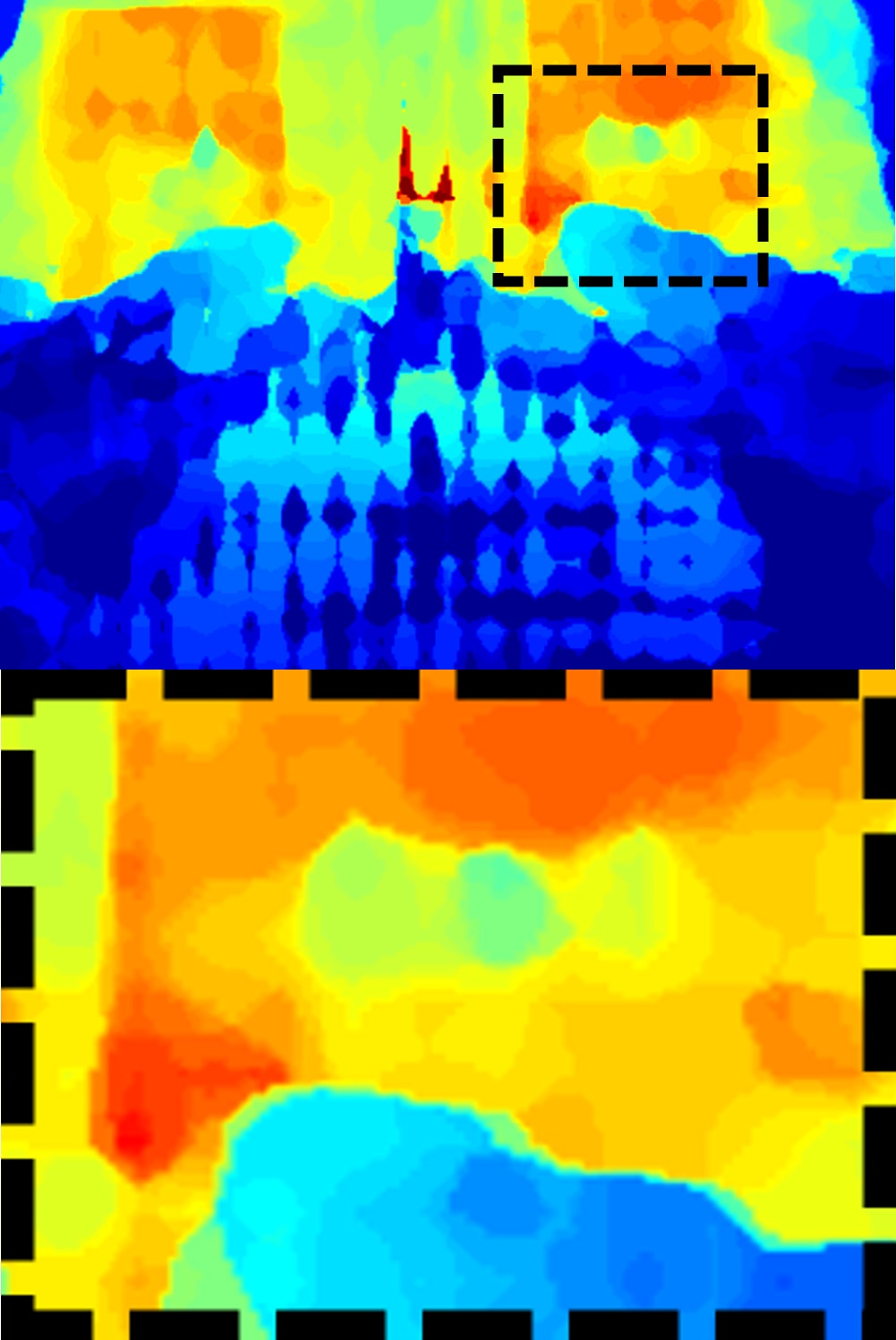}
         \hspace{1em}
         \includegraphics[width=\textwidth]{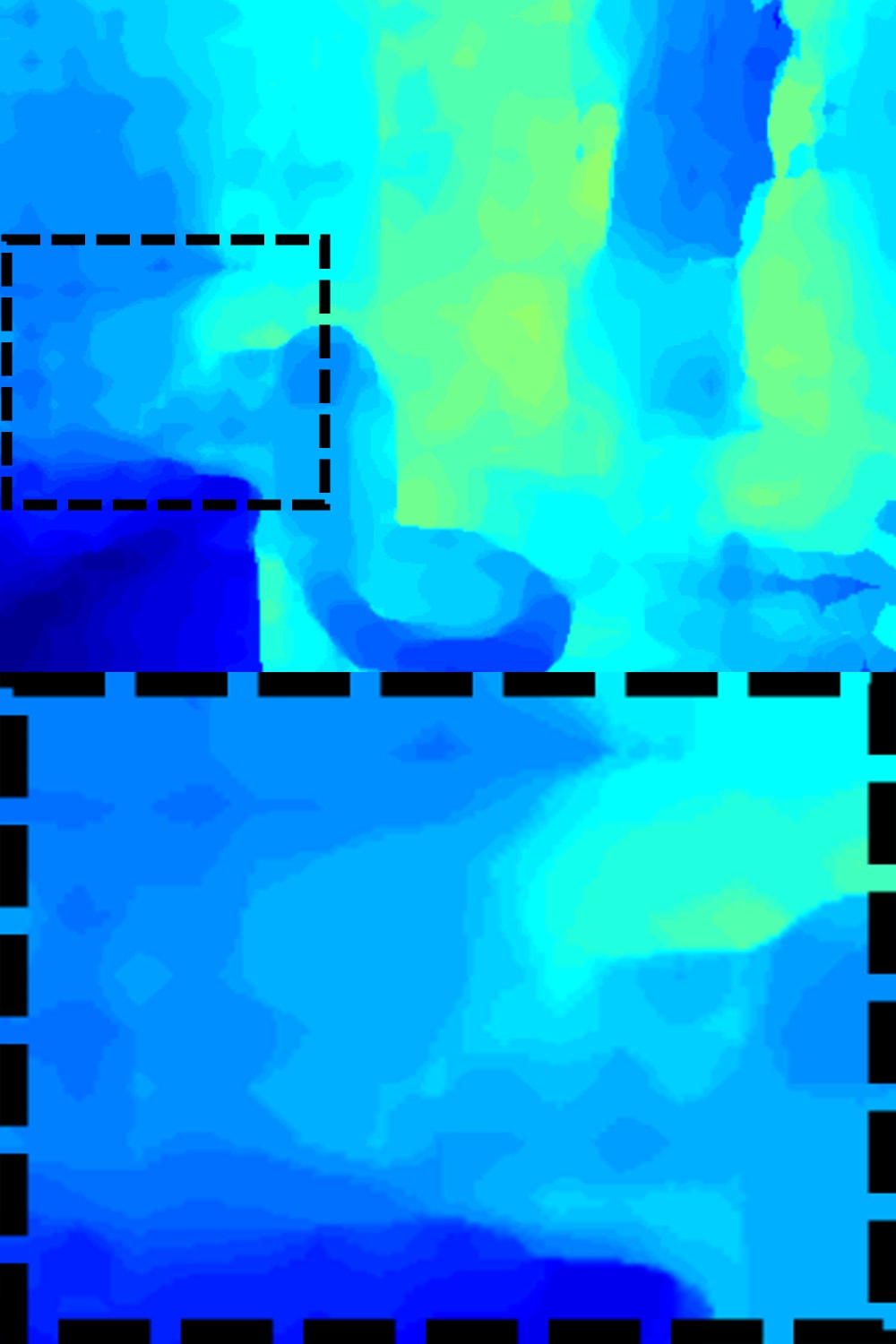}
         \label{subfig2:dorn}
     \end{subfigure}
          \begin{subfigure}[t]{0.16\textwidth}
         \centering
         \caption{DenseDepth \cite{alhashim2018high}}
         \includegraphics[width=\textwidth]{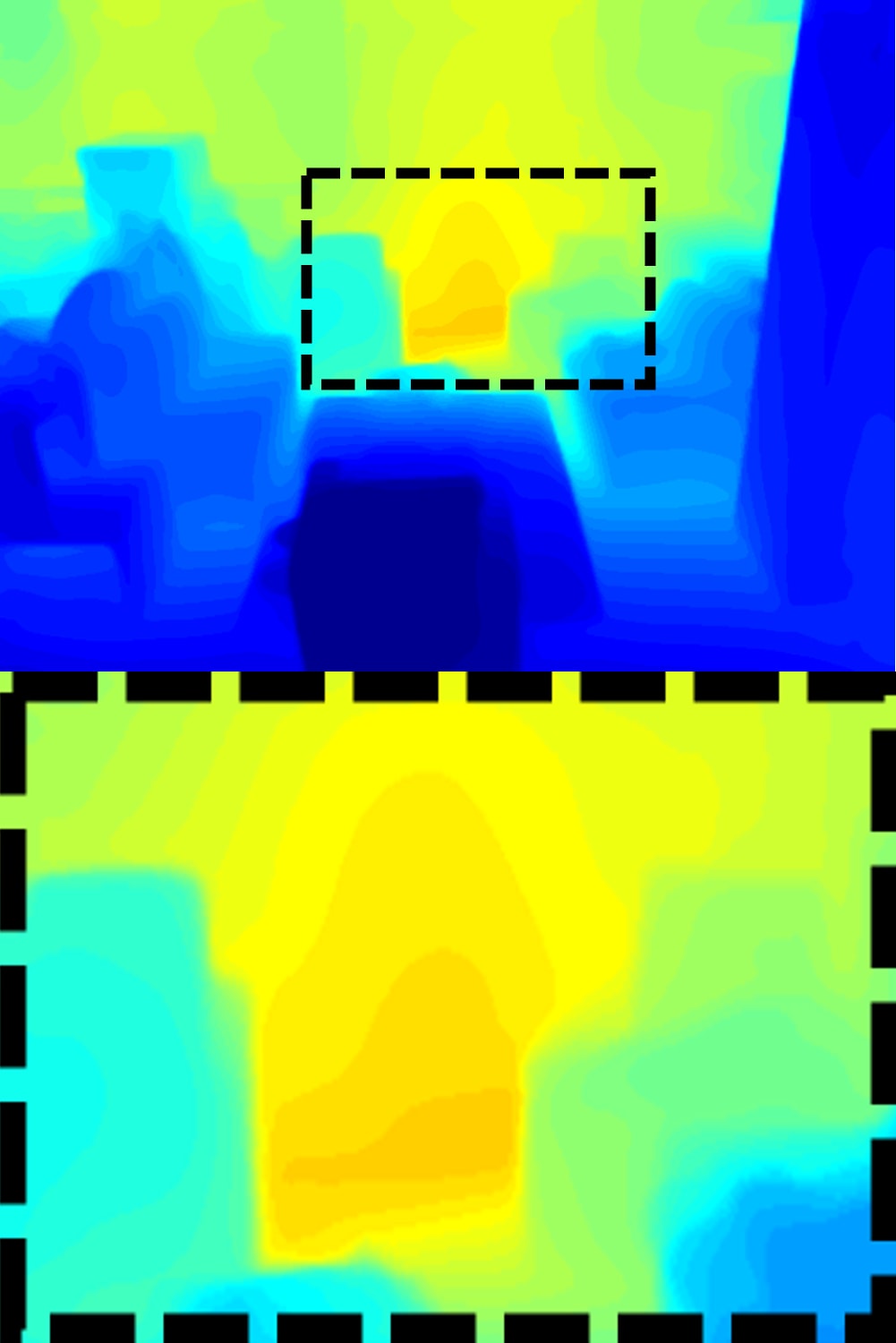}
         \hspace{1em}
         \includegraphics[width=\textwidth]{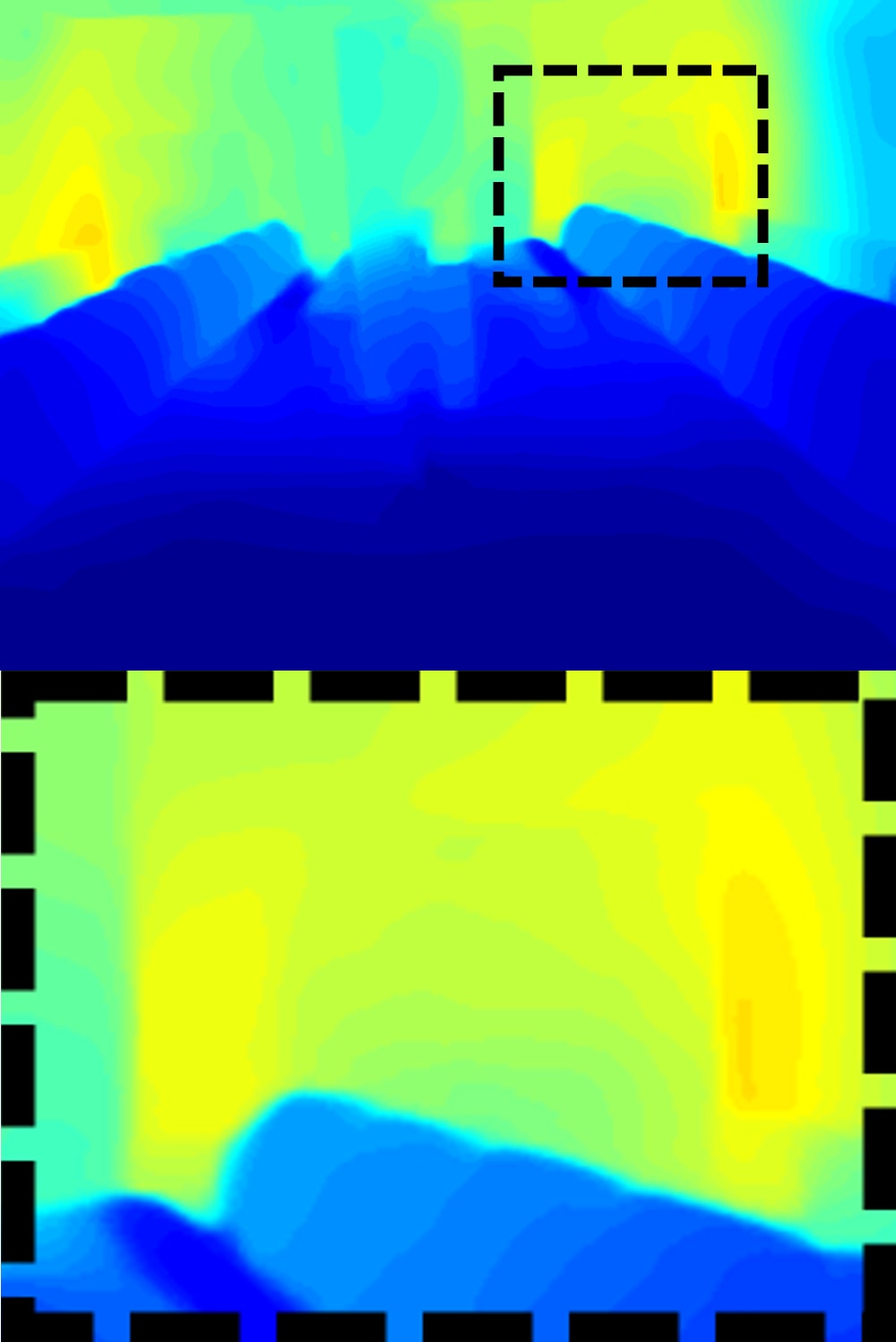}
         \hspace{1em}
         \includegraphics[width=\textwidth]{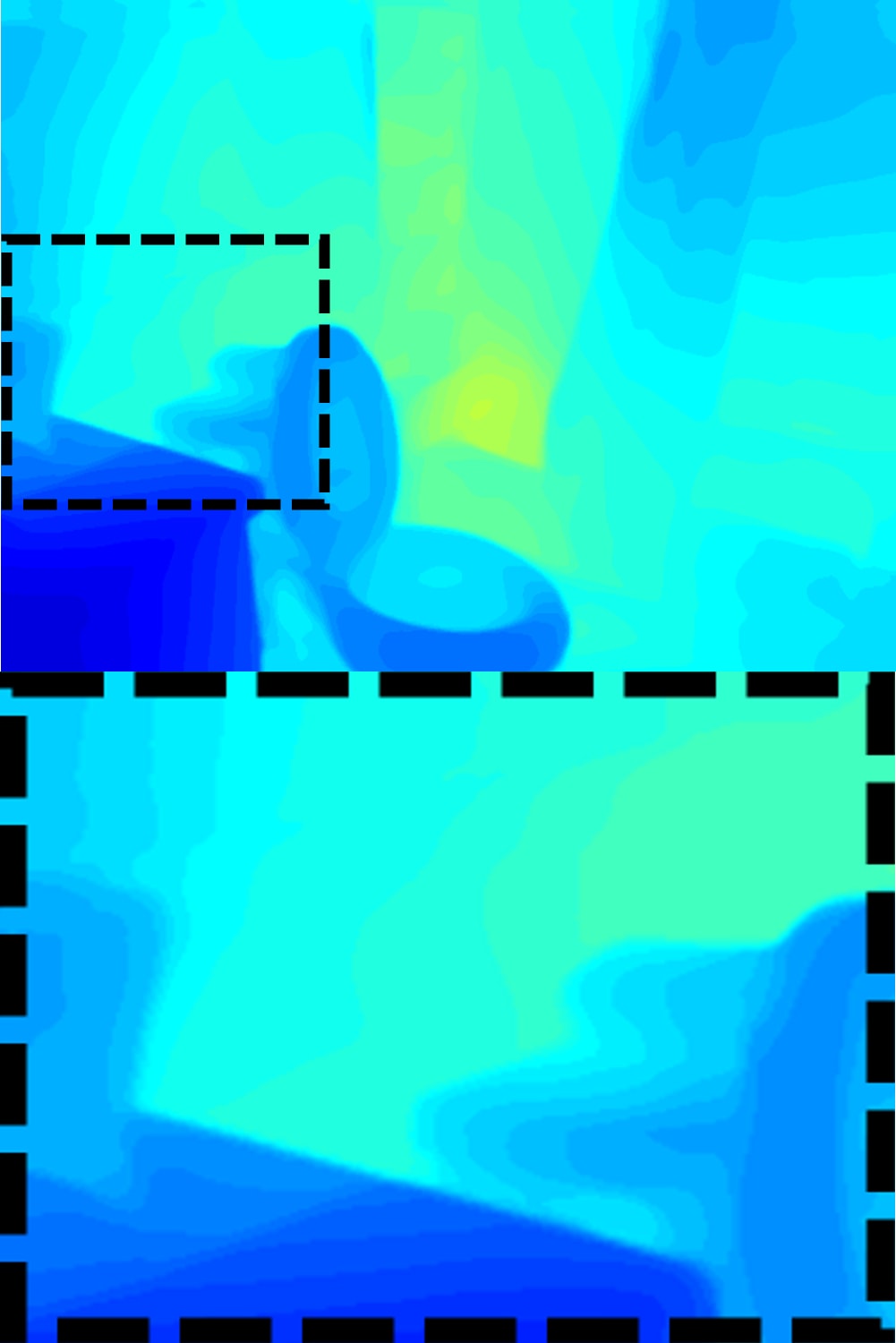}
         \label{subfig2:densed}
     \end{subfigure}
          \begin{subfigure}[t]{0.16\textwidth}
         \centering
         \caption{DSN$\dagger$}
         \includegraphics[width=\textwidth]{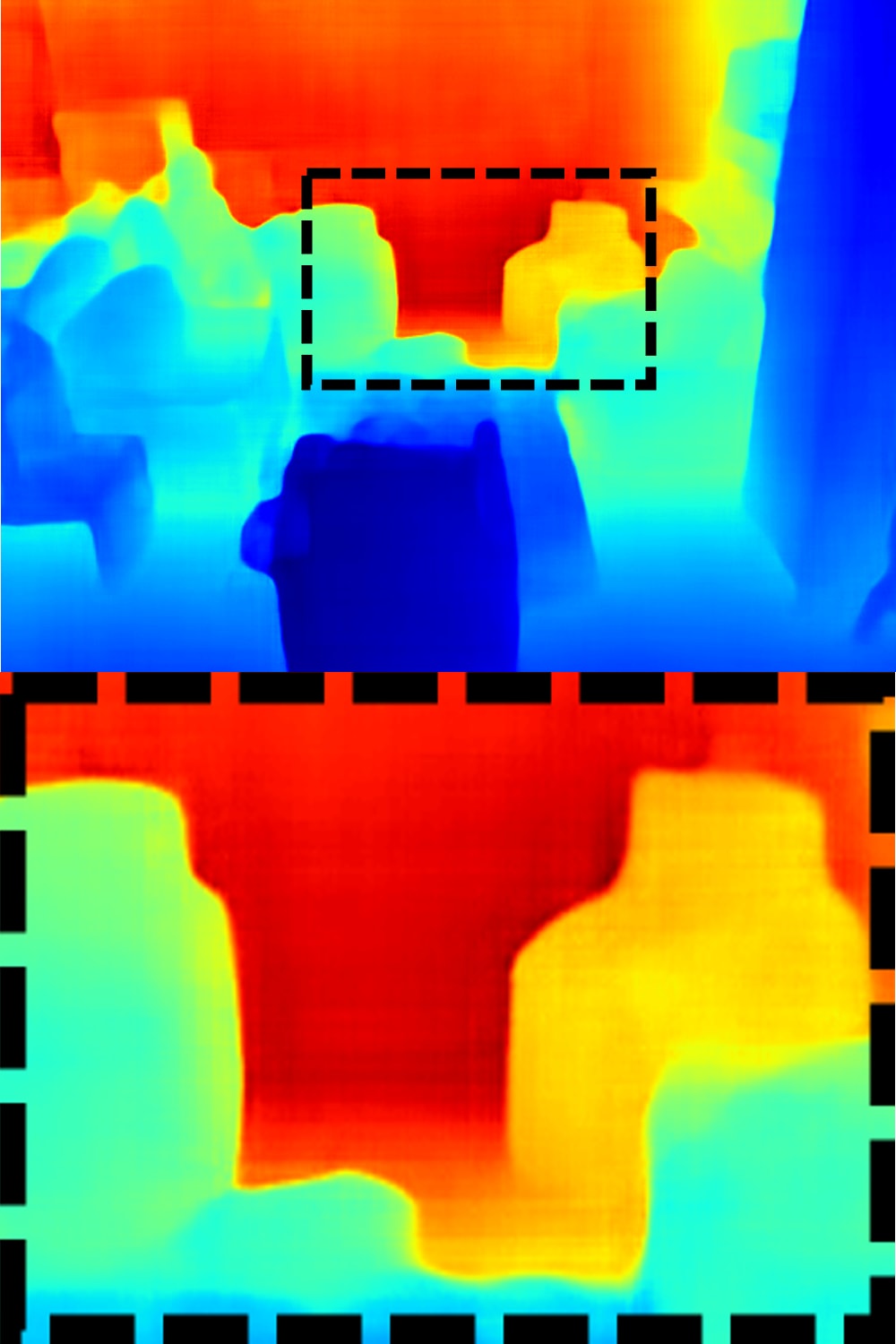}
         \hspace{1em}
         \includegraphics[width=\textwidth]{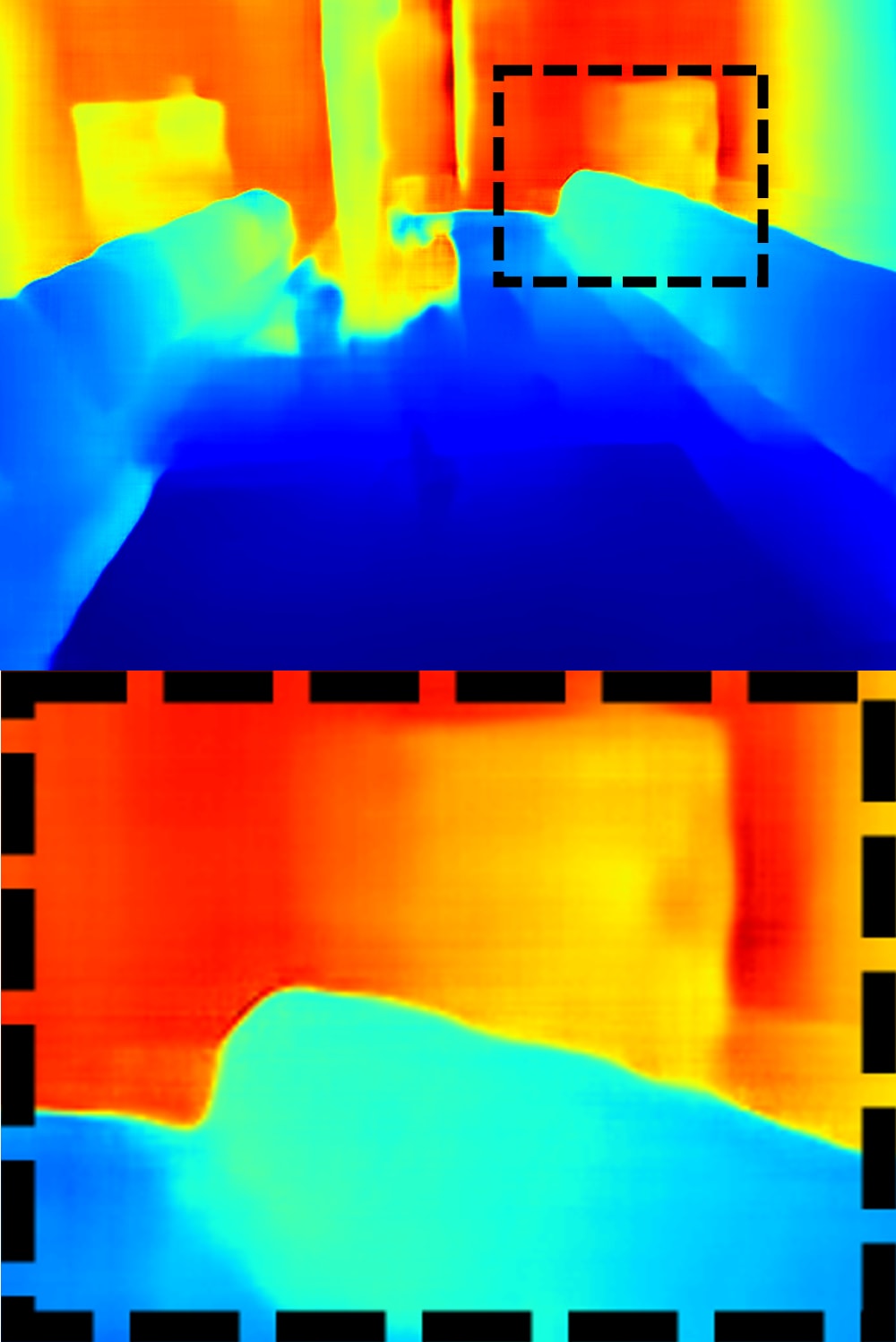}
         \hspace{1em}
         \includegraphics[width=\textwidth]{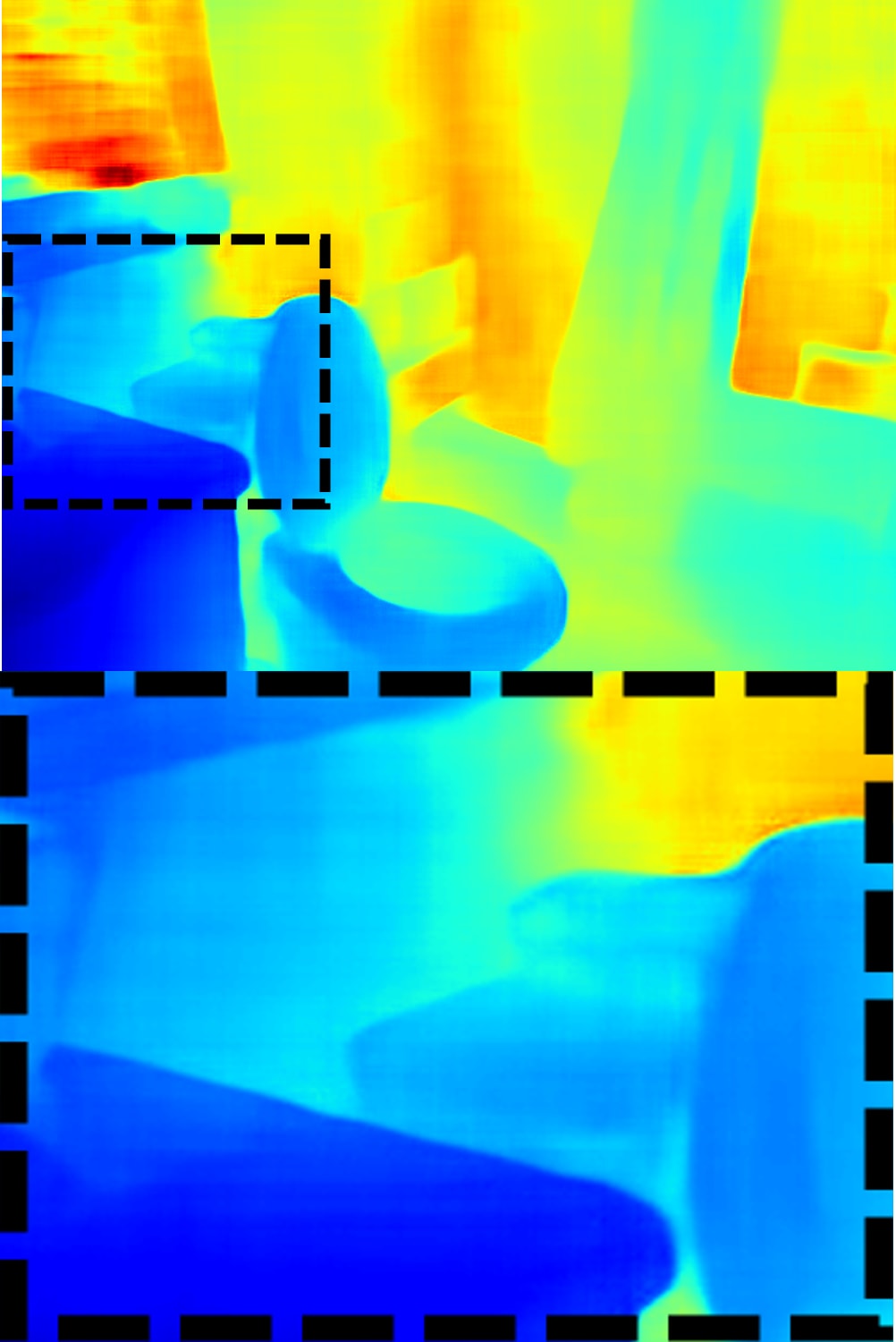}
         \label{subfig2:dsn}
     \end{subfigure}
     \caption{Detailed evaluation of the predictions of our method and the ones retrieved by recent supervised approaches. DSN$\dagger$ refers to the best DSN setup and the RGB inputs compose the NYU Depth V2 dataset \cite{silberman2012indoor}}
     \label{subfig2NYU:all}
\end{figure*}

\begin{figure*}[t!]
     \centering
     \begin{subfigure}[t]{0.13\textwidth}
         \centering
         \caption{RGB Input}
         \includegraphics[width=\textwidth]{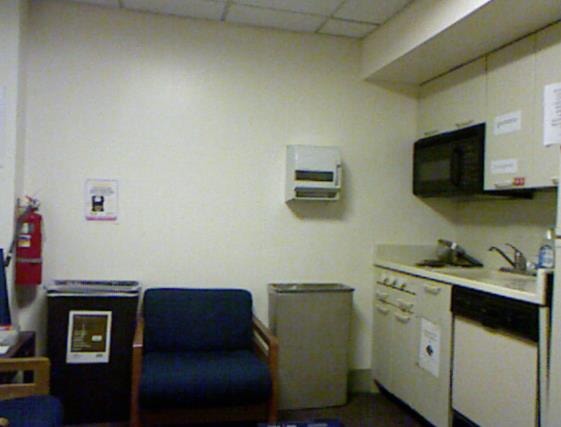}
         \hspace{1em}
         \includegraphics[width=\textwidth]{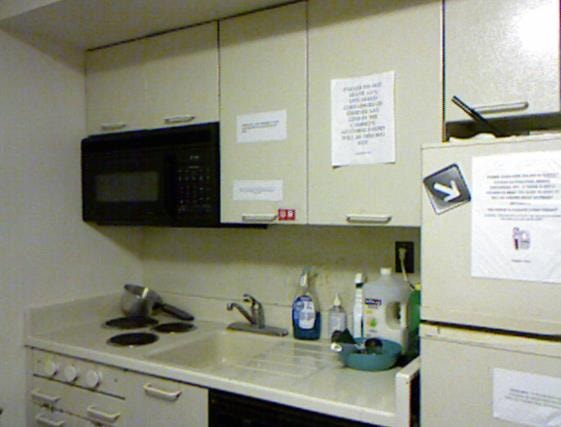}
         \hspace{1em}
         \includegraphics[width=\textwidth]{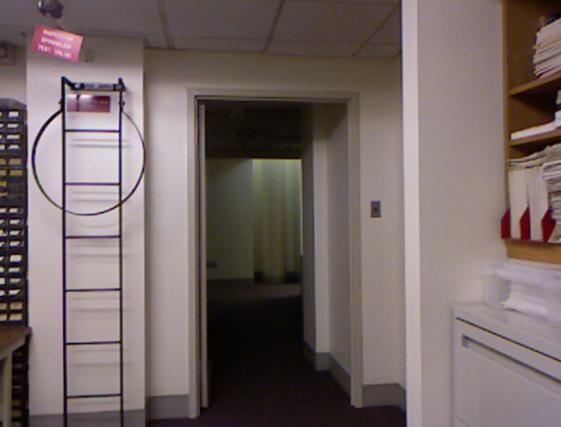}
         \label{subfig3:rgb}
     \end{subfigure}
         \begin{subfigure}[t]{0.13\textwidth}
         \centering
         \caption{Ground Truth}
         \includegraphics[width=\textwidth]{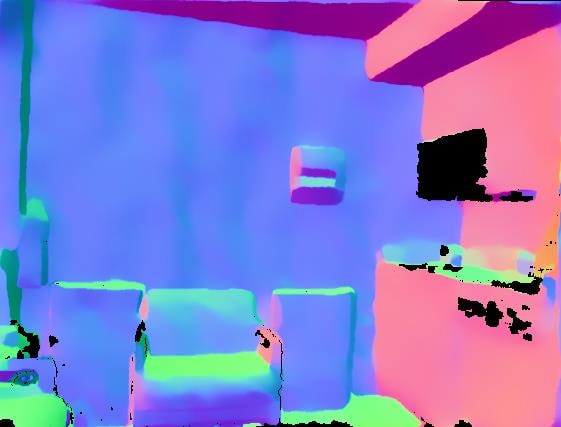}
         \hspace{1em}
         \includegraphics[width=\textwidth]{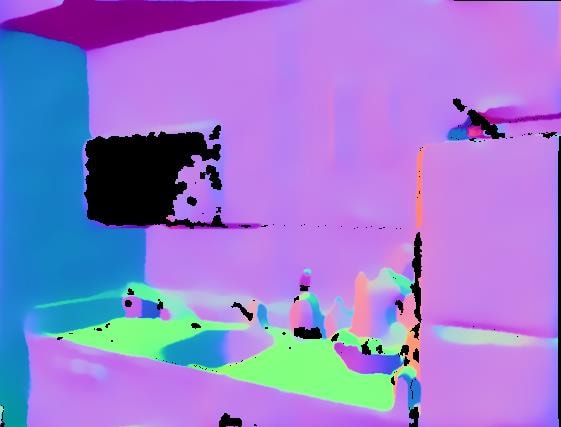}
         \hspace{1em}
         \includegraphics[width=\textwidth]{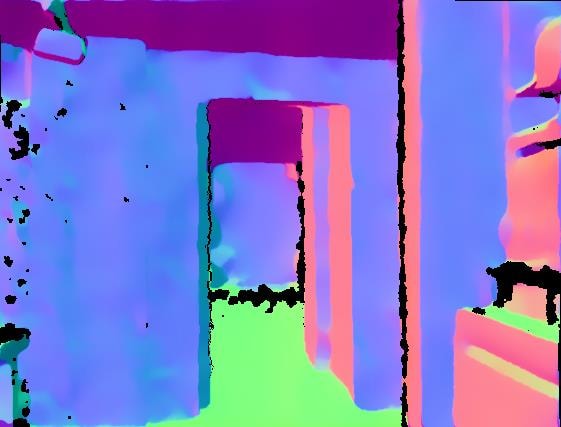}
         \label{subfig3:gt}
     \end{subfigure}
     \begin{subfigure}[t]{0.13\textwidth}
         \centering
         \caption{MS-CNN \cite{eigen2015predicting}}
         \includegraphics[width=\textwidth]{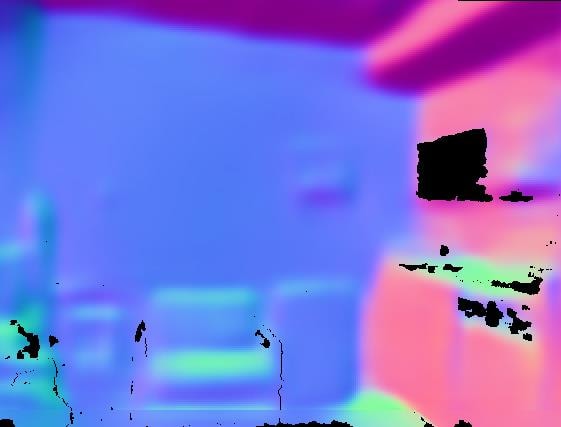}
         \hspace{1em}
         \includegraphics[width=\textwidth]{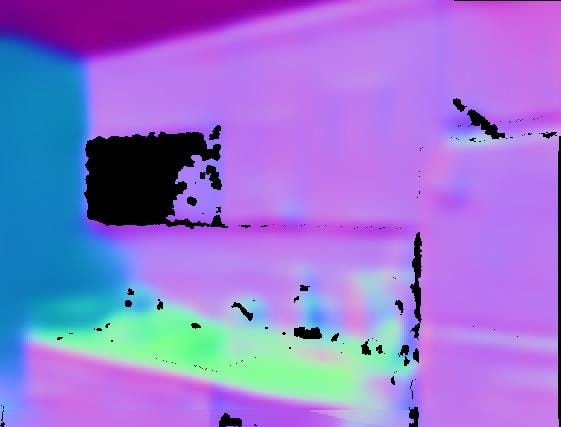}
         \hspace{1em}
         \includegraphics[width=\textwidth]{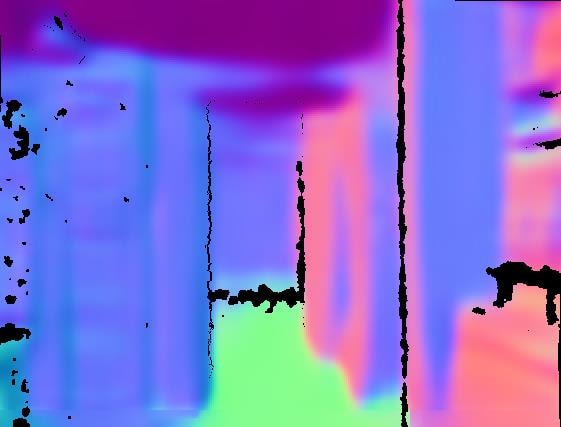}
         \label{subfig3:mscnn}
     \end{subfigure}
          \begin{subfigure}[t]{0.13\textwidth}
         \centering
         \caption{SkipNet \cite{bansal2016marr}}
         \includegraphics[width=\textwidth]{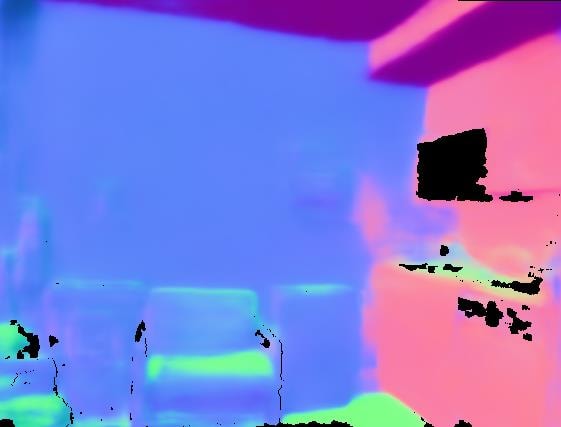}
         \hspace{1em}
         \includegraphics[width=\textwidth]{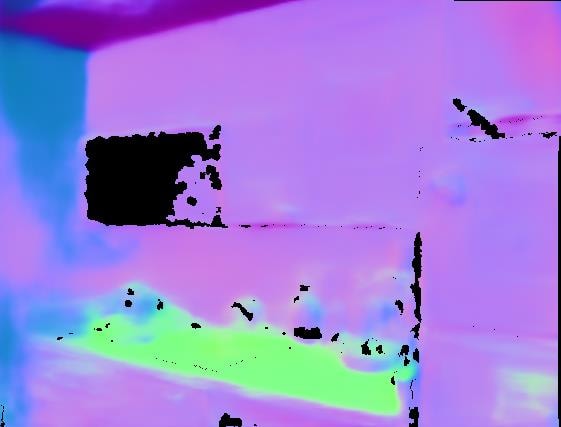}
         \hspace{1em}
         \includegraphics[width=\textwidth]{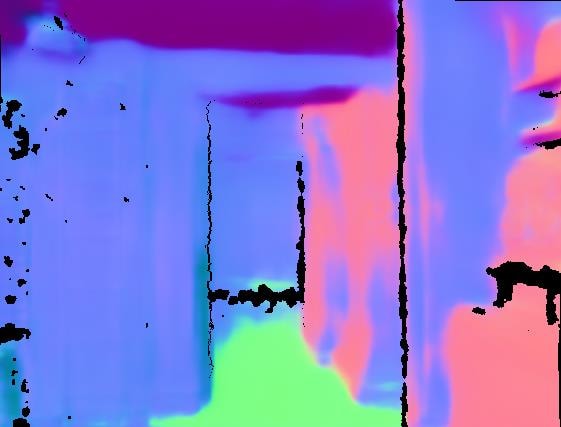}
         \label{subfig3:skipnet}
     \end{subfigure}
         \begin{subfigure}[t]{0.13\textwidth}
         \centering
         \caption{Qi \etal \cite{qi2018geonet}}
         \includegraphics[width=\textwidth]{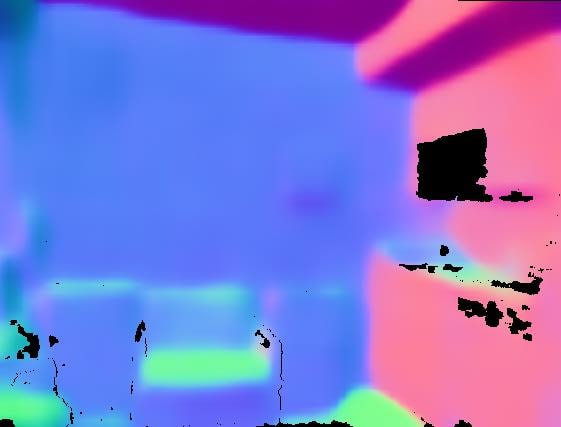}
         \hspace{1em}
         \includegraphics[width=\textwidth]{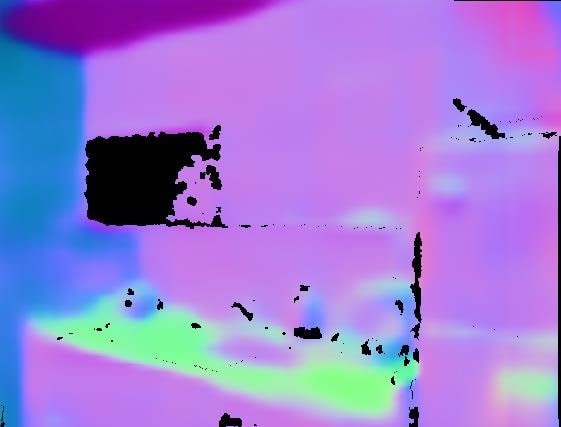}
         \hspace{1em}
         \includegraphics[width=\textwidth]{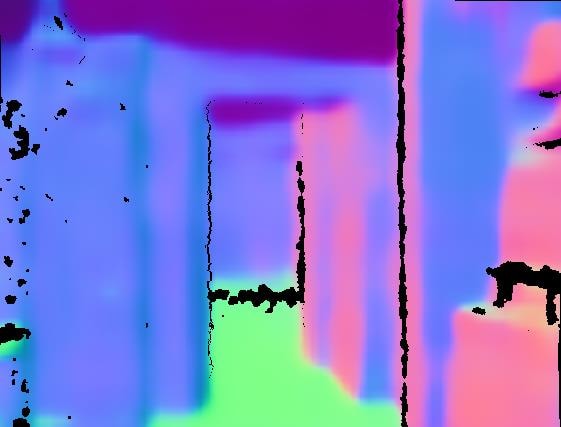}
         \label{subfig3:geonet}
     \end{subfigure}
         \begin{subfigure}[t]{0.13\textwidth}
         \centering
         \caption{PAP \cite{zhang2019pattern}}
         \includegraphics[width=\textwidth]{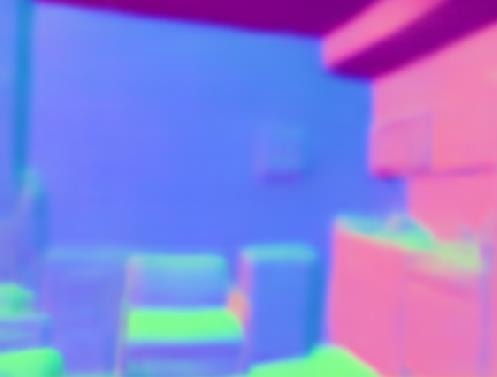}
         \hspace{1em}
         \includegraphics[width=\textwidth]{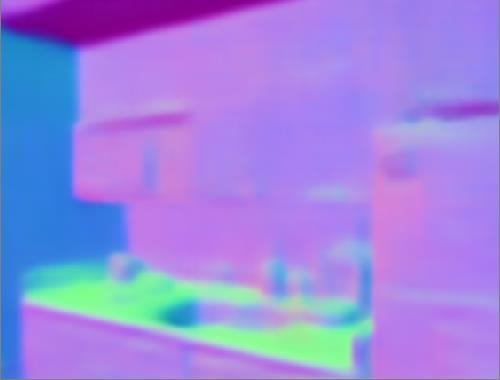}
         \hspace{1em}
         \includegraphics[width=\textwidth]{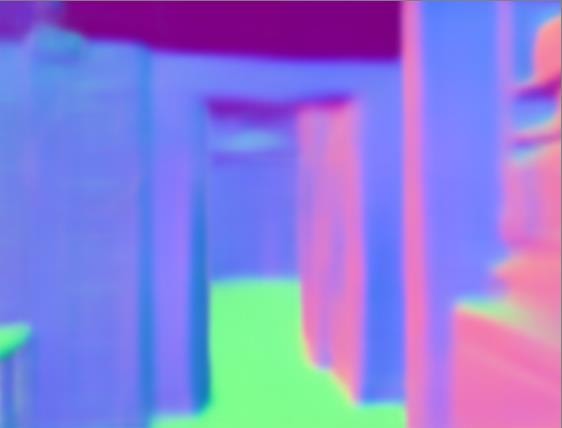}
         \label{subfig3:geonet}
     \end{subfigure}
          \begin{subfigure}[t]{0.13\textwidth}
         \centering
         \caption{DSN$\dagger$}
         \includegraphics[width=\textwidth]{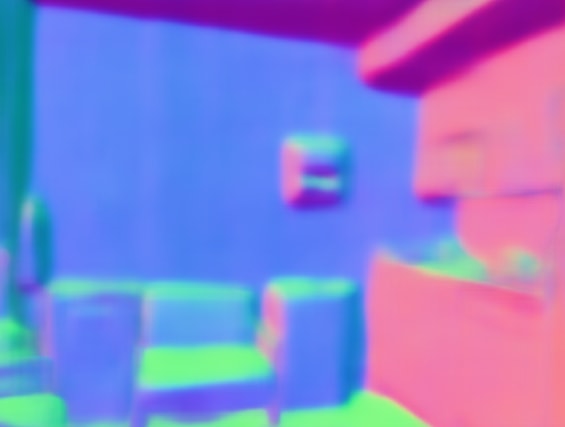}
         \hspace{1em}
         \includegraphics[width=\textwidth]{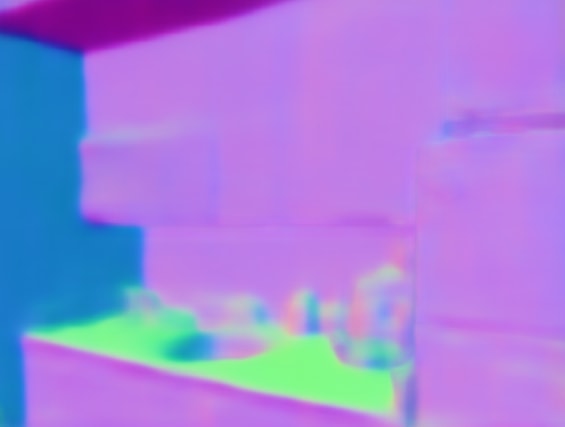}
         \hspace{1em}
         \includegraphics[width=\textwidth]{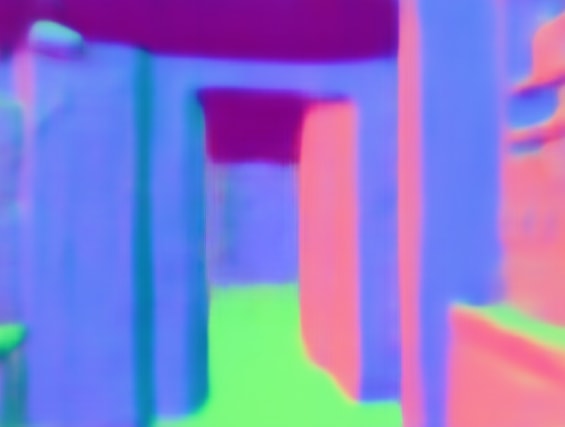}
         \label{subfig3:dsn}
     \end{subfigure}
     \caption{Qualitative comparison of the surface normals maps retrieved by our framework and by state-of-the-art techniques. DSN$\dagger$ corresponds to the best DSN configuration. The RGB inputs are part of the NYU Depth V2 dataset \cite{silberman2012indoor}. We produced the ground truth images through the method of Fouhey \etal\cite{fouhey2013data}}
     \label{subfig3NYU:all}
\end{figure*}

Through the indoor scenes depicted in Figs. \ref{subfig1NYU:all}, \ref{subfig2NYU:all} and \ref{subfig3NYU:all}, we visually compare the most accurate depth and surface normals predictions of the proposed CNN with those obtained by other supervised pipelines from the SIDE literature. By Fig. \ref{subfig1:DSN}, we can notice that our method is capable of better estimating the depth of finer objects and the geometric structure of the elements that compose the scenes, making the predictions closer to the ground truth.

Comparing the depth maps of Fig. \ref{subfig2:dorn} with those illustrated in Figs. \ref{subfig2:dorn} and \ref{subfig2:densed}, it is viable to conclude that our results present greater details for both the closest and the most distant artifacts from the camera. Moreover, Fig. \ref{subfig3NYU:all} shows that the DSN completes the ground truth pixels without information and outputs surface normals vectors more consistent with the local and global layout of the scene.

\begin{figure*}[t!]
     \centering
         \begin{subfigure}[t]{0.19\textwidth}
         \centering
         \caption{RGB input}
         \includegraphics[width=\textwidth]{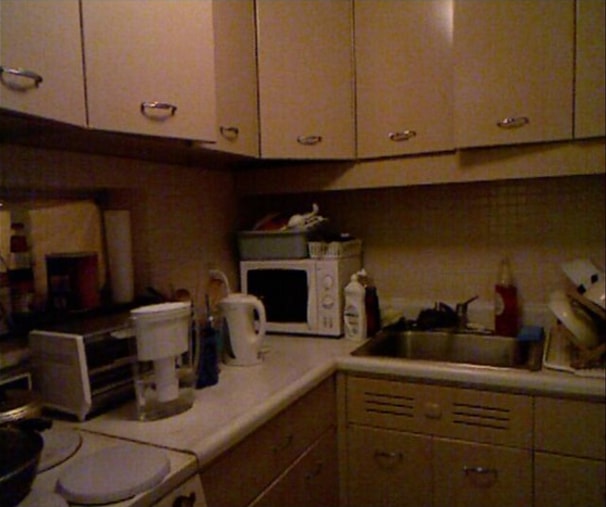}
         \hspace{1em}
         \includegraphics[width=\textwidth]{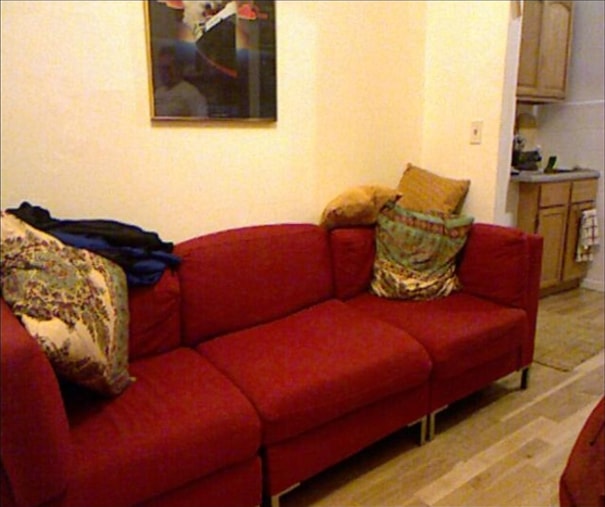}
         \hspace{1em}
         \includegraphics[width=\textwidth]{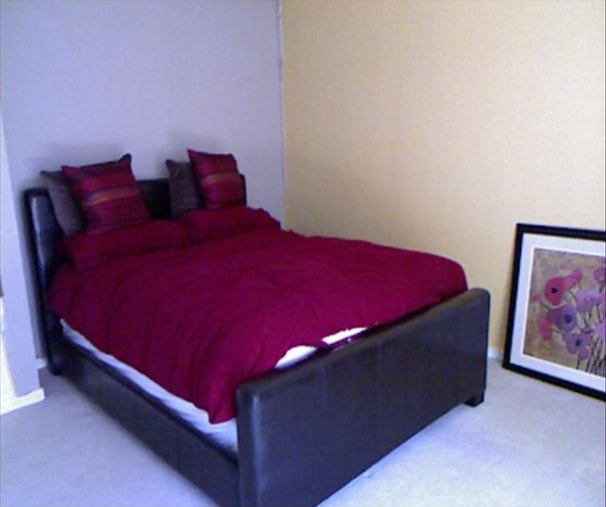}
         \label{subfig3d:0}
     \end{subfigure}
     \begin{subfigure}[t]{0.19\textwidth}
         \centering
         \caption{Depth GT}
         \includegraphics[width=\textwidth]{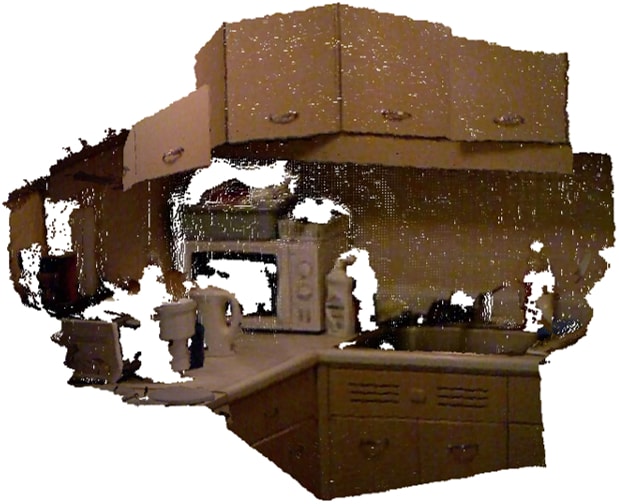}
         \hspace{1em}
         \includegraphics[width=\textwidth]{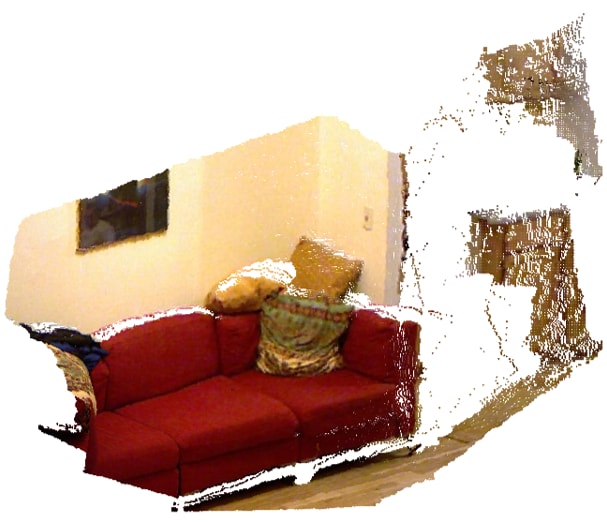}
         \hspace{1em}
         \includegraphics[width=\textwidth]{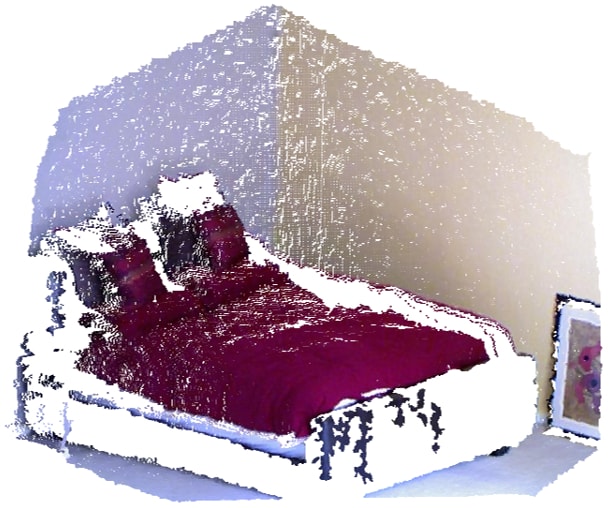}
         \label{subfig3d:1}
     \end{subfigure}
         \begin{subfigure}[t]{0.19\textwidth}
         \centering
         \caption{Depth prediction}
         \includegraphics[width=\textwidth]{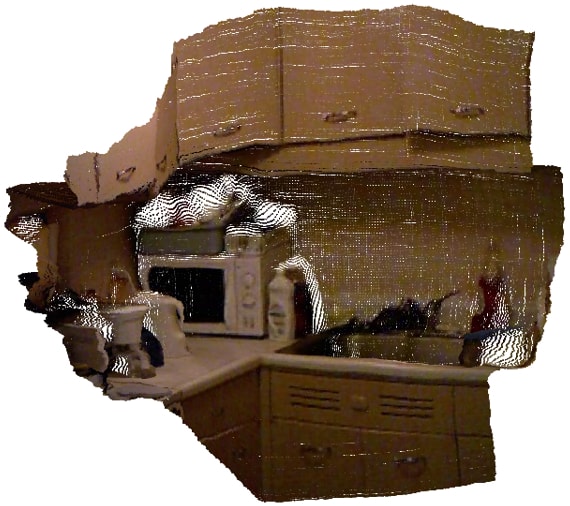}
         \hspace{1em}
         \includegraphics[width=\textwidth]{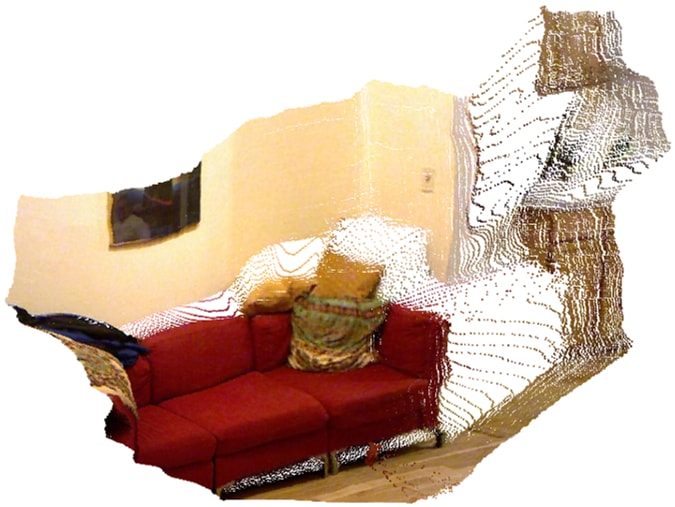}
         \hspace{1em}
         \includegraphics[width=\textwidth]{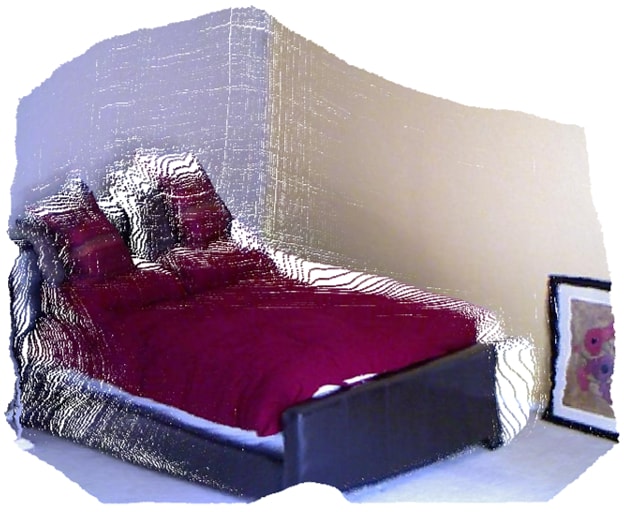}
         \label{subfig3d:2}
     \end{subfigure}
     \begin{subfigure}[t]{0.19\textwidth}
         \centering
         \caption{SN GT}
         \includegraphics[width=\textwidth]{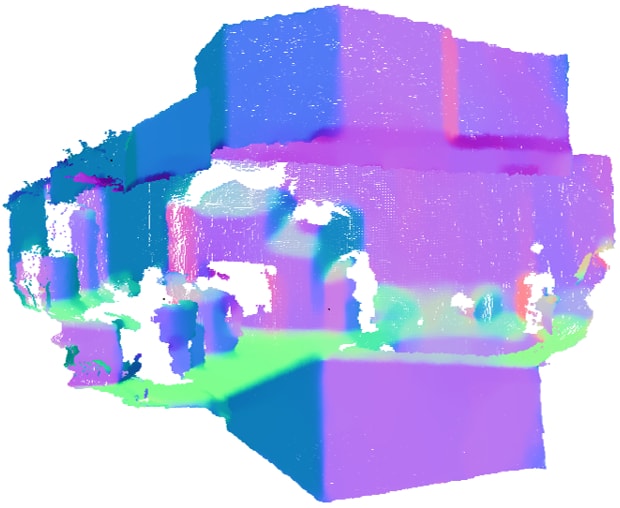}
         \hspace{1em}
         \includegraphics[width=\textwidth]{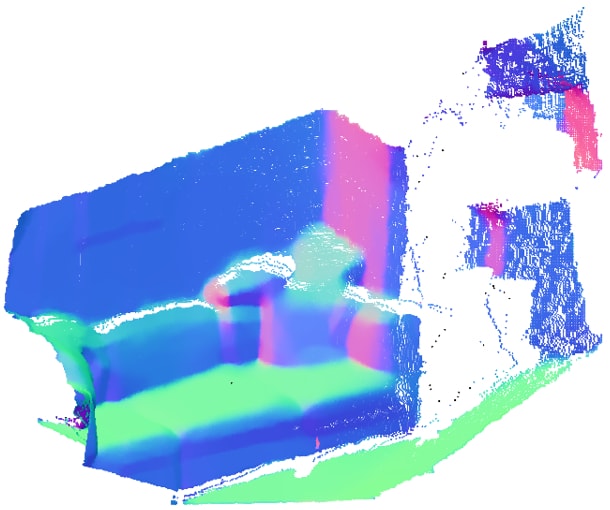}
         \hspace{1em}
         \includegraphics[width=\textwidth]{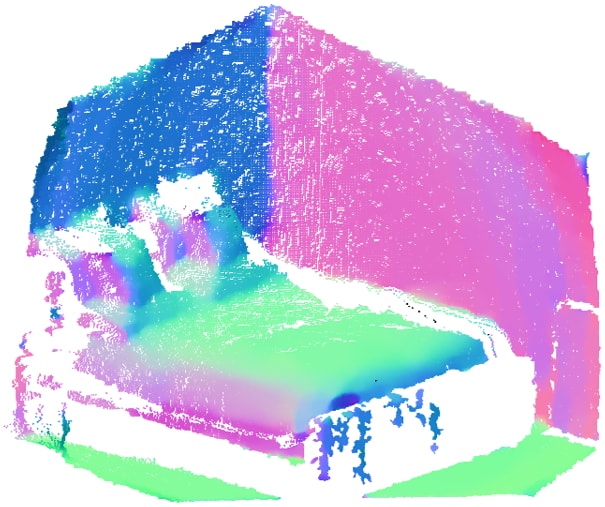}
         \label{subfig3d:3}
     \end{subfigure}
          \begin{subfigure}[t]{0.19\textwidth}
         \centering
         \caption{SN prediction}
         \includegraphics[width=\textwidth]{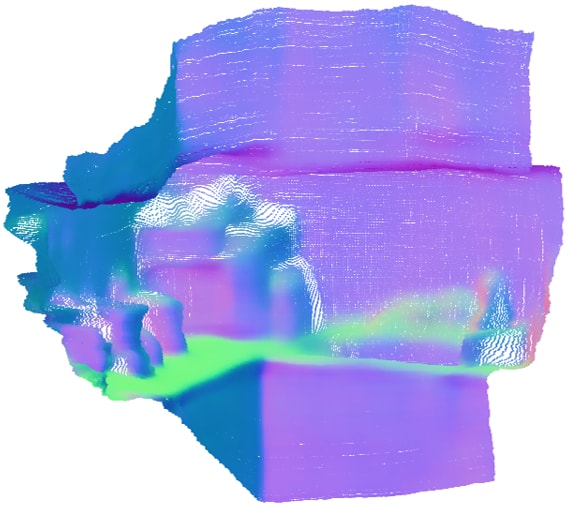}
         \hspace{1em}
         \includegraphics[width=\textwidth]{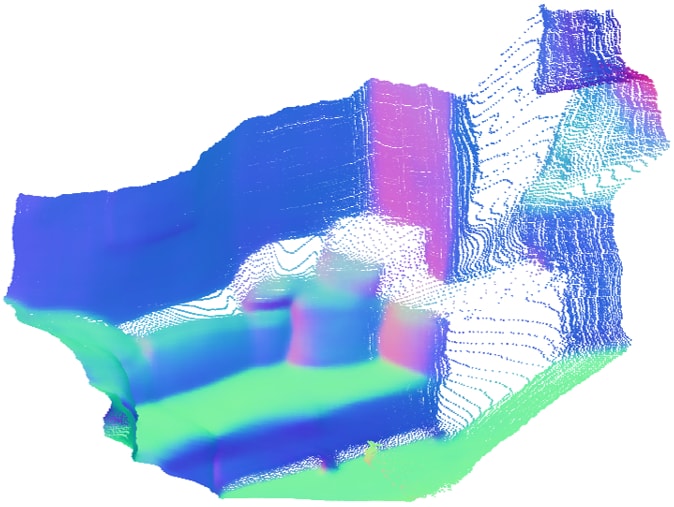}
         \hspace{1em}
         \includegraphics[width=\textwidth]{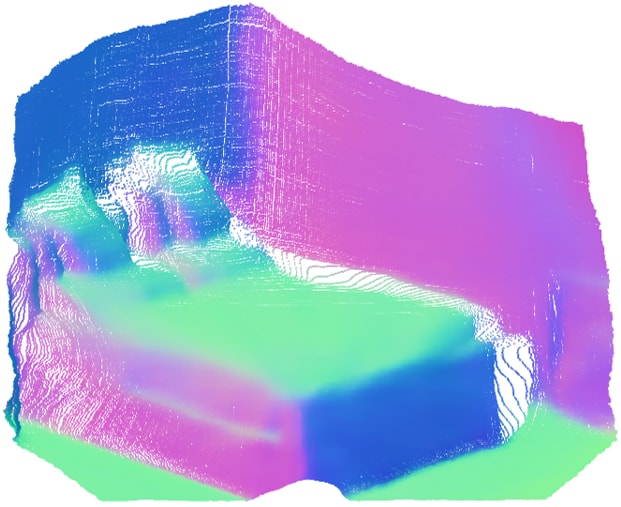}
         \label{subfig3d:4}
     \end{subfigure}
     \caption{Qualitative evaluation of the point clouds reconstructed from the most accurate DSN predictions. The input images and the depth ground truth compose the KITTI Depth dataset \cite{uhrig2017sparsity}. However, we created the reference surface normals data through the algorithm developed by Fouhey \etal\cite{fouhey2013data}}
     \label{subfig3d:all}
\end{figure*}

The point clouds illustrated in Figs. \ref{subfig3d:2} and \ref{subfig3d:4} reveal that our predictions are denser than the reference point clouds and that the estimated pixels do not obey a distorted distribution of the 3D structure which portrays the environments exposed in Fig. \ref{subfig3d:0}. 

\subsubsection{Depth Completion Studies}

\begin{figure}[!htb]
\centering
\begin{tikzpicture}
  \begin{axis}[title={Abs Rel [m]},
    ylabel near ticks,
    xlabel near ticks,
    xlabel={Samples},
    ylabel={Error},
    xmin=0, xmax=500,
    ymin=0.01, ymax=0.05,
    legend pos=north east,
    ymajorgrids=true,
    ymajorgrids=true,
    grid=major,
    name=absrel,
    height=4cm,
    width=4cm]
    \addplot[color=red,mark=triangle*] coordinates{(20,0.046)(50,0.032) (100,0.024)(200,0.018)(500,0.013)};
    \addlegendentry{\tiny Cap 10}
    \addplot[color=green,mark=triangle*] coordinates{(20,0.045)(50,0.031) (100,0.023)(200,0.018)(500,0.013)};
    \addlegendentry{\tiny Cap 5}
    \end{axis}
    \begin{axis}[title={$\delta<1.25$ $[\%]$},
    ylabel near ticks,
    xlabel near ticks,
    xlabel={Samples},
    ylabel={Accuracy},
    xmin=0, xmax=500,
    ymin=96.0, ymax=100.0,
    legend pos=south east,
    ymajorgrids=true,
    ymajorgrids=true,
    grid=major,
    name=sigma1,
    at={($(absrel.east)+(2.0cm,0)$)},
    anchor=west,
    height=4cm,
    width=4cm]
    \addplot[color=blue,mark=square*] coordinates{(20,96.4)(50,97.9) (100,98.6)(200,99.0)(500,99.4)};
    \addlegendentry{\tiny Cap 10}
    \addplot[color=yellow,mark=square*] coordinates{(20,96.5)(50,98.0) (100,98.7)(200,99.0)(500,99.4)};
    \addlegendentry{\tiny Cap 5}
    \end{axis}
    \begin{axis}[title={RMSE [m]},
    ylabel near ticks,
    xlabel near ticks,
    xlabel={Samples},
    ylabel={Error},
    xmin=0, xmax=500,
    ymin=0.09, ymax=0.25,
    legend pos=north east,
    ymajorgrids=true,
    ymajorgrids=true,
    grid=major,
    name=rmse,
    at={($(absrel.south)-(0,2cm)$)},
    anchor=north,
    height=4cm,
    width=4cm]
    \addplot[color=red,mark=triangle*] coordinates{(20,0.221)(50,0.179) (100,0.151)(200,0.132)(500,0.106)};
    \addlegendentry{\tiny Cap 10}
    \addplot[color=green,mark=triangle*] coordinates{(20,0.191)(50,0.153) (100,0.129)(200,0.111)(500,0.090)};
    \addlegendentry{\tiny Cap 5}
    \end{axis}
    \begin{axis}[title={$\delta<1.25^2$ $[\%]$},
    ylabel near ticks,
    xlabel near ticks,
    xlabel={Samples},
    ylabel={Accuracy},
    xmin=0, xmax=500,
    ymin=99.0, ymax=100,
    legend pos=south east,
    ymajorgrids=true,
    ymajorgrids=true,
    grid=major,
    name=sigma2,
    at={($(sigma1.south)-(0,2cm)$)},
    anchor=north,
    height=4cm,
    width=4cm]
    \addplot[color=blue,mark=square*] coordinates{(20,99.2)(50,99.5) (100,99.6)(200,99.7)(500,99.8)};
    \addlegendentry{\tiny Cap 10}
    \addplot[color=yellow,mark=square*] coordinates{(20,99.2)(50,99.5) (100,99.7)(200,99.8)(500,99.8)};
    \addlegendentry{\tiny Cap 5}
  \end{axis}
\end{tikzpicture}
\caption{Evolution of the depth completion metrics obtained by the DSN. We employed the reference depth maps of the NYU Depth V2 dataset \cite{silberman2012indoor} to plot the results}
\label{graph:nyulidar}
\end{figure}
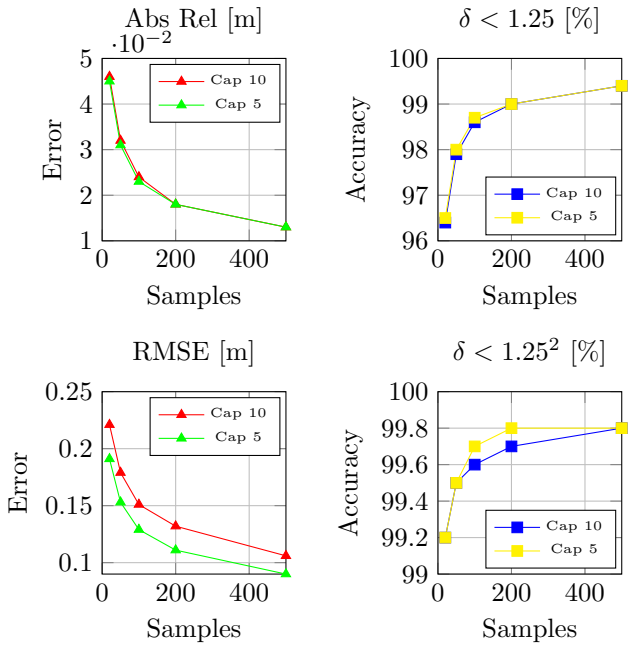

To analyze the performance of our depth completion framework in indoor scenarios, we conducted studies with the NYU Depth V2 dataset \cite{silberman2012indoor} wherein we fed the DSN with sparse depth data, sampled from the ground truth, and RGB images to retrieve dense depth maps. The results of our CNN are presented in Table \ref{tab:nyulidar}. We may observe a similar behavior of the errors and accuracy metrics than those acquired in the outdoor studies with the same depth completion network. Fig. \ref{graph:nyulidar} illustrates the DSNs response to the application of the increasing amount of sparse samples in its input.

We compare the results generated by the best DSN setup (DSN$\dagger$) with the ones acquired from recent depth completion methods through Table \ref{tab:nyulidarsoa}. Considering both Abs Rel and RMSE metrics, we achieve the top-2 in the benchmark ranking of Table \ref{tab:nyulidarsoa}. However, the Non-local CSPN technique \cite{park2020non} has a higher number of trainable parameters (25\ M) and it applies a more complex feature extraction CNN (ResNet-34).

\begin{figure*}[t!]
     \centering
          \begin{subfigure}[t]{0.12\textwidth}
         \centering
         \caption{RGB}
         \includegraphics[width=\textwidth]{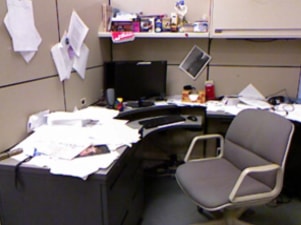}
         \hspace{1em}
         \includegraphics[width=\textwidth]{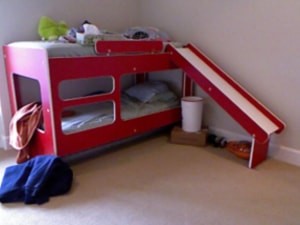}
         \label{nyulidarsubfig0:rgb}
     \end{subfigure}
     \begin{subfigure}[t]{0.12\textwidth}
         \centering
         \caption{500d}
         \includegraphics[width=\textwidth]{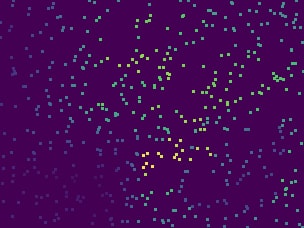}
         \hspace{1em}
         \includegraphics[width=\textwidth]{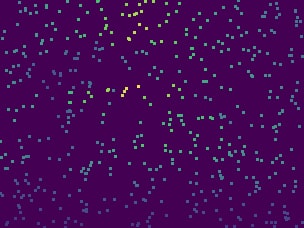}
         \label{nyulidarsubfig0:lidar}
     \end{subfigure}
         \begin{subfigure}[t]{0.12\textwidth}
         \centering
         \caption{GT}
         \includegraphics[width=\textwidth]{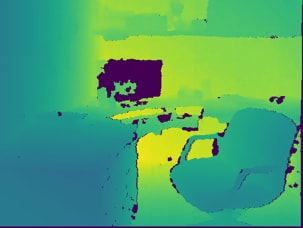}
         \hspace{1em}
         \includegraphics[width=\textwidth]{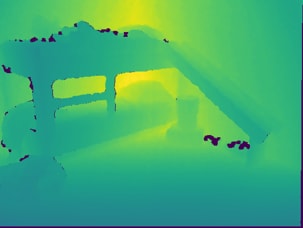}
         \label{nyulidarsubfig0:gt}
     \end{subfigure}
         \begin{subfigure}[t]{0.12\textwidth}
         \centering
         \caption{Ma \etal\cite{ma2018sparse}}
         \includegraphics[width=\textwidth]{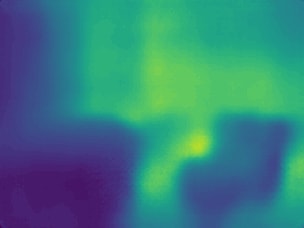}
         \hspace{1em}
         \includegraphics[width=\textwidth]{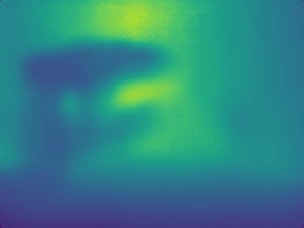}
         \label{nyulidarsubfig0:ma}
     \end{subfigure}
         \begin{subfigure}[t]{0.12\textwidth}
         \centering
         \caption{CSPN \cite{cheng2018depth}}
         \includegraphics[width=\textwidth]{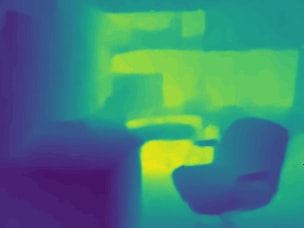}
         \hspace{1em}
         \includegraphics[width=\textwidth]{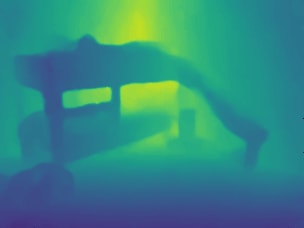}
         \label{nyulidarsubfig0:cheng}
     \end{subfigure}
         \begin{subfigure}[t]{0.12\textwidth}
         \centering
         \caption{NLSPN \cite{park2020non}}
         \includegraphics[width=\textwidth]{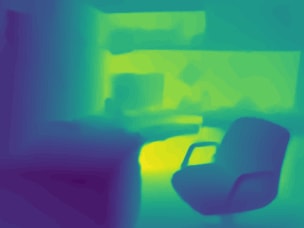}
         \hspace{1em}
         \includegraphics[width=\textwidth]{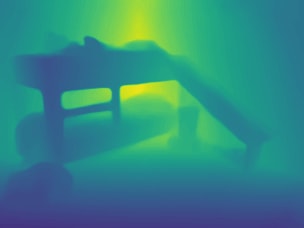}
         \label{nyulidarsubfig0:park}
     \end{subfigure}
         \begin{subfigure}[t]{0.12\textwidth}
         \centering
         \caption{DSN$\dagger$}
         \includegraphics[width=\textwidth]{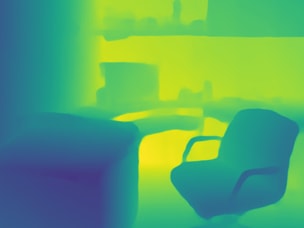}
         \hspace{1em}
         \includegraphics[width=\textwidth]{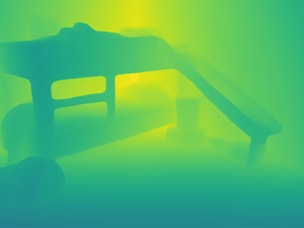}
         \label{nyulidarsubfig0:dsn}
     \end{subfigure}
         \begin{subfigure}[t]{0.12\textwidth}
         \centering
         \caption{DSN$\dagger$}
         \includegraphics[width=\textwidth]{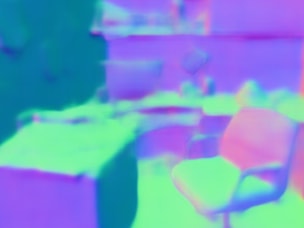}
         \hspace{1em}
         \includegraphics[width=\textwidth]{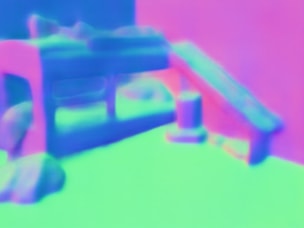}
         \label{nyulidarsubfig0:dsn}
     \end{subfigure}
     \caption{Comparison between the dense depth maps retrieved by our best depth completion CNN (DSN$\dagger$) and by other methods in the literature. We also present the surface normals maps estimated by the DSN$\dagger$. The NYU Depth V2 \cite{silberman2012indoor} is employed by all approaches. Legend: RGB - input image; $500$d - sparse input data; GT - sparse ground truth} 
     \label{nyulidarsubfig0:all}
\end{figure*}

Figure \ref{nyulidarsubfig0:all} shows the predictions produced by the DSN and by state-of-the-art techniques that address the depth completion task. Through this figure, it is plausible to state that our results are the ones that come closest to the reference. We also present the surface normals maps estimated by the DSN which are visually consistent with the output depth maps. Furthermore, the metrics presented in Table \ref{tab:nyusnlidar} indicate that, as with depth predictions, the quality of the surface normals maps increases as more sparse depth data is added to the network's input. 

\begin{table}[]
\centering
\caption{NYU Depth V2 \cite{silberman2012indoor} metrics obtained by our depth completion pipeline. The pattern DSN $3$ $+$ Pyramid$^1$ $+$ Indoor, jointly with the $\mathcal{L}_1^+$ function, is selected for all the experiments that do not involve the surface normals ground truth. DSN $3$ $+$ Pyramid$^1$ $+$ Indoor $+$ SN$\dagger$ is the DSN setup with the surface normals module which yields the most accurate results}
\label{tab:nyulidar}
\resizebox{\linewidth}{!}{
\begin{tabular}{c|c|c|ccc|ccc}
\hline
\textbf{Method} & \textbf{Cap} & \textbf{Samples} & \textbf{Abs Rel}$\downarrow$ & \textbf{RMSE}$\downarrow$ & \textbf{log10}$\downarrow$ & $\bm{\delta < 1.25}$$\uparrow$ & $\bm{\delta < 1.25^2}$$\uparrow$ & $\bm{\delta < 1.25^3}$$\uparrow$ \\\hline \hline
DSN 3 $+$ Pyramid$^1$ $+$ Indoor & 10 & 20 & 0.046 & 0.221 & 0.020 & 0.964 & 0.992 & 0.998\\
DSN 3 $+$ Pyramid$^1$ $+$ Indoor & 10 & 50 & 0.032 & 0.179 & 0.014 & 0.979 & 0.995 & 0.998\\
DSN 3 $+$ Pyramid$^1$ $+$ Indoor & 10 & 100 & 0.024 & 0.151 & 0.010 & 0.986 & 0.996 & 0.999\\
DSN 3 $+$ Pyramid$^1$ $+$ Indoor & 10 & 200 & 0.018 & 0.132 & 0.008 & 0.990 & 0.997 & 0.999\\
DSN 3 $+$ Pyramid$^1$ $+$ Indoor & 10 & 500 & 0.013 & 0.106 & 0.006 & \textbf{0.994} & 0.998 & 0.999\\
DSN 3 $+$ Pyramid$^1$ $+$ Indoor $+$ SN$\dagger$ & 10 & 200 & 0.017 & 0.128 & 0.007 & 0.991 & 0.998 & 0.999\\
DSN 3 $+$ Pyramid$^1$ $+$ Indoor $+$ SN$\dagger$ & 10 & 500 & \textbf{0.012} & \textbf{0.102} & \textbf{0.005} & \textbf{0.994} & \textbf{0.999} & \textbf{1.000}\\\hline\hline
DSN 3 $+$ Pyramid$^1$ $+$ Indoor & 5 & 20 & 0.045 & 0.191 & 0.019 & 0.965 & 0.992 & 0.998\\
DSN 3 $+$ Pyramid$^1$ $+$ Indoor & 5 & 50 & 0.031 & 0.153 & 0.013 & 0.980 & 0.995 & 0.999\\
DSN 3 $+$ Pyramid$^1$ $+$ Indoor & 5 & 100 & 0.023 & 0.129 & 0.010 & 0.987 & 0.997 & 0.999\\
DSN 3 $+$ Pyramid$^1$ $+$ Indoor & 5 & 200 & 0.018 & 0.111 & 0.008 & 0.990 & 0.998 & 0.999\\
DSN 3 $+$ Pyramid$^1$ $+$ Indoor & 5 & 500 & 0.013 & 0.090 & \textbf{0.005} & \textbf{0.994} & 0.998 & \textbf{1.000}\\
DSN 3 $+$ Pyramid$^1$ $+$ Indoor $+$ SN$\dagger$ & 5 & 200 & 0.017 & 0.107 & 0.007 & 0.991 & 0.998 & 0.999\\
DSN 3 $+$ Pyramid$^1$ $+$ Indoor $+$ SN$\dagger$ & 5 & 500 & \textbf{0.012} & \textbf{0.086} & \textbf{0.005} & \textbf{0.994} & \textbf{0.999} & \textbf{1.000}\\\hline
\end{tabular}
}
\end{table}

\begin{table}[]
\centering
\caption{Quantitative evaluation of our best depth completion framework (DSN$\dagger$) and other techniques in the state-of-the-art. The NYU Depth V2 \cite{silberman2012indoor} is used to compute all the metrics}
\label{tab:nyulidarsoa}
\resizebox{\linewidth}{!}{
\begin{tabular}{c|c|cc|ccc}
\hline
\textbf{Method} & \textbf{Samples} & \textbf{Abs Rel}$\downarrow$ & \textbf{RMSE}$\downarrow$ & $\bm{\delta < 1.25}$$\uparrow$ & $\bm{\delta < 1.25^2}$$\uparrow$ & $\bm{\delta < 1.25^3}$$\uparrow$\\\hline \hline
Ma \etal \cite{ma2018sparse} & 200 & 0.044 & 0.230 & 0.971 & 0.994 & 0.998\\
NConv \cite{eldesokey2019confidence} & 200 & 0.027 & 0.173 & 0.982 & 0.996 & 0.999\\
GuideNet \cite{tang2019learning} & 200 & 0.024 & 0.142 & 0.988 & \textbf{0.998} & \textbf{1.000}\\
DSN$\dagger$ & 200 & \textbf{0.017} & \textbf{0.128} & \textbf{0.991} & \textbf{0.998} & 0.999\\ \hline\hline
Silberman \etal \cite{silberman2012indoor} & 500 & 0.084 & 0.479 & 0.924 & 0.976 & 0.989\\
TGV \cite{ferstl2013image} & 500 & 0.123 & 0.635 & 0.819 & 0.930 & 0.968\\
Zhang \etal \cite{zhang2018deepd} & 500 & 0.042 & 0.228 & 0.971 & 0.993 & 0.997\\
Ma \etal \cite{ma2018sparse} & 500 & 0.043 & 0.204 & 0.978 & 0.996 & 0.999\\
Ma \etal \cite{ma2018sparse} + Bilateral \cite{barron2016fast} & 500 & 0.084 & 0.479 & 0.924 & 0.976 & 0.989\\
Ma \etal \cite{ma2018sparse} + SPN \cite{liu2017learning} & 500 & 0.031 & 0.172 & 0.983 & 0.997 & 0.999\\
NConv \cite{eldesokey2019confidence} & 500 & 0.018 & 0.129 & 0.990 & 0.998 & \textbf{1.000}\\
CSPN \cite{cheng2018depth} & 500 & 0.016 & 0.117 & 0.992 & \textbf{0.999} & \textbf{1.000}\\
Imran \etal \cite{imran2019depth} & 500 & 0.013 & 0.118 & 0.994 & \textbf{0.999} & -\\
CSPN++ \cite{cheng2019cspn++} & 500 & - & 0.116 & - & - & - \\
DeepLiDAR \cite{qiu2019deeplidar} & 500 & 0.022 & 0.115 & 0.993 & \textbf{0.999} & \textbf{1.000}\\
Xu \etal \cite{xu2019depth} & 500 & 0.018 & 0.112 & 0.995 & \textbf{0.999} &  \textbf{1.000}\\
GuideNet \cite{tang2019learning} & 500 & 0.015 & 0.101 & 0.995 & \textbf{0.999} & \textbf{1.000}\\
DSN$\dagger$ & 500 & \textbf{0.012} & 0.102 & 0.994 & \textbf{0.999} & \textbf{1.000}\\
NLSPN \cite{park2020non} & 500 & \textbf{0.012} & \textbf{0.092} & \textbf{0.996} & \textbf{0.999} & \textbf{1.000}\\
\hline
\end{tabular}
}
\end{table}

\begin{table}[]
\centering
\caption{Analysis of the surface normals predictions from our best depth completion CNN (DSN$\dagger$) when it receives different amounts of sparse samples as input. Mask stands for the operation of masking out the invalid pixels corresponding to the surface normals ground truth}
\label{tab:nyusnlidar}
\resizebox{\linewidth}{!}{
\begin{tabular}{c|c|ccc|ccc}
\hline
\textbf{Method} & \textbf{Samples} & \textbf{Mean}$\downarrow$ & \textbf{Median}$\downarrow$ & \textbf{RMSE}$\downarrow$ & $\bm{11.25^{\circ}}$$\uparrow$ & $\bm{22.50^{\circ}}$$\uparrow$ & $\bm{30^{\circ}}$$\uparrow$\\\hline \hline
DSN$\dagger$ $+$ Mask & 200 & 7.6 & 3.5 & 12.2 & 77.1 & 91.0 & 95.6\\
DSN$\dagger$ & 200 & 17.8 & 4.6 & 33.7 & 67.6 & 79.7 & 83.7\\
DSN$\dagger$ $+$ Mask & 500 & \textbf{6.1} & \textbf{3.0} & \textbf{9.6} & \textbf{82.9} & \textbf{94.9} & \textbf{98.1}\\
DSN$\dagger$ & 500 & 16.5 & 3.9 & 33.0 & 72.6 & 83.2 & 86\\
\hline
\end{tabular}
}%
\end{table}


\section{Conclusions}

Through the conducted surveys, we can imply that the monocular depth estimation and depth completion areas have significantly advanced in recent years due to the emergence of Deep Learning methods. These methods may overcome the performance of classical Computer Vision techniques and also generate better predictions for other fundamental tasks such as semantic segmentation, surface normals estimation and VO.

Particularly, considering all the experimental analysis, we designed a lightweight CNN architecture, whose number of trainable parameters can vary from 2\ M up to 12\ M. Our framework can also be used in real-world applications since it achieves prediction speeds from 32 fps up to 88 fps.

For the tests involving outdoor scenarios, the DenseSIDENet configuration that includes the proposed surface normals module, the 2.5D geometric loss function, the pyramid block and the pre-training with the cityscapes \cite{cordts2016cityscapes} dataset provided results in the state-of-the-art for the SIDE ($Abs\ Rel = 0.075$), depth completion ($Abs\ Rel = 0.019$) and VO ($seq00\ RMSE = 16.9438$) tasks, regarding the KITTI Benchmark datasets \cite{uhrig2017sparsity,Geiger2012CVPR,Alhaija2018IJCV}.

On the other hand, for the experiments involving indoor environments, the same DSN setup aforementioned, with the exception of the pre-training step that is conducted with the aid of indoor datasets \cite{song2015sun,janoch2013category,xiao2013sun3d}, produced promising results for the single-view depth prediction ($RMSE = 0.429$), depth completion ($RMSE = 0.102$) and surface normals estimation ($RMSE = 12.2$) tasks on the NYU Benchmark datasets \cite{silberman2012indoor}.

Therefore, we were able to determine optimal training conditions, using different deep learning techniques, as well as optimized network structures that enabled the generation of high-quality predictions in reduced processing time. As future work, we intend to employ the developed CNNs for other tasks such as robotic grasping and visual servoing control.

\section*{Acknowledgment}

This research was financed in part by the Brazilian National Council for Scientific and Technological Development - CNPq (grants 465755/2014-3 and 130602/2018-3), by the Coordination of Improvement of Higher Education Personnel - Brazil - CAPES (Finance Code 001, grants 88887.136349/2017-00 and 88887.339070/2019-00), and the S\~{a}o Paulo Research Foundation - FAPESP (grant {2014/50851-0}).


\begin{thebibliography}{100}
\expandafter\ifx\csname url\endcsname\relax
  \def\url#1{\texttt{#1}}\fi
\expandafter\ifx\csname urlprefix\endcsname\relax\def\urlprefix{URL }\fi
\expandafter\ifx\csname href\endcsname\relax
  \def\href#1#2{#2} \def\path#1{#1}\fi

\bibitem{mancini2017toward}
M.~Mancini, G.~Costante, P.~Valigi, T.~A. Ciarfuglia, J.~Delmerico,
  D.~Scaramuzza, Toward domain independence for learning-based monocular depth
  estimation, IEEE Robotics and Automation Letters 2~(3) (2017) 1778--1785.

\bibitem{wang2016surge}
P.~Wang, X.~Shen, B.~Russell, S.~Cohen, B.~Price, A.~L. Yuille, Surge: Surface
  regularized geometry estimation from a single image, in: Advances in Neural
  Information Processing Systems, 2016, pp. 172--180.

\bibitem{ramamonjisoa2019sharpnet}
M.~Ramamonjisoa, V.~Lepetit, Sharpnet: Fast and accurate recovery of occluding
  contours in monocular depth estimation, in: Proceedings of the IEEE
  International Conference on Computer Vision Workshops, 2019, pp. 0--0.

\bibitem{chang2019deep}
J.~Chang, G.~Wetzstein, Deep optics for monocular depth estimation and 3d
  object detection, in: Proceedings of the IEEE International Conference on
  Computer Vision, 2019, pp. 10193--10202.

\bibitem{loo2019cnn}
S.~Y. Loo, A.~J. Amiri, S.~Mashohor, S.~H. Tang, H.~Zhang, Cnn-svo: Improving
  the mapping in semi-direct visual odometry using single-image depth
  prediction, in: 2019 International Conference on Robotics and Automation
  (ICRA), IEEE, 2019, pp. 5218--5223.

\bibitem{tateno2017cnn}
K.~Tateno, F.~Tombari, I.~Laina, N.~Navab, Cnn-slam: Real-time dense monocular
  slam with learned depth prediction, in: Proceedings of the IEEE Conference on
  Computer Vision and Pattern Recognition, 2017, pp. 6243--6252.

\bibitem{ma2018sparse}
F.~Ma, S.~Karaman, Sparse-to-dense: Depth prediction from sparse depth samples
  and a single image, in: 2018 IEEE International Conference on Robotics and
  Automation (ICRA), IEEE, 2018, pp. 1--8.

\bibitem{van2019sparse}
W.~Van~Gansbeke, D.~Neven, B.~De~Brabandere, L.~Van~Gool, Sparse and noisy
  lidar completion with rgb guidance and uncertainty, in: 2019 16th
  International Conference on Machine Vision Applications (MVA), IEEE, 2019,
  pp. 1--6.

\bibitem{ma2019self}
F.~Ma, G.~V. Cavalheiro, S.~Karaman, Self-supervised sparse-to-dense:
  Self-supervised depth completion from lidar and monocular camera, in: 2019
  International Conference on Robotics and Automation (ICRA), IEEE, 2019, pp.
  3288--3295.

\bibitem{guizilini2020semantically}
V.~Guizilini, R.~Hou, J.~Li, R.~Ambrus, A.~Gaidon, Semantically-guided
  representation learning for self-supervised monocular depth, arXiv preprint
  arXiv:2002.12319.

\bibitem{yin2019enforcing}
W.~Yin, Y.~Liu, C.~Shen, Y.~Yan, Enforcing geometric constraints of virtual
  normal for depth prediction, in: Proceedings of the IEEE International
  Conference on Computer Vision, 2019, pp. 5684--5693.

\bibitem{chen2017deeplab}
L.-C. Chen, G.~Papandreou, I.~Kokkinos, K.~Murphy, A.~L. Yuille, Deeplab:
  Semantic image segmentation with deep convolutional nets, atrous convolution,
  and fully connected crfs, IEEE transactions on pattern analysis and machine
  intelligence 40~(4) (2017) 834--848.

\bibitem{lee2019big}
J.~H. Lee, M.-K. Han, D.~W. Ko, I.~H. Suh, From big to small: Multi-scale local
  planar guidance for monocular depth estimation, arXiv preprint
  arXiv:1907.10326.

\bibitem{xu2017multi}
D.~Xu, E.~Ricci, W.~Ouyang, X.~Wang, N.~Sebe, Multi-scale continuous crfs as
  sequential deep networks for monocular depth estimation, in: Proceedings of
  the IEEE Conference on Computer Vision and Pattern Recognition, 2017, pp.
  5354--5362.

\bibitem{alhashim2018high}
I.~Alhashim, P.~Wonka, High quality monocular depth estimation via transfer
  learning, arXiv preprint arXiv:1812.11941.

\bibitem{silberman2011indoor}
N.~Silberman, R.~Fergus, Indoor scene segmentation using a structured light
  sensor, in: 2011 IEEE international conference on computer vision workshops
  (ICCV workshops), IEEE, 2011, pp. 601--608.

\bibitem{uhrig2017sparsity}
J.~Uhrig, N.~Schneider, L.~Schneider, U.~Franke, T.~Brox, A.~Geiger, Sparsity
  invariant cnns, in: 2017 International Conference on 3D Vision (3DV), IEEE,
  2017, pp. 11--20.

\bibitem{silberman2012indoor}
N.~Silberman, D.~Hoiem, P.~Kohli, R.~Fergus, Indoor segmentation and support
  inference from rgbd images, in: European conference on computer vision,
  Springer, 2012, pp. 746--760.

\bibitem{hoiem2008closing}
D.~Hoiem, A.~A. Efros, M.~Hebert, Closing the loop in scene interpretation, in:
  2008 IEEE Conference on Computer Vision and Pattern Recognition, IEEE, 2008,
  pp. 1--8.

\bibitem{hoiem2007recovering}
D.~Hoiem, A.~A. Efros, M.~Hebert, Recovering surface layout from an image,
  International Journal of Computer Vision 75~(1) (2007) 151--172.

\bibitem{hedau2009recovering}
V.~Hedau, D.~Hoiem, D.~Forsyth, Recovering the spatial layout of cluttered
  rooms, in: 2009 IEEE 12th international conference on computer vision, IEEE,
  2009, pp. 1849--1856.

\bibitem{huang2017densely}
G.~Huang, Z.~Liu, L.~Van Der~Maaten, K.~Q. Weinberger, Densely connected
  convolutional networks, in: Proceedings of the IEEE conference on computer
  vision and pattern recognition, 2017, pp. 4700--4708.

\bibitem{howard2017mobilenets}
A.~G. Howard, M.~Zhu, B.~Chen, D.~Kalenichenko, W.~Wang, T.~Weyand,
  M.~Andreetto, H.~Adam, Mobilenets: Efficient convolutional neural networks
  for mobile vision applications, arXiv preprint arXiv:1704.04861.

\bibitem{xie2017aggregated}
S.~Xie, R.~Girshick, P.~Doll{\'a}r, Z.~Tu, K.~He, Aggregated residual
  transformations for deep neural networks, in: Proceedings of the IEEE
  conference on computer vision and pattern recognition, 2017, pp. 1492--1500.

\bibitem{hu2018squeeze}
J.~Hu, L.~Shen, G.~Sun, Squeeze-and-excitation networks, in: Proceedings of the
  IEEE conference on computer vision and pattern recognition, 2018, pp.
  7132--7141.

\bibitem{fu2018deep}
H.~Fu, M.~Gong, C.~Wang, K.~Batmanghelich, D.~Tao, Deep ordinal regression
  network for monocular depth estimation, in: Proceedings of the IEEE
  Conference on Computer Vision and Pattern Recognition, 2018, pp. 2002--2011.

\bibitem{godard2017unsupervised}
C.~Godard, O.~Mac~Aodha, G.~J. Brostow, Unsupervised monocular depth estimation
  with left-right consistency, in: Proceedings of the IEEE Conference on
  Computer Vision and Pattern Recognition, 2017, pp. 270--279.

\bibitem{eigen2014depth}
D.~Eigen, C.~Puhrsch, R.~Fergus, Depth map prediction from a single image using
  a multi-scale deep network, in: Advances in neural information processing
  systems, 2014, pp. 2366--2374.

\bibitem{laina2016deeper}
I.~Laina, C.~Rupprecht, V.~Belagiannis, F.~Tombari, N.~Navab, Deeper depth
  prediction with fully convolutional residual networks, in: 2016 Fourth
  international conference on 3D vision (3DV), IEEE, 2016, pp. 239--248.

\bibitem{he2016deep}
K.~He, X.~Zhang, S.~Ren, J.~Sun, Deep residual learning for image recognition,
  in: Proceedings of the IEEE conference on computer vision and pattern
  recognition, IEEE, Las Vegas, USA, 2016, pp. 770--778.

\bibitem{zwald2012berhu}
L.~Zwald, S.~Lambert-Lacroix, The berhu penalty and the grouped effect, arXiv
  preprint arXiv:1207.6868.

\bibitem{girshick2015fast}
R.~Girshick, Fast r-cnn, in: Proceedings of the IEEE international conference
  on computer vision, 2015, pp. 1440--1448.

\bibitem{liu2015deep}
F.~Liu, C.~Shen, G.~Lin, Deep convolutional neural fields for depth estimation
  from a single image, in: Proceedings of the IEEE conference on computer
  vision and pattern recognition, 2015, pp. 5162--5170.

\bibitem{cao2017estimating}
Y.~Cao, Z.~Wu, C.~Shen, Estimating depth from monocular images as
  classification using deep fully convolutional residual networks, IEEE
  Transactions on Circuits and Systems for Video Technology 28~(11) (2017)
  3174--3182.

\bibitem{liao2017parse}
Y.~Liao, L.~Huang, Y.~Wang, S.~Kodagoda, Y.~Yu, Y.~Liu, Parse geometry from a
  line: Monocular depth estimation with partial laser observation, in: 2017
  IEEE International Conference on Robotics and Automation (ICRA), IEEE, 2017,
  pp. 5059--5066.

\bibitem{russakovsky2015imagenet}
O.~Russakovsky, J.~Deng, H.~Su, J.~Krause, S.~Satheesh, S.~Ma, Z.~Huang,
  A.~Karpathy, A.~Khosla, M.~Bernstein, et~al., Imagenet large scale visual
  recognition challenge, International journal of computer vision 115~(3)
  (2015) 211--252.

\bibitem{hochreiter1997long}
S.~Hochreiter, J.~Schmidhuber, Long short-term memory, Neural computation 9~(8)
  (1997) 1735--1780.

\bibitem{xingjian2015convolutional}
S.~Xingjian, Z.~Chen, H.~Wang, D.-Y. Yeung, W.-K. Wong, W.-c. Woo,
  Convolutional lstm network: A machine learning approach for precipitation
  nowcasting, in: Advances in neural information processing systems, 2015, pp.
  802--810.

\bibitem{cs2018depthnet}
A.~CS~Kumar, S.~M. Bhandarkar, M.~Prasad, Depthnet: A recurrent neural network
  architecture for monocular depth prediction, in: Proceedings of the IEEE
  Conference on Computer Vision and Pattern Recognition Workshops, 2018, pp.
  283--291.

\bibitem{wang2019recurrent}
R.~Wang, S.~M. Pizer, J.-M. Frahm, Recurrent neural network for (un-)
  supervised learning of monocular video visual odometry and depth, in:
  Proceedings of the IEEE Conference on Computer Vision and Pattern
  Recognition, 2019, pp. 5555--5564.

\bibitem{xu2018structured}
D.~Xu, W.~Wang, H.~Tang, H.~Liu, N.~Sebe, E.~Ricci, Structured attention guided
  convolutional neural fields for monocular depth estimation, in: Proceedings
  of the IEEE Conference on Computer Vision and Pattern Recognition, 2018, pp.
  3917--3925.

\bibitem{chen2019attention}
Y.~Chen, H.~Zhao, Z.~Hu, Attention-based context aggregation network for
  monocular depth estimation, arXiv preprint arXiv:1901.10137.

\bibitem{jiao2018look}
J.~Jiao, Y.~Cao, Y.~Song, R.~Lau, Look deeper into depth: Monocular depth
  estimation with semantic booster and attention-driven loss, in: Proceedings
  of the European Conference on Computer Vision (ECCV), 2018, pp. 53--69.

\bibitem{wang2015towards}
P.~Wang, X.~Shen, Z.~Lin, S.~Cohen, B.~Price, A.~L. Yuille, Towards unified
  depth and semantic prediction from a single image, in: Proceedings of the
  IEEE Conference on Computer Vision and Pattern Recognition, 2015, pp.
  2800--2809.

\bibitem{hoiem2011recovering}
D.~Hoiem, A.~A. Efros, M.~Hebert, Recovering occlusion boundaries from an
  image, International Journal of Computer Vision 91~(3) (2011) 328--346.

\bibitem{liu2010single}
B.~Liu, S.~Gould, D.~Koller, Single image depth estimation from predicted
  semantic labels, in: 2010 IEEE Computer Society Conference on Computer Vision
  and Pattern Recognition, IEEE, 2010, pp. 1253--1260.

\bibitem{gupta2014learning}
S.~Gupta, R.~Girshick, P.~Arbel{\'a}ez, J.~Malik, Learning rich features from
  rgb-d images for object detection and segmentation, in: European conference
  on computer vision, Springer, 2014, pp. 345--360.

\bibitem{eigen2015predicting}
D.~Eigen, R.~Fergus, Predicting depth, surface normals and semantic labels with
  a common multi-scale convolutional architecture, in: Proceedings of the IEEE
  international conference on computer vision, 2015, pp. 2650--2658.

\bibitem{mousavian2016joint}
A.~Mousavian, H.~Pirsiavash, J.~Ko{\v{s}}eck{\'a}, Joint semantic segmentation
  and depth estimation with deep convolutional networks, in: 2016 Fourth
  International Conference on 3D Vision (3DV), IEEE, 2016, pp. 611--619.

\bibitem{zhang1999shape}
R.~Zhang, P.-S. Tsai, J.~E. Cryer, M.~Shah, Shape-from-shading: a survey, IEEE
  transactions on pattern analysis and machine intelligence 21~(8) (1999)
  690--706.

\bibitem{binford1971visual}
T.~Binford, Visual perception by computer. leee cont. on systems and control
  (1971).

\bibitem{saxena2007learning}
A.~Saxena, M.~Sun, A.~Y. Ng, Learning 3-d scene structure from a single still
  image, in: 2007 IEEE 11th International Conference on Computer Vision, IEEE,
  2007, pp. 1--8.

\bibitem{fouhey2013data}
D.~F. Fouhey, A.~Gupta, M.~Hebert, Data-driven 3d primitives for single image
  understanding, in: Proceedings of the IEEE International Conference on
  Computer Vision, 2013, pp. 3392--3399.

\bibitem{fouhey2014unfolding}
D.~F. Fouhey, A.~Gupta, M.~Hebert, Unfolding an indoor origami world, in:
  European Conference on Computer Vision, Springer, 2014, pp. 687--702.

\bibitem{qi2018geonet}
X.~Qi, R.~Liao, Z.~Liu, R.~Urtasun, J.~Jia, Geonet: Geometric neural network
  for joint depth and surface normal estimation, in: Proceedings of the IEEE
  Conference on Computer Vision and Pattern Recognition, 2018, pp. 283--291.

\bibitem{zhang2019pattern}
Z.~Zhang, Z.~Cui, C.~Xu, Y.~Yan, N.~Sebe, J.~Yang, Pattern-affinitive
  propagation across depth, surface normal and semantic segmentation, in:
  Proceedings of the IEEE Conference on Computer Vision and Pattern
  Recognition, 2019, pp. 4106--4115.

\bibitem{garg2016unsupervised}
R.~Garg, V.~K. BG, G.~Carneiro, I.~Reid, Unsupervised cnn for single view depth
  estimation: Geometry to the rescue, in: European Conference on Computer
  Vision, Springer, 2016, pp. 740--756.

\bibitem{pillai2019superdepth}
S.~Pillai, R.~Ambru{\c{s}}, A.~Gaidon, Superdepth: Self-supervised,
  super-resolved monocular depth estimation, in: 2019 International Conference
  on Robotics and Automation (ICRA), IEEE, 2019, pp. 9250--9256.

\bibitem{zhou2017unsupervised}
T.~Zhou, M.~Brown, N.~Snavely, D.~G. Lowe, Unsupervised learning of depth and
  ego-motion from video, in: Proceedings of the IEEE Conference on Computer
  Vision and Pattern Recognition, 2017, pp. 1851--1858.

\bibitem{godard2019digging}
C.~Godard, O.~Mac~Aodha, M.~Firman, G.~J. Brostow, Digging into self-supervised
  monocular depth estimation, in: Proceedings of the IEEE International
  Conference on Computer Vision, 2019, pp. 3828--3838.

\bibitem{kuznietsov2017semi}
Y.~Kuznietsov, J.~Stuckler, B.~Leibe, Semi-supervised deep learning for
  monocular depth map prediction, in: Proceedings of the IEEE conference on
  computer vision and pattern recognition, 2017, pp. 6647--6655.

\bibitem{amiri2019semi}
A.~J. Amiri, S.~Y. Loo, H.~Zhang, Semi-supervised monocular depth estimation
  with left-right consistency using deep neural network, in: 2019 IEEE
  International Conference on Robotics and Biomimetics (ROBIO), IEEE, 2019, pp.
  602--607.

\bibitem{yang2018deep}
N.~Yang, R.~Wang, J.~Stuckler, D.~Cremers, Deep virtual stereo odometry:
  Leveraging deep depth prediction for monocular direct sparse odometry, in:
  Proceedings of the European Conference on Computer Vision (ECCV), 2018, pp.
  817--833.

\bibitem{andraghetti2019enhancing}
L.~Andraghetti, P.~Myriokefalitakis, P.~L. Dovesi, B.~Luque, M.~Poggi,
  A.~Pieropan, S.~Mattoccia, Enhancing self-supervised monocular depth
  estimation with traditional visual odometry, in: 2019 International
  Conference on 3D Vision (3DV), IEEE, 2019, pp. 424--433.

\bibitem{ramirez2018geometry}
P.~Z. Ramirez, M.~Poggi, F.~Tosi, S.~Mattoccia, L.~Di~Stefano, Geometry meets
  semantics for semi-supervised monocular depth estimation, in: Asian
  Conference on Computer Vision, Springer, 2018, pp. 298--313.

\bibitem{luo2018single}
Y.~Luo, J.~Ren, M.~Lin, J.~Pang, W.~Sun, H.~Li, L.~Lin, Single view stereo
  matching, in: Proceedings of the IEEE Conference on Computer Vision and
  Pattern Recognition, 2018, pp. 155--163.

\bibitem{tosi2019learning}
F.~Tosi, F.~Aleotti, M.~Poggi, S.~Mattoccia, Learning monocular depth
  estimation infusing traditional stereo knowledge, in: Proceedings of the IEEE
  Conference on Computer Vision and Pattern Recognition, 2019, pp. 9799--9809.

\bibitem{li2018undeepvo}
R.~Li, S.~Wang, Z.~Long, D.~Gu, Undeepvo: Monocular visual odometry through
  unsupervised deep learning, in: 2018 IEEE international conference on
  robotics and automation (ICRA), IEEE, 2018, pp. 7286--7291.

\bibitem{babu2018undemon}
V.~M. Babu, K.~Das, A.~Majumdar, S.~Kumar, Undemon: Unsupervised deep network
  for depth and ego-motion estimation, in: 2018 IEEE/RSJ International
  Conference on Intelligent Robots and Systems (IROS), IEEE, 2018, pp.
  1082--1088.

\bibitem{charbonnier1994two}
P.~Charbonnier, L.~Blanc-Feraud, G.~Aubert, M.~Barlaud, Two deterministic
  half-quadratic regularization algorithms for computed imaging, in:
  Proceedings of 1st International Conference on Image Processing, Vol.~2,
  IEEE, 1994, pp. 168--172.

\bibitem{poggi2018learning}
M.~Poggi, F.~Tosi, S.~Mattoccia, Learning monocular depth estimation with
  unsupervised trinocular assumptions, in: 2018 International Conference on 3D
  Vision (3DV), IEEE, 2018, pp. 324--333.

\bibitem{guo2018learning}
X.~Guo, H.~Li, S.~Yi, J.~Ren, X.~Wang, Learning monocular depth by distilling
  cross-domain stereo networks, in: Proceedings of the European Conference on
  Computer Vision (ECCV), 2018, pp. 484--500.

\bibitem{cs2018monocular}
A.~CS~Kumar, S.~M. Bhandarkar, M.~Prasad, Monocular depth prediction using
  generative adversarial networks, in: Proceedings of the IEEE Conference on
  Computer Vision and Pattern Recognition Workshops, 2018, pp. 300--308.

\bibitem{aleotti2018generative}
F.~Aleotti, F.~Tosi, M.~Poggi, S.~Mattoccia, Generative adversarial networks
  for unsupervised monocular depth prediction, in: Proceedings of the European
  Conference on Computer Vision (ECCV), 2018, pp. 0--0.

\bibitem{gupta2019unsupervised}
H.~Gupta, K.~Mitra, Unsupervised single image underwater depth estimation, in:
  2019 IEEE International Conference on Image Processing (ICIP), IEEE, 2019,
  pp. 624--628.

\bibitem{chen2020self}
L.~Chen, W.~Tang, T.~R. Wan, N.~W. John, Self-supervised monocular image depth
  learning and confidence estimation, Neurocomputing 381 (2020) 272--281.

\bibitem{poggi2020uncertainty}
M.~Poggi, F.~Aleotti, F.~Tosi, S.~Mattoccia, On the uncertainty of
  self-supervised monocular depth estimation, in: Proceedings of the IEEE/CVF
  Conference on Computer Vision and Pattern Recognition, 2020, pp. 3227--3237.

\bibitem{gonzalezbello2020forget}
J.~L. GonzalezBello, M.~Kim, Forget about the lidar: Self-supervised depth
  estimators with med probability volumes, Advances in Neural Information
  Processing Systems 33.

\bibitem{pilzer2018unsupervised}
A.~Pilzer, D.~Xu, M.~Puscas, E.~Ricci, N.~Sebe, Unsupervised adversarial depth
  estimation using cycled generative networks, in: 2018 International
  Conference on 3D Vision (3DV), IEEE, 2018, pp. 587--595.

\bibitem{poggi2018towards}
M.~Poggi, F.~Aleotti, F.~Tosi, S.~Mattoccia, Towards real-time unsupervised
  monocular depth estimation on cpu, in: 2018 IEEE/RSJ International Conference
  on Intelligent Robots and Systems (IROS), IEEE, 2018, pp. 5848--5854.

\bibitem{peluso2019enabling}
V.~Peluso, A.~Cipolletta, A.~Calimera, M.~Poggi, F.~Tosi, S.~Mattoccia,
  Enabling energy-efficient unsupervised monocular depth estimation on
  armv7-based platforms, in: 2019 Design, Automation \& Test in Europe
  Conference \& Exhibition (DATE), IEEE, 2019, pp. 1703--1708.

\bibitem{elkerdawy2019lightweight}
S.~Elkerdawy, H.~Zhang, N.~Ray, Lightweight monocular depth estimation model by
  joint end-to-end filter pruning, in: 2019 IEEE International Conference on
  Image Processing (ICIP), IEEE, 2019, pp. 4290--4294.

\bibitem{yang2018unsupervised}
Z.~Yang, P.~Wang, W.~Xu, L.~Zhao, R.~Nevatia, Unsupervised learning of geometry
  from videos with edge-aware depth-normal consistency, in: Thirty-Second AAAI
  Conference on Artificial Intelligence, 2018.

\bibitem{yang2018lego}
Z.~Yang, P.~Wang, Y.~Wang, W.~Xu, R.~Nevatia, Lego: Learning edge with geometry
  all at once by watching videos, in: Proceedings of the IEEE Conference on
  Computer Vision and Pattern Recognition, 2018, pp. 225--234.

\bibitem{yin2018geonet}
Z.~Yin, J.~Shi, Geonet: Unsupervised learning of dense depth, optical flow and
  camera pose, in: Proceedings of the IEEE Conference on Computer Vision and
  Pattern Recognition, 2018, pp. 1983--1992.

\bibitem{mahjourian2018unsupervised}
R.~Mahjourian, M.~Wicke, A.~Angelova, Unsupervised learning of depth and
  ego-motion from monocular video using 3d geometric constraints, in:
  Proceedings of the IEEE Conference on Computer Vision and Pattern
  Recognition, 2018, pp. 5667--5675.

\bibitem{wang2018learning}
C.~Wang, J.~Miguel~Buenaposada, R.~Zhu, S.~Lucey, Learning depth from monocular
  videos using direct methods, in: Proceedings of the IEEE Conference on
  Computer Vision and Pattern Recognition, 2018, pp. 2022--2030.

\bibitem{casser2019depth}
V.~Casser, S.~Pirk, R.~Mahjourian, A.~Angelova, Depth prediction without the
  sensors: Leveraging structure for unsupervised learning from monocular
  videos, in: Proceedings of the AAAI Conference on Artificial Intelligence,
  Vol.~33, 2019, pp. 8001--8008.

\bibitem{teed2018deepv2d}
Z.~Teed, J.~Deng, Deepv2d: Video to depth with differentiable structure from
  motion, arXiv preprint arXiv:1812.04605.

\bibitem{casser2019unsupervised}
V.~Casser, S.~Pirk, R.~Mahjourian, A.~Angelova, Unsupervised monocular depth
  and ego-motion learning with structure and semantics, in: Proceedings of the
  IEEE Conference on Computer Vision and Pattern Recognition Workshops, 2019,
  pp. 0--0.

\bibitem{ranjan2019competitive}
A.~Ranjan, V.~Jampani, L.~Balles, K.~Kim, D.~Sun, J.~Wulff, M.~J. Black,
  Competitive collaboration: Joint unsupervised learning of depth, camera
  motion, optical flow and motion segmentation, in: Proceedings of the IEEE
  Conference on Computer Vision and Pattern Recognition, 2019, pp.
  12240--12249.

\bibitem{chen2019self}
Y.~Chen, C.~Schmid, C.~Sminchisescu, Self-supervised learning with geometric
  constraints in monocular video: Connecting flow, depth, and camera, in:
  Proceedings of the IEEE International Conference on Computer Vision, 2019,
  pp. 7063--7072.

\bibitem{guizilini2019packnet}
V.~Guizilini, R.~Ambrus, S.~Pillai, A.~Gaidon, Packnet-sfm: 3d packing for
  self-supervised monocular depth estimation, arXiv preprint arXiv:1905.02693.

\bibitem{sharma2019unsupervised}
A.~Sharma, J.~Ventura, Unsupervised learning of depth and ego-motion from
  cylindrical panoramic video, arXiv preprint arXiv:1901.00979.

\bibitem{liu2020mininet}
J.~Liu, Q.~Li, R.~Cao, W.~Tang, G.~Qiu, Mininet: An extremely lightweight
  convolutional neural network for real-time unsupervised monocular depth
  estimation, ISPRS Journal of Photogrammetry and Remote Sensing 166 (2020)
  255--267.

\bibitem{cheng2019cspn++}
X.~Cheng, P.~Wang, C.~Guan, R.~Yang, Cspn++: Learning context and resource
  aware convolutional spatial propagation networks for depth completion, arXiv
  preprint arXiv:1911.05377.

\bibitem{herrera2013depth}
D.~Herrera, J.~Kannala, J.~Heikkil{\"a}, et~al., Depth map inpainting under a
  second-order smoothness prior, in: Scandinavian Conference on Image Analysis,
  Springer, 2013, pp. 555--566.

\bibitem{zhang2018probability}
H.-T. Zhang, J.~Yu, Z.-F. Wang, Probability contour guided depth map inpainting
  and superresolution using non-local total generalized variation, Multimedia
  Tools and Applications 77~(7) (2018) 9003--9020.

\bibitem{zhang2016fast}
X.~Zhang, R.~Wu, Fast depth image denoising and enhancement using a deep
  convolutional network, in: 2016 IEEE International Conference on Acoustics,
  Speech and Signal Processing (ICASSP), IEEE, 2016, pp. 2499--2503.

\bibitem{shen2013layer}
J.~Shen, S.-C.~S. Cheung, Layer depth denoising and completion for
  structured-light rgb-d cameras, in: Proceedings of the IEEE conference on
  computer vision and pattern recognition, 2013, pp. 1187--1194.

\bibitem{yu2013shading}
L.-F. Yu, S.-K. Yeung, Y.-W. Tai, S.~Lin, Shading-based shape refinement of
  rgb-d images, in: Proceedings of the IEEE Conference on Computer Vision and
  Pattern Recognition, 2013, pp. 1415--1422.

\bibitem{lu2015sparse}
J.~Lu, D.~Forsyth, Sparse depth super resolution, in: Proceedings of the IEEE
  Conference on Computer Vision and Pattern Recognition, 2015, pp. 2245--2253.

\bibitem{qiu2019deeplidar}
J.~Qiu, Z.~Cui, Y.~Zhang, X.~Zhang, S.~Liu, B.~Zeng, M.~Pollefeys, Deeplidar:
  Deep surface normal guided depth prediction for outdoor scene from sparse
  lidar data and single color image, in: Proceedings of the IEEE Conference on
  Computer Vision and Pattern Recognition, 2019, pp. 3313--3322.

\bibitem{ku2018defense}
J.~Ku, A.~Harakeh, S.~L. Waslander, In defense of classical image processing:
  Fast depth completion on the cpu, in: 2018 15th Conference on Computer and
  Robot Vision (CRV), IEEE, 2018, pp. 16--22.

\bibitem{schneider2016semantically}
N.~Schneider, L.~Schneider, P.~Pinggera, U.~Franke, M.~Pollefeys, C.~Stiller,
  Semantically guided depth upsampling, in: German conference on pattern
  recognition, Springer, 2016, pp. 37--48.

\bibitem{eldesokey2019confidence}
A.~Eldesokey, M.~Felsberg, F.~S. Khan, Confidence propagation through cnns for
  guided sparse depth regression, IEEE transactions on pattern analysis and
  machine intelligence.

\bibitem{chodosh2018deep}
N.~Chodosh, C.~Wang, S.~Lucey, Deep convolutional compressed sensing for lidar
  depth completion, in: Asian Conference on Computer Vision, Springer, 2018,
  pp. 499--513.

\bibitem{huang2019hms}
Z.~Huang, J.~Fan, S.~Cheng, S.~Yi, X.~Wang, H.~Li, Hms-net: Hierarchical
  multi-scale sparsity-invariant network for sparse depth completion, IEEE
  Transactions on Image Processing.

\bibitem{yan2020revisiting}
L.~Yan, K.~Liu, E.~Belyaev, Revisiting sparsity invariant convolution: A
  network for image guided depth completion, IEEE Access.

\bibitem{lu2020depth}
K.~Lu, N.~Barnes, S.~Anwar, L.~Zheng, From depth what can you see? depth
  completion via auxiliary image reconstruction, in: Proceedings of the
  IEEE/CVF Conference on Computer Vision and Pattern Recognition, 2020, pp.
  11306--11315.

\bibitem{tang2019learning}
J.~Tang, F.-P. Tian, W.~Feng, J.~Li, P.~Tan, Learning guided convolutional
  network for depth completion, arXiv preprint arXiv:1908.01238.

\bibitem{cheng2018depth}
X.~Cheng, P.~Wang, R.~Yang, Depth estimation via affinity learned with
  convolutional spatial propagation network, in: Proceedings of the European
  Conference on Computer Vision (ECCV), 2018, pp. 103--119.

\bibitem{park2020non}
J.~Park, K.~Joo, Z.~Hu, C.-K. Liu, I.~S. Kweon, Non-local spatial propagation
  network for depth completion, arXiv preprint arXiv:2007.10042.

\bibitem{xu2020deformable}
Z.~Xu, Y.~Wang, J.~Yao, Deformable spatial propagation network for depth
  completion, arXiv preprint arXiv:2007.04251.

\bibitem{chen2019learning}
Y.~Chen, B.~Yang, M.~Liang, R.~Urtasun, Learning joint 2d-3d representations
  for depth completion, in: Proceedings of the IEEE International Conference on
  Computer Vision, 2019, pp. 10023--10032.

\bibitem{li2020multi}
A.~Li, Z.~Yuan, Y.~Ling, W.~Chi, C.~Zhang, et~al., A multi-scale guided cascade
  hourglass network for depth completion, in: The IEEE Winter Conference on
  Applications of Computer Vision, 2020, pp. 32--40.

\bibitem{dimitrievski2018learning}
M.~Dimitrievski, P.~Veelaert, W.~Philips, Learning morphological operators for
  depth completion, in: International Conference on Advanced Concepts for
  Intelligent Vision Systems, Springer, 2018, pp. 450--461.

\bibitem{xu2019depth}
Y.~Xu, X.~Zhu, J.~Shi, G.~Zhang, H.~Bao, H.~Li, Depth completion from sparse
  lidar data with depth-normal constraints, in: Proceedings of the IEEE
  International Conference on Computer Vision, 2019, pp. 2811--2820.

\bibitem{wu2019phasecam3d}
Y.~Wu, V.~Boominathan, H.~Chen, A.~Sankaranarayanan, A.~Veeraraghavan,
  Phasecam3d—learning phase masks for passive single view depth estimation,
  in: 2019 IEEE International Conference on Computational Photography (ICCP),
  IEEE, 2019, pp. 1--12.

\bibitem{lee2020deep}
S.~Lee, J.~Lee, D.~Kim, J.~Kim, Deep architecture with cross guidance between
  single image and sparse lidar data for depth completion, IEEE Access 8 (2020)
  79801--79810.

\bibitem{klodt2018supervising}
M.~Klodt, A.~Vedaldi, Supervising the new with the old: learning sfm from sfm,
  in: Proceedings of the European Conference on Computer Vision (ECCV), 2018,
  pp. 698--713.

\bibitem{zou2018df}
Y.~Zou, Z.~Luo, J.-B. Huang, Df-net: Unsupervised joint learning of depth and
  flow using cross-task consistency, in: Proceedings of the European Conference
  on Computer Vision (ECCV), 2018, pp. 36--53.

\bibitem{li2018deep}
R.~Li, K.~Xian, C.~Shen, Z.~Cao, H.~Lu, L.~Hang, Deep attention-based
  classification network for robust depth prediction, in: Asian Conference on
  Computer Vision, Springer, 2018, pp. 663--678.

\bibitem{kong2019pixel}
S.~Kong, C.~Fowlkes, Pixel-wise attentional gating for scene parsing, in: 2019
  IEEE Winter Conference on Applications of Computer Vision (WACV), IEEE, 2019,
  pp. 1024--1033.

\bibitem{li2018monocular}
B.~Li, Y.~Dai, M.~He, Monocular depth estimation with hierarchical fusion of
  dilated cnns and soft-weighted-sum inference, Pattern Recognition 83 (2018)
  328--339.

\bibitem{goldman2019learn}
M.~Goldman, T.~Hassner, S.~Avidan, Learn stereo, infer mono: Siamese networks
  for self-supervised, monocular, depth estimation, in: Proceedings of the IEEE
  Conference on Computer Vision and Pattern Recognition Workshops, 2019, pp.
  0--0.

\bibitem{zhang2018deeph}
Z.~Zhang, C.~Xu, J.~Yang, Y.~Tai, L.~Chen, Deep hierarchical guidance and
  regularization learning for end-to-end depth estimation, Pattern Recognition
  83 (2018) 430--442.

\bibitem{ochs2019sdnet}
M.~Ochs, A.~Kretz, R.~Mester, Sdnet: Semantically guided depth estimation
  network, in: German Conference on Pattern Recognition, Springer, 2019, pp.
  288--302.

\bibitem{zhou2019unsupervised}
J.~Zhou, Y.~Wang, K.~Qin, W.~Zeng, Unsupervised high-resolution depth learning
  from videos with dual networks, in: Proceedings of the IEEE International
  Conference on Computer Vision, 2019, pp. 6872--6881.

\bibitem{sandler2018mobilenetv2}
M.~Sandler, A.~Howard, M.~Zhu, A.~Zhmoginov, L.-C. Chen, Mobilenetv2: Inverted
  residuals and linear bottlenecks, in: Proceedings of the IEEE conference on
  computer vision and pattern recognition, 2018, pp. 4510--4520.

\bibitem{ribeiro2019fast}
E.~G. Ribeiro, V.~Grassi, Fast convolutional neural network for real-time
  robotic grasp detection, in: 2019 19th International Conference on Advanced
  Robotics (ICAR), IEEE, 2019, pp. 49--54.

\bibitem{yang2018denseaspp}
M.~Yang, K.~Yu, C.~Zhang, Z.~Li, K.~Yang, Denseaspp for semantic segmentation
  in street scenes, in: Proceedings of the IEEE Conference on Computer Vision
  and Pattern Recognition, 2018, pp. 3684--3692.

\bibitem{agarap2018deep}
A.~F. Agarap, Deep learning using rectified linear units (relu), arXiv preprint
  arXiv:1803.08375.

\bibitem{clevert2015fast}
D.-A. Clevert, T.~Unterthiner, S.~Hochreiter, Fast and accurate deep network
  learning by exponential linear units (elus), arXiv preprint arXiv:1511.07289.

\bibitem{cordts2016cityscapes}
M.~Cordts, M.~Omran, S.~Ramos, T.~Rehfeld, M.~Enzweiler, R.~Benenson,
  U.~Franke, S.~Roth, B.~Schiele, The cityscapes dataset for semantic urban
  scene understanding, in: Proceedings of the IEEE conference on computer
  vision and pattern recognition, 2016, pp. 3213--3223.

\bibitem{song2015sun}
S.~Song, S.~P. Lichtenberg, J.~Xiao, Sun rgb-d: A rgb-d scene understanding
  benchmark suite, in: Proceedings of the IEEE conference on computer vision
  and pattern recognition, 2015, pp. 567--576.

\bibitem{janoch2013category}
A.~Janoch, S.~Karayev, Y.~Jia, J.~T. Barron, M.~Fritz, K.~Saenko, T.~Darrell, A
  category-level 3d object dataset: Putting the kinect to work, in: Consumer
  depth cameras for computer vision, Springer, 2013, pp. 141--165.

\bibitem{xiao2013sun3d}
J.~Xiao, A.~Owens, A.~Torralba, Sun3d: A database of big spaces reconstructed
  using sfm and object labels, in: Proceedings of the IEEE international
  conference on computer vision, 2013, pp. 1625--1632.

\bibitem{carvalho2018regression}
M.~Carvalho, B.~Le~Saux, P.~Trouv{\'e}-Peloux, A.~Almansa, F.~Champagnat, On
  regression losses for deep depth estimation, in: 2018 25th IEEE International
  Conference on Image Processing (ICIP), IEEE, 2018, pp. 2915--2919.

\bibitem{sun2014quantitative}
D.~Sun, S.~Roth, M.~J. Black, A quantitative analysis of current practices in
  optical flow estimation and the principles behind them, International Journal
  of Computer Vision 106~(2) (2014) 115--137.

\bibitem{chen2018log}
P.~Chen, G.~Chen, S.~Zhang, Log hyperbolic cosine loss improves variational
  auto-encoder.

\bibitem{nixon2019feature}
M.~Nixon, A.~Aguado, Feature extraction and image processing for computer
  vision, Academic press, 2019.

\bibitem{afshari2017gaussian}
H.~H. Afshari, S.~A. Gadsden, S.~Habibi, Gaussian filters for parameter and
  state estimation: A general review of theory and recent trends, Signal
  Processing 135 (2017) 218--238.

\bibitem{kingma2014adam}
D.~P. Kingma, J.~Ba, Adam: A method for stochastic optimization (2014).
\newblock \href {http://arxiv.org/abs/1412.6980} {\path{arXiv:1412.6980}}.

\bibitem{geiger2013vision}
A.~Geiger, P.~Lenz, C.~Stiller, R.~Urtasun, Vision meets robotics: The kitti
  dataset, The International Journal of Robotics Research 32~(11) (2013)
  1231--1237.

\bibitem{Geiger2012CVPR}
A.~Geiger, P.~Lenz, R.~Urtasun, Are we ready for autonomous driving? the kitti
  vision benchmark suite, in: Conference on Computer Vision and Pattern
  Recognition (CVPR), 2012.

\bibitem{bansal2016marr}
A.~Bansal, B.~Russell, A.~Gupta, Marr revisited: 2d-3d alignment via surface
  normal prediction, in: Proceedings of the IEEE conference on computer vision
  and pattern recognition, 2016, pp. 5965--5974.

\bibitem{dos2019sparse}
N.~dos Santos~Rosa, V.~Guizilini, V.~Grassi, Sparse-to-continuous: Enhancing
  monocular depth estimation using occupancy maps, in: 2019 19th International
  Conference on Advanced Robotics (ICAR), IEEE, 2019, pp. 793--800.

\bibitem{shivakumar2019dfusenet}
S.~S. Shivakumar, T.~Nguyen, I.~D. Miller, S.~W. Chen, V.~Kumar, C.~J. Taylor,
  Dfusenet: Deep fusion of rgb and sparse depth information for image guided
  dense depth completion, in: 2019 IEEE Intelligent Transportation Systems
  Conference (ITSC), IEEE, 2019, pp. 13--20.

\bibitem{wong2020unsupervised}
A.~Wong, X.~Fei, S.~Tsuei, S.~Soatto, Unsupervised depth completion from visual
  inertial odometry, IEEE Robotics and Automation Letters 5~(2) (2020)
  1899--1906.

\bibitem{imran2019depth}
S.~Imran, Y.~Long, X.~Liu, D.~Morris, Depth coefficients for depth completion,
  in: 2019 IEEE/CVF Conference on Computer Vision and Pattern Recognition
  (CVPR), IEEE, 2019, pp. 12438--12447.

\bibitem{hekmatian2019conf}
H.~Hekmatian, S.~Al-Stouhi, J.~Jin, Conf-net: Predicting depth completion
  error-map forhigh-confidence dense 3d point-cloud, arXiv preprint
  arXiv:1907.10148.

\bibitem{zhang2019dfinenet}
Y.~Zhang, T.~Nguyen, I.~D. Miller, S.~S. Shivakumar, S.~Chen, C.~J. Taylor,
  V.~Kumar, Dfinenet: Ego-motion estimation and depth refinement from sparse,
  noisy depth input with rgb guidance, arXiv preprint arXiv:1903.06397.

\bibitem{jaritz2018sparse}
M.~Jaritz, R.~De~Charette, E.~Wirbel, X.~Perrotton, F.~Nashashibi, Sparse and
  dense data with cnns: Depth completion and semantic segmentation, in: 2018
  International Conference on 3D Vision (3DV), IEEE, 2018, pp. 52--60.

\bibitem{yang2019dense}
Y.~Yang, A.~Wong, S.~Soatto, Dense depth posterior (ddp) from single image and
  sparse range, in: Proceedings of the IEEE Conference on Computer Vision and
  Pattern Recognition, 2019, pp. 3353--3362.

\bibitem{Alhaija2018IJCV}
H.~Alhaija, S.~Mustikovela, L.~Mescheder, A.~Geiger, C.~Rother, Augmented
  reality meets computer vision: Efficient data generation for urban driving
  scenes, International Journal of Computer Vision (IJCV).

\bibitem{saxena2008make3d}
A.~Saxena, M.~Sun, A.~Y. Ng, Make3d: Depth perception from a single still
  image., in: AAAI, Vol.~3, 2008, pp. 1571--1576.

\bibitem{liu2015learning}
F.~Liu, C.~Shen, G.~Lin, I.~Reid, Learning depth from single monocular images
  using deep convolutional neural fields, IEEE transactions on pattern analysis
  and machine intelligence 38~(10) (2015) 2024--2039.

\bibitem{yang2017unsupervised}
Z.~Yang, P.~Wang, W.~Xu, L.~Zhao, R.~Nevatia, Unsupervised learning of geometry
  with edge-aware depth-normal consistency, arXiv preprint arXiv:1711.03665.

\bibitem{nath2018adadepth}
J.~Nath~Kundu, P.~Krishna~Uppala, A.~Pahuja, R.~Venkatesh~Babu, Adadepth:
  Unsupervised content congruent adaptation for depth estimation, in:
  Proceedings of the IEEE Conference on Computer Vision and Pattern
  Recognition, 2018, pp. 2656--2665.

\bibitem{zhang2018progressive}
Z.~Zhang, C.~Xu, J.~Yang, J.~Gao, Z.~Cui, Progressive hard-mining network for
  monocular depth estimation, IEEE Transactions on Image Processing 27~(8)
  (2018) 3691--3702.

\bibitem{zhan2018unsupervised}
H.~Zhan, R.~Garg, C.~Saroj~Weerasekera, K.~Li, H.~Agarwal, I.~Reid,
  Unsupervised learning of monocular depth estimation and visual odometry with
  deep feature reconstruction, in: Proceedings of the IEEE Conference on
  Computer Vision and Pattern Recognition, 2018, pp. 340--349.

\bibitem{wong2019bilateral}
A.~Wong, S.~Soatto, Bilateral cyclic constraint and adaptive regularization for
  unsupervised monocular depth prediction, in: Proceedings of the IEEE
  Conference on Computer Vision and Pattern Recognition, 2019, pp. 5644--5653.

\bibitem{mehta2018structured}
I.~Mehta, P.~Sakurikar, P.~Narayanan, Structured adversarial training for
  unsupervised monocular depth estimation, in: 2018 International Conference on
  3D Vision (3DV), IEEE, 2018, pp. 314--323.

\bibitem{luo2018every}
C.~Luo, Z.~Yang, P.~Wang, Y.~Wang, W.~Xu, R.~Nevatia, A.~Yuille, Every pixel
  counts++: Joint learning of geometry and motion with 3d holistic
  understanding, arXiv preprint arXiv:1810.06125.

\bibitem{gordon2019depth}
A.~Gordon, H.~Li, R.~Jonschkowski, A.~Angelova, Depth from videos in the wild:
  Unsupervised monocular depth learning from unknown cameras, in: Proceedings
  of the IEEE International Conference on Computer Vision, 2019, pp.
  8977--8986.

\bibitem{chen2019towards}
P.-Y. Chen, A.~H. Liu, Y.-C. Liu, Y.-C.~F. Wang, Towards scene understanding:
  Unsupervised monocular depth estimation with semantic-aware representation,
  in: Proceedings of the IEEE Conference on Computer Vision and Pattern
  Recognition, 2019, pp. 2624--2632.

\bibitem{su2020monocular}
W.~Su, H.~Zhang, Q.~Zhou, W.~Yang, Z.~Wang, Monocular depth estimation using
  information exchange network, IEEE Transactions on Intelligent Transportation
  Systems.

\bibitem{gur2019single}
S.~Gur, L.~Wolf, Single image depth estimation trained via depth from defocus
  cues, in: Proceedings of the IEEE Conference on Computer Vision and Pattern
  Recognition, 2019, pp. 7683--7692.

\bibitem{gurram2018monocular}
A.~Gurram, O.~Urfalioglu, I.~Halfaoui, F.~Bouzaraa, A.~M. L{\'o}pez, Monocular
  depth estimation by learning from heterogeneous datasets, in: 2018 IEEE
  Intelligent Vehicles Symposium (IV), IEEE, 2018, pp. 2176--2181.

\bibitem{pilzer2019refine}
A.~Pilzer, S.~Lathuiliere, N.~Sebe, E.~Ricci, Refine and distill: Exploiting
  cycle-inconsistency and knowledge distillation for unsupervised monocular
  depth estimation, in: Proceedings of the IEEE Conference on Computer Vision
  and Pattern Recognition, 2019, pp. 9768--9777.

\bibitem{zhu2019structure}
J.~Zhu, Y.~Shi, M.~Ren, Y.~Fang, K.-C. Lien, J.~Gu, Structure-attentioned
  memory network for monocular depth estimation, arXiv preprint
  arXiv:1909.04594.

\bibitem{watson2019self}
J.~Watson, M.~Firman, G.~J. Brostow, D.~Turmukhambetov, Self-supervised
  monocular depth hints, in: Proceedings of the IEEE International Conference
  on Computer Vision, 2019, pp. 2162--2171.

\bibitem{tang2018ba}
C.~Tang, P.~Tan, Ba-net: Dense bundle adjustment network, arXiv preprint
  arXiv:1806.04807.

\bibitem{kim2018deep}
Y.~Kim, H.~Jung, D.~Min, K.~Sohn, Deep monocular depth estimation via
  integration of global and local predictions, IEEE transactions on Image
  Processing 27~(8) (2018) 4131--4144.

\bibitem{gan2018monocular}
Y.~Gan, X.~Xu, W.~Sun, L.~Lin, Monocular depth estimation with affinity,
  vertical pooling, and label enhancement, in: Proceedings of the European
  Conference on Computer Vision (ECCV), 2018, pp. 224--239.

\bibitem{cadena2016multi}
C.~Cadena, A.~R. Dick, I.~D. Reid, Multi-modal auto-encoders as joint
  estimators for robotics scene understanding., in: Robotics: Science and
  Systems, Vol.~5, 2016, p.~1.

\bibitem{liu2017learning}
S.~Liu, S.~De~Mello, J.~Gu, G.~Zhong, M.-H. Yang, J.~Kautz, Learning affinity
  via spatial propagation networks, in: Advances in Neural Information
  Processing Systems, 2017, pp. 1520--1530.

\bibitem{engel2017direct}
J.~Engel, V.~Koltun, D.~Cremers, Direct sparse odometry, IEEE transactions on
  pattern analysis and machine intelligence 40~(3) (2017) 611--625.

\bibitem{forster2014svo}
C.~Forster, M.~Pizzoli, D.~Scaramuzza, Svo: Fast semi-direct monocular visual
  odometry, in: 2014 IEEE international conference on robotics and automation
  (ICRA), IEEE, 2014, pp. 15--22.

\bibitem{mur2017orb}
R.~Mur-Artal, J.~D. Tard{\'o}s, Orb-slam2: An open-source slam system for
  monocular, stereo, and rgb-d cameras, IEEE Transactions on Robotics 33~(5)
  (2017) 1255--1262.

\bibitem{karsch2014depth}
K.~Karsch, C.~Liu, S.~B. Kang, Depth transfer: Depth extraction from video
  using non-parametric sampling, IEEE transactions on pattern analysis and
  machine intelligence 36~(11) (2014) 2144--2158.

\bibitem{liu2014discrete}
M.~Liu, M.~Salzmann, X.~He, Discrete-continuous depth estimation from a single
  image, in: Proceedings of the IEEE Conference on Computer Vision and Pattern
  Recognition, 2014, pp. 716--723.

\bibitem{ladicky2014pulling}
L.~Ladicky, J.~Shi, M.~Pollefeys, Pulling things out of perspective, in:
  Proceedings of the IEEE conference on computer vision and pattern
  recognition, 2014, pp. 89--96.

\bibitem{li2015depth}
B.~Li, C.~Shen, Y.~Dai, A.~Van Den~Hengel, M.~He, Depth and surface normal
  estimation from monocular images using regression on deep features and
  hierarchical crfs, in: Proceedings of the IEEE conference on computer vision
  and pattern recognition, 2015, pp. 1119--1127.

\bibitem{roy2016monocular}
A.~Roy, S.~Todorovic, Monocular depth estimation using neural regression
  forest, in: Proceedings of the IEEE conference on computer vision and pattern
  recognition, 2016, pp. 5506--5514.

\bibitem{chakrabarti2016depth}
A.~Chakrabarti, J.~Shao, G.~Shakhnarovich, Depth from a single image by
  harmonizing overcomplete local network predictions, in: Advances in Neural
  Information Processing Systems, 2016, pp. 2658--2666.

\bibitem{li2017two}
J.~Li, R.~Klein, A.~Yao, A two-streamed network for estimating fine-scaled
  depth maps from single rgb images, in: Proceedings of the IEEE International
  Conference on Computer Vision, 2017, pp. 3372--3380.

\bibitem{xu2018pad}
D.~Xu, W.~Ouyang, X.~Wang, N.~Sebe, Pad-net: Multi-tasks guided
  prediction-and-distillation network for simultaneous depth estimation and
  scene parsing, in: Proceedings of the IEEE Conference on Computer Vision and
  Pattern Recognition, 2018, pp. 675--684.

\bibitem{lee2018single}
J.-H. Lee, M.~Heo, K.-R. Kim, C.-S. Kim, Single-image depth estimation based on
  fourier domain analysis, in: Proceedings of the IEEE Conference on Computer
  Vision and Pattern Recognition, 2018, pp. 330--339.

\bibitem{lu2019index}
H.~Lu, Y.~Dai, C.~Shen, S.~Xu, Index network, arXiv preprint arXiv:1908.09895.

\bibitem{nekrasov2019real}
V.~Nekrasov, T.~Dharmasiri, A.~Spek, T.~Drummond, C.~Shen, I.~Reid, Real-time
  joint semantic segmentation and depth estimation using asymmetric
  annotations, in: 2019 International Conference on Robotics and Automation
  (ICRA), IEEE, 2019, pp. 7101--7107.

\bibitem{lee2019monocular}
J.-H. Lee, C.-S. Kim, Monocular depth estimation using relative depth maps, in:
  Proceedings of the IEEE Conference on Computer Vision and Pattern
  Recognition, 2019, pp. 9729--9738.

\bibitem{bian2020unsupervised}
J.-W. Bian, H.~Zhan, N.~Wang, T.-J. Chin, C.~Shen, I.~Reid, Unsupervised depth
  learning in challenging indoor video: Weak rectification to rescue, arXiv
  preprint arXiv:2006.02708.

\bibitem{hu2019revisiting}
J.~Hu, M.~Ozay, Y.~Zhang, T.~Okatani, Revisiting single image depth estimation:
  Toward higher resolution maps with accurate object boundaries, in: 2019 IEEE
  Winter Conference on Applications of Computer Vision (WACV), IEEE, 2019, pp.
  1043--1051.

\bibitem{chen2019structure}
X.~Chen, X.~Chen, Z.-J. Zha, Structure-aware residual pyramid network for
  monocular depth estimation, arXiv preprint arXiv:1907.06023.

\bibitem{zhang2018joint}
Z.~Zhang, Z.~Cui, C.~Xu, Z.~Jie, X.~Li, J.~Yang, Joint task-recursive learning
  for semantic segmentation and depth estimation, in: Proceedings of the
  European Conference on Computer Vision (ECCV), 2018, pp. 235--251.

\bibitem{wang2020sdc}
L.~Wang, J.~Zhang, O.~Wang, Z.~Lin, H.~Lu, Sdc-depth: Semantic
  divide-and-conquer network for monocular depth estimation, in: Proceedings of
  the IEEE/CVF Conference on Computer Vision and Pattern Recognition, 2020, pp.
  541--550.

\bibitem{ladicky2014discriminatively}
L.~Ladick{\`y}, B.~Zeisl, M.~Pollefeys, Discriminatively trained dense surface
  normal estimation, in: ECCV, Vol.~2, 2014, p.~4.

\bibitem{wang2015designing}
X.~Wang, D.~Fouhey, A.~Gupta, Designing deep networks for surface normal
  estimation, in: Proceedings of the IEEE Conference on Computer Vision and
  Pattern Recognition, 2015, pp. 539--547.

\bibitem{ferstl2013image}
D.~Ferstl, C.~Reinbacher, R.~Ranftl, M.~R{\"u}ther, H.~Bischof, Image guided
  depth upsampling using anisotropic total generalized variation, in:
  Proceedings of the IEEE International Conference on Computer Vision, 2013,
  pp. 993--1000.

\bibitem{zhang2018deepd}
Y.~Zhang, T.~Funkhouser, Deep depth completion of a single rgb-d image, in:
  Proceedings of the IEEE Conference on Computer Vision and Pattern
  Recognition, 2018, pp. 175--185.

\bibitem{barron2016fast}
J.~T. Barron, B.~Poole, The fast bilateral solver, in: European Conference on
  Computer Vision, Springer, 2016, pp. 617--632.

\end{thebibliography}

\end{document}